%% file: arxiv/main_arxiv.tex
\pdfoutput=1
\documentclass[10pt,logo,copyright]{arxiv/nvidiatechreport}
\input{arxiv/preamble_arxiv}

\renewcommand{\today}{2026-09-29}
\title{\methodname: Segment Anything\texorpdfstring{\protect\\}{ }in Feed-Forward 4D Visual Geometry}
\author{%
Jingdong Zhang$^{1,2,*}$, Xin Li$^1$, Jan Kautz$^2$, Wenping Wang$^1$, Chris Choy$^2$\\
$^1$\href{https://www.tamu.edu}{\textcolor{black}{Texas A\&M University, College Station}},\quad
$^2$\href{https://www.nvidia.com}{\textcolor{black}{NVIDIA}}\\
{\normalfont\small
\href{https://evergreen0929.github.io/projects/s4vy.html}{Project Page}\quad\textbar\quad
\href{https://youtu.be/rvIcANaEws4}{Video Demo}\quad\textbar\quad
\href{https://github.com/Evergreen0929/S4VY}{Code}}%
}

\hypersetup{
  pdftitle={S4VY: Segment Anything in Feed-Forward 4D Visual Geometry},
  pdfauthor={Jingdong Zhang, Xin Li, Jan Kautz, Wenping Wang, Chris Choy}
}

\renewcommand{\copyrightext}{\footerfont
  \textsuperscript{*}This work was done during Jingdong Zhang's internship at NVIDIA.\\
  \textcopyright\,\the\year{} NVIDIA. All rights reserved.}

\begin{document}
\maketitle
\fancyhead[C]{\footerfont S4VY: Segment Anything in Feed-Forward 4D Visual Geometry}

\input{arxiv/teaser_arxiv}
\input{sections/0_abstract}
\abscontent

\input{sections/1_intro}
\input{sections/2_related}

\input{arxiv/sections/3_method_report}

\input{arxiv/sections/4_grounding_report}

\input{arxiv/sections/5_data_training_report}
\input{arxiv/sections/6_experiments_report}
\input{sections/8_conclusion}

\bibliographystyle{plainnat}
\bibliography{references/main}

\clearpage
\appendix
\input{arxiv/sections/A_method_details}
\input{arxiv/sections/B_experimental_details}
\input{arxiv/sections/C_additional_results}

\end{document}

%% file: arxiv/preamble_arxiv.tex
\usepackage[authoryear,sort&compress,round]{natbib}
\usepackage{xspace}
\usepackage{multirow}
\usepackage{makecell}
\usepackage{soul}
\usepackage{pifont}
\usepackage{subcaption}
\usepackage{float}
\usepackage{placeins}
\usepackage[capitalize,nameinlink]{cleveref}

\floatstyle{ruled}
\newfloat{algorithm}{tbp}{loa}
\floatname{algorithm}{Algorithm}

\input{macros}

\DeclareUnicodeCharacter{03C0}{\ensuremath{\pi}}
\DeclareUnicodeCharacter{00D7}{\ensuremath{\times}}
\DeclareUnicodeCharacter{2192}{\ensuremath{\rightarrow}}
\DeclareUnicodeCharacter{2013}{--}
\DeclareUnicodeCharacter{2014}{---}
\DeclareUnicodeCharacter{2019}{'}
\DeclareUnicodeCharacter{00B3}{\ensuremath{^3}}

\crefname{equation}{Eq.}{Eqs.}
\crefname{figure}{Fig.}{Figs.}
\crefname{section}{Sec.}{Secs.}
\crefname{table}{Tab.}{Tabs.}
\crefname{algorithm}{Algorithm}{Algorithms}

%% file: macros.tex
\newcommand{\methodname}{S4VY\xspace}                   

\newcommand{\JF}{$\mathcal{J}\&\mathcal{F}$\xspace}

\newcommand{\tmiou}{T\text{-}mIoU\xspace}
\newcommand{\tsr}{T\text{-}SR\xspace}
\newcommand{\fdiou}{4D\text{-}IoU\xspace}
\newcommand{\fdsr}{4D\text{-}SR\xspace}
\newcommand{\stq}{STQ\xspace}

\newif\ifnotes\notestrue


%% file: arxiv/teaser_arxiv.tex
\begin{figure}[h]
\centering
\includegraphics[width=0.90\textwidth]{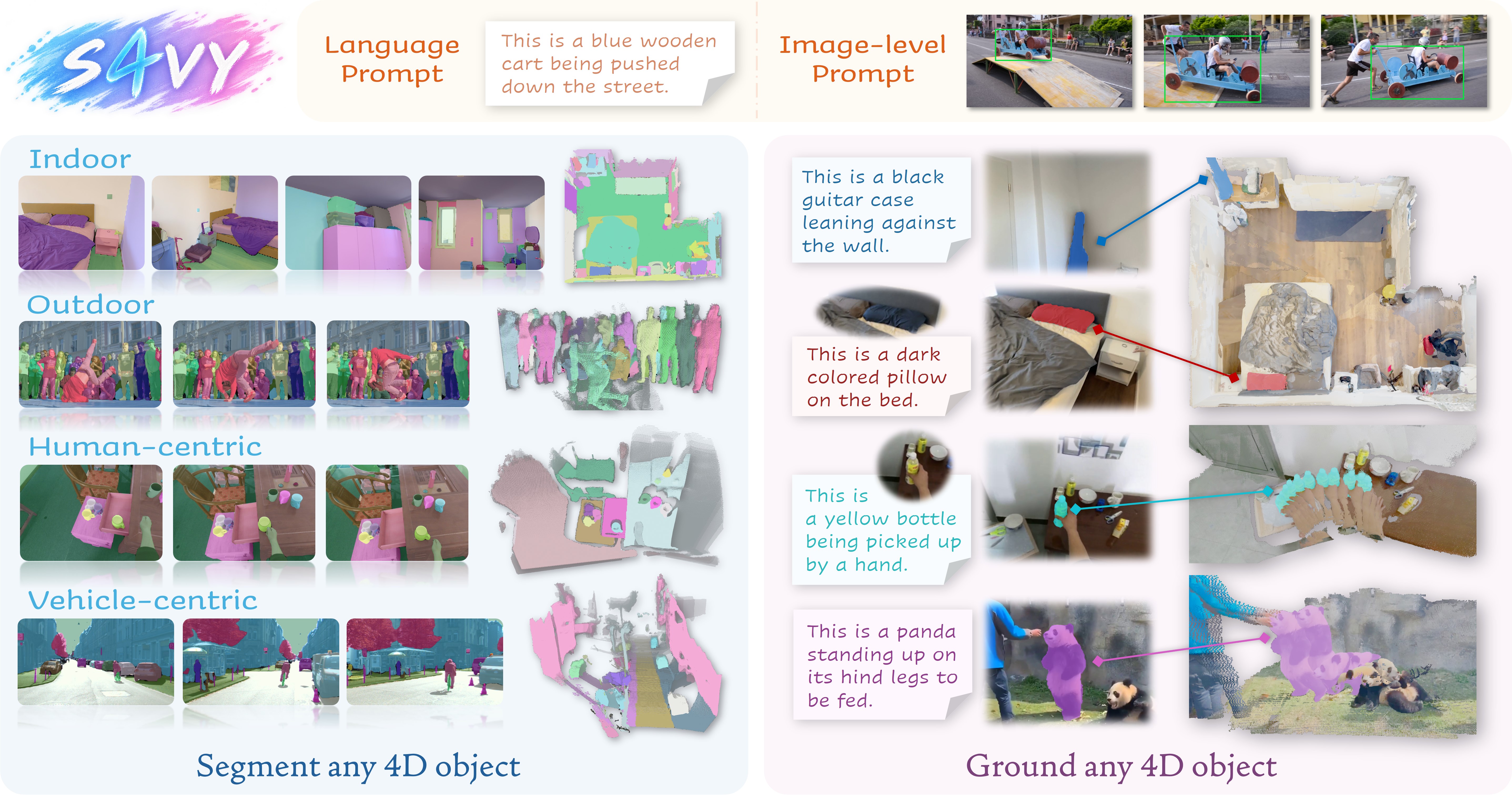}
  \caption{\textbf{Overview of \methodname{}.}
  \emph{Left:} Given a set of RGB observations, \methodname{} predicts a consistent 4D mask for every instance in the scene.
  \emph{Right:} Any instance can be grounded via either a natural-language prompt or an image-level point/box prompt.}
\label{fig:teaser}
\end{figure}

%% file: sections/0_abstract.tex
\begin{abstract}
Accurate instance segmentation in dynamic scenes is important for downstream applications such as
robotics and autonomous driving.
Existing Segment Anything models operate primarily on
2D image or video masks and preserve identity through sequential memory, while promptable 4D instance segmentation built upon feed-forward visual geometry remains underexplored. We introduce
\methodname{}, a Segment Anything model built on feed-forward 4D visual geometry. From a set of RGB
observations, \methodname{} transforms shared visual-geometric features into an
exhaustive set of class-agnostic 4D instance masks through a space-time query decoder, with each
persistent object query binding one entity across all observations. This representation supports
prompt-independent segmentation as well as point- and box-conditioned selection, without requiring
a seed mask or temporal ordering. We further develop an agentic harness for natural-language
grounding in the large observation space of a 4D scene. Active tree search identifies relevant
frames without scanning every fixed window; a dual-stream grounder combines fine-grained VLM
visual priors with geometry-consistent instance features through complementary bounding-box prediction and
object-query matching; and an independent critic selects the final 4D instance mask from their
predictions. Extensive experiments demonstrate state-of-the-art 4D instance segmentation and strong
language-guided grounding performance under a unified evaluation spanning static and dynamic scenes.

\end{abstract}

%% file: sections/1_intro.tex

\section{Introduction}
\label{sec:intro}

Instance-level segmentation of dynamic scenes is fundamental to perception and planning in embodied
systems such as humanoid robots and autonomous vehicles.
The SAM family~\citep{kirillov2023segment,ravi2025sam,carion2026sam} supports promptable
segmentation across images and videos. Yet these models remain centered on predicting sets of 2D masks
and propagating temporal memory to track instances among frames, without explicitly modeling the spatial geometry
shared across observations. This becomes a critical bottleneck in challenging dynamic scenes, where
multiple visually similar objects move simultaneously and must be re-identified after extended
occlusion or absence from view. Real-world observations may also be sparse or unordered, further
challenging sequential identity propagation. These settings call for object identities established through shared
3D/4D geometry, rather than inherited solely along an ordered sequence of 2D frames.

Segmentation on explicit 3D representations offers a direct route to spatial scene
understanding~\citep{jiang2020pointgroup,schult2023mask3d,kolodiazhnyi2024oneformer3d,
choy2026spaceformer}, but typically assumes the inputs to be point clouds or RGB-D observations
obtained from separate reconstruction pipelines or specialized sensors. This dependence potentially accumulates errors and limits data scalability. Recent feed-forward visual geometry models~\citep{wang2024dust3r,wang2025vggt,wang2026pi,wang2026vggt} recover scene geometry directly from RGB observations in a single forward pass. Their consistent geometry representations have enabled methods that couple feed-forward reconstruction with semantic, panoptic, or instance segmentation~\citep{zust2025panst3r,li2025iggt,qu2026segvggt,zou2026iggt4d}. Within this visual-geometry paradigm, most of them target static scenes. IGGT4D~\citep{zou2026iggt4d} extends instance segmentation to dynamic 4D scenes, but associates local masks with an additional feature clustering, while flexible prompt interfaces remain underexplored.

To address these limitations, we present \methodname{} (pronounced ``savvy''), named for its goal of segmenting anything in 4D visual geometry. 
As shown in \cref{fig:teaser}, \methodname{} is a Segment Anything model that unifies prompt-free, exhaustive 4D instance segmentation with prompt-conditioned selection of any discovered instance in a shared 4D visual-geometric representation. Our segmentation model operates on geometry features jointly contextualized across all input observations by a feed-forward visual geometry backbone~\citep{wang2026vggt}, regardless of their order. A Space-Time Query Decoder uses global learnable object queries to predict instance masks, with each active query maintaining one instance identity across the entire observation set. A Prompt Encoder maps point or box prompts from one or multiple observations to a promptable object query,enabling any object in the observed scene to be selected.

 Beyond point and box prompts, natural-language grounding provides an important interface for users and downstream systems to specify target instances in 4D scenes. A referent may appear in only a limited subset of a large observation set, making target discovery and consistent 4D association challenging. Vision-Language Models (VLMs) have advanced reasoning segmentation and visual grounding~\citep{lai2024lisa,rasheed2024glamm,li2025seeground,liu2025reasongrounder}, while MVGGT~\citep{wu2026mvggt} extends referring segmentation to sparse multiview RGB inputs. These methods assume static scenes or fixed sparse views, leaving target discovery and consistent grounding over large dynamic observation sets unresolved. We develop an agentic grounding harness specifically designed for feed-forward 4D visual geometry. Its dual-stream Grounder combines the geometry-consistent feature space of the segmentation model with the fine-grained visual priors of the VLM, producing complementary object-query and bounding-box predictions. Guided by the Grounder’s target-presence estimates, Active Tree Search recursively locates relevant observations across the frame space without evaluating every fixed window. A comparative Critic jointly evaluates representative overlays of the two mask candidates and selects the final 4D instance.

Our contributions are:

{\setlength{\leftmargini}{1.5em}
\begin{itemize}
\setlength{\itemsep}{1pt}
\setlength{\topsep}{2pt}
\setlength{\parsep}{0pt}
\setlength{\parskip}{0pt}
\item \textbf{Feed-forward Segment Anything in 4D.} We introduce a unified model that transforms
shared visual-geometric representations into exhaustive class-agnostic 4D instances with persistent
identities, while allowing any discovered instance to be selected by point or box prompts.
\item \textbf{Agentic language grounding in 4D.} We develop a grounding harness for large 4D
observation spaces that combines VLM visual priors with geometry-consistent instance features through
dual-stream prediction, active tree search, and comparative critic selection.
\item \textbf{Evaluation across static and dynamic scenes.} Extensive experiments establish
state-of-the-art 4D instance segmentation and strong language grounding performance under
unified evaluation.
\end{itemize}}

%% file: sections/2_related.tex

\section{Related Work}
\label{sec:related}
\vspace{-2mm}
\paragraph{Feed-forward multi-view geometry and understanding.}
With the scaling of vision transformers and training data, multi-view geometry has rapidly shifted from
optimization pipelines to feed-forward reconstruction. DUSt3R~\citep{wang2024dust3r} initiated
the modern line through pointmap regression, after which stronger correspondence, whole-set
aggregation, and greater scale steadily widened its reach~\citep{leroy2024grounding,wang20253d,
cabon2025must3r,yang2025fast3r}.  VGGT~\citep{wang2025vggt} then brought cameras,
  depth, pointmaps, and tracks into one transformer. Geometry became a shared multi-view feature space.
  $\pi^3$~\citep{wang2026pi} further removes reference-view dependence through permutation-equivariant visual geometry prediction, while VGGT-$\Omega$~\citep{wang2026vggt} extended this shared
geometric representation across static and dynamic scenes. These representations have quickly become 
substrates for scene understanding, with semantic and
panoptic prediction increasingly coupled to feed-forward reconstruction~\citep{fan2024large,
sun2026uni3r,zust2025panst3r}. At the instance level, IGGT~\citep{li2025iggt} groups
geometry-aware features, whereas SegVGGT~\citep{qu2026segvggt} and
FAST3DIS~\citep{li2026fast3dis} decode object queries; IGGT4D~\citep{zou2026iggt4d} is among the
few methods to extend joint reconstruction and instance understanding explicitly to 4D. This line
differs from methods that segment an existing point cloud or space--time scan~\citep{choy20194d,
jiang2020pointgroup,aygun20214d,vu2022softgroup,kreuzberg20224d,schult2023mask3d,
kolodiazhnyi2024oneformer3d,choy2026spaceformer,steiner2026rescene4d}, as well as approaches that
lift 2D masks into a preconstructed neural or Gaussian scene~\citep{siddiqui2023panoptic,
bhalgat2023contrastive,ye2024gaussian,li2026trase,wu2026consistent}. Within this paradigm, most methods remain limited to static multi-view scenes, while the closest dynamic extension relies on clustering-based identity association. Moreover, existing methods do not unify exhaustive 4D instance discovery with point, box, and language interfaces. S4VY addresses these limitations within a shared feed-forward 4D representation.

\vspace{-2mm}
\paragraph{Segment Anything models.}
SAM~\citep{kirillov2023segment} established promptable image segmentation, while subsequent models
extended its interface to text and controllable mask granularity and improved segmentation quality,
efficiency, and robustness~\citep{zou2023segment,li2024segment,ke2023segment,
xiong2024efficientsam,chen2023sam,zhang2024quantifying}. Video extensions either attach tracking or
temporal memory to image masks~\citep{yang2023track,cheng2023tracking,rajivc2025segment,ravi2025sam},
with SAM~3~\citep{carion2026sam} further supporting text and exemplar concepts, or represent
instances with queries shared across a clip~\citep{cheng2021mask2former,wu2022seqformer}; both remain
defined over 2D image or video representations. SAM-inspired 3D and 4D methods consume explicit
point clouds, RGB-D observations, or camera--LiDAR streams~\citep{cen2023segment,yin2024sai3d,
zhou2024point,xu2025sam4d,liu2023segment}, or lift 2D masks into preconstructed neural or Gaussian
scenes~\citep{ye2024gaussian,ji2024segment}. SAM-V~\citep{gong2026samv} integrates feed-forward
geometry into SAM but targets prompted multi-view segmentation of static scenes. \methodname{}
instead unifies exhaustive 4D instance discovery with point- and box-conditioned selection over
potentially unordered RGB observations.

\vspace{-2mm}
\paragraph{Language-guided segmentation and 4D scene grounding.}
Grounding DINO~\citep{liu2024grounding} and language-driven segmentation models such as
CRIS~\citep{wang2022cris} and Language as Queries~\citep{wu2022language} align free-form text
with image regions, masks, and video tracks. Related 3D and 4D work grounds text or open-vocabulary
features in scene representations~\citep{prabhudesai2020embodied,chen2020scanrefer,
achlioptas2020referit3d,zhao20213dvg,wu2023eda,peng2023openscene,takmaz2023openmask3d,
fiebelman20254,li20254d,zhou2025feature4x,wu20254dlangvggt}. With stronger VLMs, recent work has expanded grounding from direct semantic matching toward free-form descriptions, compositional relations, and implicit intent through reasoning segmentation, geometry-aware 3D localization, and language-guided 4D segmentation~\citep{lai2024lisa,xia2024gsva,rasheed2024glamm,ren2024pixellm,yan2024visa,
zhu2024scanreason,chen2024grounded,yang2024llm,xu2024vlm,li2025seeground,
liu2025reasongrounder,wu2026mvggt,hu2026g,li2025sam3,yuan2026instructsam,
tang2026panopticquery,li20264dvlt}.
Agentic long-video methods iteratively retrieve query-relevant observations or refine visual
evidence through hierarchical and active observation~\citep{wang2024videoagent,wang2025videotree,
li2026lenswalk,wang2026active}. Most existing scene-grounding methods operate on static scenes or fixed sparse views, while agentic video systems search an ordered timeline for
video-level reasoning. \methodname{} instead grounds persistent instances across large 4D observation sets, incorporating Active Tree Search with VLM visual priors,
geometry-consistent features, and Critic selection.

\input{figures_final/main_arch}

%% file: figures_final/main_arch.tex
\begin{figure}[t]
  \centering
  \includegraphics[width=\textwidth]{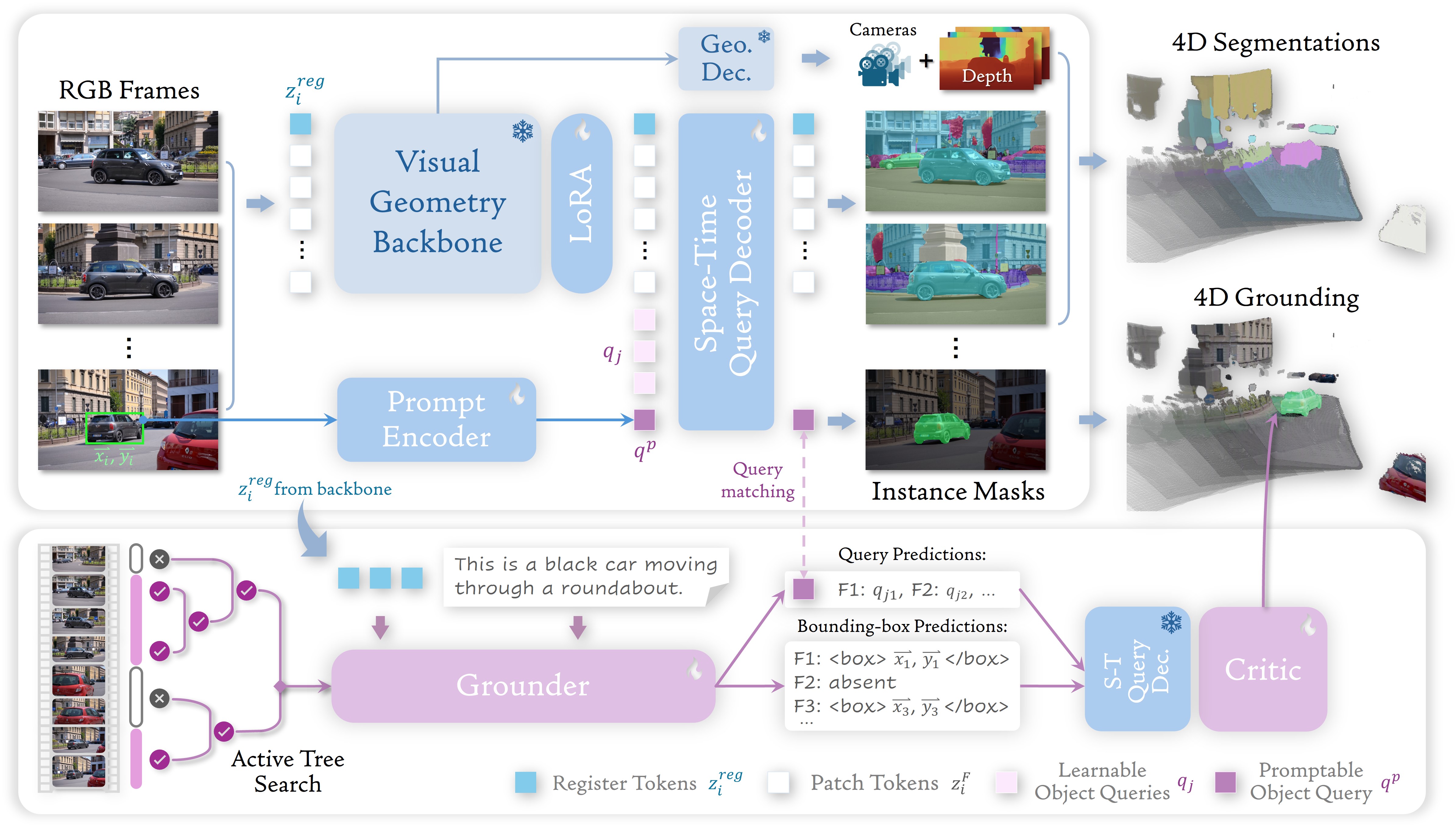}
  \vspace{-6mm}
  \caption{\textbf{Architecture of \methodname{}.}
  \emph{Top:} The backbone encodes RGB observations into register and patch tokens. The Space-Time
  Query Decoder applies learnable object queries to the patch tokens to produce exhaustive instance masks
  across observations, while the Prompt Encoder maps image-level prompts to a promptable query.
  \emph{Bottom:} For a language expression, Active Tree Search selects observation windows for the
  dual-stream Grounder, which predicts object-query matches and bounding boxes. The two predictions
  are decoded into mask candidates and compared by the Critic.}
  \label{fig:overview}
  \vspace{-3mm}
\end{figure}

%% file: arxiv/sections/3_method_report.tex
\section{Segment Anything in 4D}
\label{sec:method}

As illustrated in \cref{fig:overview}, \methodname{} builds a query-based 4D segmenter on
feed-forward visual geometry, supporting exhaustive 4D instance segmentation as well as point-
and box-prompted segmentation.

\subsection{Preliminaries and problem formulation}

\paragraph{Feed-forward visual geometry.}
Given $T$ RGB observations $\mathcal{I}=\{I_i\}_{i=1}^{T}$, where
$I_i\in\mathbb{R}^{3\times H\times W}$, our feed-forward visual geometry backbone~\citep{wang2026vggt} represents observation
$i$ with patch tokens $\mathbf{z}_i^{F}$, a camera token $\mathbf{z}_i^{\mathrm{cam}}$, and register tokens $\mathbf{z}_i^{\mathrm{reg}}$, and jointly contextualizes them through alternating
within-frame and cross-frame attention. The segmenter decodes the contextualized patch tokens,
while the register tokens are retained for the language-grounding interface in
\cref{sec:grounding}.

\paragraph{Exhaustive 4D instance segmentation.}
Let $\mathcal{S}\subset\mathbb{R}^{3}\times\{t_i\}_{i=1}^{T}$ denote the spatiotemporal scene
observed by $\mathcal{I}$, where $t_i$ is the acquisition time of $I_i$. The input observations may
be temporally sparse or presented out of chronological order. The task is to recover
\begin{equation}
  \mathcal{I}
  \mapsto \mathcal{M}=\bigl\{\mathbf{M}_j\bigr\}_{j=1}^{N},
  \qquad \mathbf{M}_j:\mathcal{S}\rightarrow\{0,1\},
  \label{eq:seg4d-formulation}
\end{equation}
where $\mathbf{M}_j$ denotes the 4D mask of instance j; exhaustive segmentation recovers the
complete set $\mathcal{M}$ of all $N$ instances in the observed scene. Following the SAM
paradigm~\citep{chen2023sam,ravi2025sam}, a point or box prompt can instead specify any individual
$\mathbf{M}_j$.

\subsection{Segmentation design}

\paragraph{Space-Time Query Decoder.}
Our Space-Time Query Decoder exploits the rich cross-view and temporal correspondences encoded by feed-forward visual geometry to recover persistent 4D instances. We formulate this process as end-to-end set prediction following Mask2Former~\citep{cheng2022masked}, avoiding the post-clustering identity association used by the IGGT series~\citep{li2025iggt,zou2026iggt4d}. A set of global learnable object queries $\{\mathbf{q}_j\}$ jointly attends
to the LoRA-adapted~\citep{hu2021lora} contextualized patch tokens
$\{\mathbf{z}_i^{F}\}_{i=1}^{T}$. The Space-Time Query Decoder $f_{\mathrm{ST}}$ predicts the
complete 4D mask of each query as
\begin{equation}
  \hat{\mathbf{M}}_j=f_{\mathrm{ST}}\!\left(\mathbf{q}_j,\{\mathbf{z}_i^{F}\}\right).
\end{equation}
During training, bipartite matching assigns each ground-truth 4D instance to a unique learnable
object query, with unmatched predictions assigned to the no-object class.

\paragraph{Prompt encoder.}
To support image-level prompting, the prompt encoder $f_{\mathrm{P}}$ combines point or box prompts
$p$ from one or more observations by sampling local visual features from
$\{\mathbf{z}_i^{F}\}_{i=1}^{T}$ to produce a promptable object query $\mathbf{q}^{p}$, which is
appended to the learnable object queries $\{\mathbf{q}_j\}$ at the decoder input:
\begin{equation}
  \mathbf{q}^{p}=f_{\mathrm{P}}\!\left(\{\mathbf{z}_i^{F}\},p\right), \qquad
  \hat{\mathbf{M}}^{p}=f_{\mathrm{ST}}\!\left(
  \mathbf{q}^{p},\{\mathbf{q}_j\},\{\mathbf{z}_i^{F}\}\right).
\end{equation}
Unidirectional self-attention lets $\mathbf{q}^{p}$ match the corresponding learnable object query
while preventing potentially ambiguous prompt signals from affecting $\{\mathbf{q}_j\}$, preserving
prompt-free instance discovery. For prompts from multiple observations, learned relevance weights
suppress occluded, imprecise, or inconsistent local evidence before forming $\mathbf{q}^{p}$.
Since prompts supplied by users or downstream localization tools may be spatially imprecise, we perturb
point locations and box coordinates during training to improve tolerance to localization errors.

\input{figures_final/prompt_st_decoder}

%% file: figures_final/prompt_st_decoder.tex
\begin{figure}[t]
  \centering
  \includegraphics[width=\textwidth]{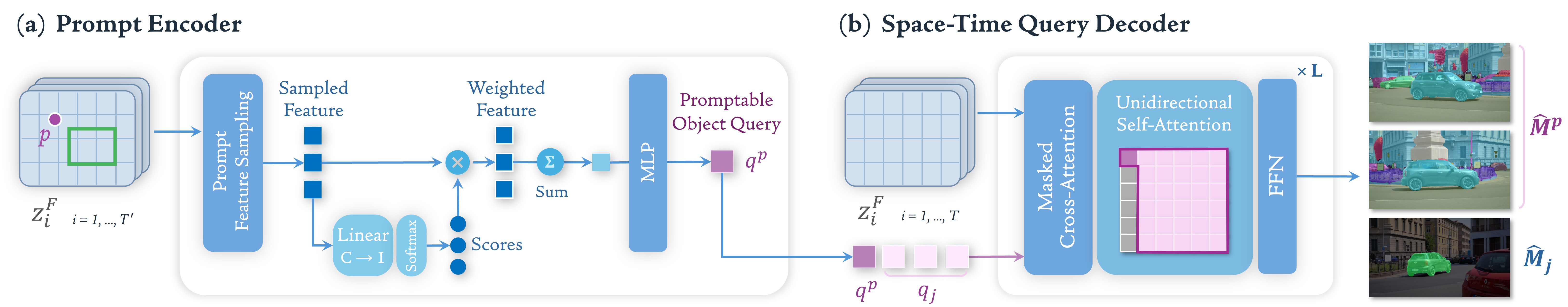}
  \caption{
  \emph{(a)}\textbf{ Prompt Encoder:} Point or box prompts sample dense features derived from the contextualized patch
  tokens and are aggregated into a promptable object query; $T'$ is the number of observations
  containing prompts. \emph{(b)} \textbf{Space-Time Query Decoder:} The promptable and learnable object queries decode persistent
  instance masks over the complete observation set. The attention mask blocks learnable
  queries from reading the prompt query while retaining the reverse direction.}
  \label{fig:supp-prompt-st-decoder}
\end{figure}

%% file: arxiv/sections/4_grounding_report.tex
\section{Ground Anything in 4D}
\label{sec:grounding}

Given a natural-language prompt $x$, 4D grounding aims to select the corresponding 4D mask
$\mathbf{M}^{*}$ of the referred object throughout the observed scene masklets.

\subsection{Grounding components}

\paragraph{Grounder.}

Vision-Language Models (VLMs) provide strong spatial priors for grounding referred objects in individual
images, yet their native RGB interface does not exploit the cross-view consistency encoded by
feed-forward visual geometry. The backbone register tokens form a compact, language-conditioned
scene representation~\citep{wang2026vggt}. Supplying all
patch tokens from multiple observations would introduce prohibitive context redundancy; we
therefore inject the per-frame camera and register tokens as compact geometric context. We further
supervise the grounder with a scene-geometry-guided chain of thought (CoT)~\citep{wei2022chain}
distilled from visual and metric scene cues, including object size, height, and inter-object distance,
assisting scene reasoning and query and box predictions.

Our grounder retains the input flexibility of the base segmenter, accommodating conventional
temporal sequences as well as sparse, non-consecutive observations. Each training window is formed from multiple observations sampled across $\mathcal{I}$ rather than restricted to consecutive frames. Let $N_{\mathrm{w}}$ denote the window size and $s_1,\ldots,s_{N_{\mathrm{w}}}\in\{1,\ldots,T\}$ the sampled observation indices. The
dual-stream prediction is summarized as
\begin{equation}
  \left\{\left(\hat{\mathbf{q}}_{s_k},\hat{p}_{s_k}\right)\right\}_{k=1}^{N_{\mathrm{w}}}
  = f_{\mathrm{G}}\!\left(
    x,
    \left\{I_{s_k},\mathbf{z}_{s_k}^{\mathrm{cam}},\mathbf{z}_{s_k}^{\mathrm{reg}}\right\}_{k=1}^{N_{\mathrm{w}}}
  \right),
  \label{eq:dual-stream-grounder}
\end{equation}
where $\hat{\mathbf{q}}_{s_k}$ denotes the query prediction and $\hat{p}_{s_k}$ the predicted box
prompt; a null query or absent prompt indicates that the referent is not present. In the
query-matching stream, the query predictions are matched to a persistent object query; in the box
stream, the predicted box prompts are jointly encoded by $f_{\mathrm{P}}$ into $\mathbf{q}^{p}$.
The two streams yield complementary 4D mask candidates through $f_{\mathrm{ST}}$.

\paragraph{Critic.}

Given the 4D mask candidates $\hat{\mathbf{M}}^{q}$ and $\hat{\mathbf{M}}^{p}$ produced by the
query-matching stream and box stream, respectively, we render each candidate over the observed
frames. An independent VLM-based critic $f_{\mathrm{C}}$ predicts their pairwise preference and
selects the candidate that better matches the expression $x$:
\begin{equation}
  s_{\mathrm{C}}=f_{\mathrm{C}}\!\left(x,\hat{\mathbf{M}}^{q},\hat{\mathbf{M}}^{p}\right),
  \qquad
  \hat{\mathbf{M}}^{*}=\begin{cases}
    \hat{\mathbf{M}}^{q}, & s_{\mathrm{C}}>0,\\
    \hat{\mathbf{M}}^{p}, & \text{otherwise},
  \end{cases}
  \label{eq:grounding-critic}
\end{equation}
where $s_{\mathrm{C}}$ is the pairwise preference logit, with a positive value favoring
$\hat{\mathbf{M}}^{q}$. The critic is fine-tuned on candidate pairs, with the candidate that better
agrees with the ground-truth mask providing the selection target.

\paragraph{Optimization.}

The grounder is optimized with cross-entropy for query matching and an autoregressive
language-modeling loss for its generated outputs, while the critic uses binary cross-entropy to
select the candidate with greater ground-truth overlap:
\begin{equation}
  \mathcal{L}_{\mathrm{G}}=\mathcal{L}_{\mathrm{query}}+\mathcal{L}_{\mathrm{LM}},
  \qquad
  \mathcal{L}_{\mathrm{C}}=\operatorname{BCE}(s_{\mathrm{C}},y_{\mathrm{C}}),
  \label{eq:grounding-objectives}
\end{equation}
where $y_{\mathrm{C}}=1$ when $\hat{\mathbf{M}}^{q}$ has greater ground-truth overlap and $0$
otherwise.

\subsection{Agentic grounding harness}

Grounding in a large 4D scene requires locating the observations that contain the referent and
associating them with a consistent instance. Processing large observation sets with
fixed sliding windows repeatedly evaluates irrelevant content. Agentic video methods mitigate this
cost by exploiting temporal continuity to retrieve relevant evidence along an ordered
timeline~\citep{wang2024videoagent,wang2025videotree}. However, our segmenter also accepts sparse or non-consecutive observations, for which temporal prior information is weakened. We therefore introduce Active Tree Search (ATS), driven by predicted target presence, to explore the observation set adaptively.

As illustrated in \cref{fig:ats}, ATS forms each grounder window by uniformly sampling
$N_{\mathrm{w}}$ observations from the current search range for joint evaluation. It begins with a
global window spanning the full observation set. At
each level, the grounder predicts target presence for every sampled observation; the neighboring
sample positions around each positive prediction delimit the subranges examined at the next level.
Each selected subrange is recursively resampled as a new grounder window, allowing the grounder to
determine the next search step from its current predictions. The recursion stops once each grounder window contains only adjacent frames. Predictions from all examined windows
are aggregated by voting to form two mask candidates from two streams respectively. Representative observations are then
selected, and the candidate masks are rendered as overlays and presented jointly with the expression
$x$ to the critic for comparative evaluation, determining the final mask $\hat{\mathbf{M}}^{*}$.

\begin{algorithm}[!tbp]
\caption{Active Tree Search}
\label{alg:ats}
\small
\textbf{Input:} observations $\mathcal{I}=\{I_i\}_{i=1}^{T}$, expression $x$, window size
$N_{\mathrm{w}}$, and maximum depth $D$\par
\textbf{Output:} grounded 4D mask $\hat{\mathbf{M}}^{*}$
\vspace{0.25em}

\begin{tabular}{@{}r@{\hspace{0.5em}}p{0.90\linewidth}@{}}
1: & $\mathcal{E}\leftarrow\textsc{Search}([1,T],0)$. \\
2: & $(\hat{\mathbf{M}}^{q},\hat{\mathbf{M}}^{p})\leftarrow$ aggregate and decode both
prediction streams in $\mathcal{E}$. \\
3: & $\hat{\mathbf{M}}^{*}\leftarrow f_{\mathrm{C}}
(x,\hat{\mathbf{M}}^{q},\hat{\mathbf{M}}^{p})$; \textbf{return}
$\hat{\mathbf{M}}^{*}$. \\
\multicolumn{2}{@{}l}{\textbf{function} \textsc{Search}($[a,b],d$)} \\
4: & \hspace{1em}Uniformly sample $N_{\mathrm{w}}$ indices $S$ from $[a,b]$; use every index if
$b-a+1\leq N_{\mathrm{w}}$. \\
5: & \hspace{1em}$\{(\hat{\mathbf{q}}_i,\hat{p}_i)\}_{i\in S}\leftarrow
f_{\mathrm{G}}(x,\{I_i\}_{i\in S})$; add the predictions to $\mathcal{E}$. \\
6: & \hspace{1em}\textbf{if} the indices in $S$ are adjacent, or $d=D$,
\textbf{then return} $\mathcal{E}$. \\
7: & \hspace{1em}$P\leftarrow\{i\in S\mid\hat{p}_i\neq\varnothing\}$;
$\mathcal{C}\leftarrow\textsc{Expand}([a,b],S,P,d)$. \\
8: & \hspace{1em}\textbf{for each} $C\in\mathcal{C}$, set
$\mathcal{E}\leftarrow\mathcal{E}\cup\textsc{Search}(C,d+1)$. \\
9: & \hspace{1em}\textbf{return} $\mathcal{E}$. \\
\multicolumn{2}{@{}l}{\textbf{function} \textsc{Expand}($[a,b],S,P,d$)} \\
10: & \hspace{1em}\textbf{if} $P=\varnothing$, \textbf{return}
$\textsc{UniformPartition}([a,b])$ when $d=0$, and $\varnothing$ otherwise. \\
11: & \hspace{1em}Group consecutive indices in $P$ and form one child range per group, bounded by
its nearest sampled neighbors in $S$. \\
12: & \hspace{1em}\textbf{return} the child ranges $\mathcal{C}$. \\
\end{tabular}
\end{algorithm}

\input{arxiv/figures/ats_report}

%% file: arxiv/figures/ats_report.tex
\begin{figure}[!tbp]
  \centering
  \includegraphics[width=\textwidth]{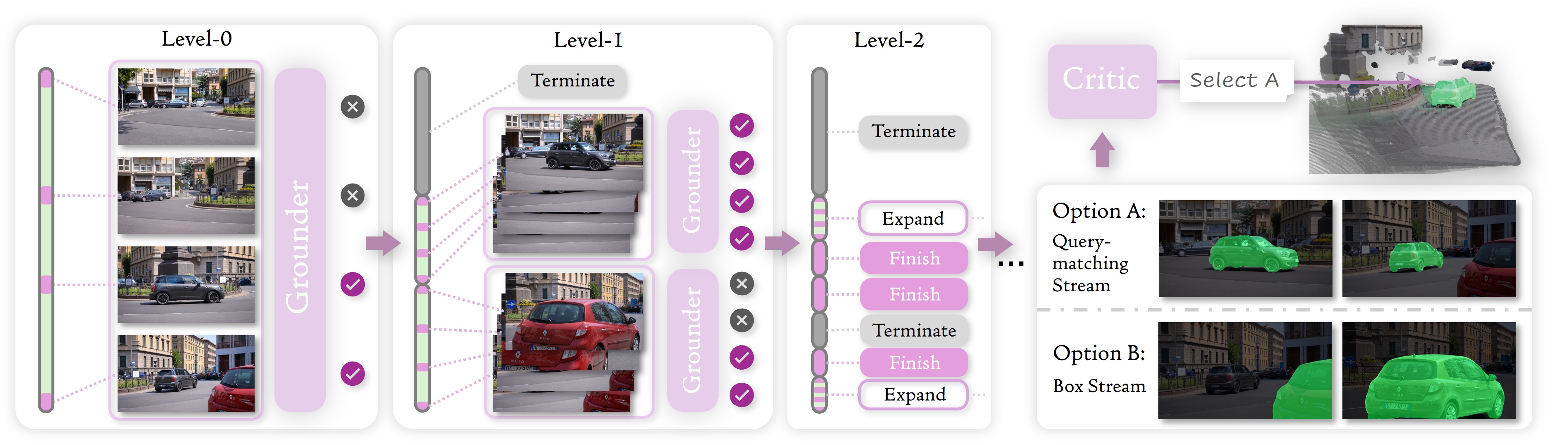}
  \vspace{-4mm}
  \caption{\textbf{Agentic grounding harness.} Active tree search locates relevant observations,
  the dual-stream grounder produces two 4D mask candidates, and the critic selects the final result.
  The illustrated example uses $N_w=4$ observations per grounder window.}
  \label{fig:ats}
  \vspace{-2mm}
\end{figure}

%% file: arxiv/sections/5_data_training_report.tex
\section{Data Curation and Training}
\label{sec:data-training}

\input{figures_final/data_curation}

Learning to segment any instance in 4D requires broad supervision across scene domains and motion
patterns. We consolidate 19 multi-view and video datasets into four groups: exhaustively annotated
multi-view scenes, exhaustively annotated videos, partially annotated video masklets, and
unlabelled multi-view observations. As shown in \cref{fig:data-curation}, the collection spans
static and dynamic scenes, real and synthetic imagery, and diverse indoor and outdoor content.
\Cref{tab:supp-training-data} lists the sampled portion of every source; together they contribute
6.05M observations from 33,220 scenes or clips.

\input{tables/tab_supp_training_data}

\paragraph{Self-distillation.}
Exhaustive set prediction treats every unmatched query as no-object, which would suppress
unlabelled instances if applied directly to partial annotations. We therefore train a teacher on
the exhaustive sources and use its temporally consistent masklets to densify partial annotations
and supervise unlabelled data. Human annotations take precedence, overlapping teacher predictions
are removed, and uncertain regions are ignored. This transfers exhaustive instance coverage to the
broader corpus without treating missing annotations as background.

\paragraph{Optimization.}
After Hungarian matching~\citep{cheng2022masked,carion2020end}, the segmenter is trained with object
classification, binary cross-entropy, and soft Dice losses~\citep{milletari2016v}:
\begin{equation}
  \mathcal{L}_{\mathrm{seg}}
  = \mathcal{L}_{\mathrm{obj}}
  + \mathcal{L}_{\mathrm{BCE}}
  + \mathcal{L}_{\mathrm{Dice}}.
  \label{eq:segmentation-objective}
\end{equation}
Training progressively shifts from the diverse pseudo-labelled pool toward exhaustive data to
reduce the effect of noisy labels near convergence. The Prompt Encoder can be trained jointly with
the segmenter or separately while the segmenter is frozen; all reported results use the latter.
Full data and training details appear in \cref{sec:supp-data-details,sec:supp-segmenter-training}.

%% file: figures_final/data_curation.tex
\begin{figure}[t]
  \centering
  \begin{minipage}[c]{0.54\linewidth}
    \centering
    \includegraphics[width=\linewidth]{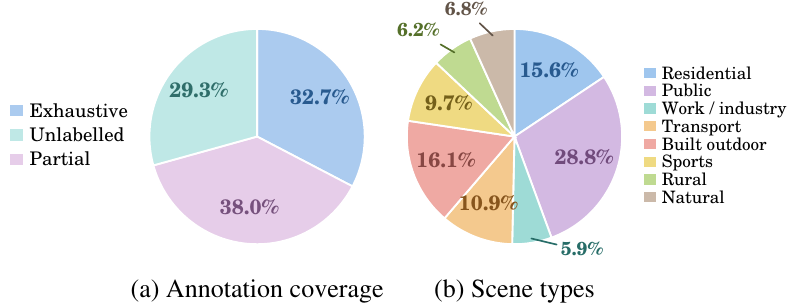}
  \end{minipage}\hfill%
  \begin{minipage}[c]{0.44\linewidth}
    \centering
    \setlength{\tabcolsep}{1.2pt}
    \renewcommand{\arraystretch}{1.02}
    \scalebox{0.75}{%
      \begin{tabular}{@{}lccccrr@{}}
        \toprule
        \multirow{2}{*}{Supervision} & \multicolumn{2}{c}{Type} & \multicolumn{2}{c}{Origin} & \multicolumn{2}{c}{Scale} \\
        \cmidrule(lr){2-3}\cmidrule(lr){4-5}\cmidrule(l){6-7}
        & Sta. & Dyn. & Real & Syn. & Scenes & Frames \\
        \midrule
        Exhaustive multi-view & \ding{51} & -- & \ding{51} & \ding{51} & 7.50k & 1.50M \\
        Exhaustive video & -- & \ding{51} & \ding{51} & -- & 3.35k & 83.1k \\
        Partial video & -- & \ding{51} & \ding{51} & \ding{51} & 12.63k & 958.6k \\
        Unlabelled multi-view & \ding{51} & -- & \ding{51} & -- & 9.74k & 3.51M \\
        \bottomrule
      \end{tabular}%
    }
    \vspace{0.15em}

    {\scriptsize (c) Data types and scale}
  \end{minipage}

  \vspace{-1mm}
  \caption{\textbf{Training corpus.} (a $\&$ b) Annotation coverage and dominant scene types among
  the curated data. (c) Data composition and scale by supervision type.}
  \label{fig:data-curation}
  \vspace{-4mm}
\end{figure}

%% file: tables/tab_supp_training_data.tex
\begin{table}[t]
  \centering
  \caption{\textbf{Training-data composition.} Counts report the sampled scenes or video clips and
  RGB observations used for training, rather than the full sizes of the source datasets.}
  \label{tab:supp-training-data}
  \setlength{\tabcolsep}{3.2pt}
  \renewcommand{\arraystretch}{1.04}
  \scalebox{0.75}{%
  \begin{tabular*}{1.3333\textwidth}{@{\extracolsep{\fill}}lccccrr@{}}
    \toprule
    Dataset & Supervision & Motion & Scene & Origin & Scenes/clips & Frames \\
    \midrule
    Infinigen~\citep{raistrick2023infinite} & Exhaustive & Static & Indoor & Synthetic & 1,466 & 146,034 \\
    ScanNet++~\citep{yeshwanth2023scannet++} & Exhaustive & Static & Indoor & Real & 907 & 760,226 \\
    RealEstate10K~\citep{zhou2018stereo} & Exhaustive & Static & Mixed & Real & 5,127 & 593,212 \\
    \midrule
    VIPSeg~\citep{miao2022large} & Exhaustive & Dynamic & Mixed & Real & 2,806 & 66,767 \\
    Cityscapes-VPS~\citep{kim2020video} & Exhaustive & Dynamic & Outdoor & Real & 394 & 1,970 \\
    KITTI-STEP~\citep{weber2021step} & Exhaustive & Dynamic & Outdoor & Real & 12 & 5,027 \\
    JRDB-PanoTrack~\citep{le2024jrdb} & Exhaustive & Dynamic & Mixed & Real & 135 & 9,365 \\
    \midrule
    SA-V~\citep{ravi2025sam} & Partial & Dynamic & Mixed & Real & 4,635 & 362,391 \\
    YouTube-VOS~\citep{xu2018youtube} & Partial & Dynamic & Mixed & Real & 3,471 & 94,588 \\
    MeViS~\citep{ding2023mevis} & Partial & Dynamic & Mixed & Real & 1,662 & 112,170 \\
    MOSE~\citep{ding2023mose} & Partial & Dynamic & Mixed & Real & 1,496 & 92,715 \\
    UVO-Dense~\citep{wang2021unidentified} & Partial & Dynamic & Mixed & Real & 759 & 68,310 \\
    DynamicReplica~\citep{karaev2023dynamicstereo} & Partial & Dynamic & Indoor & Synthetic & 236 & 70,800 \\
    SAIL-VOS~\citep{hu2019sail} & Partial & Dynamic & Mixed & Synthetic & 163 & 86,248 \\
    LVOS~\citep{hong2023lvos} & Partial & Dynamic & Mixed & Real & 120 & 65,237 \\
    DAVIS~\citep{perazzi2016benchmark} & Partial & Dynamic & Mixed & Real & 60 & 4,209 \\
    VOST~\citep{tokmakov2023breaking} & Partial & Dynamic & Mixed & Real & 32 & 1,920 \\
    \midrule
    DL3DV~\citep{ling2024dl3dv} & Pseudo & Static & Mixed & Real & 9,649 & 3,316,851 \\
    Matterport3D~\citep{chang2017matterport3d} & Pseudo & Static & Indoor & Real & 90 & 194,400 \\
    \midrule
    \textbf{Total} & & & & & \textbf{33,220} & \textbf{6,052,440} \\
    \bottomrule
  \end{tabular*}%
  }
\end{table}

%% file: arxiv/sections/6_experiments_report.tex
\section{Experiments}
\label{sec:exp}

\subsection{Experimental setup}
\label{sec:exp-setup}

\paragraph{Benchmarks and metrics.}
We evaluate prompt-free instance segmentation on the ScanNet validation setting
from~\citep{li2025iggt}, DAVIS~\citep{pont20172017}, LVOS~\citep{hong2023lvos}, and
VIPSeg~\citep{miao2022large}. ScanNet contains eight sampled observations per scene, following
IGGT, and LVOS is uniformly sampled to 100 observations per video. We report T-mIoU, T-SR, and
\JF{} on ScanNet; \fdiou{}, \fdsr{}, and \JF{} on DAVIS and LVOS; and
\stq{}~\citep{weber2021step} and mAP on VIPSeg. The grounding evaluation comprises 700 ScanRefer
queries from 121 validation scenes, 3,320 expressions from 80 held-out ScanNet++ scenes, all 122
Ref-DAVIS expressions, and 616 single-referent MeViS expressions from 47 videos. Static scenes use
3D mask mIoU and box/mask accuracy at IoU 0.25; videos use \JF{}.

For video benchmarks with 2D annotations, we lift both annotated and predicted masks through the
same estimated depth and cameras, then voxelize each observation in a distinct $(x,y,z,t)$ slice.
After one-to-one instance matching, let $\iota_j$ be the voxel IoU of the prediction matched to
ground-truth instance $j$, with $\iota_j=0$ when it is unmatched. The clip-level metrics are
\begin{equation}
  \operatorname{4D\text{-}IoU}=\frac{1}{N}\sum_{j=1}^{N}\iota_j,
  \qquad
  \operatorname{4D\text{-}SR}=\frac{1}{N}\sum_{j=1}^{N}\mathbb{1}[\iota_j>0.5].
  \label{eq:4d-metrics}
\end{equation}
Scores are averaged across clips. Complete benchmark construction and metric definitions appear in
\cref{sec:supp-evaluation-protocols}.

\paragraph{Baselines.}
For segmentation, we compare lifted 3D segmentation~\citep{yang2023sam3d}, image/video
systems~\citep{cheng2023tracking,wu2024general,li2024omg,ravi2025sam}, and feed-forward
visual-geometry methods~\citep{zust2025panst3r,li2025iggt,qu2026segvggt,zou2026iggt4d,
gong2026samv}. Official checkpoints receive the same observations without semantic labels, target
prompts, or first-frame ground truth. SAM~2 uses automatic mask discovery with periodic
re-discovery, while memory-intensive baselines retain all observations through full-sequence or
streaming inference. Grounding baselines cover VLM--segmenter and 3D-proposal
pipelines~\citep{xu2024vlm,drozdov2026z3d,ravi2025sam}, end-to-end image/video referring
models~\citep{carion2026sam,yuan2026sa2va}, and multi-view geometry--language
alignment~\citep{wu2026mvggt}. VLM$+$SAM~2 jointly processes eight uniformly sampled video
observations and predicts boxes or absence before bidirectional propagation; MVGGT receives the
same number of observations. Z3D uses reconstructed geometry except in the explicitly marked GT
oracle rows.

\paragraph{Implementation details.}
The segmenter initializes from the VGGT-$\Omega$-1B checkpoint at 512-pixel resolution. We use
rank-256 LoRA~\citep{hu2021lora} adapters and train the segmenter for 50,000 steps. The Grounder
initializes from Qwen3-VL-4B~\citep{bai2025qwen3}, uses eight-observation windows during training
and inference, and is trained on approximately 80k expressions. The Critic initializes from
Qwen3-VL-2B and is trained separately on paired mask candidates. Reported prompting results use the
separately trained Prompt Encoder. Additional training and baseline settings are given in
\cref{sec:supp-experimental-details}.

\subsection{Quantitative comparisons}

\paragraph{Class-agnostic segmentation.}
\Cref{tab:master} compares prompt-free instance segmentation across static multi-view scenes and
dynamic videos. \methodname{} leads every benchmark: its 0.803 T-mIoU and 0.792 T-SR on ScanNet
reflect accurate cross-view instance recovery, while 0.803 and 0.602 \JF{} on DAVIS and LVOS show
that persistent identities are maintained in dynamic scenes without temporal memory propagation.
The gains in STQ and mAP extend this result to dense panoptic evaluation.

\input{arxiv/tables/tab_master_seg}

\Cref{tab:supp-ego-driving} evaluates transfer to egocentric and driving perception. HOI4D and
VISOR are zero-shot evaluations, while KITTI-STEP uses a held-out driving split. \methodname{}
leads all reported metrics on the three datasets. T-SR remains lower than the other measures for
every method because small, similar targets, rapid motion, occlusion, and re-entry make persistent
full-clip recovery harder.

\input{arxiv/tables/tab_ego_driving_full}

The ten-clip ScanNet setting is complemented by the complete 312-scene ScanNet200 validation split
in \cref{tab:scannet200}. \methodname{} improves over IGGT4D by 2.9, 4.6, and 2.8 points in
T-mIoU, T-SR, and \JF{}, respectively. \Cref{tab:supp-robustness} further compares sparse
observations with shuffled versions of the same inputs. Video baselines that propagate information
along the sequence decline after shuffling, whereas feed-forward geometry methods do not exhibit a
consistent decline. On the appearance-diverse, shuffled LVOS setting, \methodname{} reaches 0.588
\JF{}, 12.6 points above the strongest video baseline.

\input{tables/tab_supp_scannet_robustness}

\Cref{tab:supp-prompted} supplies every method with the same sampled prompts. \methodname{} is
strongest with point prompts, whereas SAM~2 is generally stronger with clean multi-observation box
prompts. Under corrupted prompts from annotation or upstream localization errors, the Prompt
Encoder's relevance scoring suppresses inconsistent evidence and substantially improves robustness
over propagation-based prompting.

\input{tables/tab_supp_prompted}

\paragraph{Language-guided grounding.}
\Cref{tab:grounding} compares \methodname{} with VLM--segmenter pipelines, end-to-end referring
models, and Z3D's explicit 3D-proposal grounding under the unified protocol. \methodname{} performs
best across the four evaluations. Z3D approaches our static-scene results only with ground-truth
point clouds; replacing them with reconstructed point clouds causes a large decline, showing the
sensitivity of proposal-first 3D grounding to imperfect geometry.

\input{arxiv/tables/tab_master_ground}

\Cref{tab:supp-grounding-tools} compares grounding tools with the same Prompt Encoder and mask
decoder. General-purpose MLLMs and single-image grounding tools provide strong local priors, but
their coordinate predictions are not aligned with the persistent instance space. Our query stream
uses that representation directly and obtains the strongest result, while the box stream retains
the VLM's native spatial prediction interface.

\input{tables/tab_supp_grounding_tools}

\Cref{tab:supp-frame-grounding} evaluates frame-level target presence in scenes averaging 80.3
observations, with targets present in 28.6\% of them. ATS achieves the highest F1 and frame-set IoU;
its returned frame set covers substantially more target observations than conservative selection
while remaining much smaller than the full input.

\input{tables/tab_supp_frame_grounding}

\subsection{Ablation studies}

\input{arxiv/tables/tab_main_ablation}

\paragraph{Grounding architecture.}
\Cref{tab:main-ablation}(a) shows that register tokens provide the largest gain by exposing compact,
scene-wide geometric information to the Grounder. Multi-observation inference further improves
over single-view grounding by resolving references from cross-view context, and geometry-guided CoT
provides an additional gain. Across all four benchmarks in panels (c--d), the Critic consistently
improves over either stream alone, confirming that query matching and box prediction capture
complementary evidence.

\paragraph{Active Tree Search.}
Panel (b) relates target presence to search cost. ATS uses substantially fewer Grounder windows on
ScanRefer and ScanNet++, where the target appears in fewer than 30\% of observations. Its mean
window count also remains below sliding-window traversal on both video benchmarks, while preserving
or improving the reported stream scores in most settings.

\subsection{Qualitative comparisons}

\input{arxiv/figures/segmentation_qualitative_extended}

\Cref{fig:segmentation-qualitative} shows more complete, identity-consistent masks in static and
dynamic scenes. In panel (c), several basketball players have similar appearances and repeatedly
occlude one another during rapid motion. The purple player's identity remains stable through the
shot, while the ball is still tracked after possession passes to the red player. Panel (d) shows
the same persistence under the rapid motion and deformation of the kangaroo and cat. In contrast,
SAM~2 can miss newly visible objects or confuse identities across non-contiguous indoor views, and
existing feed-forward geometry baselines remain weaker on dynamic objects.

\input{arxiv/figures/grounding_qualitative_full}

\Cref{fig:grounding-qualitative} shows that the same persistent representation supports robust
language grounding, including expressions that require spatial reasoning rather than appearance
matching alone.

The panoramic and driving examples in \cref{fig:frame-4d-qualitative} show coherent mask identities
when visualized in reconstructed 4D space. This behavior transfers to unconstrained indoor,
outdoor, and driving sequences in \cref{fig:inthewild-qualitative}, despite large changes in
viewpoint, scale, motion, and scene content. These visualizations assess instance consistency in the
reconstructed scene rather than independent geometric accuracy.

\input{arxiv/figures/frame_4d_qualitative}
\input{arxiv/figures/inthewild_main}

\clearpage

%% file: arxiv/tables/tab_master_seg.tex
\begin{table}[t]
\centering
\caption{\textbf{Class-agnostic instance segmentation.} All methods are evaluated without target
prompts or first-frame ground truth. Higher is better for segmentation metrics; latency is measured
over 100 observations on an A100-80GB (s/clip).}
\vspace{-2mm}
\label{tab:master}
\setlength{\tabcolsep}{3.5pt}
\renewcommand{\arraystretch}{1.05}
\scalebox{0.75}{%
\begin{tabular*}{1.3333\linewidth}{@{\extracolsep{\fill}}lcccccccccccc@{}}
\toprule
\multirow{2}{*}{Method}
  & \multicolumn{3}{c}{ScanNet}
  & \multicolumn{3}{c}{DAVIS}
  & \multicolumn{3}{c}{LVOS}
  & \multicolumn{2}{c}{VIPSeg}
  & Latency \\
\cmidrule(lr){2-4}\cmidrule(lr){5-7}\cmidrule(lr){8-10}\cmidrule(lr){11-12}
  & \tmiou & \tsr & \JF
  & \fdiou & \fdsr & \JF
  & \fdiou & \fdsr & \JF
  & \stq & mAP & @100 \\
\midrule
SAM3D~\citep{yang2023sam3d}
  & 0.686 & 0.296 & 0.574 & -- & -- & -- & -- & -- & -- & -- & -- & -- \\
DEVA~\citep{cheng2023tracking}
  & 0.180 & 0.029 & 0.219 & 0.531 & 0.584 & 0.531 & 0.373 & 0.312 & 0.412 & 0.583 & 0.047 & 32.3 \\
GLEE~\citep{wu2024general}
  & 0.127 & 0.038 & 0.178 & 0.710 & 0.870 & 0.754 & 0.329 & 0.225 & 0.367 & 0.207 & 0.214 & 36.8 \\
OMG-Seg~\citep{li2024omg}
  & 0.189 & 0.067 & 0.218 & 0.632 & 0.789 & 0.691 & 0.167 & 0.102 & 0.191 & 0.554 & 0.205 & 13.7 \\
SAM~2~\citep{ravi2025sam}
  & 0.635 & 0.438 & 0.588 & 0.703 & 0.823 & 0.750 & 0.450 & 0.460 & 0.444 & 0.700 & 0.249 & 175 \\
PanSt3R~\citep{zust2025panst3r}
  & 0.747 & 0.708 & 0.719 & 0.484 & 0.506 & 0.513 & 0.297 & 0.120 & 0.315 & 0.698 & 0.197 & 54.0 \\
IGGT~\citep{li2025iggt}
  & 0.517 & 0.408 & 0.534 & 0.348 & 0.302 & 0.407 & 0.139 & 0.040 & 0.156 & 0.622 & 0.089 & 272 \\
SegVGGT~\citep{qu2026segvggt}
  & 0.561 & 0.471 & 0.494 & 0.178 & 0.022 & 0.259 & 0.069 & 0.000 & 0.069 & 0.340 & 0.069 & -- \\
IGGT4D~\citep{zou2026iggt4d}
  & 0.781 & 0.708 & 0.726 & 0.543 & 0.667 & 0.604 & 0.154 & 0.047 & 0.192 & 0.729 & 0.190 & 18.5 \\
SAM-V~\citep{gong2026samv}
  & 0.729 & 0.692 & 0.685 & 0.307 & 0.145 & 0.336 & 0.197 & 0.070 & 0.174 & 0.344 & 0.083 & -- \\
\midrule
\textbf{\methodname{} (Ours)}
  & \textbf{0.803} & \textbf{0.792} & \textbf{0.781}
  & \textbf{0.715} & \textbf{0.914} & \textbf{0.803}
  & \textbf{0.533} & \textbf{0.573} & \textbf{0.602}
  & \textbf{0.795} & \textbf{0.352} & \textbf{10.1} \\
\bottomrule
\end{tabular*}%
}
\label{tab:iggtbench}
\label{tab:vos}
\label{tab:vipseg}
\vspace{-3mm}
\end{table}

%% file: arxiv/tables/tab_ego_driving_full.tex
\begin{table}[htbp]
\centering
\caption{\textbf{Egocentric and driving-scene segmentation.} Left: results on HOI4D and
VISOR. Right: results on the KITTI-STEP evaluation split. IGGT4D is omitted on HOI4D because its
training mixture contains part of the evaluation split.}
\label{tab:supp-ego-driving}
\begin{minipage}[b]{0.64\linewidth}
\vspace{0pt}
\centering
\renewcommand{\arraystretch}{1.05}
\scalebox{0.75}{%
\begin{tabular*}{1.3333\linewidth}{@{\extracolsep{\fill}}lcccccc@{}}
\toprule
\multirow{2}{*}{Method} & \multicolumn{3}{c}{HOI4D} & \multicolumn{3}{c}{VISOR} \\
\cmidrule(lr){2-4}\cmidrule(lr){5-7}
& \JF & \tmiou & \tsr & \JF & \tmiou & \tsr \\
\midrule
DEVA~\citep{cheng2023tracking}       & 0.594 & 0.557 & 0.250 & 0.213 & 0.111 & 0.018 \\
GLEE~\citep{wu2024general}           & 0.427 & 0.404 & 0.261 & -- & -- & -- \\
OMG-Seg~\citep{li2024omg}            & 0.272 & 0.248 & 0.192 & 0.020 & 0.016 & 0.000 \\
SAM~2~\citep{ravi2025sam}            & 0.572 & 0.542 & 0.219 & 0.295 & 0.176 & 0.060 \\
PanSt3R~\citep{zust2025panst3r}      & 0.523 & 0.524 & 0.352 & 0.195 & 0.142 & 0.044 \\
IGGT~\citep{li2025iggt}              & 0.378 & 0.374 & 0.069 & 0.148 & 0.099 & 0.027 \\
SegVGGT~\citep{qu2026segvggt}        & 0.266 & 0.280 & 0.131 & 0.097 & 0.078 & 0.037 \\
IGGT4D~\citep{zou2026iggt4d}         & -- & -- & -- & 0.261 & 0.179 & 0.068 \\
SAM-V~\citep{gong2026samv}           & 0.465 & 0.462 & 0.122 & 0.199 & 0.157 & 0.035 \\
\midrule
\textbf{\methodname{} (Ours)}        & \textbf{0.624} & \textbf{0.602} & \textbf{0.516}
                                      & \textbf{0.313} & \textbf{0.204} & \textbf{0.118} \\
\bottomrule
\end{tabular*}%
}
\end{minipage}\hfill
\begin{minipage}[b]{0.34\linewidth}
\vspace{0pt}
\centering
\renewcommand{\arraystretch}{1.05}
\scalebox{0.75}{%
\begin{tabular*}{1.3333\linewidth}{@{\extracolsep{\fill}}lccc@{}}
\toprule
\multirow{2}{*}{Method} & \multicolumn{3}{c}{KITTI-STEP} \\
\cmidrule(lr){2-4}
& \JF & \tmiou & \tsr \\
\midrule
DEVA       & 0.181 & 0.171 & 0.037 \\
OMG-Seg    & 0.164 & 0.161 & 0.088 \\
SAM~2      & 0.274 & 0.266 & 0.075 \\
IGGT       & 0.113 & 0.110 & 0.017 \\
SegVGGT    & 0.065 & 0.071 & 0.002 \\
IGGT4D     & 0.268 & 0.250 & 0.105 \\
SAM-V      & 0.203 & 0.196 & 0.021 \\
\midrule
\textbf{\methodname{} (Ours)}        & \textbf{0.501} & \textbf{0.519} & \textbf{0.191} \\
\bottomrule
\end{tabular*}%
}
\end{minipage}
\end{table}

%% file: tables/tab_supp_scannet_robustness.tex
\begin{table}[htbp]
\centering
\input{tables/tab_scannet200}\hfill
\input{tables/tab_supp_robustness}
\end{table}

%% file: tables/tab_scannet200.tex
\begin{minipage}[b]{0.49\linewidth}
\vspace{0pt}
\centering
\caption{\textbf{Complete ScanNet200 validation split.} RGB-only evaluation.}
\label{tab:scannet200}
\setlength{\tabcolsep}{2.2pt}
\renewcommand{\arraystretch}{1.05}
\scalebox{0.75}{%
\begin{tabular*}{1.32\linewidth}{@{\extracolsep{\fill}}lccc@{}}
\toprule
Method & \tmiou & \tsr & \JF \\
\midrule
VGGT-$\Omega$ + Mask3D~\citep{schult2023mask3d}
                                                & 0.053 & 0.017 & 0.050 \\
SAM3D~\citep{yang2023sam3d}                    & 0.463 & 0.207 & 0.338 \\
DEVA~\citep{cheng2023tracking}                 & 0.088 & 0.029 & 0.086 \\
GLEE~\citep{wu2024general}                     & 0.042 & 0.024 & 0.053 \\
OMG-Seg~\citep{li2024omg}                      & 0.101 & 0.050 & 0.101 \\
SAM~2~\citep{ravi2025sam}                      & 0.327 & 0.153 & 0.288 \\
PanSt3R~\citep{zust2025panst3r}                & 0.461 & 0.318 & 0.403 \\
IGGT~\citep{li2025iggt}                         & 0.339 & 0.210 & 0.298 \\
SegVGGT~\citep{qu2026segvggt}                  & 0.425 & 0.319 & 0.362 \\
IGGT4D~\citep{zou2026iggt4d}                   & 0.488 & 0.330 & 0.426 \\
SAM-V~\citep{gong2026samv}                    & 0.376 & 0.205 & 0.333 \\
\midrule
\textbf{\methodname{}}                          & \textbf{0.517} & \textbf{0.376} & \textbf{0.454} \\
\bottomrule
\end{tabular*}%
}
\end{minipage}

%% file: tables/tab_supp_robustness.tex
\begin{minipage}[b]{0.49\linewidth}
\vspace{0pt}
\centering
\caption{\textbf{Robustness to sparse and unordered inputs.} All entries are class-agnostic
\JF{}.}
\label{tab:supp-robustness}
\setlength{\tabcolsep}{5.5pt}
\renewcommand{\arraystretch}{1.05}
\scalebox{0.75}{%
\begin{tabular*}{1.32\linewidth}{@{\extracolsep{\fill}}lccccc@{}}
\toprule
\multirow{2}{*}{Method} & \multicolumn{2}{c}{DAVIS} & \multicolumn{3}{c}{LVOS} \\
\cmidrule(lr){2-3}\cmidrule(lr){4-6}
& Sparse & Shuffle & Sparse & Shuffle & Hard \\
\midrule
DEVA       & 0.532 & 0.526 & 0.433 & 0.416 & 0.307 \\
GLEE       & 0.749 & 0.743 & 0.560 & 0.547 & 0.462 \\
OMG-Seg    & 0.693 & 0.665 & 0.452 & 0.435 & 0.311 \\
SAM~2      & 0.766 & 0.749 & 0.579 & 0.559 & 0.418 \\
PanSt3R    & 0.579 & 0.581 & 0.447 & 0.464 & 0.359 \\
IGGT       & 0.448 & 0.459 & 0.269 & 0.273 & 0.171 \\
SegVGGT    & 0.328 & 0.332 & 0.111 & 0.106 & 0.078 \\
IGGT4D     & 0.613 & 0.600 & 0.268 & 0.268 & 0.248 \\
\midrule
\textbf{\methodname{} (Ours)}        & \textbf{0.807} & \textbf{0.800} & \textbf{0.645} & \textbf{0.656} & \textbf{0.588} \\
\bottomrule
\end{tabular*}%
}
\end{minipage}

%% file: tables/tab_supp_prompted.tex
\begin{table}[htbp]
\centering
\caption{\textbf{Point- and box-prompted segmentation} on ScanNet and DAVIS. $K$ is the maximum
number of evenly spaced target-present observations used for prompting; corruption replaces the
specified fraction with target-irrelevant prompts. All entries report \JF{} for \methodname{} and
SAM~2~\citep{ravi2025sam}; SAM-V~\citep{gong2026samv} supports point prompts only. For $K>1$,
SAM-V independently decodes each prompted observation and retains the output with the highest
predicted IoU. $K=8$ covers all visible ScanNet views.}
\label{tab:supp-prompted}
\setlength{\tabcolsep}{4.5pt}
\renewcommand{\arraystretch}{1.05}
\scalebox{0.75}{%
\begin{tabular*}{1.32\linewidth}{@{\extracolsep{\fill}}cc*{10}{c}@{}}
\toprule
\multirow{3}{*}{$K$} & \multirow{3}{*}{Corruption}
& \multicolumn{5}{c}{ScanNet} & \multicolumn{5}{c}{DAVIS} \\
\cmidrule(lr){3-7}\cmidrule(lr){8-12}
& & \multicolumn{2}{c}{Box} & \multicolumn{3}{c}{Point}
    & \multicolumn{2}{c}{Box} & \multicolumn{3}{c}{Point} \\
\cmidrule(lr){3-4}\cmidrule(lr){5-7}\cmidrule(lr){8-9}\cmidrule(lr){10-12}
& & \methodname{} & SAM~2 & \methodname{} & SAM~2 & SAM-V
    & \methodname{} & SAM~2 & \methodname{} & SAM~2 & SAM-V \\
\midrule
1  & 0\%  & 0.806 & 0.753 & 0.793 & 0.649 & 0.567 & 0.790 & 0.809 & 0.793 & 0.771 & 0.134 \\
2  & 0\%  & 0.804 & 0.844 & 0.825 & 0.713 & 0.598 & 0.789 & 0.818 & 0.802 & 0.764 & 0.124 \\
4  & 0\%  & 0.800 & 0.877 & 0.825 & 0.760 & 0.589 & 0.779 & 0.827 & 0.805 & 0.760 & 0.145 \\
8  & 0\%  & 0.801 & 0.878 & 0.825 & 0.769 & 0.589 & 0.807 & 0.829 & 0.810 & 0.780 & 0.129 \\
16 & 0\%  & --    & --    & --    & --    & --    & 0.796 & 0.828 & 0.805 & 0.775 & 0.127 \\
\midrule
8  & 25\% & 0.768 & 0.401 & 0.662 & 0.346 & 0.488 & 0.733 & 0.733 & 0.800 & 0.659 & 0.091 \\
8  & 50\% & 0.665 & 0.253 & 0.507 & 0.256 & --    & 0.651 & 0.532 & 0.761 & 0.445 & -- \\
16 & 25\% & --    & --    & --    & --    & --    & 0.727 & 0.722 & 0.808 & 0.668 & 0.082 \\
\bottomrule
\end{tabular*}%
}
\end{table}

%% file: arxiv/tables/tab_master_ground.tex
\begin{table}[t]
\centering
\caption{\textbf{Language-guided grounding.} Static benchmarks report 3D mIoU and
box/mask accuracy at IoU 0.25; dynamic benchmarks report J\&F. Z3D is evaluated with either
benchmark GT point clouds or reconstructed point clouds.}
\vspace{-2mm}
\label{tab:grounding}
\setlength{\tabcolsep}{1.5pt}
\renewcommand{\arraystretch}{1.05}
\scalebox{0.75}{%
\begin{tabular*}{1.3333\linewidth}{@{\extracolsep{\fill}}lccccccccc@{}}
\toprule
\multirow{2}{*}{Method}
  & \multirow{2}{*}{VLM Size}
  & \multicolumn{3}{c}{ScanRefer}
  & \multicolumn{3}{c}{ScanNet++}
  & \multicolumn{1}{c}{Ref-DAVIS}
  & \multicolumn{1}{c}{MeViS} \\
\cmidrule(lr){3-5}\cmidrule(lr){6-8}\cmidrule(lr){9-9}\cmidrule(lr){10-10}
  & & 3D mIoU & Box@.25 & Mask@.25
  & 3D mIoU & Box@.25 & Mask@.25
  & J\&F
  & J\&F \\
\midrule
VLM-Grounder~\citep{xu2024vlm}
  & 30.5B & 0.105 & 26.0 & 17.0 & 0.063 & 12.4 & 9.9 & 0.183 & 0.141 \\
VLM$+$SAM~2~\citep{ravi2025sam}
  & 30.5B & 0.308 & 34.9 & 42.6 & 0.170 & 15.1 & 26.2 & 0.709 & 0.483 \\
Sa2VA-8B~\citep{yuan2026sa2va}
  & 8B & 0.124 & 14.9 & 17.9 & 0.040 & 3.4 & 6.4 & 0.732 & 0.542 \\
SAM~3~\citep{carion2026sam}
  & N/A & 0.247 & 29.7 & 33.3 & 0.172 & 19.5 & 26.3 & 0.567 & 0.369 \\
MVGGT~\citep{wu2026mvggt}
  & N/A & 0.325 & 35.3 & 50.1 & 0.090 & 7.8 & 14.8 & 0.005 & 0.003 \\
Z3D (GT)~\citep{drozdov2026z3d}
  & 30.5B & 0.431 & 52.4 & 58.1 & 0.266 & \textbf{33.7} & 40.2 & -- & -- \\
Z3D (recon.)~\citep{drozdov2026z3d}
  & 30.5B & 0.114 & 26.1 & 16.1 & 0.157 & 31.1 & 28.6 & 0.147 & 0.180 \\
\midrule
\textbf{\methodname{} (Ours)}
  & 4B$+$2B
  & \textbf{0.461} & \textbf{56.2} & \textbf{59.4}
  & \textbf{0.293} & 30.4 & \textbf{44.9}
  & \textbf{0.755} & \textbf{0.573} \\
\bottomrule
\end{tabular*}%
}
\vspace{-1mm}
\end{table}

%% file: tables/tab_supp_grounding_tools.tex
\begin{table}[htbp]
\centering
\caption{\textbf{Grounding-tool comparison} on $160$ queries randomly sampled from ScanRefer.
Coordinates from each tool are decoded by the same promptable segmenter. Both configurations predict boxes. Model size refers to the grounding
model only.}
\label{tab:supp-grounding-tools}
\setlength{\tabcolsep}{6pt}
\renewcommand{\arraystretch}{1.04}
\scalebox{0.75}{%
\begin{tabular*}{1.32\linewidth}{@{\extracolsep{\fill}}lllccc@{}}
\toprule
Method & Model Size & Configuration & 3D mask mIoU & Mask@0.5 & Box@0.5 \\
\midrule
Molmo 2~\citep{clark2026molmo2}       & 4B & Multi-frame point tracking & 0.432 & 45.4 & 46.1 \\
Molmo 2~\citep{clark2026molmo2}       & 4B & Multi-frame point localization & 0.407 & 43.4 & 42.1 \\
Qwen3-VL~\citep{bai2025qwen3}         & 4B & Single-view box & 0.316 & 32.2 & 32.9 \\
Qwen3-VL~\citep{bai2025qwen3}         & 4B & Multi-view box & 0.202 & 21.1 & 20.4 \\
GLM-4.1V~\citep{hong2025glm}              & 9B & Per-view box & 0.303 & 32.9 & 29.6 \\
LocateAnything~\citep{wang2026locateanything} & 3B & Per-view box & 0.287 & 31.6 & 31.6 \\
GPT-5.6~\citep{openai2026gpt56}       & N/A & Per-view box & 0.421 & 47.4 & 42.8 \\
\midrule
\textbf{Ours}                            & 4B & Box stream & 0.424 & 45.3 & 42.0 \\
\textbf{Ours}                            & 4B & Query stream & \textbf{0.468} & \textbf{50.0} & \textbf{50.7} \\
\bottomrule
\end{tabular*}%
}
\end{table}

%% file: tables/tab_supp_frame_grounding.tex
\begin{table}[htbp]
\centering
\caption{\textbf{Agentic frame localization} on ScanRefer. Model size denotes the VLM used for
frame selection; selected frames reports the returned fraction rather than a ranked metric.}
\label{tab:supp-frame-grounding}
\setlength{\tabcolsep}{2.5pt}
\renewcommand{\arraystretch}{1.04}
\scalebox{0.75}{%
\begin{tabular*}{1.32\linewidth}{@{\extracolsep{\fill}}llrrrrr@{}}
\toprule
Method & Model Size & Precision & Recall & F1 & IoU & Selected frames (\%) \\
\midrule
VideoAgent~\citep{wang2024videoagent} & 235B & 0.305 & 0.108 & 0.144 & 0.082 & 9.5\% \\
VideoTree~\citep{wang2025videotree}   & 235B & 0.283 & 0.562 & 0.341 & 0.221 & 56.6\% \\
Vidi 1.5~\citep{team2025vidi}          & 9B & 0.626 & 0.144 & 0.192 & 0.121 & 7.0\% \\
LensWalk~\citep{li2026lenswalk}        & 235B & 0.343 & 0.181 & 0.192 & 0.128 & 12.6\% \\
\midrule
Uniform top-8                          & N/A & 0.291 & 0.144 & 0.173 & 0.099 & 13.8\% \\
Uniform top-16                         & N/A & 0.285 & 0.274 & 0.250 & 0.151 & 27.4\% \\
Select all                             & N/A & 0.286 & 1.000 & 0.416 & 0.286 & 100.0\% \\
\midrule
\textbf{Ours}                          & 4B & 0.708 & 0.535 & \textbf{0.573} & \textbf{0.455} & 21.6\% \\
\bottomrule
\end{tabular*}%
}
\end{table}

%% file: arxiv/tables/tab_main_ablation.tex
\begin{table}[t]
\centering
\caption{\textbf{Ablation studies.} (a) Grounder inputs on ScanRefer. (b) Sliding-window and ATS
traversal; Presence is the fraction of observations containing the target, and $\Delta Q$ and
$\Delta B$ are the changes in query- and box-stream scores when using ATS. (c--d) Dual-stream and
Critic decomposition on static and dynamic grounding benchmarks.}
\label{tab:main-ablation}
\renewcommand{\arraystretch}{1.05}
\begin{minipage}[b]{0.42\linewidth}
\centering
\scalebox{0.75}{%
\begin{tabular*}{1.32\linewidth}{@{\extracolsep{\fill}}lrrr@{}}
\toprule
\multicolumn{4}{c}{(a) Grounder inputs} \\
\cmidrule(lr){1-4}
Input & 3D mIoU & Mask@.5 & Box@.5 \\
\midrule
Single w/o $\mathbf{z}^{\mathrm{reg}}$ & 0.206 & 22.9 & 20.7 \\
Single                     & 0.370 & 42.0 & 37.7 \\
Multi                      & 0.398 & 45.2 & 40.3 \\
Multi $+$ CoT              & \textbf{0.407} & \textbf{46.0} & \textbf{41.1} \\
\bottomrule
\end{tabular*}%
}
\end{minipage}\hfill
\begin{minipage}[b]{0.56\linewidth}
\centering
\scalebox{0.75}{%
\begin{tabular*}{1.32\linewidth}{@{\extracolsep{\fill}}lrrrrr@{}}
\toprule
\multicolumn{6}{c}{(b) Active Tree Search} \\
\cmidrule(lr){1-6}
Benchmark & Presence & Sliding & ATS & $\Delta Q$ & $\Delta B$ \\
\midrule
ScanNet++ & 19.1 & 23.0 & 8.2  & $-1.9$ & $-1.7$ \\
ScanRefer & 27.9 & 28.5 & 9.7  & $-1.7$ & $+2.7$ \\
MeViS     & 88.1 & 15.5 & 11.4 & $+0.7$ & $+0.1$ \\
Ref-DAVIS & 93.7 & 17.6 & 15.8 & $+3.7$ & $+0.3$ \\
\bottomrule
\end{tabular*}%
}
\end{minipage}

\vspace{1.6mm}
\begin{minipage}[b]{0.49\linewidth}
\centering
\scalebox{0.75}{%
\begin{tabular*}{1.32\linewidth}{@{\extracolsep{\fill}}llrrr@{}}
\toprule
\multicolumn{5}{c}{(c) Static 3D grounding} \\
\cmidrule(lr){1-5}
\multirow{2}{*}{Benchmark} & \multirow{2}{*}{Output} & \multicolumn{3}{c}{3D Mask} \\
\cmidrule(lr){3-5}
& & mIoU & @.25 & @.5 \\
\midrule
\multirow{3}{*}{ScanRefer}
  & Query     & 0.406 & 51.8 & 45.0 \\
  & Box       & 0.434 & 55.0 & 47.6 \\
  & $+$Critic & \textbf{0.461} & \textbf{59.4} & \textbf{50.5} \\
\midrule
\multirow{3}{*}{ScanNet++}
  & Query     & 0.268 & 41.7 & 30.9 \\
  & Box       & 0.271 & 41.0 & 29.8 \\
  & $+$Critic & \textbf{0.293} & \textbf{44.9} & \textbf{33.1} \\
\bottomrule
\end{tabular*}%
}
\end{minipage}\hfill
\begin{minipage}[b]{0.49\linewidth}
\centering
\scalebox{0.75}{%
\begin{tabular*}{1.32\linewidth}{@{\extracolsep{\fill}}llrrr@{}}
\toprule
\multicolumn{5}{c}{(d) Dynamic 4D grounding} \\
\cmidrule(lr){1-5}
\multirow{2}{*}{Benchmark} & \multirow{2}{*}{Output} & \multicolumn{3}{c}{Video Mask} \\
\cmidrule(lr){3-5}
& & $\mathcal{J}$ & $\mathcal{F}$ & \JF{} \\
\midrule
\multirow{3}{*}{Ref-DAVIS}
  & Query     & 0.626 & 0.730 & 0.678 \\
  & Box       & 0.695 & 0.793 & 0.744 \\
  & $+$Critic & \textbf{0.706} & \textbf{0.805} & \textbf{0.755} \\
\midrule
\multirow{3}{*}{MeViS}
  & Query     & 0.470 & 0.579 & 0.524 \\
  & Box       & 0.494 & 0.591 & 0.543 \\
  & $+$Critic & \textbf{0.522} & \textbf{0.623} & \textbf{0.573} \\
\bottomrule
\end{tabular*}%
}
\end{minipage}
\end{table}

%% file: arxiv/figures/segmentation_qualitative_extended.tex
\begin{figure}[!tbp]
  \centering
  \setlength{\tabcolsep}{0pt}
  \renewcommand{\arraystretch}{1.02}
  \newcommand{\mainqualrowlabel}[1]{%
    \raisebox{\dimexpr4.75mm-.5\height+.5\depth\relax}{%
      \rotatebox[origin=c]{90}{\fontsize{5.6}{6.0}\selectfont #1}}}
  \newcommand{\mainqualimage}[1]{\includegraphics[width=0.445\textwidth]{#1}}
  \begin{tabular}{r@{\hspace{1.0mm}}c@{\hspace{2.2mm}}c}
    & {\fontsize{6.5}{7}\selectfont\textbf{(a) ScanNet}} &
      {\fontsize{6.5}{7}\selectfont\textbf{(b) DAVIS}} \\
    \mainqualrowlabel{RGB} &
      \mainqualimage{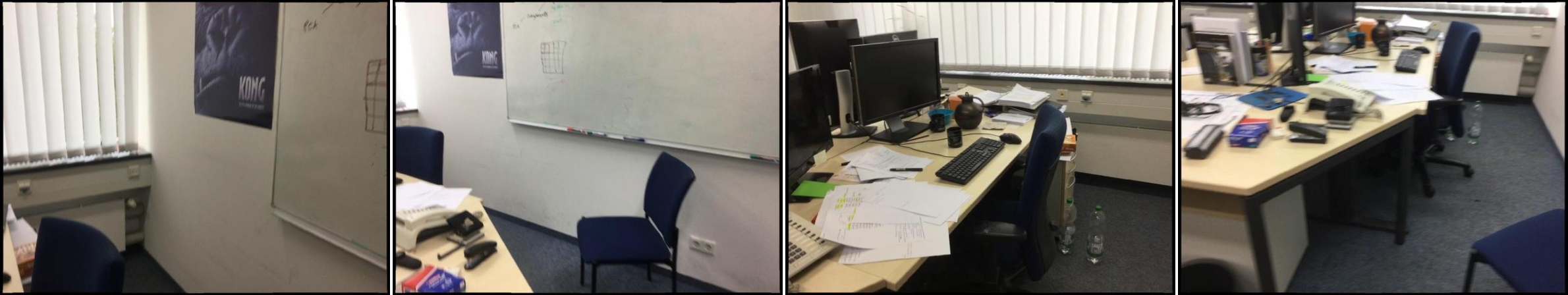} &
      \mainqualimage{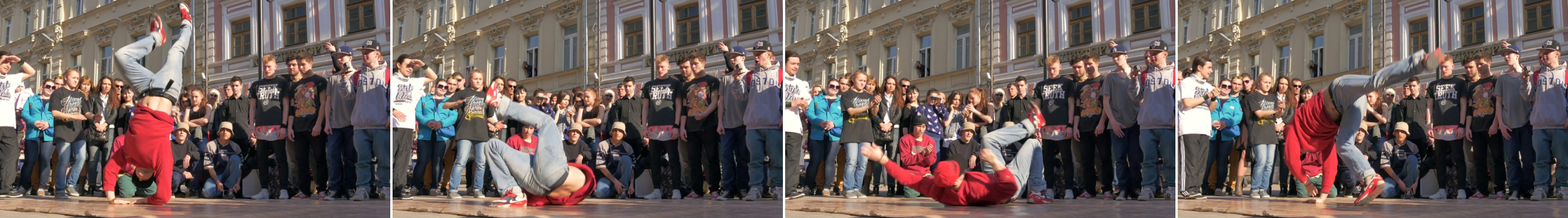} \\
    \mainqualrowlabel{SAM 2} &
      \mainqualimage{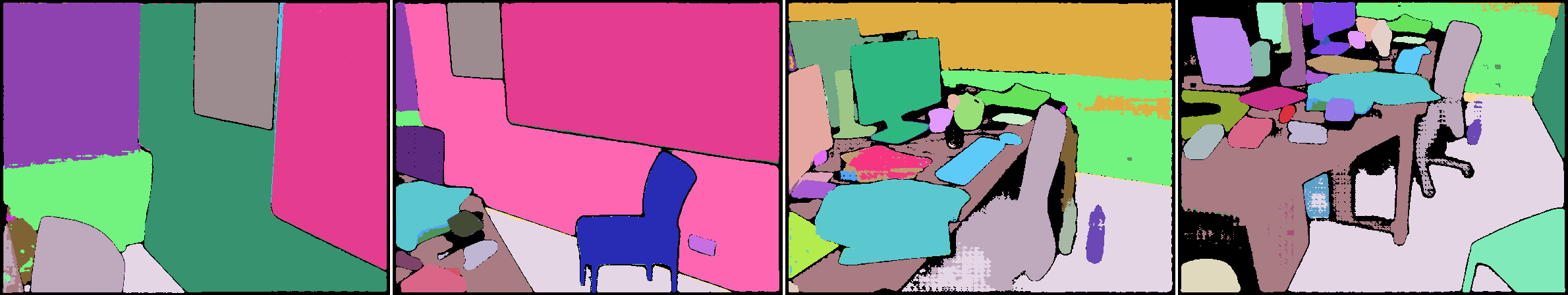} &
      \mainqualimage{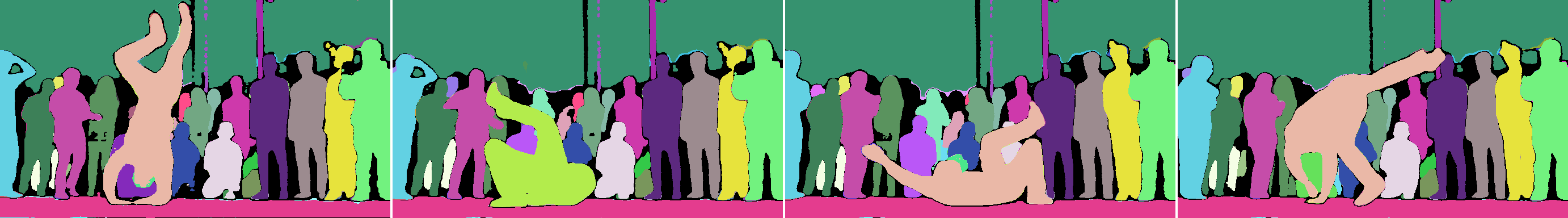} \\
    \mainqualrowlabel{IGGT} &
      \mainqualimage{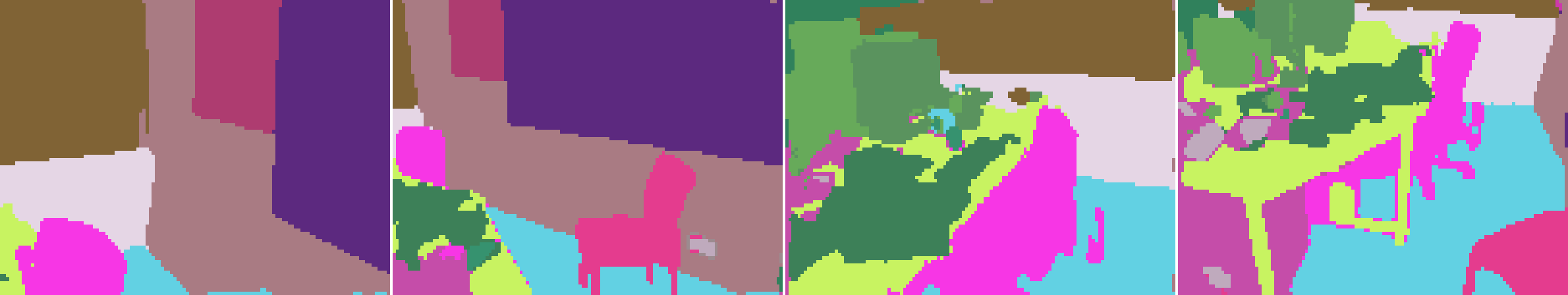} &
      \mainqualimage{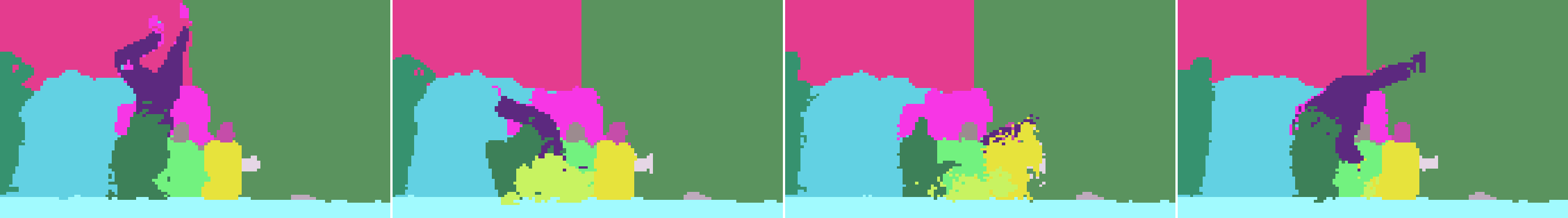} \\
    \mainqualrowlabel{IGGT4D} &
      \mainqualimage{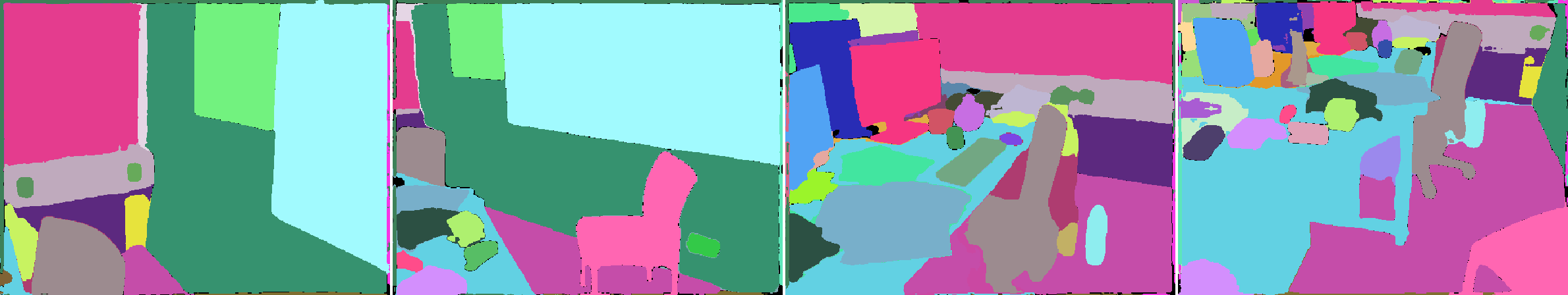} &
      \mainqualimage{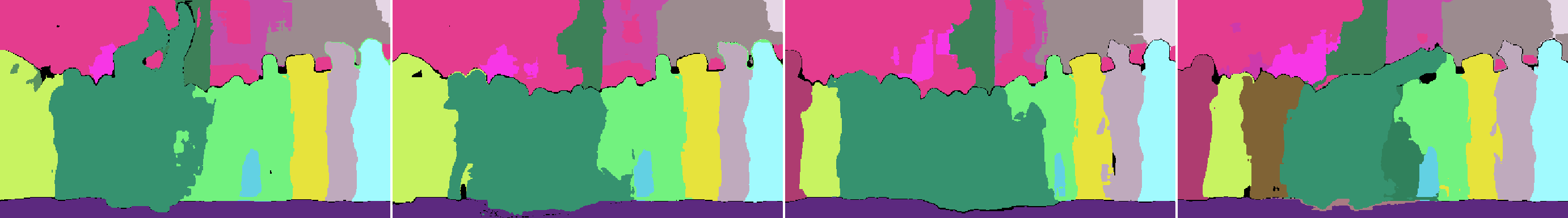} \\
    \mainqualrowlabel{\textbf{Ours}} &
      \mainqualimage{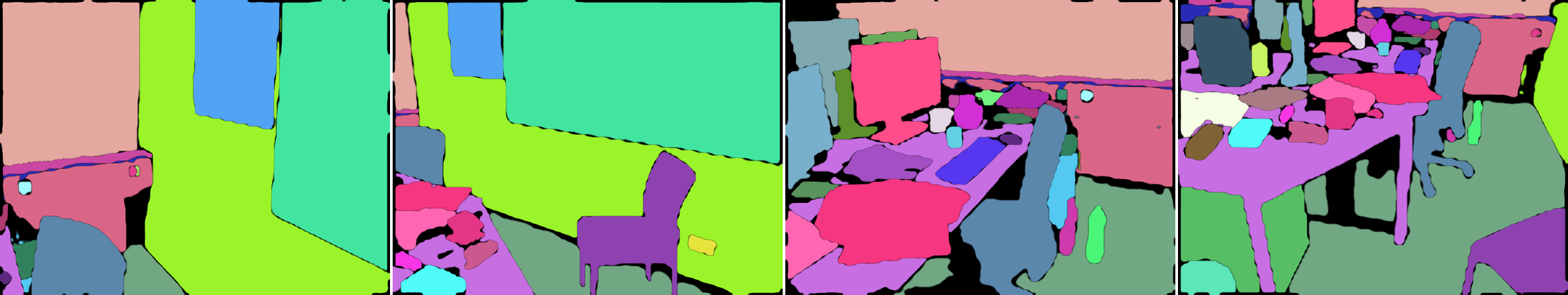} &
      \mainqualimage{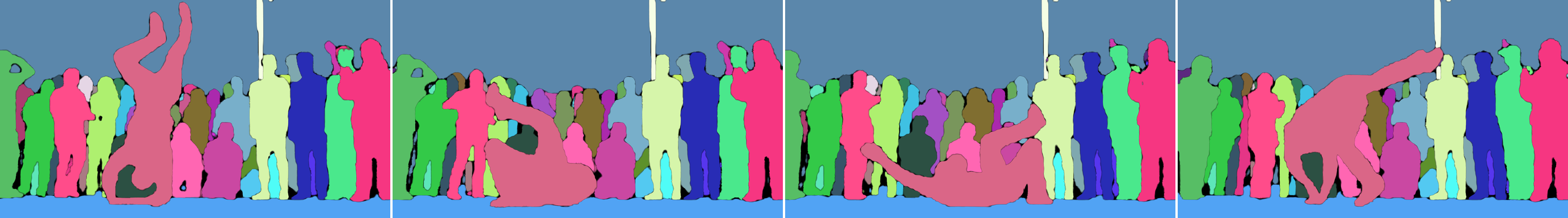} \\
    \mainqualrowlabel{GT} &
      \mainqualimage{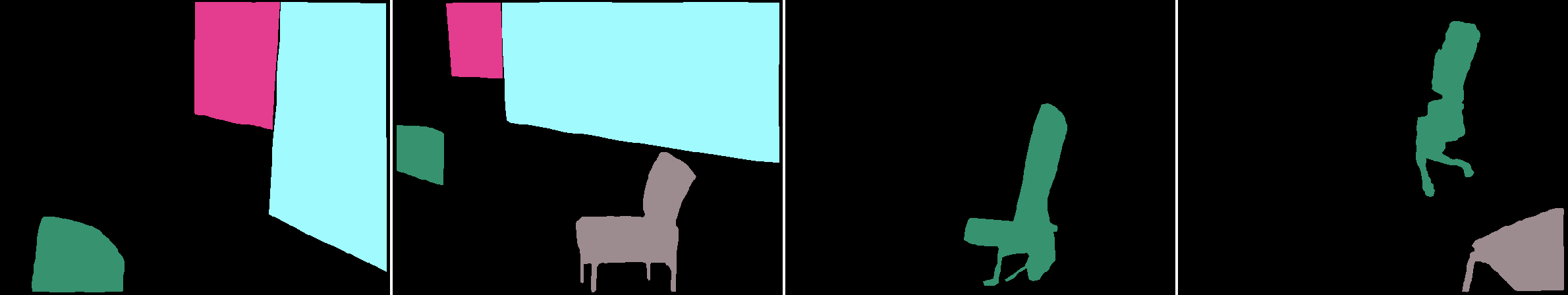} &
      \mainqualimage{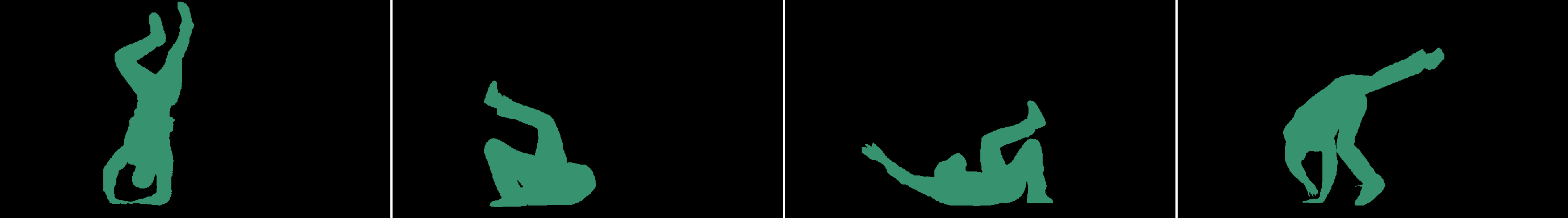} \\
    \multicolumn{3}{c}{\vspace{0.7mm}} \\
    & {\fontsize{6.5}{7}\selectfont\textbf{(c) LVOS}} &
      {\fontsize{6.5}{7}\selectfont\textbf{(d) HOI4D}} \\
    \mainqualrowlabel{RGB} &
      \mainqualimage{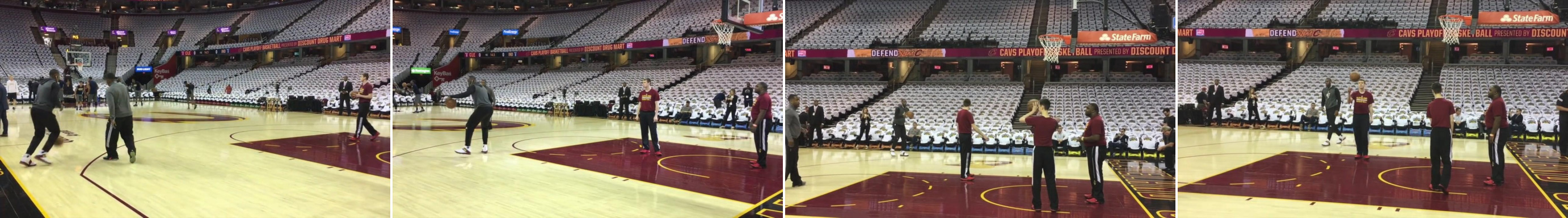} &
      \mainqualimage{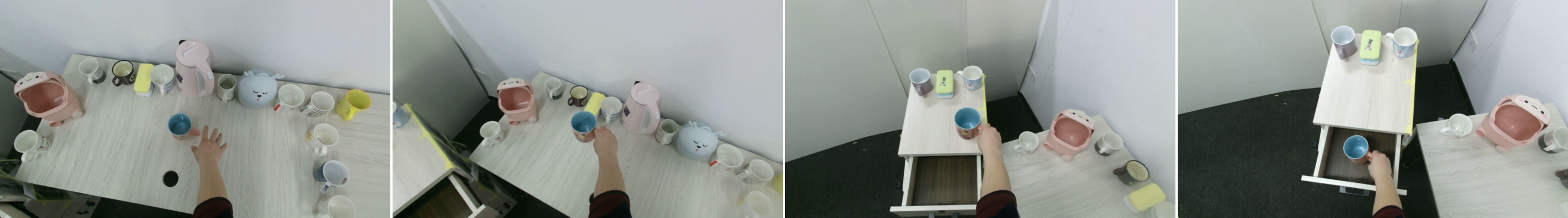} \\
    \mainqualrowlabel{SAM 2} &
      \mainqualimage{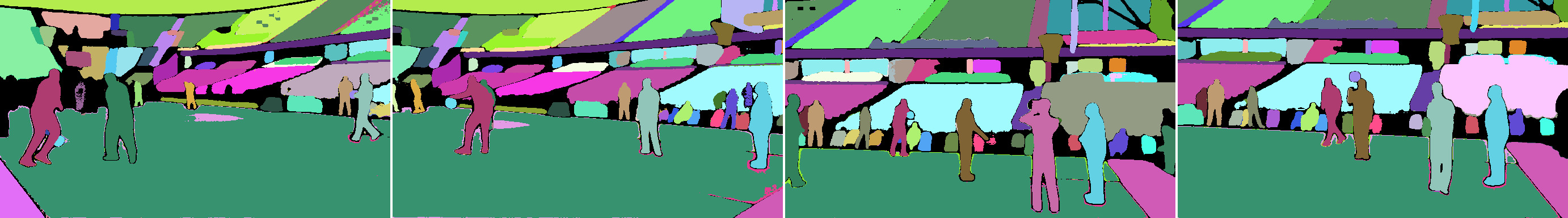} &
      \mainqualimage{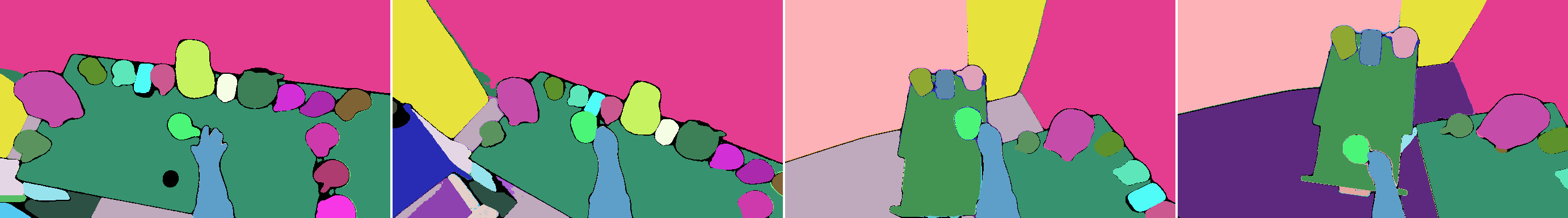} \\
    \mainqualrowlabel{IGGT} &
      \mainqualimage{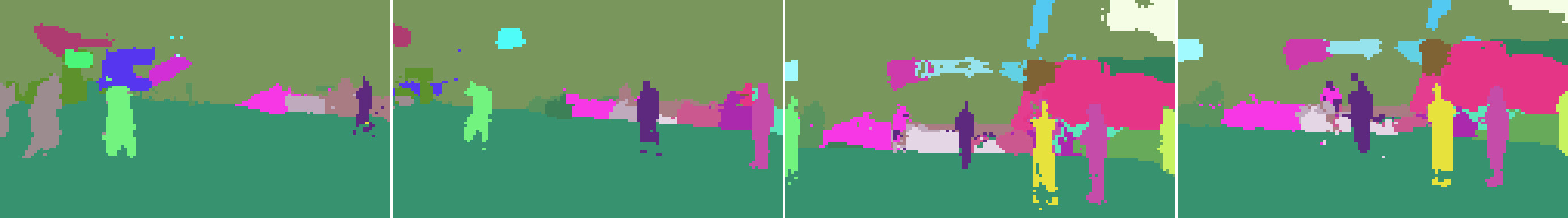} &
      \mainqualimage{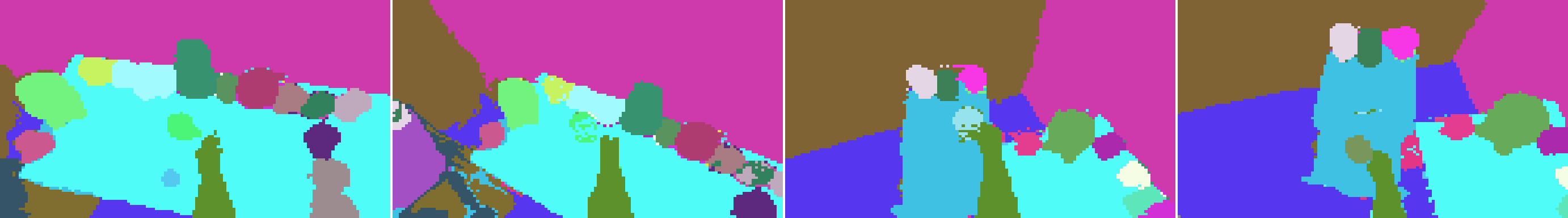} \\
    \mainqualrowlabel{IGGT4D} &
      \mainqualimage{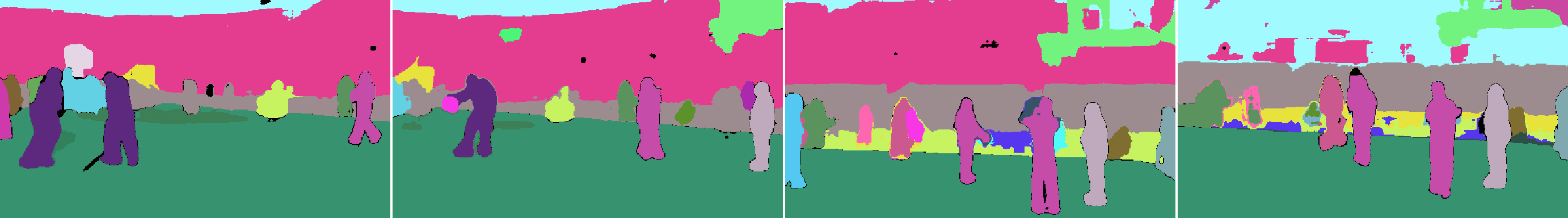} &
      \mainqualimage{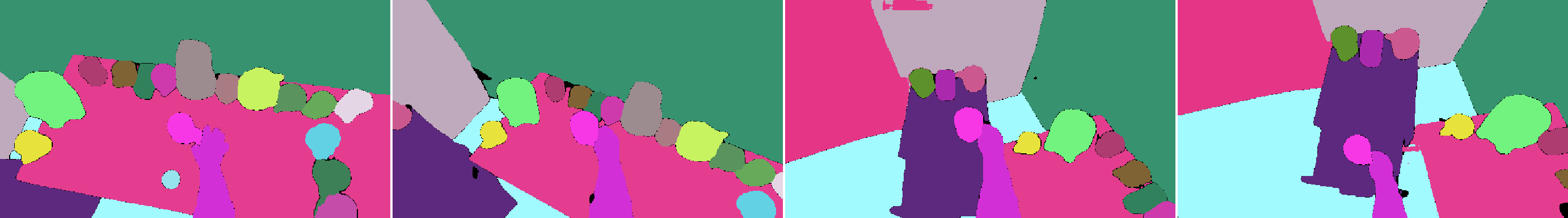} \\
    \mainqualrowlabel{\textbf{Ours}} &
      \mainqualimage{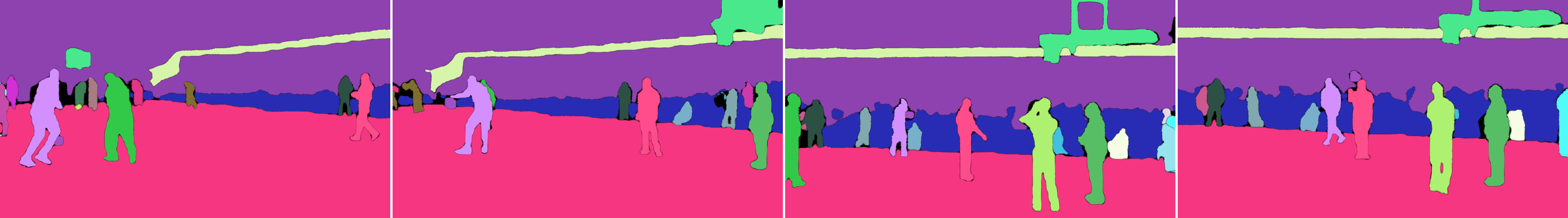} &
      \mainqualimage{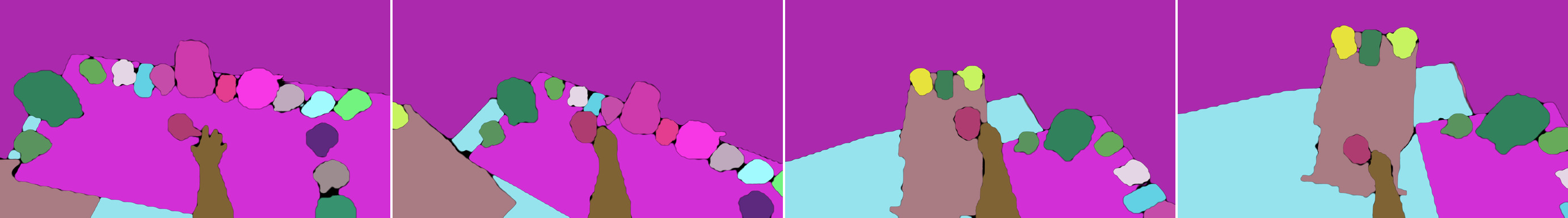} \\
    \mainqualrowlabel{GT} &
      \mainqualimage{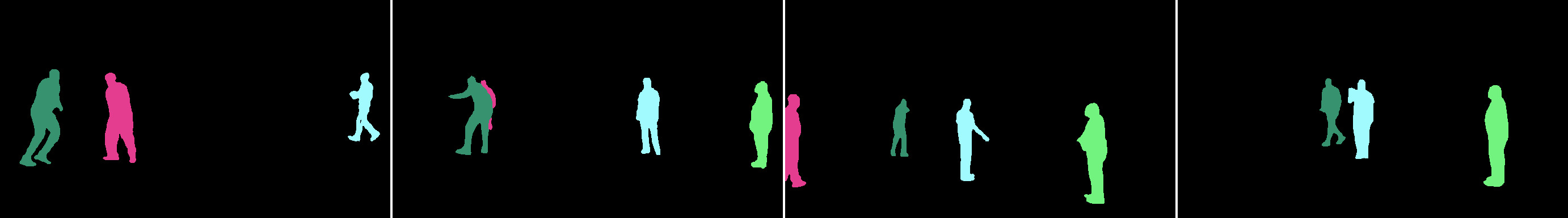} &
      \mainqualimage{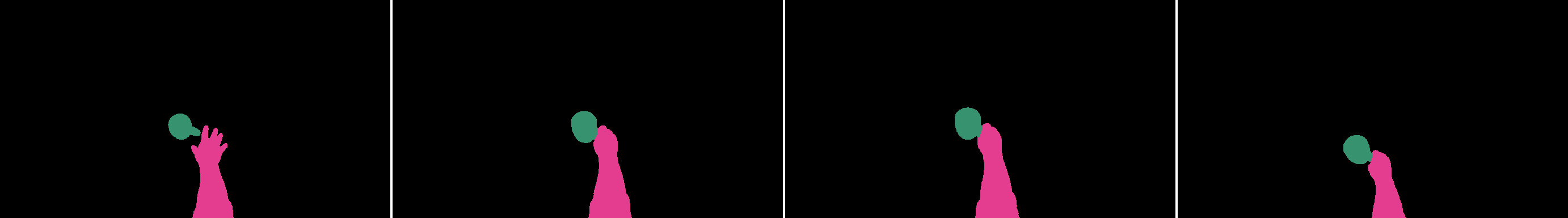}
  \end{tabular}
  \caption{\textbf{Qualitative comparison of class-agnostic segmentation.}
  A consistent mask color denotes the same instance within each sequence. Ground-truth masks cover
  only annotated instances.}
  \label{fig:segmentation-qualitative}
\end{figure}

%% file: arxiv/figures/grounding_qualitative_full.tex
\begin{figure}[!tbp]
  \centering
  \setlength{\tabcolsep}{0pt}
  \renewcommand{\arraystretch}{1.05}
  \newcommand{\groundmainrowlabel}[1]{%
    \raisebox{\dimexpr5.0mm-.5\height+.5\depth\relax}{%
      \rotatebox[origin=c]{90}{\fontsize{6}{6.4}\selectfont #1}}}
  \newcommand{\groundmainimage}[1]{\includegraphics[width=0.445\textwidth]{#1}}
  \begin{tabular}{r@{\hspace{1.0mm}}c@{\hspace{2.0mm}}c}
    & {\fontsize{7}{7.5}\selectfont\textbf{(a) ScanRefer}} &
      {\fontsize{7}{7.5}\selectfont\textbf{(b) MeViS}} \\
    & \parbox[c][9mm][c]{0.445\textwidth}{\centering\fontsize{6}{6.4}\selectfont\emph{``This is a recycling bin atop a surface. It may be empty or filled.''}} &
      \parbox[c][9mm][c]{0.445\textwidth}{\centering\fontsize{6}{6.4}\selectfont\emph{``The rabbit that changed its location.''}} \\
    \groundmainrowlabel{VLM+SAM2} &
      \groundmainimage{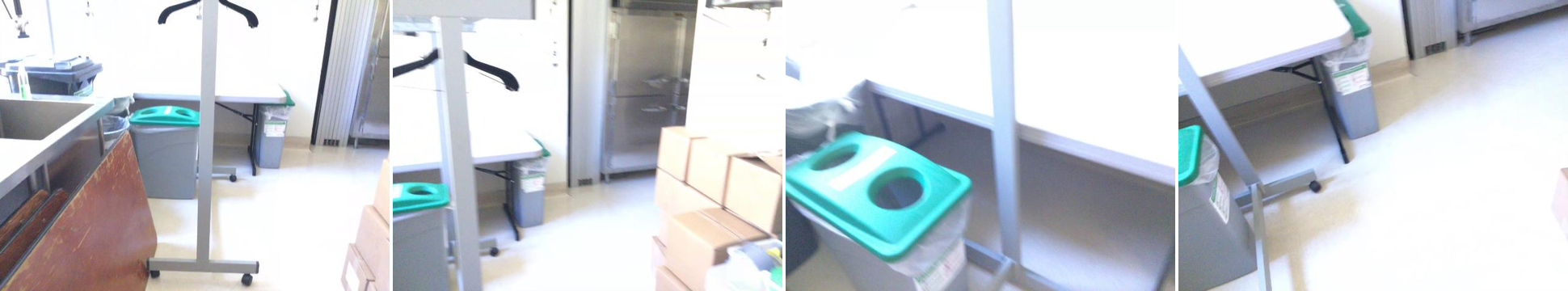} &
      \groundmainimage{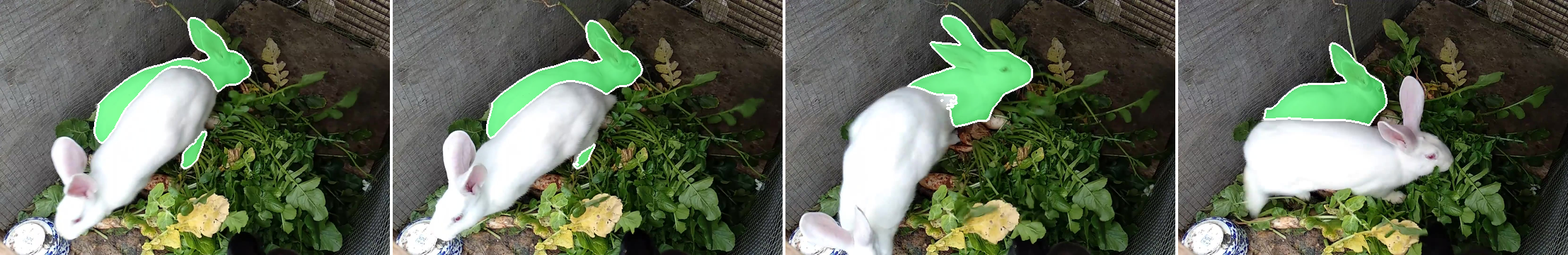} \\
    \groundmainrowlabel{SAM3} &
      \groundmainimage{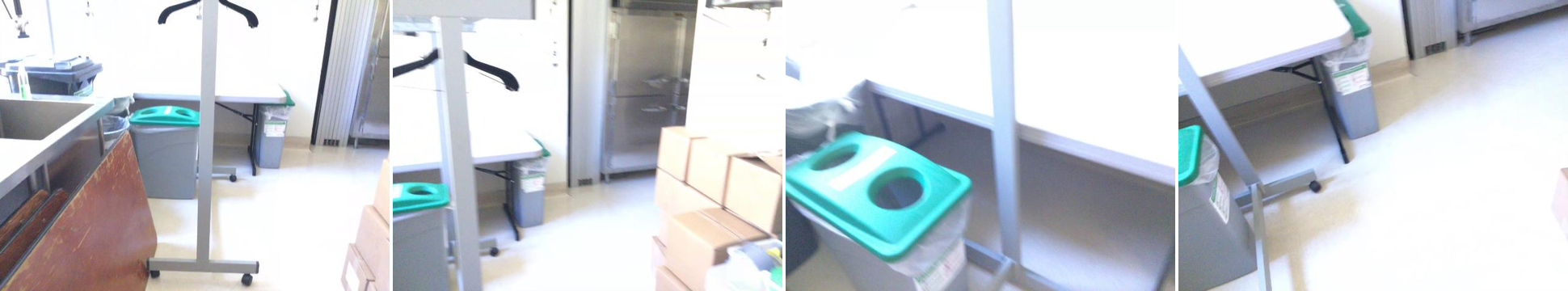} &
      \groundmainimage{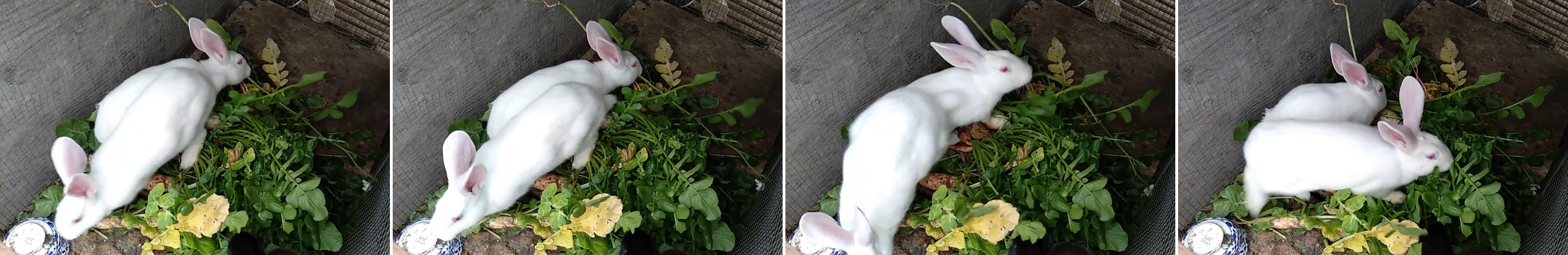} \\
    \groundmainrowlabel{\textbf{Ours}} &
      \groundmainimage{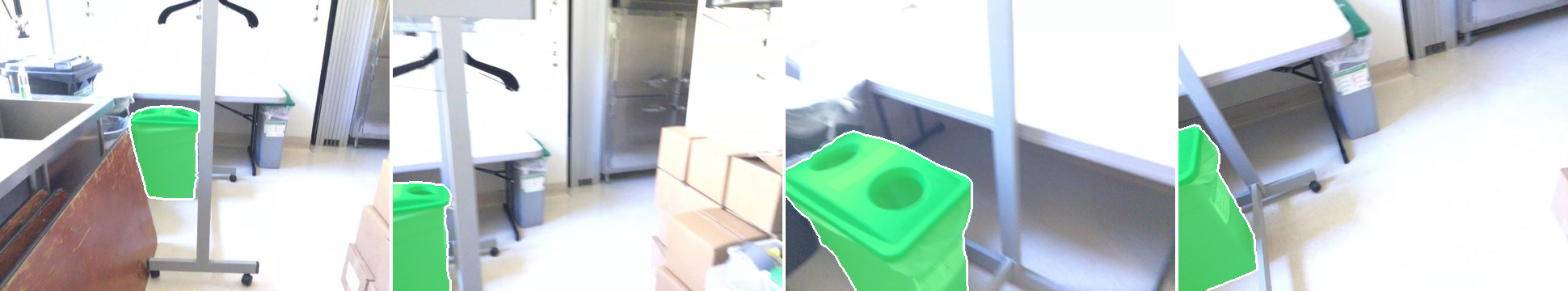} &
      \groundmainimage{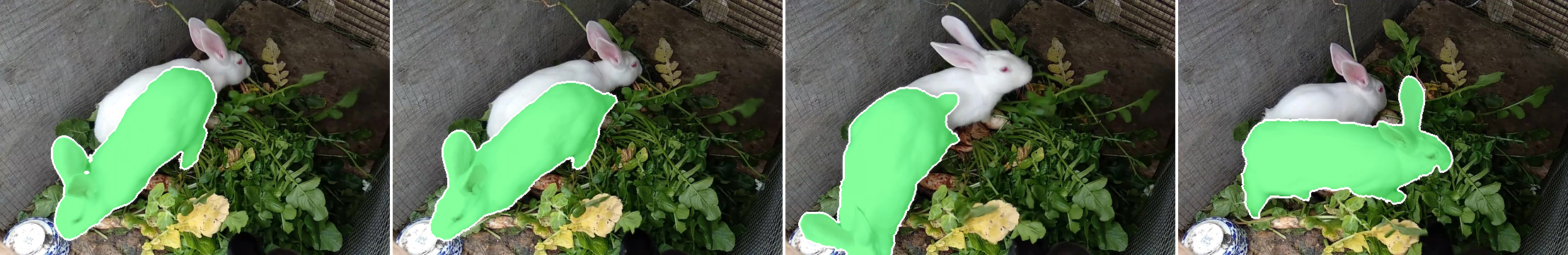} \\
    \groundmainrowlabel{GT} &
      \groundmainimage{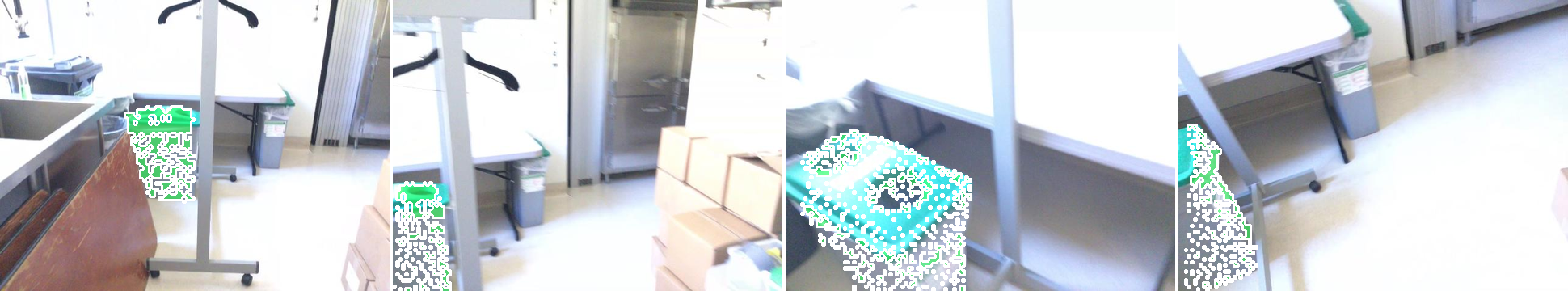} &
      \groundmainimage{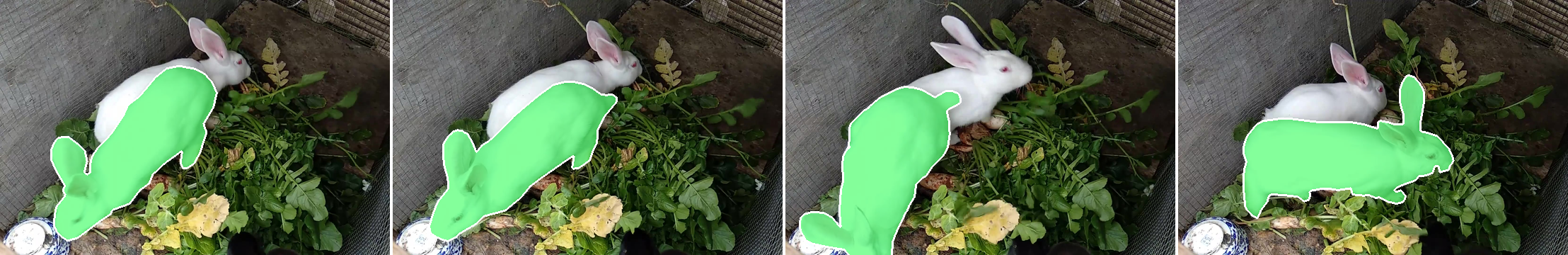}
  \end{tabular}
  \caption{\textbf{Qualitative comparison of language-guided grounding.}
  ScanRefer ground-truth masks are projected from the annotated 3D point cloud.}
  \label{fig:grounding-qualitative}
\end{figure}

%% file: arxiv/figures/frame_4d_qualitative.tex
\begin{figure}[p]
  \centering
  \setlength{\tabcolsep}{1.2pt}
  \renewcommand{\arraystretch}{0.97}
  \newcommand{\mainpcrowlabel}[1]{\raisebox{-.5\height}{\rotatebox[origin=c]{90}{\scriptsize #1}}}
  \newcommand{\mainpcpano}[1]{\raisebox{-.5\height}{\includegraphics[width=0.23\textwidth]{#1}}}
  \newcommand{\mainpcdrive}[1]{\raisebox{-.5\height}{\includegraphics[width=0.35\textwidth]{#1}}}
  \begin{tabular}{@{}>{\centering\arraybackslash}m{4mm}ccc@{}}
    \multicolumn{4}{c}{\scriptsize\textbf{(a) Panoramic indoor scene}} \\
    \multicolumn{4}{c}{\includegraphics[width=0.82\textwidth]{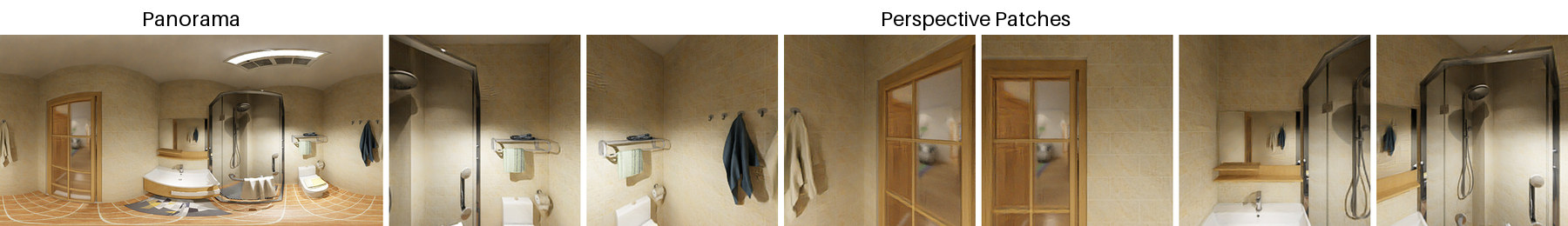}} \\[-0.2mm]
    & \scriptsize IGGT & \scriptsize IGGT4D & \scriptsize\textbf{Ours} \\
    \mainpcrowlabel{RGB} &
      \mainpcpano{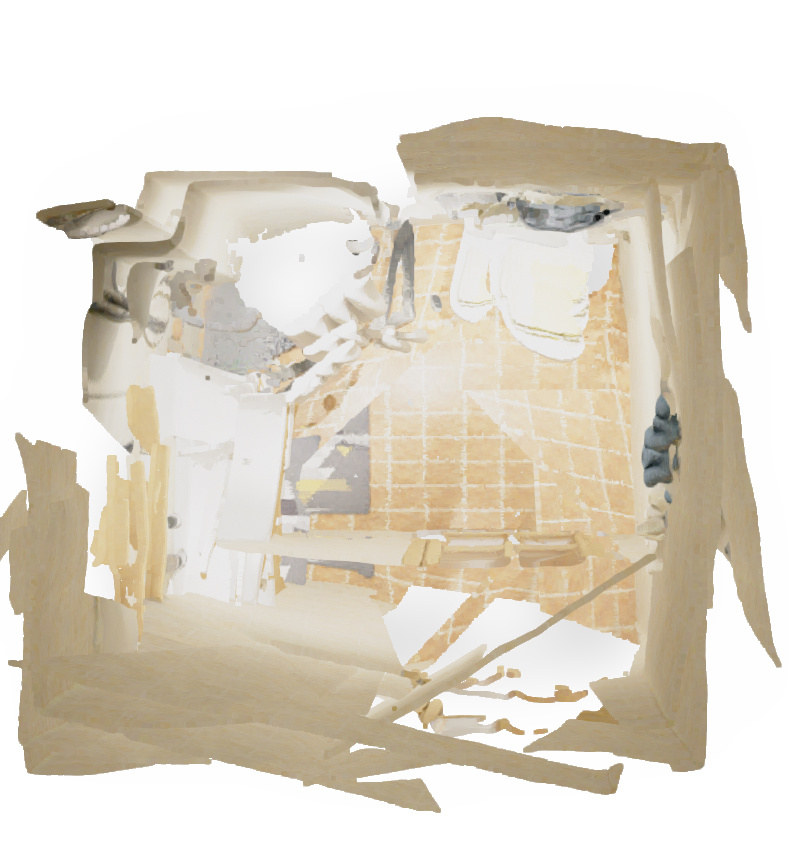} &
      \mainpcpano{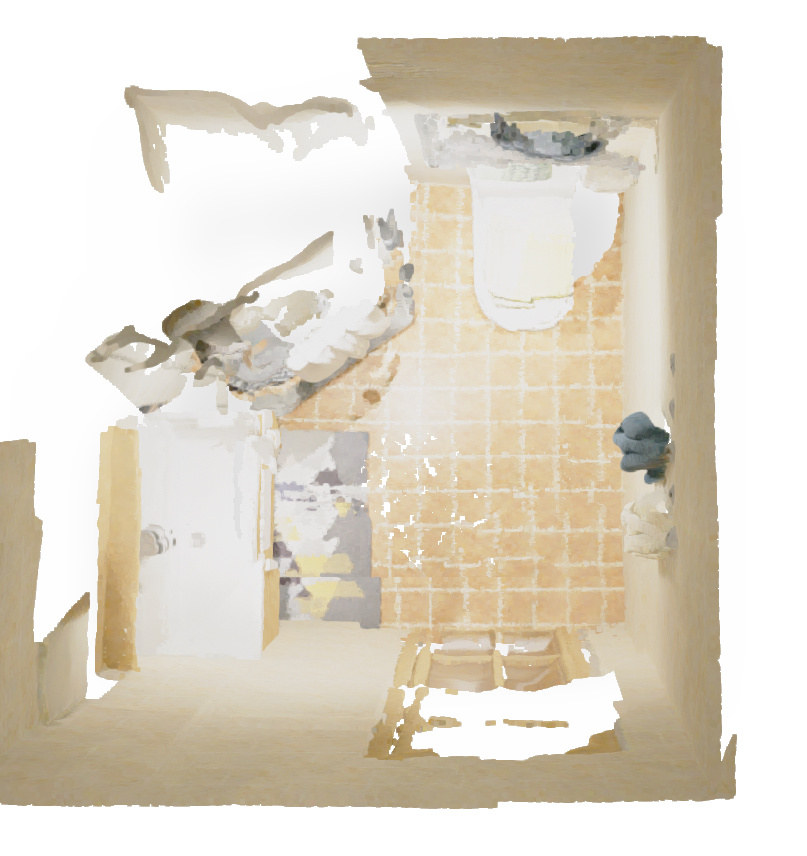} &
      \mainpcpano{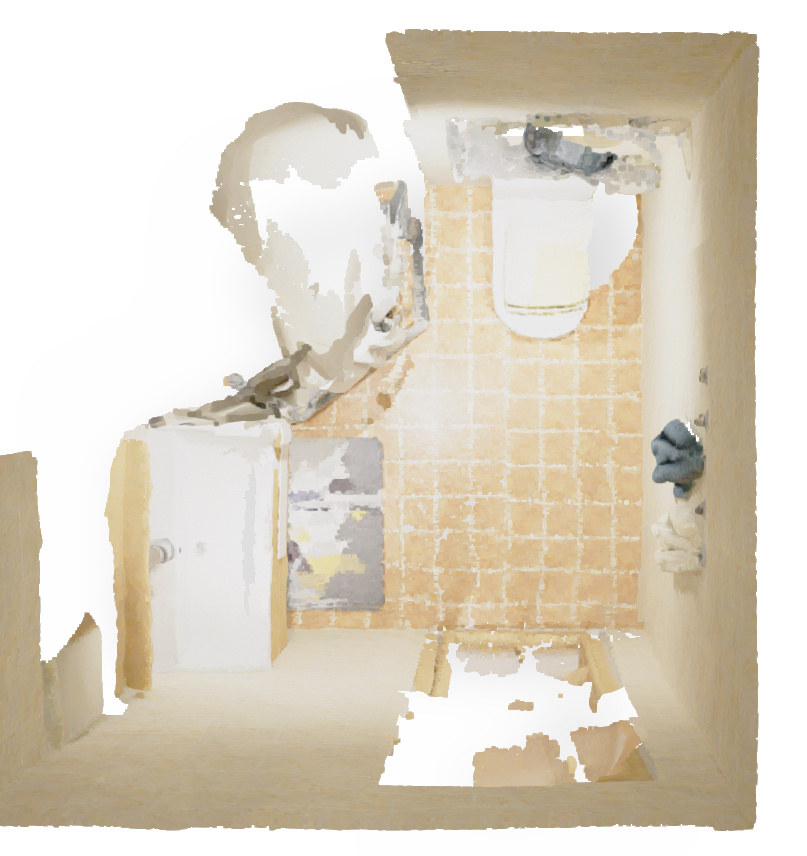} \\[-0.5mm]
    \mainpcrowlabel{Mask} &
      \mainpcpano{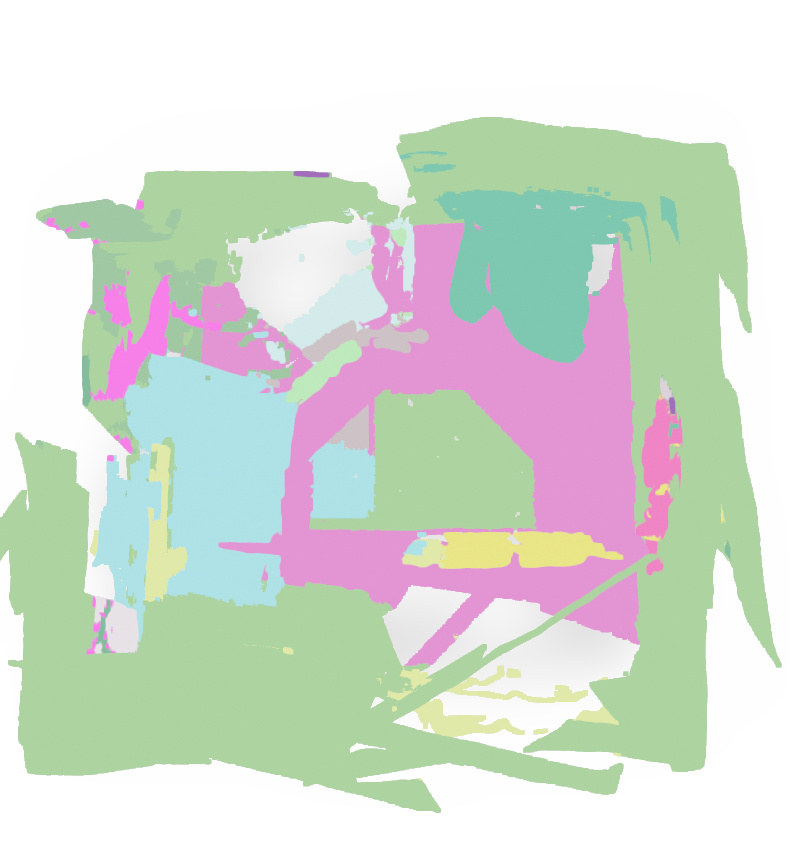} &
      \mainpcpano{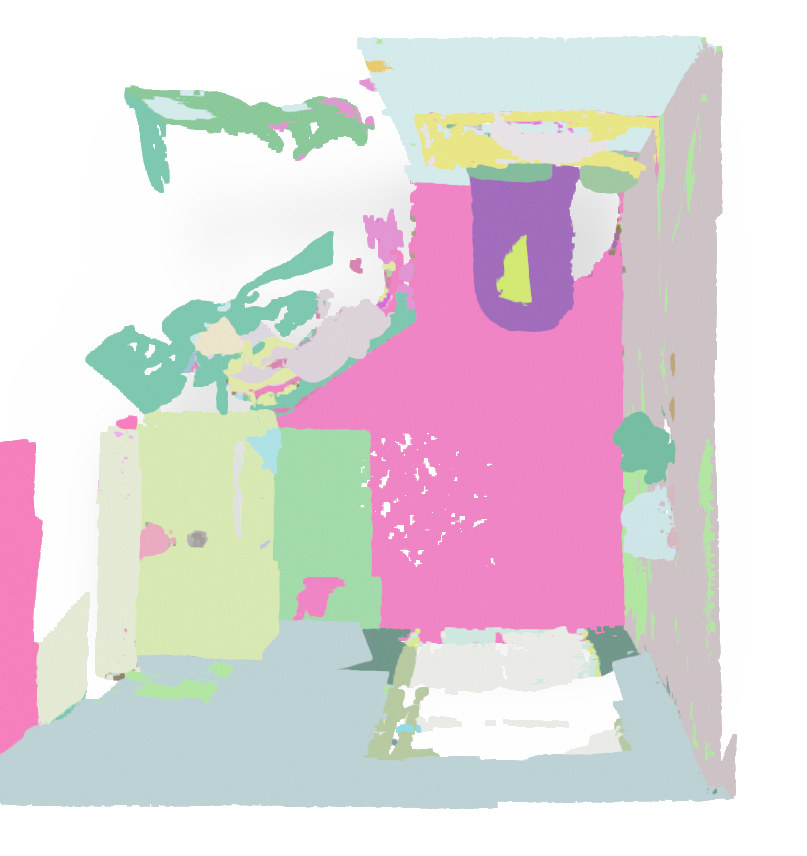} &
      \mainpcpano{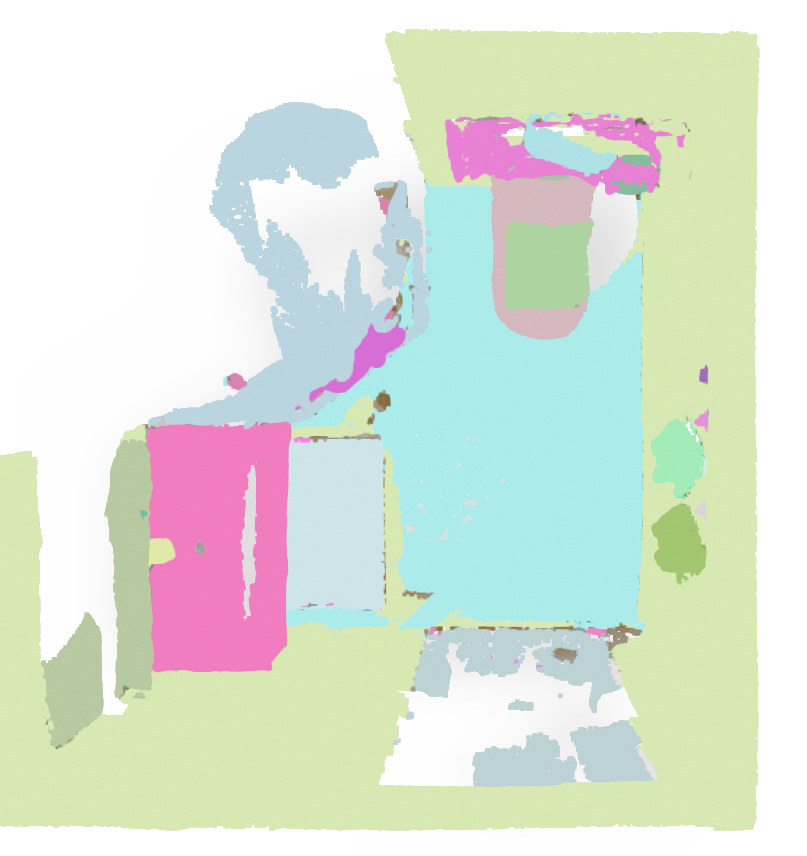} \\[1.8mm]
  \end{tabular}

  \begin{tabular}{@{}>{\centering\arraybackslash}m{4mm}cc@{}}
    \multicolumn{3}{c}{\scriptsize\textbf{(b) Dynamic driving scene}} \\
    \multicolumn{3}{c}{\includegraphics[width=0.82\textwidth]{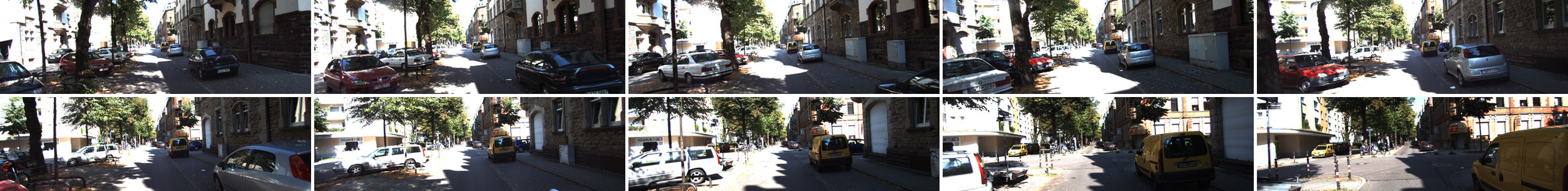}} \\[-0.2mm]
    & \scriptsize IGGT4D & \scriptsize\textbf{Ours} \\
    \mainpcrowlabel{RGB} &
      \mainpcdrive{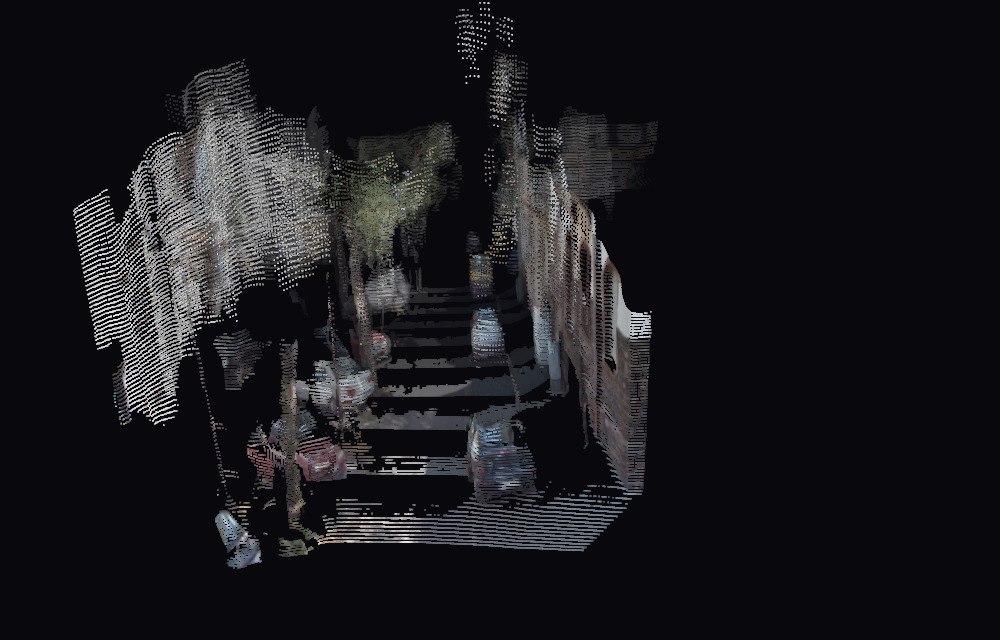} &
      \mainpcdrive{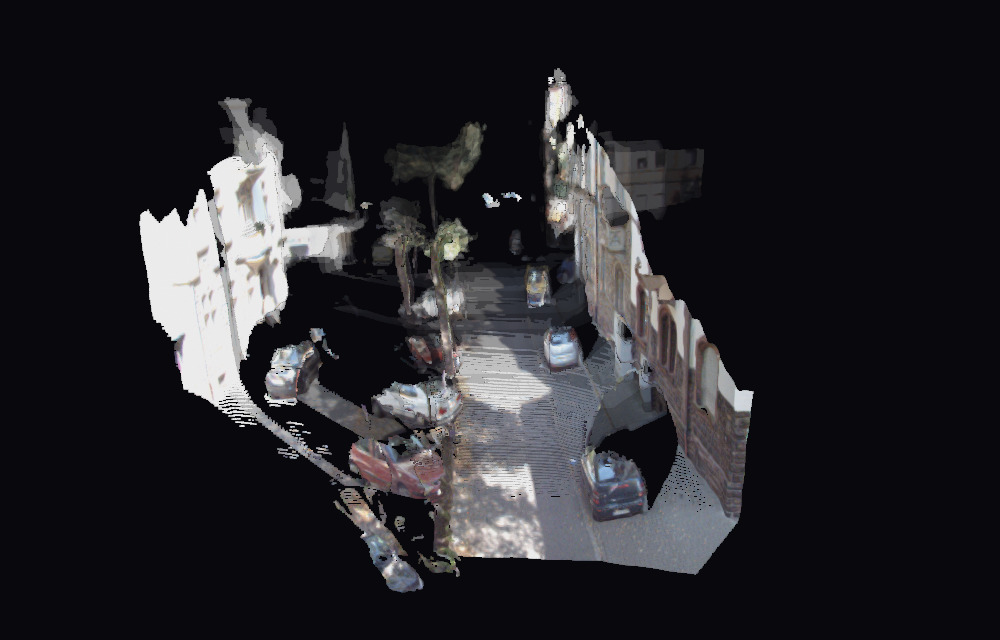} \\[-0.5mm]
    \mainpcrowlabel{Mask} &
      \mainpcdrive{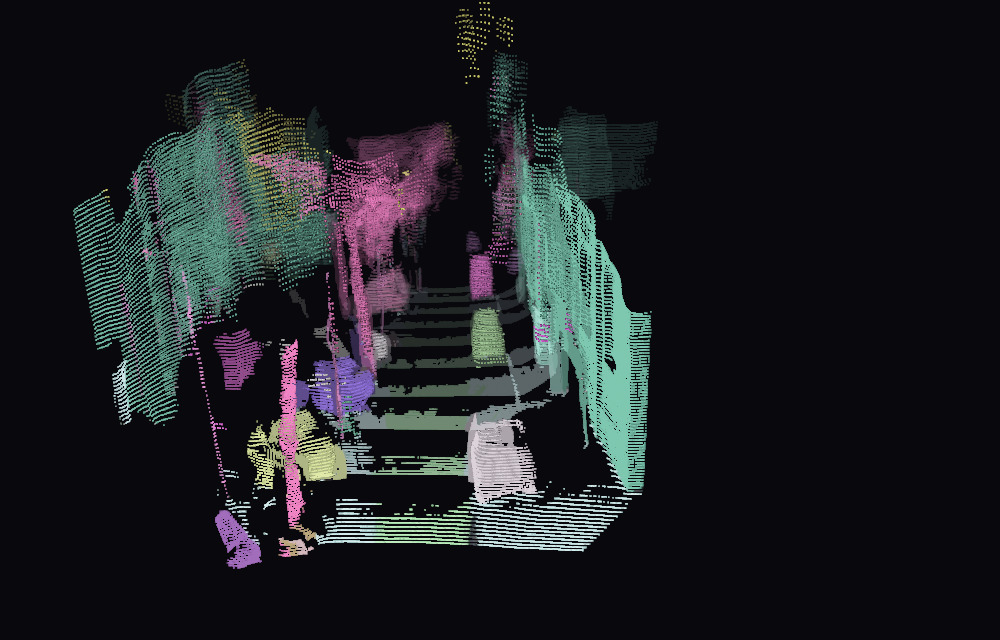} &
      \mainpcdrive{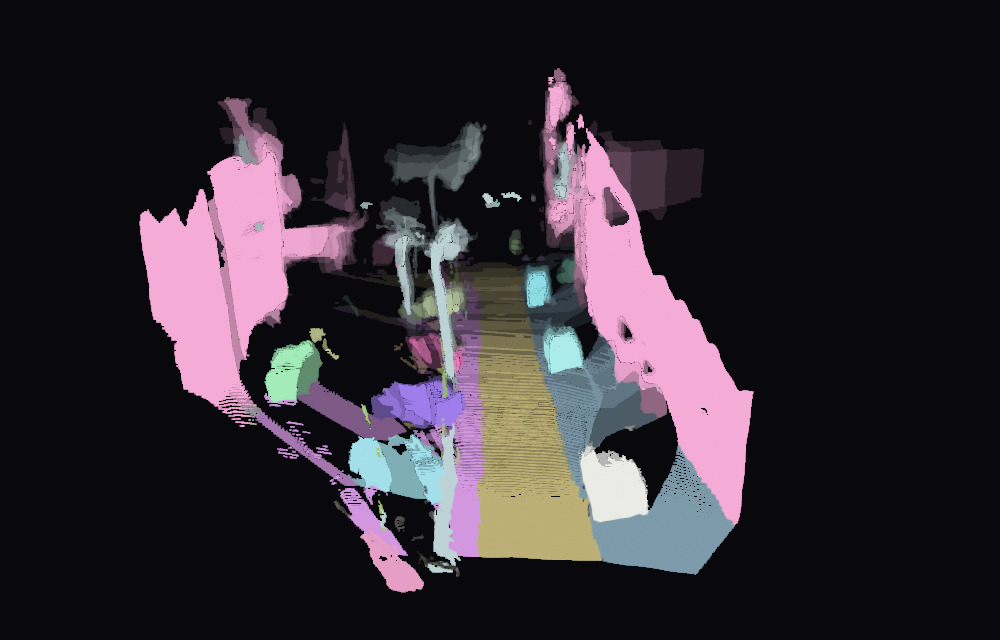}
  \end{tabular}
  \caption{\textbf{Frame-space and reconstructed 4D-space visualizations.}
  Selected observations are followed by aligned RGB and instance-colored point-cloud views. The
  results illustrate persistent instance identities across panoramic views and dynamic driving
  observations.}
  \label{fig:frame-4d-qualitative}
\end{figure}

%% file: arxiv/figures/inthewild_main.tex
\begin{figure}[p]
  \centering
  \setlength{\tabcolsep}{0pt}
  \renewcommand{\arraystretch}{0.94}
  \newcommand{\wildmainrowlabel}[1]{%
    \raisebox{-.5\height}{\rotatebox[origin=c]{90}{\fontsize{6}{6.4}\selectfont #1}}}
  \newcommand{\wildmainstrip}[1]{%
    \raisebox{-.5\height}{\includegraphics[width=0.92\textwidth]{figures_final/inthewild_qualitative/#1}}}
  \begin{tabular}{@{}>{\centering\arraybackslash}m{4mm}c@{}}
    \multicolumn{2}{c}{\fontsize{7}{7.5}\selectfont\textbf{(a) Jet Ski at a Harbor}} \\
    \wildmainrowlabel{RGB} & \wildmainstrip{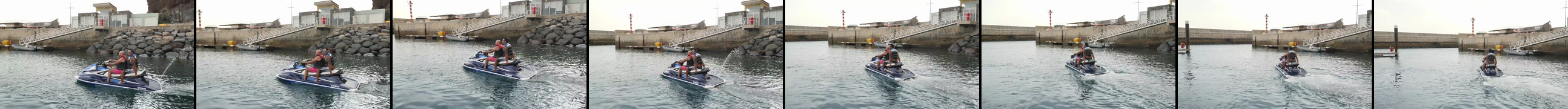} \\
    \wildmainrowlabel{Mask} & \wildmainstrip{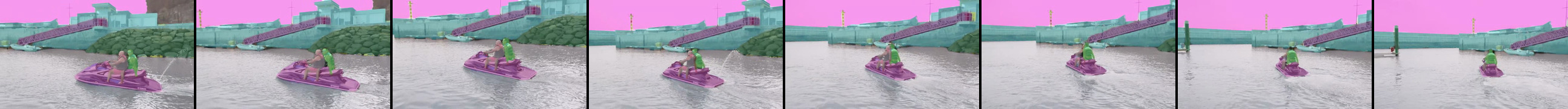} \\
    \multirow{2}{*}{\wildmainrowlabel{4D}} & \wildmainstrip{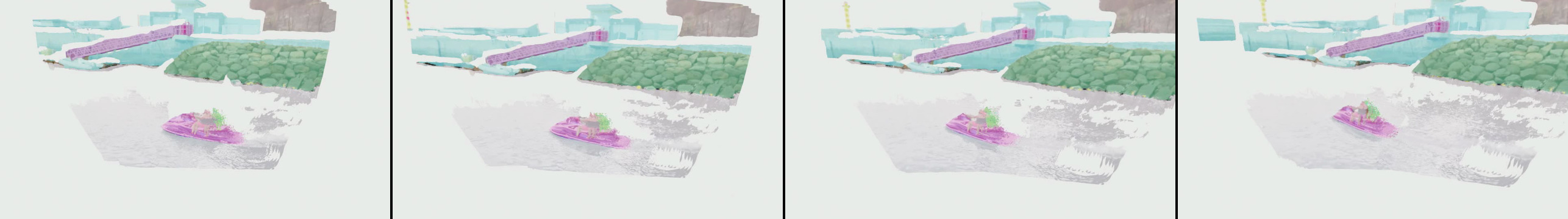} \\
    & \wildmainstrip{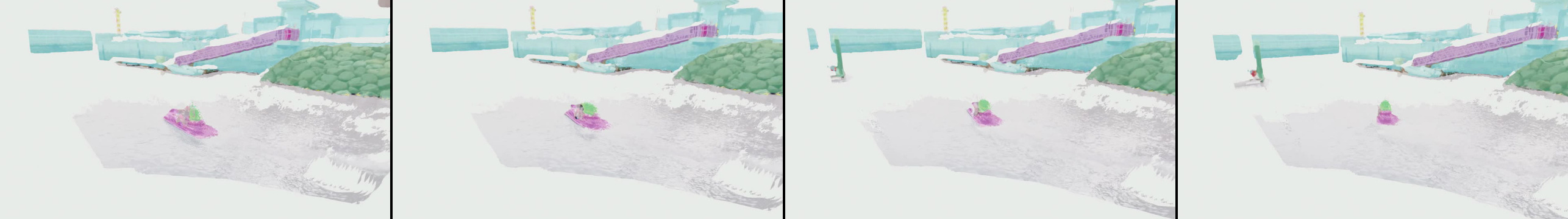} \\[0.7mm]
    \multicolumn{2}{c}{\fontsize{7}{7.5}\selectfont\textbf{(b) Indoor Bedroom}} \\
    \wildmainrowlabel{RGB} & \wildmainstrip{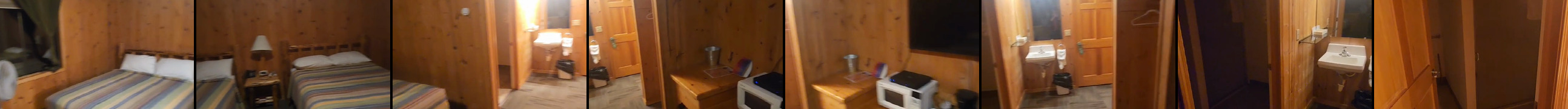} \\
    \wildmainrowlabel{Mask} & \wildmainstrip{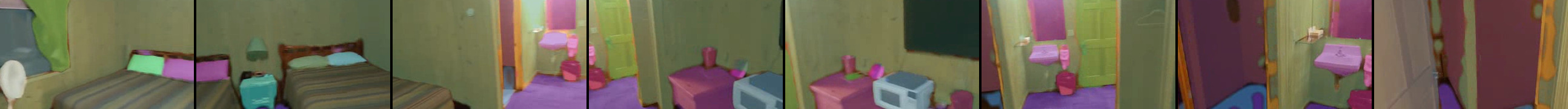} \\
    \multirow{2}{*}{\wildmainrowlabel{4D}} & \wildmainstrip{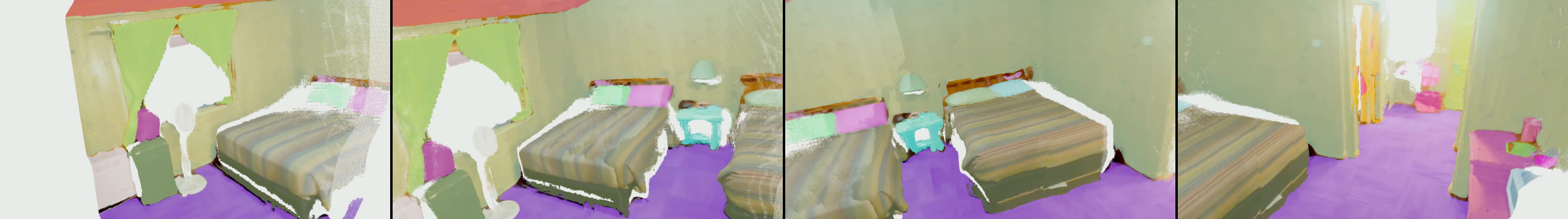} \\
    & \wildmainstrip{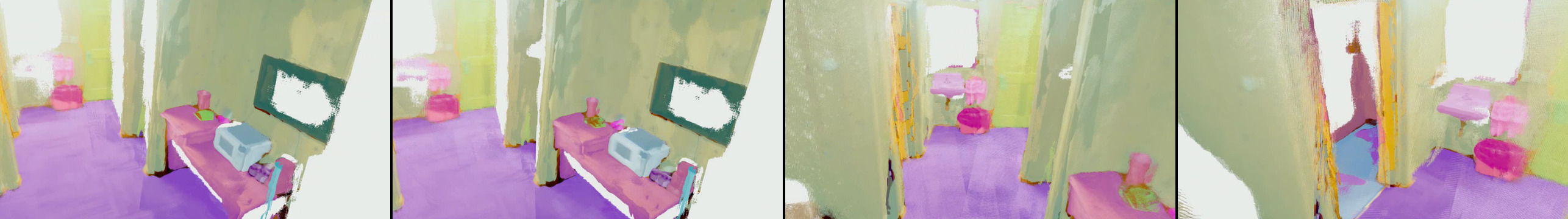} \\[0.7mm]
    \multicolumn{2}{c}{\fontsize{7}{7.5}\selectfont\textbf{(c) Cat in a Living Room}} \\
    \wildmainrowlabel{RGB} & \wildmainstrip{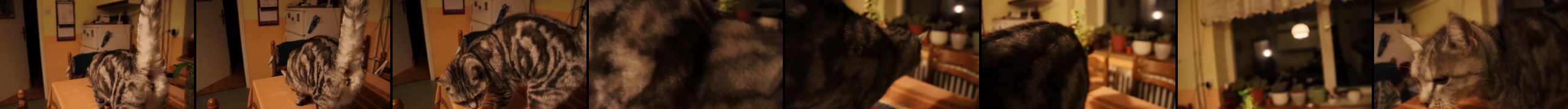} \\
    \wildmainrowlabel{Mask} & \wildmainstrip{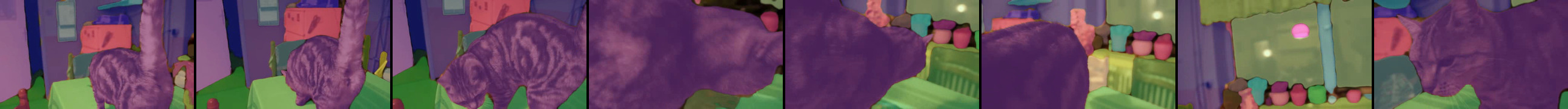} \\
    \multirow{2}{*}{\wildmainrowlabel{4D}} & \wildmainstrip{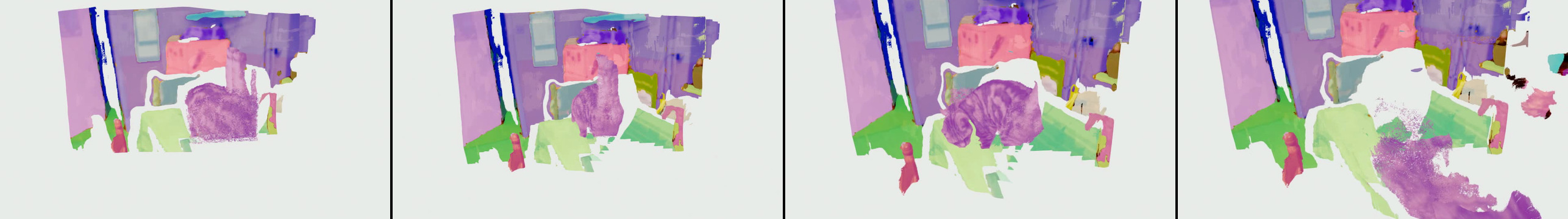} \\
    & \wildmainstrip{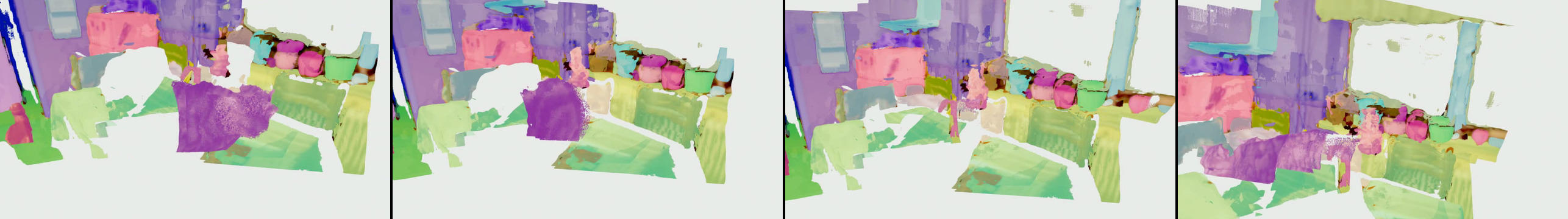}
  \end{tabular}
  \caption{\textbf{In-the-wild 4D segmentation.}
  Uniformly sampled observations are shown with their masks and reconstructed 4D instances in the
  same temporal order.}
  \label{fig:inthewild-qualitative}
\end{figure}

%% file: sections/8_conclusion.tex
\vspace{-2mm}
\section{Conclusion}
\label{sec:conclusion}
\vspace{-2mm}

We introduced \methodname{}, a feed-forward Segment Anything model for 4D scenes. By exploiting a
shared visual-geometric representation of an RGB observation set, \methodname{} exhaustively
segments class-agnostic instances with persistent identities and supports point- and box-based
selection without temporal ordering or memory propagation. We further developed an agentic
language-grounding harness that combines VLM visual priors with geometry-consistent instance
features, actively searches large observation spaces, and resolves complementary grounding
predictions. Experiments across static and dynamic scenes demonstrate state-of-the-art 4D instance
segmentation and strong language-guided grounding performance under our unified evaluation. A
limitation is that jointly processing multiple
high-resolution observations constrains the output mask resolution.

%% file: arxiv/sections/A_method_details.tex
\section{Additional Method Details}
\label{sec:supp-method-details}

\subsection{Space-Time Query Decoder}
\label{sec:supp-st-decoder}

As illustrated in \cref{fig:supp-prompt-st-decoder}(b), the Space-Time Query Decoder separates
global instance reasoning from dense mask prediction. The contextualized patch tokens
$\{\mathbf{z}_i^{F}\}_{i=1}^{T}$ follow two complementary paths. A projection $g_{\mathrm{mem}}$
maps them to a joint decoder memory
\begin{equation}
  \mathbf{Z}=\operatorname{Concat}_{i=1}^{T}
  g_{\mathrm{mem}}\!\left(\mathbf{z}_i^{F}\right),
  \label{eq:supp-decoder-memory}
\end{equation}
while a dense feature head fuses intermediate backbone features into per-observation feature maps
$\{\mathbf{F}_i\}_{i=1}^{T}$. The former supports reasoning across the complete observation set;
the latter preserves the spatial detail required for mask prediction. No chronological-order
embedding is added to $\mathbf{Z}$, allowing the same decoder to process ordered, sparse, or
shuffled observations.

Starting from $N_q$ learnable object queries
$\mathbf{Q}^{(0)}=\{\mathbf{q}_j\}_{j=1}^{N_q}$, decoder layer
$\ell$ performs a Mask2Former-style update~\citep{cheng2022masked}
\begin{equation}
  \mathbf{Q}^{(\ell)}=
  \mathcal{D}_{\ell}\!\left(
    \mathbf{Q}^{(\ell-1)},\mathbf{Z};\mathbf{A}^{(\ell-1)}
  \right),
  \label{eq:supp-st-update}
\end{equation}
where $\mathcal{D}_{\ell}$ comprises masked cross-attention to $\mathbf{Z}$, self-attention among
the object queries, and a feed-forward network. The attention mask
$\mathbf{A}^{(\ell-1)}$ is obtained from the preceding mask prediction and restricts each query to
its current spatiotemporal support; a query with empty support can again attend to the full memory.
Because one query state is shared by all observations, cross-attention gathers different
appearances of an instance into a single persistent representation rather than constructing a
separate query trajectory for each frame.

Purely prompt-driven decoding, as in SAM-style models~\citep{kirillov2023segment}, treats each
prompt independently and may favor locally coherent, low-level regions, leading to
over-segmentation. Our learnable object queries instead exchange information through self-attention
and are jointly optimized under exhaustive set supervision, establishing a global instance
partition before any prompt is introduced.

When a promptable query $\mathbf{q}^{p}$ from \cref{sec:supp-prompt-encoder} is present, it is
concatenated once with the learnable queries. For the ordering
$\overline{\mathbf{Q}}^{(0)}=[\mathbf{q}^{p};\mathbf{Q}^{(0)}]$ shown in
\cref{fig:supp-prompt-st-decoder}(b), the decoder update becomes
\begin{equation}
  \overline{\mathbf{Q}}^{(\ell)}=
  \mathcal{D}_{\ell}\!\left(
    \overline{\mathbf{Q}}^{(\ell-1)},\mathbf{Z};
    \mathbf{A}^{(\ell-1)},\mathbf{U}
  \right),
  \qquad
  \mathbf{U}=
  \begin{bmatrix}
    1 & \mathbf{1}_{1\times N_q} \\
    [-\infty]_{N_q\times1} & \mathbf{1}_{N_q\times N_q}
  \end{bmatrix},
  \label{eq:supp-unidirectional-attention}
\end{equation}
where $1$ denotes an allowed connection and $[-\infty]_{N_q\times1}$ denotes a block whose entries
are all masked. Rows denote receiving queries and columns denote source queries. The lower-left
block therefore prevents the learnable queries from reading $\mathbf{q}^{p}$, while the fully
connected first row allows $\mathbf{q}^{p}$ to read every learnable query. This unidirectional mask is
distinct from the spatiotemporal cross-attention mask $\mathbf{A}^{(\ell-1)}$ in
\cref{eq:supp-st-update}. It lets the promptable query resolve its target against the globally
discovered instance set, while preventing local prompt evidence from altering the convergence
or predictions of the learnable queries.

For a refined query $\mathbf{q}_j$, a mask head produces an embedding
$\mathbf{e}_j=f_{\mathrm{mask}}(\mathbf{q}_j)$ and evaluates it against every dense feature map:
\begin{equation}
  \hat{\mathbf{M}}_{j,i}=\mathbf{e}_j^{\mathsf T}\mathbf{F}_i,
  \qquad
  \hat{o}_j=f_{\mathrm{obj}}(\mathbf{q}_j).
  \label{eq:supp-mask-readout}
\end{equation}
The common embedding $\mathbf{e}_j$ makes the query index the instance identity throughout the
observation set, while $\hat{o}_j$ distinguishes object queries from no-object predictions.
Hungarian matching assigns each ground-truth 4D mask to one query over the full observation set;
unmatched queries are supervised as no-object. The decoder therefore predicts persistent masks
end to end, without clustering dense embeddings or associating independently decoded instances.

\subsection{Prompt Encoder}
\label{sec:supp-prompt-encoder}

The prompt encoder in \cref{fig:supp-prompt-st-decoder}(a) converts image-level points and boxes
into a promptable object query. Although the contextualized patch tokens are the shared backbone
representation, prompt sampling operates on the spatially dense maps $\mathbf{F}_i$ derived from
them. Let $\mathcal{P}=\{p_r\}_{r=1}^{R}$ collect prompts placed in one or more observations, and
let $\tau_r$ identify the prompt type. A point samples a compact neighborhood around its location,
whereas a box samples a regular grid over its enclosed region. A learned sampler then summarizes
each set of samples and adds its type embedding:
\begin{equation}
  \mathbf{v}_r=
  \mathcal{S}_{\tau_r}\!\left(\mathbf{F}_{i_r},p_r\right)
  +\mathbf{t}_{\tau_r}.
  \label{eq:supp-prompt-sample}
\end{equation}
This preserves local evidence for points while allowing boxes to aggregate evidence across their
spatial extent.

Prompts from different observations need not be equally informative: sampled regions may contain
different amounts of target evidence because of occlusion or background inclusion. Moreover,
prompts supplied by users or downstream localization tools may contain localization errors or be
inconsistent across observations. We therefore predict a relevance score for each sampled feature,
enabling the network to filter unreliable evidence before forming the promptable query:
\begin{equation}
  \alpha_r=
  \frac{\exp\!\left(a(\mathbf{v}_r)\right)}
       {\sum_{s=1}^{R}\exp\!\left(a(\mathbf{v}_s)\right)},
  \qquad
  \mathbf{q}^{p}=f_{\mathrm{P}}\!\left(
    \sum_{r=1}^{R}\alpha_r\mathbf{v}_r
  \right).
  \label{eq:supp-prompt-aggregate}
\end{equation}
A positional embedding derived from the mean prompt coordinates is supplied independently of the
content query. Consequently, evidence from multiple observations can specify one target without
collapsing its appearance and location into the same encoding. In
\cref{fig:supp-prompt-st-decoder}(a), $T'$ denotes the number of observations containing prompts.

The resulting $\mathbf{q}^{p}$ enters the decoder through the unidirectional self-attention in
\cref{eq:supp-unidirectional-attention} and uses the same joint memory and dense mask readout as the
learnable queries. It is supervised directly with the indicated target mask, rather than being
assigned to a learnable query through an additional matching objective, and decodes the target's
complete 4D mask even when prompts are supplied in only a subset of observations.

\subsection{Training Data Details}
\label{sec:supp-data-details}

\Cref{tab:supp-training-data} details the 19 sources summarized in
\cref{fig:data-curation}(c). The reported counts refer to the subsets sampled for training rather
than the full source releases. In particular, we sample 5,127 trajectories from RealEstate10K,
4,635 clips from SA-V, and 9,649 scenes from DL3DV. The resulting corpus contains 33,220 scenes,
sub-scenes, or video clips and 6,052,440 RGB observations spanning indoor, outdoor, and mixed
environments as well as real and synthetic imagery.

\input{arxiv/figures/training_data_examples_report}

The exhaustive subset provides the set-level supervision required to discover every annotated
instance. Its static component combines geometry-consistent synthetic views from
Infinigen~\citep{raistrick2023infinite}, real indoor captures from
ScanNet++~\citep{yeshwanth2023scannet++}, and large-scale camera trajectories from
RealEstate10K~\citep{zhou2018stereo}. Its dynamic component draws on the diverse panoptic videos of
VIPSeg~\citep{miao2022large}, urban driving sequences from
Cityscapes-VPS~\citep{kim2020video} and KITTI-STEP~\citep{weber2021step}, and the indoor-outdoor
pedestrian sequences of JRDB-PanoTrack~\citep{le2024jrdb}. Together, these sources expose the
decoder to both cross-view appearance changes in static geometry and persistent identities under
scene motion.

The partially annotated video subset broadens the motion and appearance distribution beyond
panoptic benchmarks. SA-V~\citep{ravi2025sam}, YouTube-VOS~\citep{xu2018youtube}, and
UVO-Dense~\citep{wang2021unidentified} contribute diverse object masklets; MeViS~\citep{ding2023mevis},
MOSE~\citep{ding2023mose}, LVOS~\citep{hong2023lvos}, DAVIS~\citep{perazzi2016benchmark,pont20172017},
and VOST~\citep{tokmakov2023breaking} emphasize motion-dependent references, dense distractors,
long-term visibility changes, and object transformations. DynamicReplica~\citep{karaev2023dynamicstereo}
and SAIL-VOS~\citep{hu2019sail} complement these real videos with controlled synthetic motion.
Their original masklets supervise annotated targets, while teacher predictions provide dense
instance coverage outside the labelled subset.

Finally, DL3DV~\citep{ling2024dl3dv} and Matterport3D~\citep{chang2017matterport3d} provide
3.51M unlabelled multi-view observations from diverse real scenes. These sources are trained solely
with teacher-generated pseudo-masklets and account for most of the corpus scale. Representative
RGB observations and their corresponding annotations or pseudo-mask overlays are shown in
\cref{fig:supp-training-data-examples}.

\subsection{Agentic Grounding in 4D}
\label{sec:supp-grounding-details}

\paragraph{Grounder details.}
For a grounder window, each observation is represented by its RGB image and the corresponding
camera and register tokens from the frozen segmentation backbone. The geometry tokens are projected
into the VLM embedding space and condition a frame-specific grounding state. In the query stream,
this state is compared with all
persistent object-query embeddings, augmented with a learned null state for target absence. In the
box stream, we retain Qwen's native text-generation interface to exploit its pretrained spatial
priors, autoregressively predicting a normalized box or an explicit absence token for each
observation. The resulting boxes are passed jointly to the prompt encoder, rather than being
used to restrict the observations decoded by the segmenter.

An observation may be visited by several overlapping search windows. Let $\mathcal{V}$ denote the
set of visited windows and $a_{w,i,j}$ the matching score assigned to persistent object query
$\mathbf{q}_j$ for observation $i$ in window $w$. The query stream selects
\begin{equation}
  j^{q}=\arg\max_j\sum_{w\in\mathcal{V}}\sum_{i\in w}a_{w,i,j},
  \label{eq:supp-query-voting}
\end{equation}
which aggregates query evidence across all visited windows without introducing an additional
temporal association step. Repeated box predictions for the same observation are consolidated by
presence voting and coordinate-wise median aggregation. The selected persistent query and the
aggregated box prompts are then decoded by the Space-Time Query Decoder into
$\hat{\mathbf{M}}^{q}$ and $\hat{\mathbf{M}}^{p}$, respectively.

\paragraph{Critic details.}
The critic operates on the final decoded masks. For each
candidate, we select representative observations according to its visible mask area and render the
original image beside a colored mask overlay. The two rendered candidates and the referring
expression are presented to a VLM followed by a scalar preference head. Because a pairwise VLM
judge may favor one presentation position irrespective of candidate quality, candidate order is
randomized during training. At inference, we additionally evaluate both candidate orders and
antisymmetrize their logits:
\begin{equation}
  s_{\mathrm{C}}=
  h_{\mathrm{C}}\!\left(x,\mathcal{R}(\hat{\mathbf{M}}^{q}),
  \mathcal{R}(\hat{\mathbf{M}}^{p})\right)
  -h_{\mathrm{C}}\!\left(x,\mathcal{R}(\hat{\mathbf{M}}^{p}),
  \mathcal{R}(\hat{\mathbf{M}}^{q})\right),
  \label{eq:supp-critic-swap}
\end{equation}
where $\mathcal{R}$ denotes representative-view rendering and a positive score selects the query
candidate. This comparison allows the critic to judge correspondence with the expression as well
as the spatial and temporal consistency visible in the decoded masks.

\paragraph{Active Tree Search.}
ATS is defined over the supplied observation order; adjacency therefore refers to neighboring input
observations, which may follow a static scan trajectory or a dynamic sequence. The complete
recursive traversal is given in \cref{alg:ats}.

The box stream therefore controls the recursive search, while predictions from every visited
window contribute to the two candidates evaluated by the critic.

\paragraph{Grounding data construction.}
\emph{Grounder data.}
We combine the human referring annotations of ScanRefer~\citep{chen2020scanrefer},
MeViS~\citep{ding2023mevis}, and Ref-DAVIS~\citep{khoreva2018video} with descriptions generated for
ScanNet++~\citep{yeshwanth2023scannet++} and LVOS~\citep{hong2023lvos}, which do not provide
referring text. For ScanNet++, we use ground-truth 3D instances and their semantic labels as class anchors.
  Qwen3-VL-32B~\citep{bai2025qwen3} receives the most visible RGB observation and up to two
  spatially separated context observations to generate up to five descriptions per instance.
  Quality control filters refusals, regenerates descriptions that omit the class anchor, checks
  hallucinations, contradictions, and neighboring-object errors with a VLM, and removes exact
  duplicates. For LVOS, we use GPT-5.6~\citep{openai2026gpt56} to generate referring-expression
  annotations for its video sequences. Each description is associated with its annotated instance or track and matched to a persistent
object query by mask overlap over the available observations. For ScanNet++ and the video sources,
we discard training pairs whose predicted masks have 2D mask IoU below $0.4$ with the annotations,
preventing incomplete or fragmented predictions from introducing incorrect grounding
correspondences. After target filtering and source balancing, the training loader contains 79,856
records from 3,065 scenes or videos: 30,352 from ScanRefer, 31,764 from ScanNet++, 16,888 from
MeViS, 435 from LVOS, and 417 from Ref-DAVIS.

Geometry-guided CoT supervision is generated for the known referent in a separate stage. For video
datasets without 3D/4D ground truth, the visual geometry backbone~\citep{wang2026vggt} reconstructs depth and camera
geometry directly from the RGB observations. Back-projecting the ground-truth masks through this
geometry yields an object-centric 4D point track, whose per-observation centroids and extents
describe the referent's location, motion, and spatial relations to nearby instances. Static-scene
geometry further provides metric object extent, height above the floor, and metric inter-object
distances. Since monocular video reconstruction has neither an absolute scale nor a gravity-aligned
frame, its facts instead
use target visibility, size-normalized motion, relative object size and distance, and variation in
camera distance.
Qwen3-VL-235B receives these facts, the referring expression, up to four uniformly sampled RGB
observations, and the known target, and produces a short rationale that must combine an appearance
cue with at least one geometric or spatiotemporal cue without revealing the target index.

\emph{Critic data.}
We construct critic training data from the grounder's full dual-stream predictions on the training
splits. For each expression, the query and box outputs are decoded into two candidate masks and
rendered as a comparison pair. Candidate quality is measured against the annotated referent using
3D box IoU for static scenes and \JF{} for videos; the higher-scoring candidate provides the
preferred label. We discard pairs in which neither
candidate reaches $0.5$ and subsample near-ties, leaving 4,005 comparisons; this prevents jointly
failed or ambiguous predictions from introducing noisy comparative supervision.

%% file: arxiv/figures/training_data_examples_report.tex
\begin{figure}[!tbp]
  \centering
  \begingroup
  \setlength{\tabcolsep}{0.25pt}
  \renewcommand{\arraystretch}{0.96}
  \newcommand{\suppdatarowlabel}[1]{%
    \raisebox{\dimexpr-.5\height+.3ex\relax}{\rotatebox[origin=c]{90}{\scriptsize #1}}}
  \newcommand{\suppdataimage}[1]{%
    \raisebox{-.5\height}{\includegraphics[width=0.118\linewidth]{#1}}}
  \begin{tabular}{@{}c@{\hspace{2pt}}*{8}{c}@{}}
    & \multicolumn{4}{c}{\scriptsize Infinigen}
    & \multicolumn{4}{c}{\scriptsize RealEstate10K} \\
    \suppdatarowlabel{RGB}
    & \suppdataimage{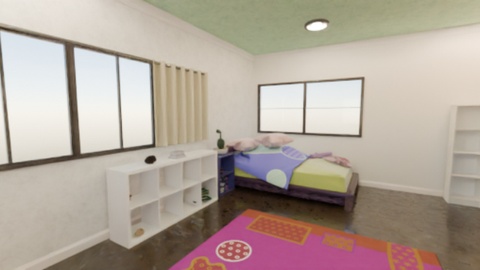}
    & \suppdataimage{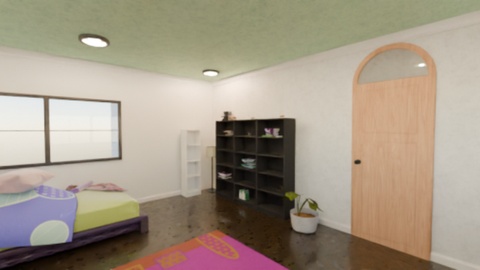}
    & \suppdataimage{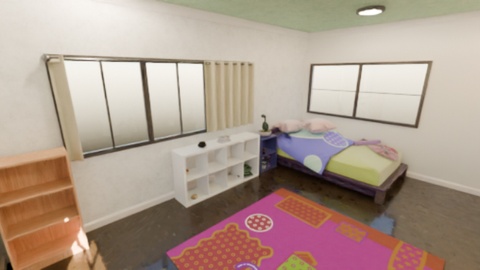}
    & \suppdataimage{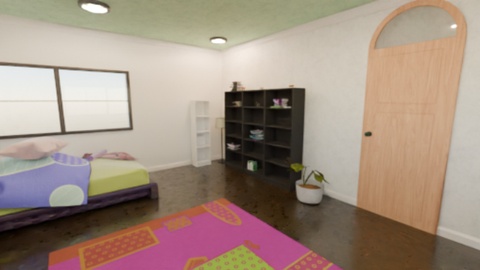}
    & \suppdataimage{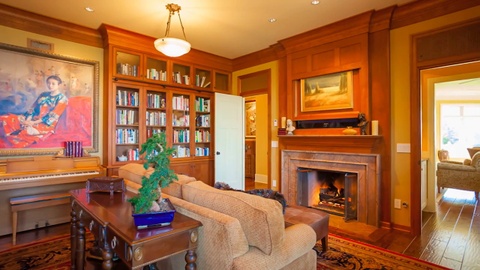}
    & \suppdataimage{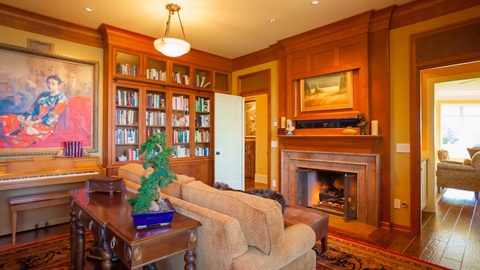}
    & \suppdataimage{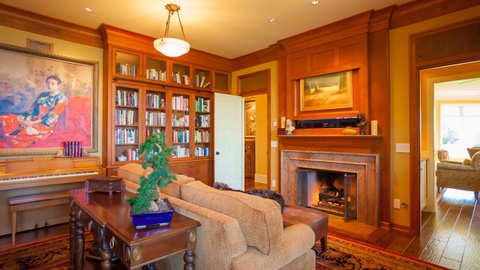}
    & \suppdataimage{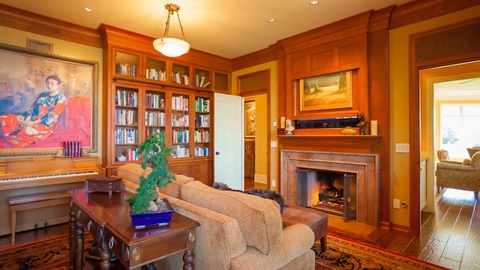} \\
    \suppdatarowlabel{Mask}
    & \suppdataimage{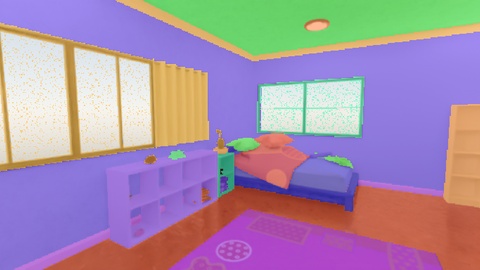}
    & \suppdataimage{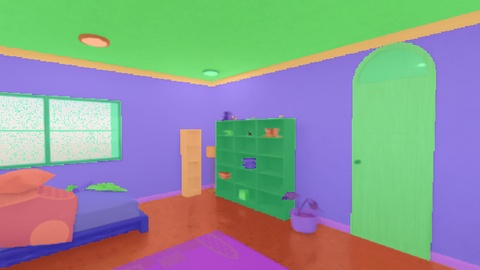}
    & \suppdataimage{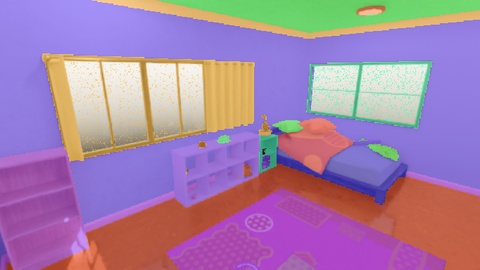}
    & \suppdataimage{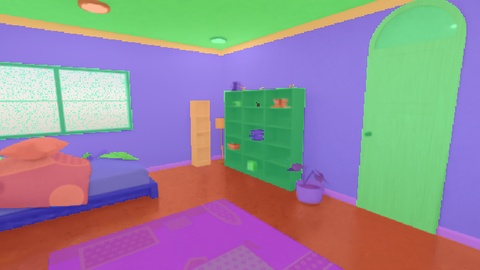}
    & \suppdataimage{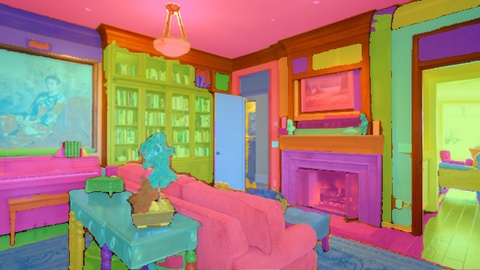}
    & \suppdataimage{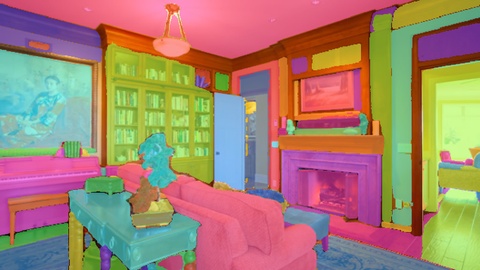}
    & \suppdataimage{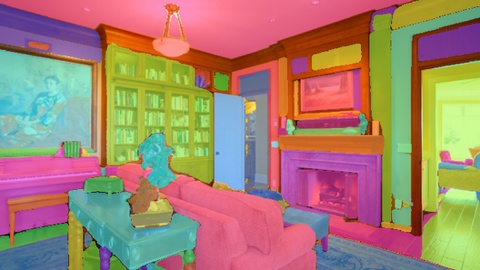}
    & \suppdataimage{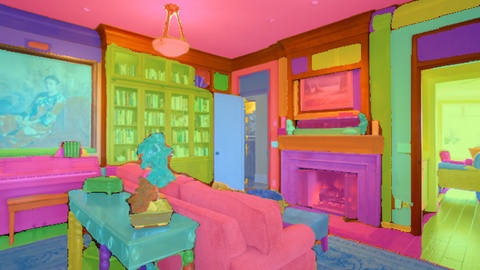} \\
    \addlinespace[0.2em]
    & \multicolumn{4}{c}{\scriptsize Cityscapes-VPS}
    & \multicolumn{4}{c}{\scriptsize JRDB-PanoTrack} \\
    \suppdatarowlabel{RGB}
    & \suppdataimage{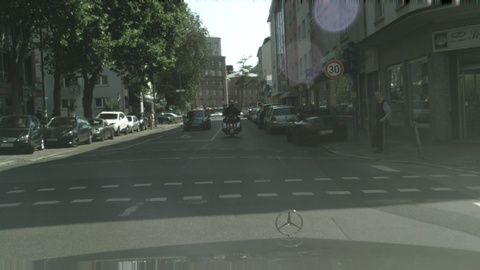}
    & \suppdataimage{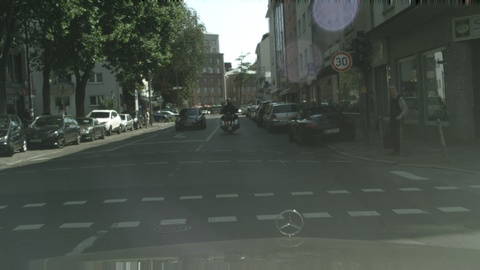}
    & \suppdataimage{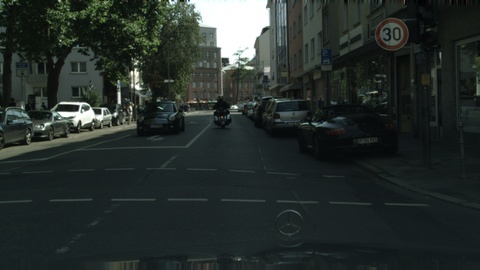}
    & \suppdataimage{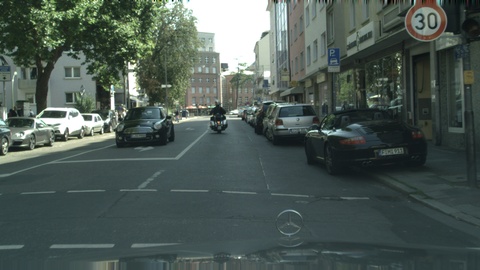}
    & \suppdataimage{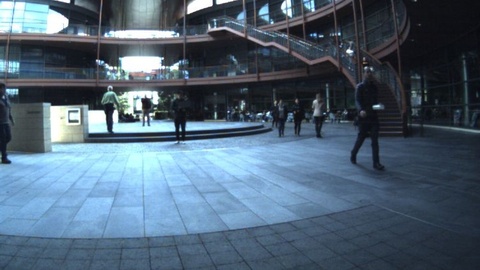}
    & \suppdataimage{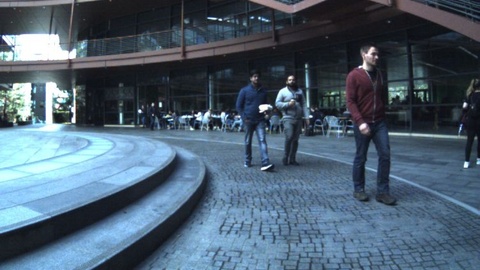}
    & \suppdataimage{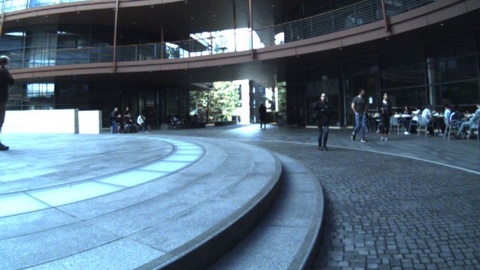}
    & \suppdataimage{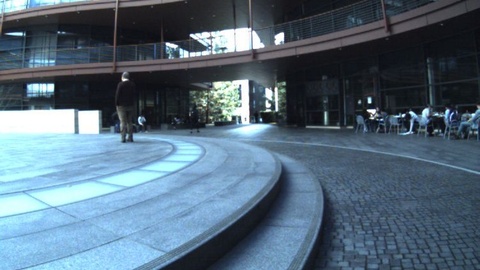} \\
    \suppdatarowlabel{Mask}
    & \suppdataimage{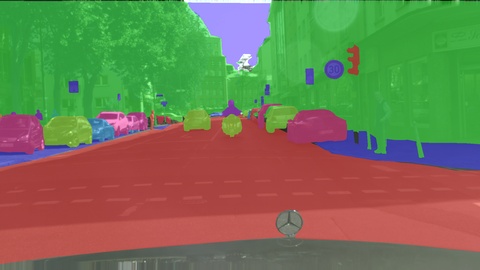}
    & \suppdataimage{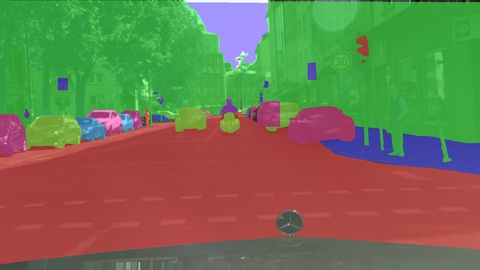}
    & \suppdataimage{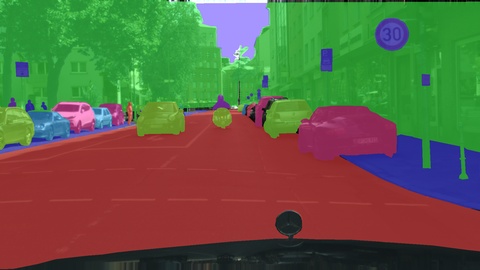}
    & \suppdataimage{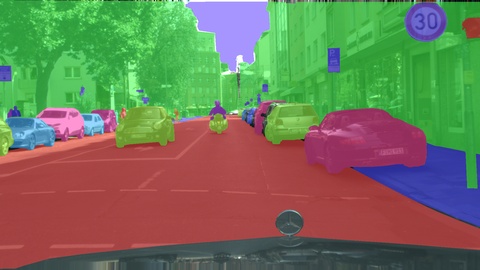}
    & \suppdataimage{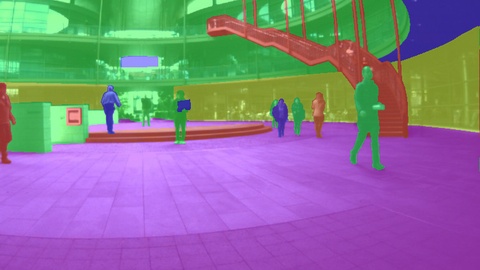}
    & \suppdataimage{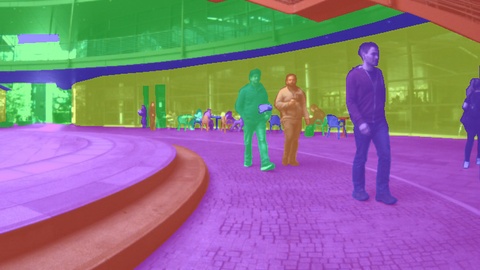}
    & \suppdataimage{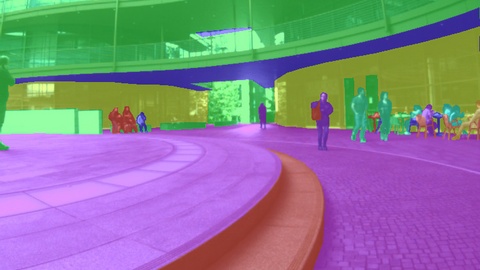}
    & \suppdataimage{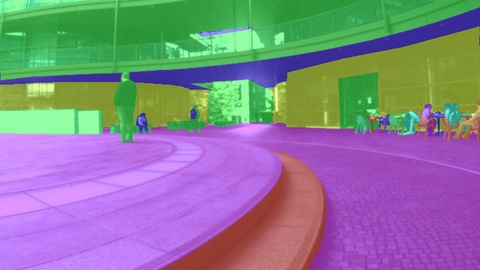} \\
    \addlinespace[0.2em]
    & \multicolumn{4}{c}{\scriptsize SA-V}
    & \multicolumn{4}{c}{\scriptsize YouTube-VOS} \\
    \suppdatarowlabel{RGB}
    & \suppdataimage{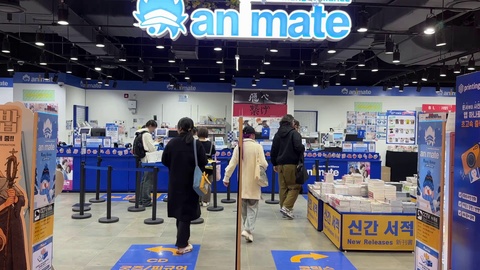}
    & \suppdataimage{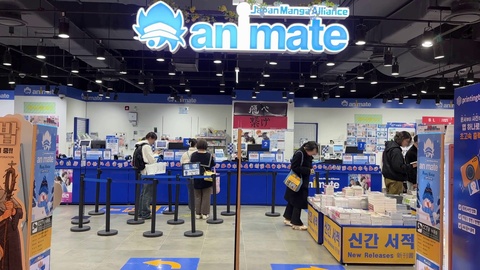}
    & \suppdataimage{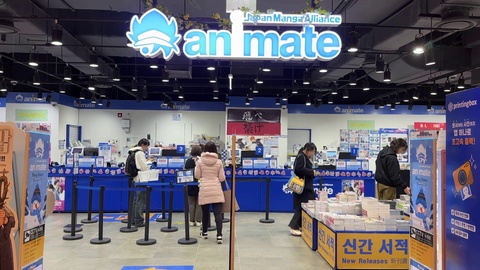}
    & \suppdataimage{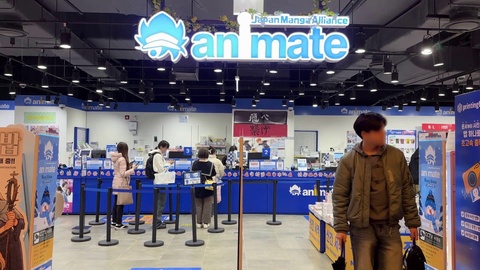}
    & \suppdataimage{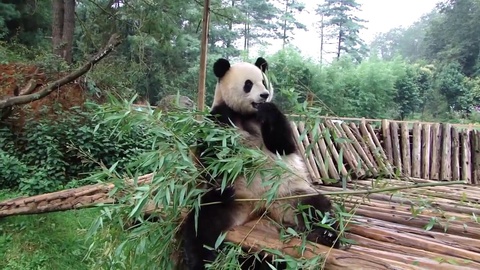}
    & \suppdataimage{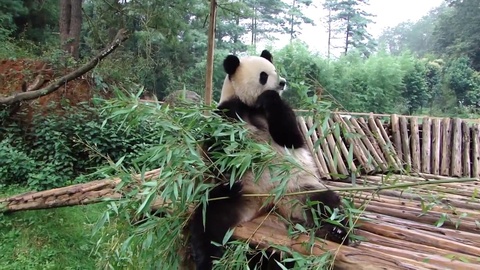}
    & \suppdataimage{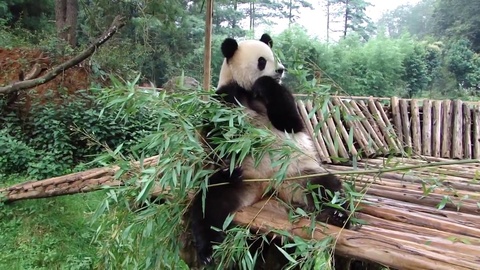}
    & \suppdataimage{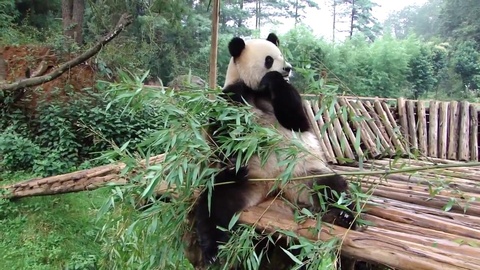} \\
    \suppdatarowlabel{Mask}
    & \suppdataimage{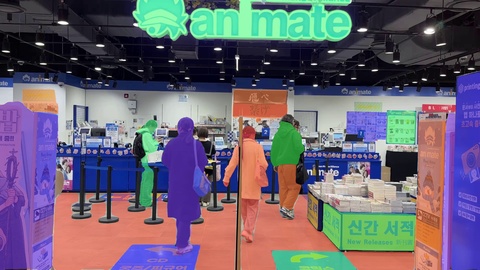}
    & \suppdataimage{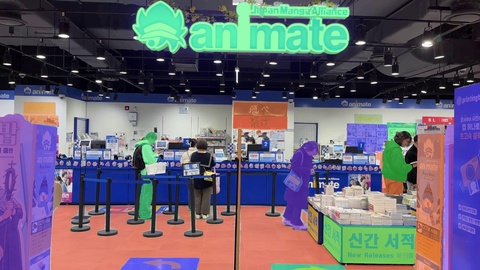}
    & \suppdataimage{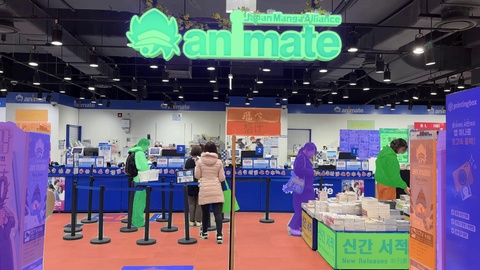}
    & \suppdataimage{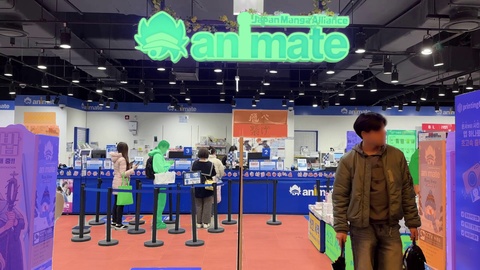}
    & \suppdataimage{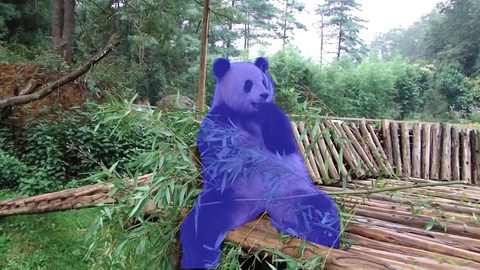}
    & \suppdataimage{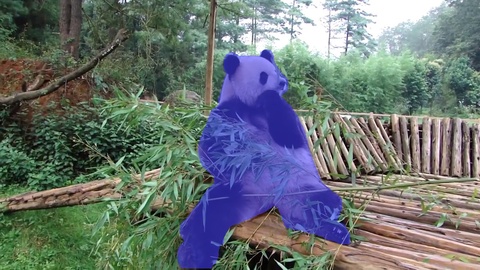}
    & \suppdataimage{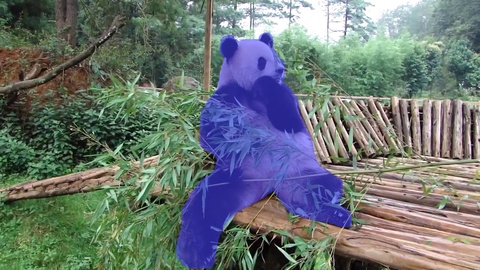}
    & \suppdataimage{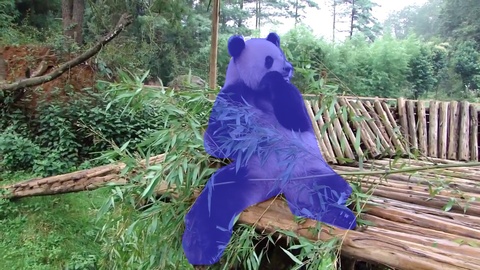} \\
    \addlinespace[0.2em]
    & \multicolumn{4}{c}{\scriptsize MOSE}
    & \multicolumn{4}{c}{\scriptsize UVO-Dense} \\
    \suppdatarowlabel{RGB}
    & \suppdataimage{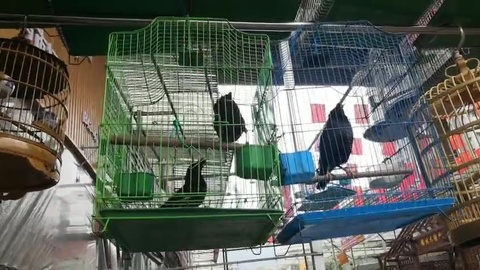}
    & \suppdataimage{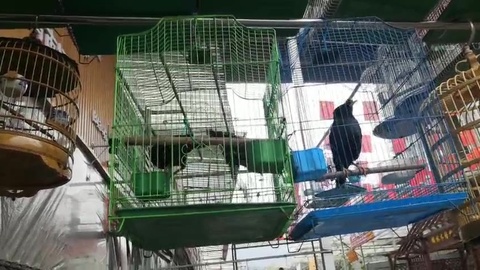}
    & \suppdataimage{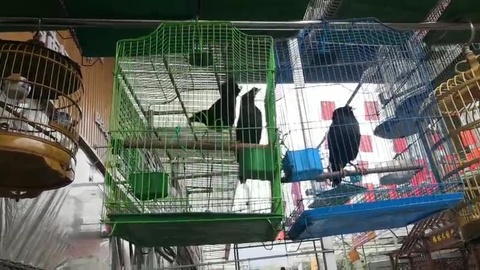}
    & \suppdataimage{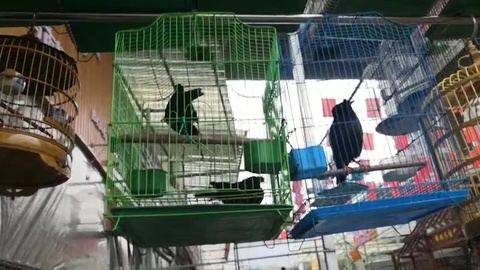}
    & \suppdataimage{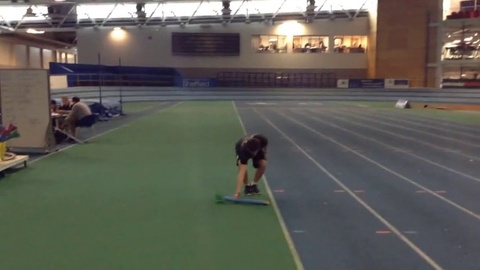}
    & \suppdataimage{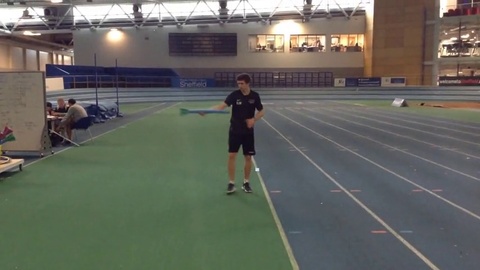}
    & \suppdataimage{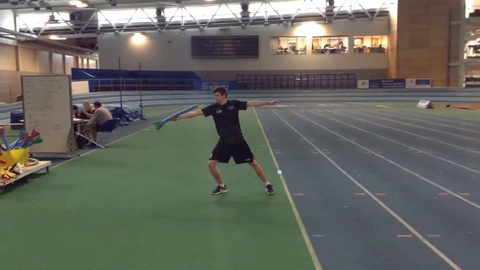}
    & \suppdataimage{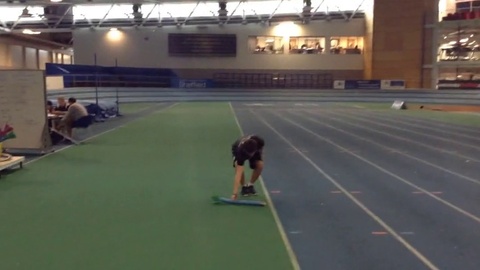} \\
    \suppdatarowlabel{Mask}
    & \suppdataimage{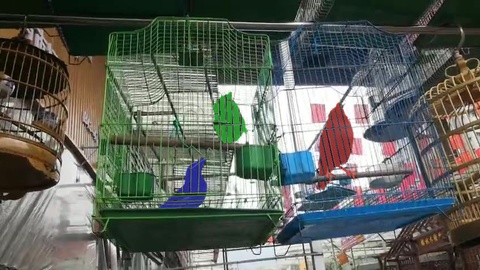}
    & \suppdataimage{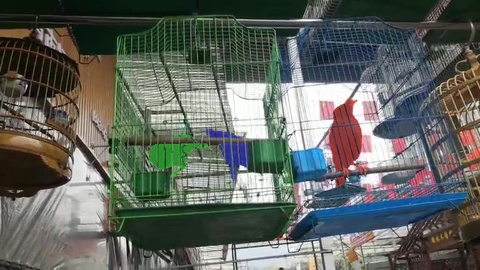}
    & \suppdataimage{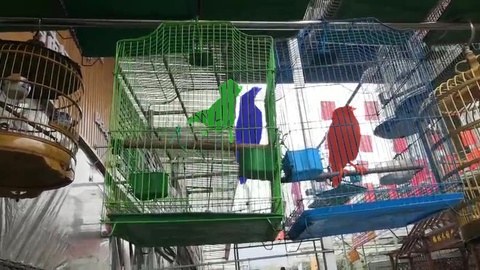}
    & \suppdataimage{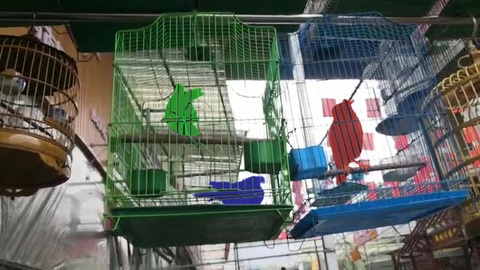}
    & \suppdataimage{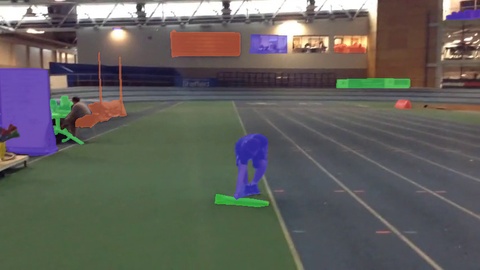}
    & \suppdataimage{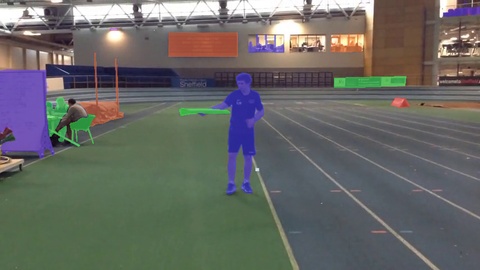}
    & \suppdataimage{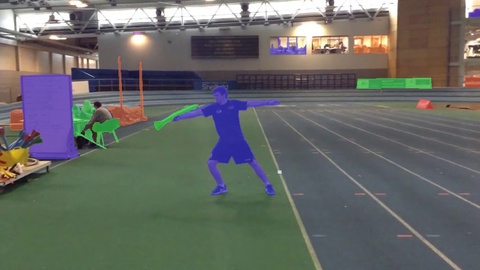}
    & \suppdataimage{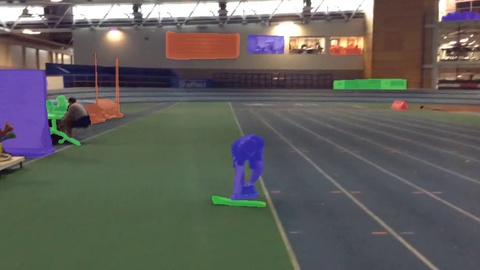} \\
    \addlinespace[0.2em]
    & \multicolumn{4}{c}{\scriptsize DynamicReplica}
    & \multicolumn{4}{c}{\scriptsize SAIL-VOS} \\
    \suppdatarowlabel{RGB}
    & \suppdataimage{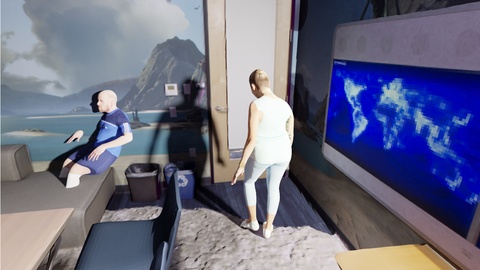}
    & \suppdataimage{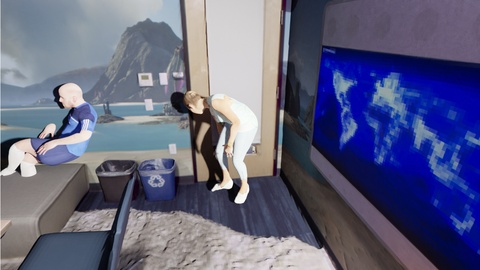}
    & \suppdataimage{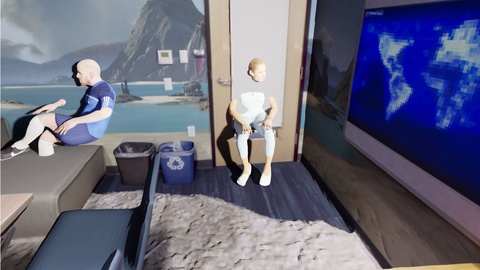}
    & \suppdataimage{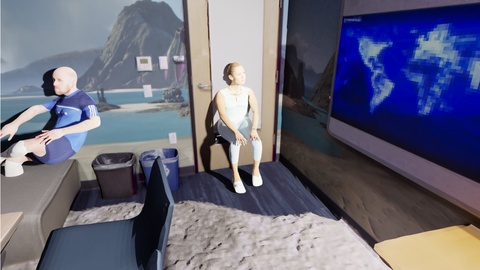}
    & \suppdataimage{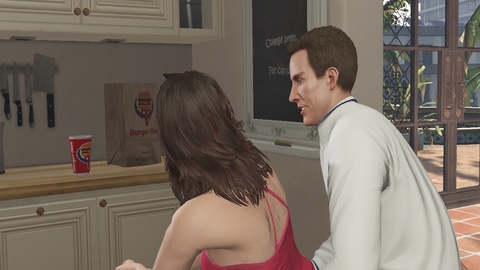}
    & \suppdataimage{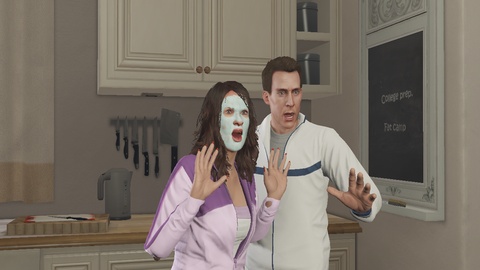}
    & \suppdataimage{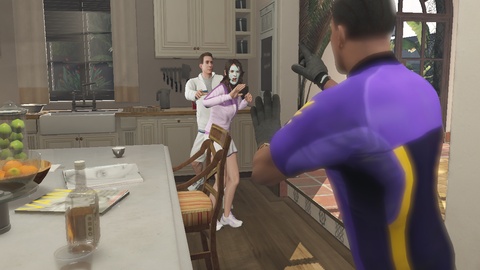}
    & \suppdataimage{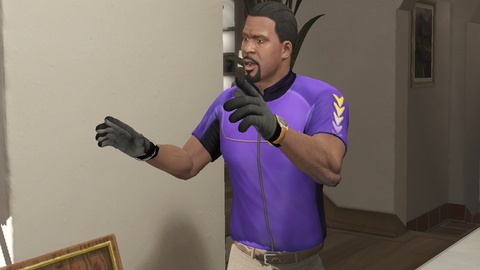} \\
    \suppdatarowlabel{Mask}
    & \suppdataimage{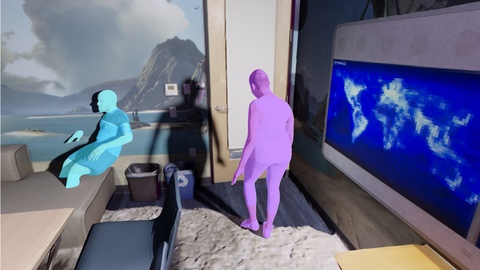}
    & \suppdataimage{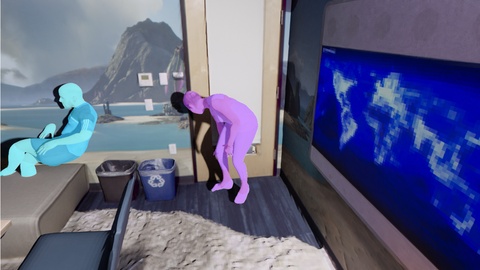}
    & \suppdataimage{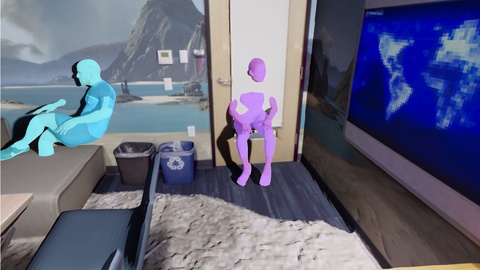}
    & \suppdataimage{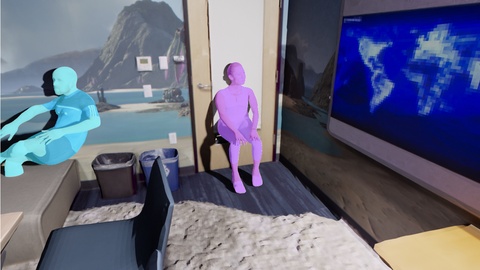}
    & \suppdataimage{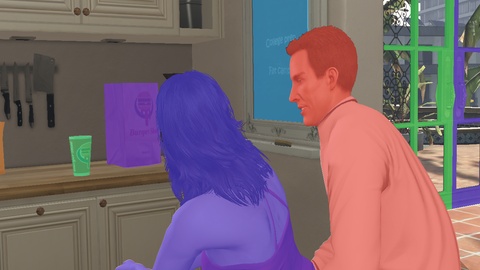}
    & \suppdataimage{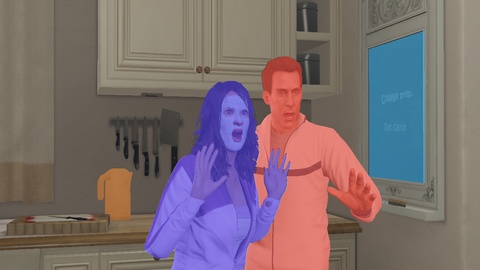}
    & \suppdataimage{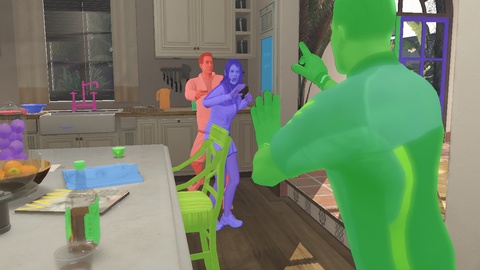}
    & \suppdataimage{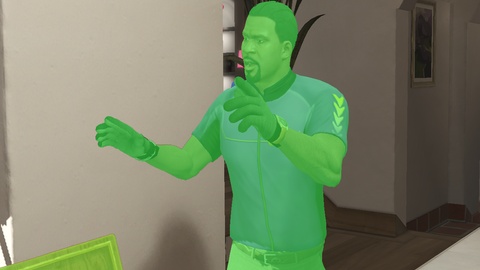} \\
    \addlinespace[0.2em]
    & \multicolumn{4}{c}{\scriptsize VOST}
    & \multicolumn{4}{c}{\scriptsize DL3DV (Pseudo)} \\
    \suppdatarowlabel{RGB}
    & \suppdataimage{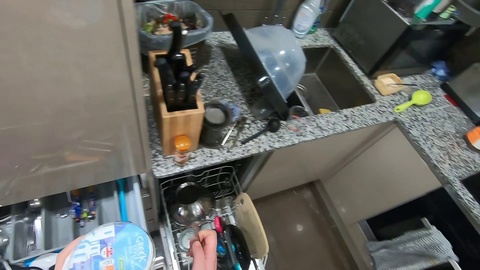}
    & \suppdataimage{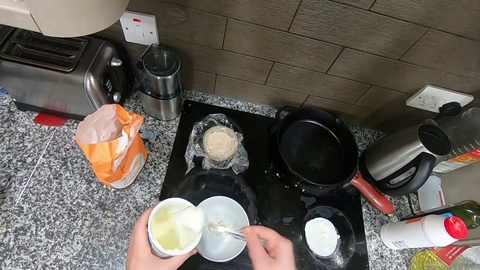}
    & \suppdataimage{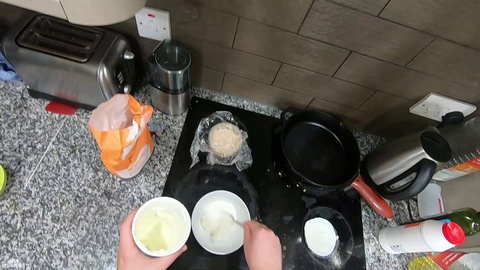}
    & \suppdataimage{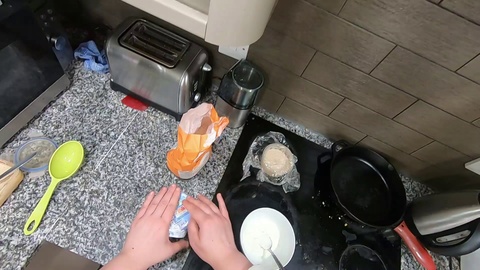}
    & \suppdataimage{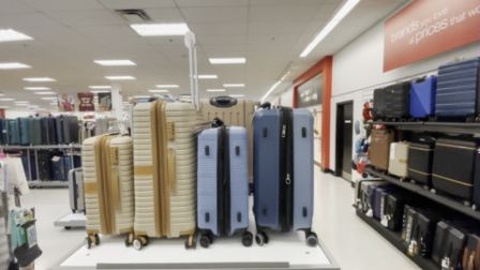}
    & \suppdataimage{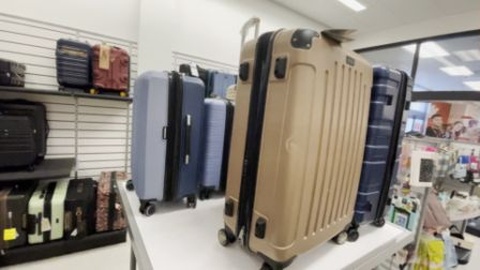}
    & \suppdataimage{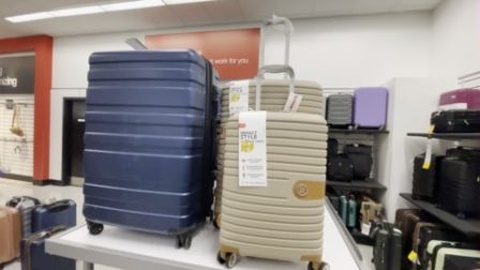}
    & \suppdataimage{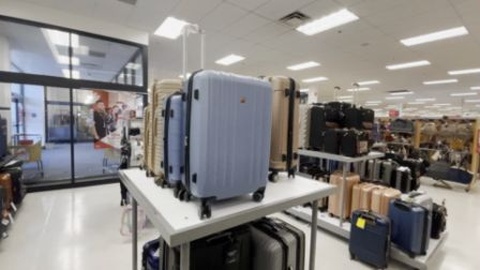} \\
    \suppdatarowlabel{Mask}
    & \suppdataimage{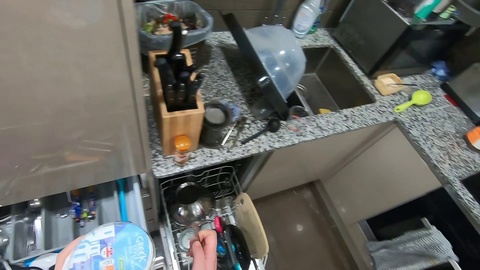}
    & \suppdataimage{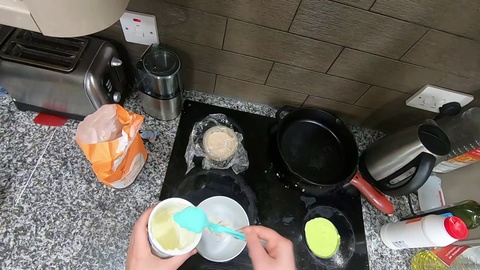}
    & \suppdataimage{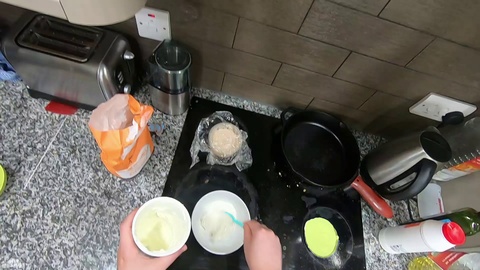}
    & \suppdataimage{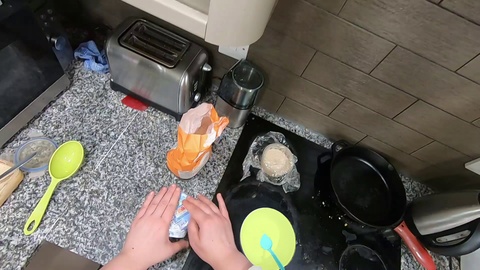}
    & \suppdataimage{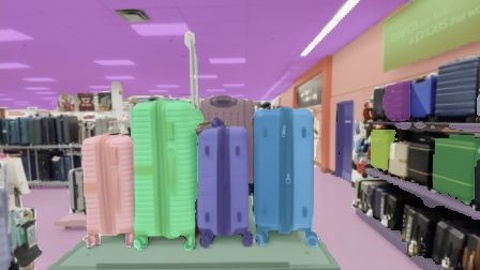}
    & \suppdataimage{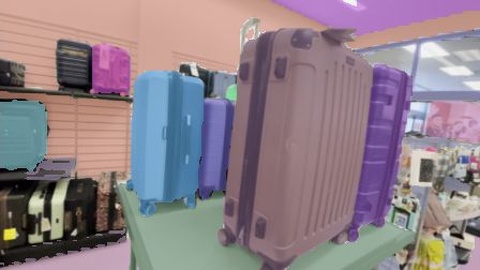}
    & \suppdataimage{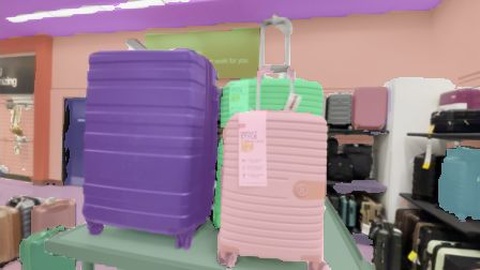}
    & \suppdataimage{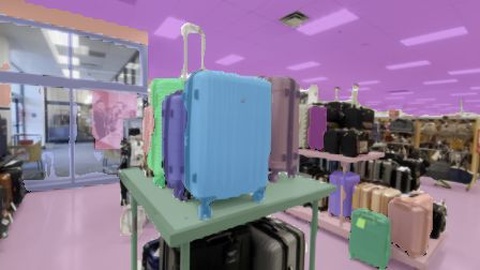} \\
  \end{tabular}
  \endgroup
  \caption{\textbf{Representative training observations.} Four RGB observations and corresponding
  masks from twelve training sources not used in our evaluation; DL3DV uses
  pseudo-masks.}
  \label{fig:supp-training-data-examples}
\end{figure}

%% file: arxiv/sections/B_experimental_details.tex
\section{Experimental Details}
\label{sec:supp-experimental-details}

\subsection{Segmenter and Prompt Encoder Training}
\label{sec:supp-segmenter-training}

\paragraph{Self-distillation with incomplete annotations.}
After Hungarian matching, exhaustive set prediction supervises every unmatched query as no-object.
Applied directly to partial annotations, this objective would suppress unlabelled instances, while
unlabelled sequences provide no positive targets. We therefore train a teacher on the exhaustive
sources and use it to densify the remaining data. Original masklets are augmented with
non-overlapping teacher predictions for partially annotated samples, whereas unlabelled samples
are supervised entirely by the teacher. The shared teacher--student architecture thus transfers
exhaustive supervision to the broader corpus through self-distillation.

Teacher predictions above $0.7$ object probability are retained, regions between $0.3$ and $0.7$
are ignored, and the no-object loss is preserved elsewhere. Human annotations take precedence;
overlapping teacher masks are discarded, and each retained pseudo-masklet is weighted by half its
teacher confidence before Hungarian matching. This three-band supervision densifies partial labels
without assigning uncertain predictions to either foreground or background.

The segmenter is optimized for $50{,}000$ steps from the same 1B visual geometry checkpoint using $8$--$16$ randomly
sampled observations at $512$-pixel resolution. AdamW uses a global batch size of $64$, weight
decay $0.05$, gradient clipping at $1.0$, and $5\%$ linear warm-up followed by cosine decay. Peak
learning rates are $2.83\times10^{-5}$ for rank-$256$ LoRA adapters and
$5.66\times10^{-4}$ for the Space-Time Query Decoder and mask heads. The number for learnable object queries is set as $400$. Object classification, BCE,
and Dice losses have weights $2$, $5$, and $5$, with a no-object weight of $0.1$. The exhaustive to
incomplete sampling ratio increases linearly from $1{:}10$ to $2{:}1$. Since incomplete samples
rely on teacher pseudo-masklets, this curriculum uses their diversity to accelerate early
convergence, then progressively favors exhaustive annotations to reduce the influence of noisy
labels. Training uses $64$ NVIDIA A100 GPUs.

\paragraph{Prompt encoder training.}
The unidirectional attention in \cref{eq:supp-unidirectional-attention} isolates the learnable
object queries from the promptable query, supporting either separate or joint training. Our
reported model freezes the backbone, LoRA adapters, Space-Time Query Decoder, and mask heads, and
optimizes only the prompt encoder for $15{,}000$ steps with AdamW on eight GPUs. The per-GPU batch
size is one, the peak learning rate is $3\times10^{-4}$, and weight decay, warm-up, and gradient
clipping are $0.05$, $5\%$, and $1.0$, respectively. The reported grounding pipeline uses the
checkpoint at step $13{,}500$.

Prompts are sampled online from annotated or pseudo-labelled masklets, with at most four target
instances per sample. We use box prompts with probability $0.4$ and point prompts otherwise. A
point prompt contains a positive point inside the target mask; for each point-prompted observation,
an additional corrective point is sampled outside it with probability $0.3$. Negative clicks thus
supplement rather than replace positive evidence, teaching the encoder to exclude surrounding
regions and nearby instances. Box prompts are not paired with negative boxes; their coordinates
are instead jittered to model localization error. Prompts span a random number of target-visible
observations. The promptable query is directly supervised by the corresponding 4D mask using BCE
and Dice losses weighted by $5$, with intermediate-layer supervision weighted by $0.4$.

Alternatively, LoRA, decoder, mask-head, and prompt-encoder parameters are jointly optimized for
$50{,}000$ steps with the segmenter settings above. We interleave exhaustive and prompt-conditioned
steps at $1{:}1$ for the first $20{,}000$ steps, $2{:}1$ for the next $20{,}000$, and $5{:}1$ for
the final $10{,}000$. Since the prompt-conditioned pool includes partially annotated and
pseudo-labelled masklets, this curriculum accelerates prompt learning early and progressively
reduces exposure to label noise toward convergence. For exhaustively annotated prompt batches, the
learnable-query set loss is retained with weight $0.5$. With probability $0.05$, up to $20\%$ of
prompted observations are corrupted: a point is moved to a random location, while a box is replaced
by one from another instance or by a random box. This augmentation directly trains the relevance
weights in \cref{eq:supp-prompt-aggregate} to suppress inconsistent evidence across observations.
The separate strategy applies an analogous noisy-to-clean curriculum during segmenter
self-distillation; its subsequent frozen-segmenter prompt stage contains only prompt-conditioned
updates. We systematically evaluate separate and joint training across all benchmarks and observe
only marginal differences. All results reported in this paper therefore use the separately trained
segmenter and Prompt Encoder checkpoints.

\subsection{Grounder and Critic Training}
\label{sec:supp-grounder-training}

\paragraph{Grounder.}
We use a Grounder window size of $N_{\mathrm{w}}=8$ during both training and inference. During
training, the observations are sampled from a span of at most $32$: $80\%$ of the windows are
centred on a target-visible observation and balance target-present and target-absent observations,
while the remainder are sampled without a presence constraint to include negative windows. We fully fine-tune Qwen3-VL-4B for 4 epochs, corresponding to approximately $5{,}000$ optimizer
updates on eight GPUs. AdamW uses an effective batch size of $64$, weight decay $0.01$, and peak
learning rates of $8\times10^{-6}$ for the VLM and $10^{-4}$ for the newly initialized projection
and prediction heads. We apply $300$ warm-up steps followed by cosine decay, bfloat16 training, and
gradient clipping at $1.0$. The query-classification and autoregressive language-modeling losses in
\cref{eq:grounding-objectives} are equally weighted.

\paragraph{Critic.}
The $4{,}005$ candidate pairs constructed in \cref{sec:supp-grounding-details} are shuffled during
training, and their presentation order is randomized on every visit to remove branch-position
bias. Near-tie examples constitute $30\%$ of the training pairs, exposing the Critic to both clear
and ambiguous comparisons. We fully fine-tune Qwen3-VL-2B with its scalar readout for three epochs
on four GPUs. AdamW uses an effective batch size of
$32$, a learning rate of $5\times10^{-6}$, weight decay $0.01$, bfloat16 precision, and gradient
clipping at $1.0$.

\subsection{Baseline Configurations}
\label{sec:supp-inference-details}

\paragraph{Class-agnostic segmentation baselines.}
We compare lifted 3D segmentation, video segmentation, and feed-forward visual-geometry methods,
using their official checkpoints on the same observations without semantic labels, target prompts,
or first-frame ground truth. For SAM~2, the Automatic Mask Generator uses a
$16\times16$ point grid to discover instances, whose masks initialize the video predictor. The
predictor propagates their identities, while periodic automatic mask generation adds masks with
less than $0.5$ IoU to every active track as new instances. Re-discovery is performed every 2
observations for the short ScanNet clips and every $15$ frames for the video benchmarks. When
full-sequence inference exceeds GPU memory, we retain all observations through memory-bounded
streaming-window inference. In particular, our IGGT implementation appends fixed scene-wide
anchor views to each sliding window, concatenates the resulting dense instance features, and applies
clustering over the complete sequence to maintain identities across windows. IGGT4D uses its
released offline full-attention configuration. SAM-V uses the released checkpoint and
automatic-mask settings; sequences longer than $32$ observations are processed in overlapping
$32$-observation windows, with identities matched by mask IoU on the shared observation.

\paragraph{Language-grounding baselines.}
VLM-Grounder uses Qwen3-VL-30B-A3B-Thinking with YOLO-World and a single-view ensemble; for videos,
it runs its 3D pipeline on the reconstructed geometry and reprojects the selected box to each
observation. VLM$+$SAM~2 uses Qwen3-VL-30B-A3B-Instruct, and Sa2VA uses its released 8B model.
SAM~3 and MVGGT use their official
end-to-end checkpoints without a separately served VLM: SAM~3 receives category prompts on static
scenes and full expressions on videos, while MVGGT receives eight uniformly sampled observations.
Z3D~\citep{drozdov2026z3d} uses
Qwen3-VL-30B-A3B-Thinking within a 3D-proposal pipeline combining MaskClustering, CLIP retrieval,
VLM filtering, SAM~3 segmentation, and 3D proposal voting. To preserve the RGB-only input setting,
its proposals are constructed from the
feed-forward geometry reconstructed for each scene rather than from ground-truth point clouds; on
videos, the selected 3D proposal is projected back to the observations for evaluation. For static
scenes, VLM$+$SAM~2 ranks the observations with CLIP and predicts a normalized box on the selected
view. For the video results in \cref{tab:grounding}, the VLM instead receives eight uniformly sampled
observations jointly and predicts one box or absence per observation; these boxes initialize SAM~2
for bidirectional propagation over the complete sequence.

\subsection{Benchmarks and Metrics}
\label{sec:supp-evaluation-protocols}

\paragraph{Benchmarks.}
For class-agnostic segmentation, we use the ScanNet setting from IGGT~\citep{li2025iggt},
DAVIS~2017, LVOS, and VIPSeg, comprising ten static multi-view clips, $30$ short videos, $50$
long-term videos, and $343$ panoptic videos, respectively. Following~\citep{li2025iggt,zou2026iggt4d}, each ScanNet clip
contains 8 sampled observations, while each LVOS video is uniformly sampled to $100$
observations. For the separate robustness evaluation, each video produces three \emph{sparse}
subsets by sampling $16$ observations without replacement from its first $32$ observations, last
$32$ observations, and full temporal extent, respectively; the sampled observations retain their
chronological order, and each subset contains annotated foreground. The \emph{shuffle} setting
randomly permutes exactly the same observations and their annotations. For the LVOS \emph{hard}
setting, farthest-point sampling over normalized RGB thumbnail descriptors selects $16$
appearance-diverse observations across the full video; at least four contain annotated foreground
when available, after which their order is shuffled. Additional evaluations further
cover two zero-shot egocentric benchmarks: $48$ HOI4D~\citep{liu2022hoi4d} clips and $42$
EPIC-KITCHENS VISOR~\citep{darkhalil2022epic} clips, as well as 9
KITTI-STEP~\citep{weber2021step} sequences for driving
scenes. For language
grounding, we sample $700$ ScanRefer queries from $121$ validation scenes and hold out $80$
ScanNet++ scenes with $3{,}320$ expressions. The video evaluation contains all $122$ Ref-DAVIS
expressions and $616$ single-referent MeViS expressions from $47$ videos.

\paragraph{Metrics.}
For class-agnostic segmentation, T-mIoU averages over ground-truth tracks the intersection-over-union
accumulated across observations, while T-SR measures the fraction whose IoU exceeds $0.5$ in every
observation where they are present. We report the standard region similarity $\mathcal{J}$, boundary accuracy $\mathcal{F}$,
and their mean \JF{}~\citep{perazzi2016benchmark,pont20172017} on DAVIS and LVOS, and STQ and
class-agnostic thing mAP on VIPSeg. Because the video benchmarks provide only per-frame 2D masks,
both annotated and predicted masks are lifted with the same depth and cameras predicted by the
visual geometry backbone~\citep{wang2026vggt}. The resulting point sets are evaluated with the strict
4D metrics in \cref{eq:4d-metrics}, using a separate temporal voxel slice for every observation and
one-to-one instance matching.
Static grounding is evaluated by 3D mask mIoU and the proportions of predicted masks and axis-aligned
boxes exceeding $0.25$ IoU; Ref-DAVIS and MeViS use \JF{}.

%% file: arxiv/sections/C_additional_results.tex
\section{Additional Results and Discussion}
\label{sec:supp-additional-results}

\subsection{Efficiency and Additional Ablations}

\paragraph{Segmentation efficiency.}
\Cref{tab:runtime} reports end-to-end seconds per clip. Joint feed-forward processing avoids repeated
per-frame detection and propagation, so the efficiency advantage grows with the observation count.
Direct set prediction also avoids the post-hoc clustering used by IGGT. \methodname{} processes an
eight-observation ScanNet scene in 0.5 seconds and a 100-observation clip in about 10 seconds on an
A100-80GB. \Cref{tab:runtime-gb200} shows that a GB200 processes four 100-frame clips in parallel at
42.11 frames/s.

\input{arxiv/tables/tab_runtime}

\paragraph{Segmenter ablations.}
\Cref{tab:supp-segmenter-ablations} reports class-agnostic instance AP on
IGGT-Benchmark~\citep{li2025iggt}; mAP averages AP over IoU thresholds from 0.50 to 0.95, while
AP$_{50}$ and AP$_{75}$ use fixed thresholds. LoRA matches or improves full fine-tuning, with the
largest gain at AP$_{75}$. Removing frame positional embeddings also preserves segmentation quality
relative to learned embeddings, whereas sinusoidal embeddings degrade all three metrics. These
results support parameter-efficient backbone adaptation and a decoder without frame-order priors.

\input{arxiv/tables/tab_segmenter_ablations}

\subsection{Additional Qualitative Results}
\label{sec:supp-additional-qualitative}

\paragraph{Frame-space and 4D-space results.}
Feed-forward 3D semantic segmentation has recently been extended to panoramic
inputs~\citep{yoon2026panoseg3r}. Following the panorama-to-perspective processing of
MTPano~\citep{zhang2026mtpano}, we decompose each panorama into overlapping perspective patches.

\input{arxiv/figures/pointcloud_panorama_remaining}
\FloatBarrier

\Cref{fig:supp-panorama-pointcloud} and \cref{fig:supp-static-pointcloud} provide further indoor results on
Structured3D~\citep{zheng2020structured3d} and ARKitScenes~\citep{baruch2021arkitscenes}, neither of
which is in the training corpus. The dynamic and driving examples in
\cref{fig:supp-dynamic-pointcloud,fig:supp-driving-pointcloud} show persistent instance identities
under camera and object motion. These visualizations assess the coherence of predicted identities in
the reconstructed scene rather than independent geometric accuracy.

\input{arxiv/figures/pointcloud_driving_remaining}
\FloatBarrier

\paragraph{More segmentation comparisons.}
\Cref{fig:supp-segmentation-qualitative} adds comparisons on static multi-view scenes and dynamic
videos. Across these cases, \methodname{} retains instance identities through viewpoint changes,
occlusion, rapid motion, and deformation.

\paragraph{More grounding comparisons.}
The results in \cref{fig:supp-grounding-qualitative-1,fig:supp-grounding-qualitative-2} include
references that require spatial reasoning rather than appearance matching alone. For example,
\cref{fig:supp-grounding-qualitative-1}(c) contains several similar laptops; the relation ``across
from an armchair'' identifies the intended instance consistently across views.

\paragraph{In-the-wild evaluation.}
\Cref{fig:supp-inthewild-qualitative} extends the main in-the-wild results to road, downtown, and
cycling scenes. Despite changes in viewpoint, scale, motion, and scene content, \methodname{}
maintains coherent identities across observations. For visualization, we follow
\citet{he2012guided} and apply guided image filtering to the predicted mask boundaries using the
corresponding RGB observations.

\input{arxiv/figures/additional_segmentation_combined}
\input{figures_final/pointcloud_static}
\input{figures_final/pointcloud_dynamic}
\input{arxiv/figures/additional_grounding_full_1}
\input{arxiv/figures/additional_grounding_full_2}
\input{arxiv/figures/inthewild_appendix}

%% file: arxiv/tables/tab_runtime.tex
\begin{table}[htbp]
\centering
\begin{minipage}[b]{0.64\linewidth}
\centering
\caption{\textbf{Segmentation efficiency.} Average wall-clock seconds per clip on a sampled
benchmark subset using an A100-80GB. Parentheses report the mean number of observations in the
sampled clips. Lower is better.}
\label{tab:runtime}
\scriptsize
\setlength{\tabcolsep}{1.7pt}
\renewcommand{\arraystretch}{1.05}
\resizebox{\linewidth}{!}{%
\begin{tabular}{@{}lccccc@{}}
\toprule
Method & ScanNet (8) & VIPSeg (41) & DAVIS (71) & LVOS (100) & HOI4D (100) \\
\midrule
DEVA~\citep{cheng2023tracking}       & 3.0 & 9.3 & 32.3 & 32.3 & 26.7 \\
GLEE~\citep{wu2024general}           & 2.5 & 13.8 & 24.6 & 36.8 & 35.9 \\
OMG-Seg~\citep{li2024omg}            & 0.9 & 5.4 & 10.1 & 13.7 & 14.6 \\
SAM~2~\citep{ravi2025sam}            & 3.4 & 46 & 183 & 175 & 168 \\
PanSt3R~\citep{zust2025panst3r}      & 3.3 & 18.3 & 35.3 & 54.0 & 50.7 \\
IGGT~\citep{li2025iggt}              & 13.3 & 83 & 162 & 272 & 230 \\
IGGT4D~\citep{zou2026iggt4d}         & 2.6 & 6.9 & 14.4 & 18.5 & 18.0 \\
\midrule
\textbf{\methodname{} (Ours)}        & \textbf{0.5} & \textbf{3.0} & \textbf{6.6} & \textbf{10.1} & \textbf{10.2} \\
\bottomrule
\end{tabular}
}
\end{minipage}\hfill
\begin{minipage}[b]{0.33\linewidth}
\centering
\caption{\textbf{Hardware scaling.} Best-tested model-forward throughput at
$384\!\times\!688$. Parentheses report batch size $B$.}
\label{tab:runtime-gb200}
\scriptsize
\setlength{\tabcolsep}{2.2pt}
\renewcommand{\arraystretch}{1.05}
\begin{tabular*}{\linewidth}{@{\extracolsep{\fill}}rccc@{}}
\toprule
Frames & A100 & GB200 & Ratio \\
\midrule
1   & 25.85 (32) & 63.15 (32) & $2.44\times$ \\
4   & 25.44 (32) & 65.19 (32) & $2.56\times$ \\
8   & 24.37 (16) & 64.40 (32) & $2.64\times$ \\
16  & 22.28 (8)  & 61.61 (16) & $2.77\times$ \\
32  & 18.98 (4)  & 56.69 (8)  & $2.99\times$ \\
64  & 14.72 (2)  & 48.64 (4)  & $3.30\times$ \\
100 & 11.64 (1)  & 42.11 (4)  & $3.62\times$ \\
\bottomrule
\end{tabular*}
\end{minipage}
\end{table}

%% file: arxiv/tables/tab_segmenter_ablations.tex
\begin{table}[t]
\centering
\caption{\textbf{Segmenter ablations} on IGGT-Benchmark~\citep{li2025iggt}.
(a) Backbone adaptation. (b) Frame positional encoding in the decoder.}
\label{tab:supp-segmenter-ablations}
\renewcommand{\arraystretch}{1.05}
\begin{minipage}[t]{0.49\linewidth}
\centering
\scalebox{0.75}{%
\begin{tabular*}{1.32\linewidth}{@{\extracolsep{\fill}}lrrr@{}}
\toprule
\multicolumn{4}{c}{(a) Backbone adaptation} \\
\cmidrule(lr){1-4}
Strategy & mAP & AP$_{50}$ & AP$_{75}$ \\
\midrule
Full fine-tuning & 0.200 & 0.286 & 0.233 \\
LoRA             & \textbf{0.210} & \textbf{0.287} & \textbf{0.260} \\
\bottomrule
\end{tabular*}%
}
\end{minipage}\hfill
\begin{minipage}[t]{0.49\linewidth}
\centering
\scalebox{0.75}{%
\begin{tabular*}{1.32\linewidth}{@{\extracolsep{\fill}}lrrr@{}}
\toprule
\multicolumn{4}{c}{(b) Frame positional encoding} \\
\cmidrule(lr){1-4}
Encoding & mAP & AP$_{50}$ & AP$_{75}$ \\
\midrule
Learned    & 0.208 & 0.288 & \textbf{0.252} \\
Sinusoidal & 0.186 & 0.262 & 0.220 \\
None       & \textbf{0.212} & \textbf{0.294} & 0.251 \\
\bottomrule
\end{tabular*}%
}
\end{minipage}
\end{table}

%% file: arxiv/figures/pointcloud_panorama_remaining.tex
\begin{figure}[!t]
  \centering
  \setlength{\tabcolsep}{1.2pt}
  \renewcommand{\arraystretch}{0.98}
  \newcommand{\pcpanoremrowlabel}[1]{\raisebox{-.5\height}{\rotatebox[origin=c]{90}{\scriptsize #1}}}
  \newcommand{\pcpanoremimage}[1]{\raisebox{-.5\height}{\includegraphics[width=0.285\textwidth]{#1}}}
  \begin{tabular}{@{}>{\centering\arraybackslash}m{4mm}ccc@{}}
    \multicolumn{4}{c}{\includegraphics[width=0.92\textwidth]{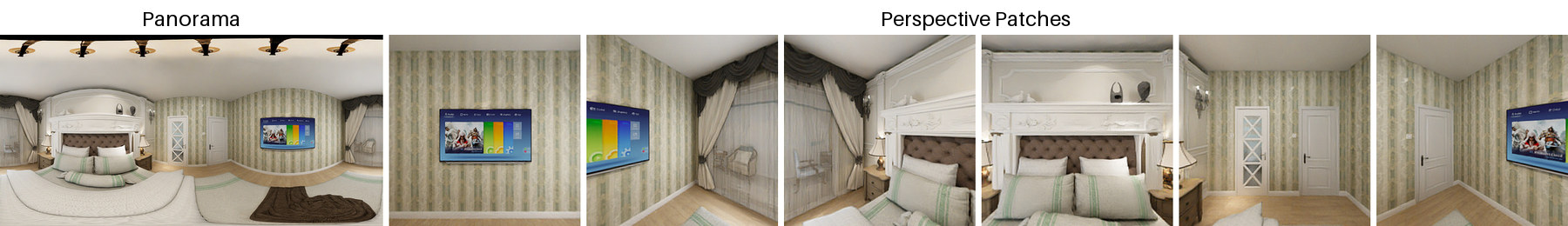}} \\[-0.2mm]
    & \scriptsize IGGT & \scriptsize IGGT4D & \scriptsize\textbf{Ours} \\
    \pcpanoremrowlabel{RGB} &
      \pcpanoremimage{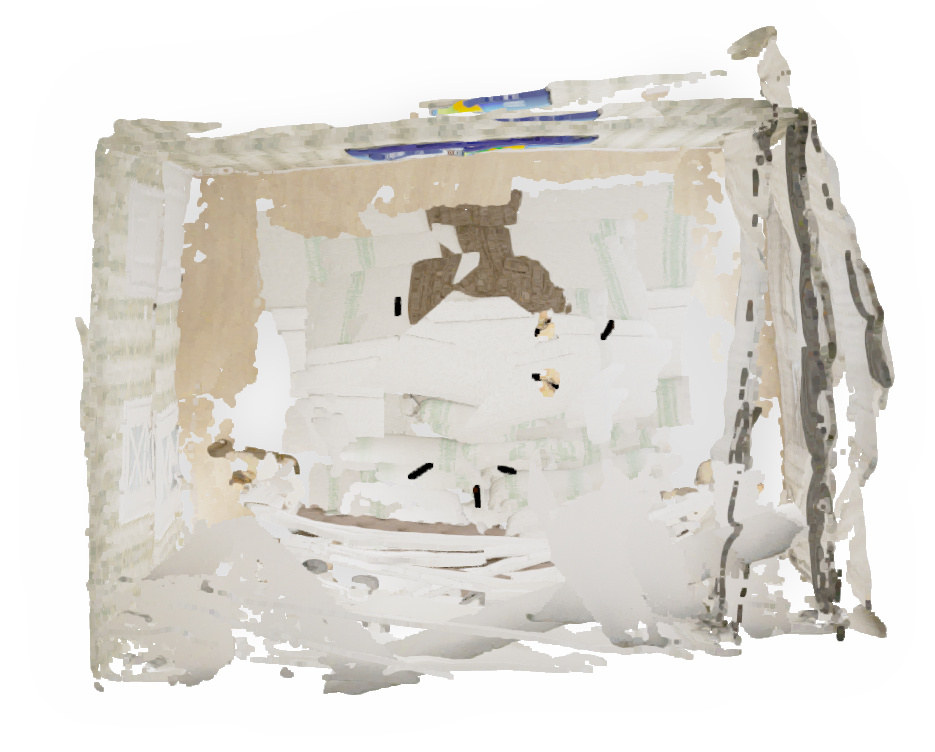} &
      \pcpanoremimage{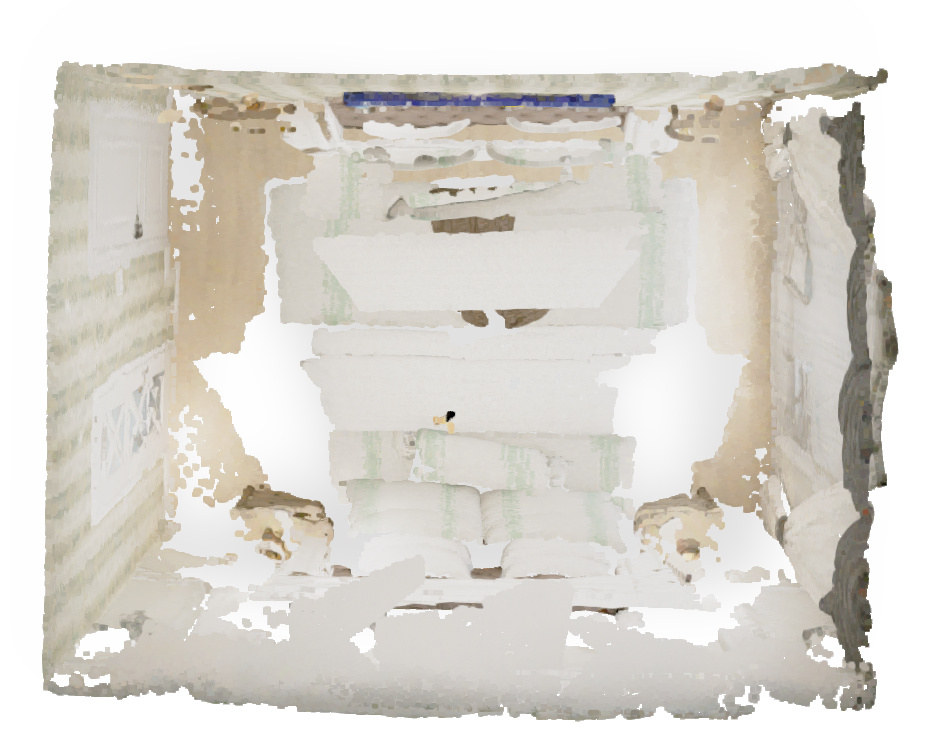} &
      \pcpanoremimage{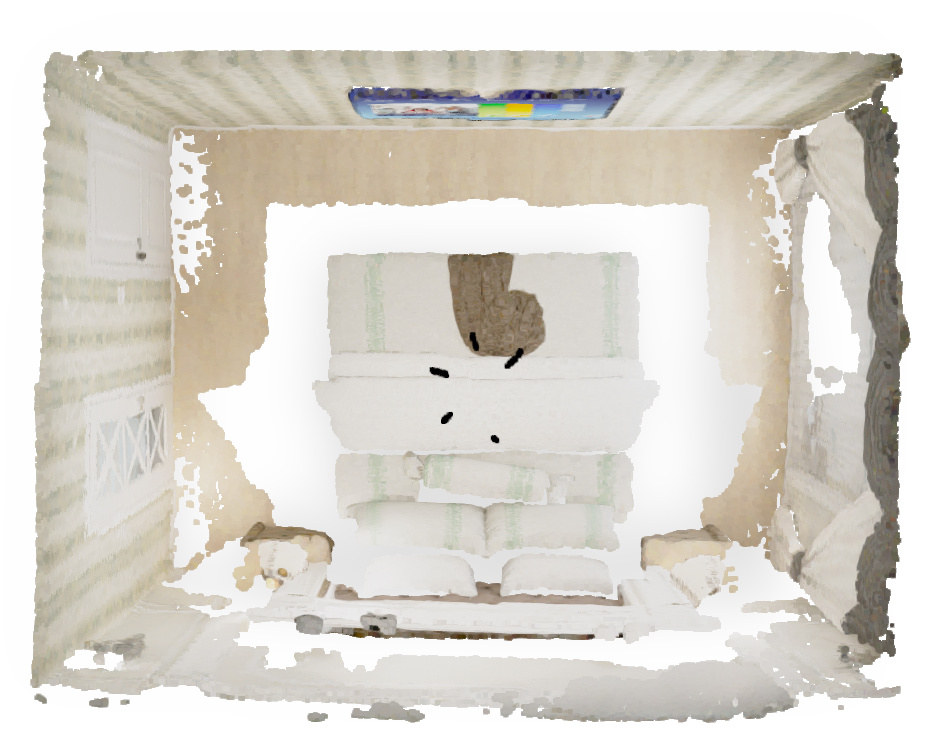} \\[-0.5mm]
    \pcpanoremrowlabel{Mask} &
      \pcpanoremimage{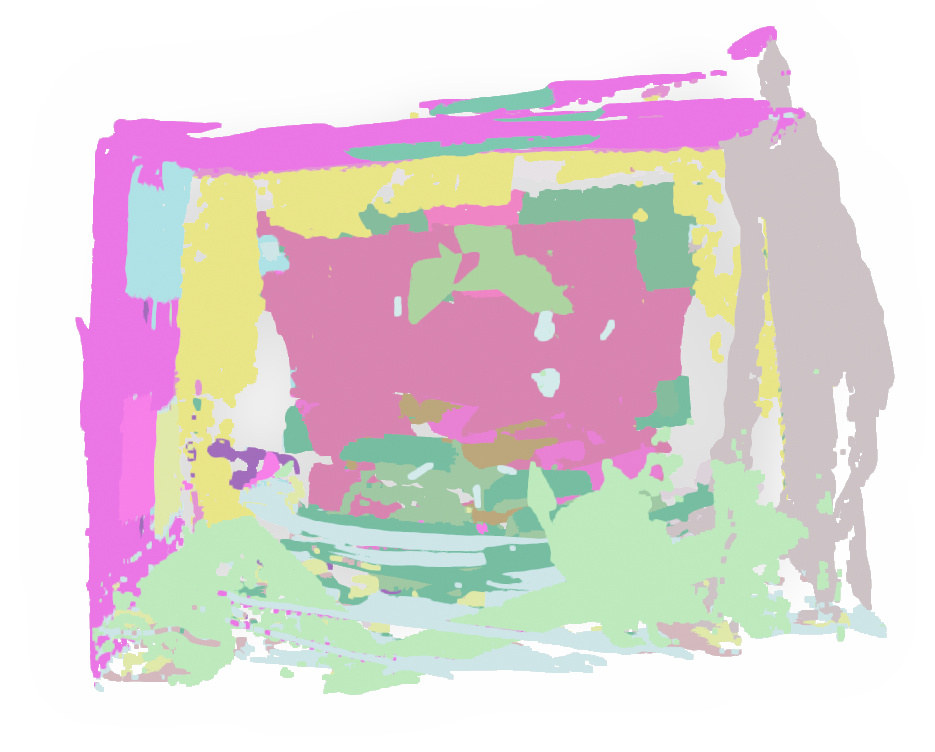} &
      \pcpanoremimage{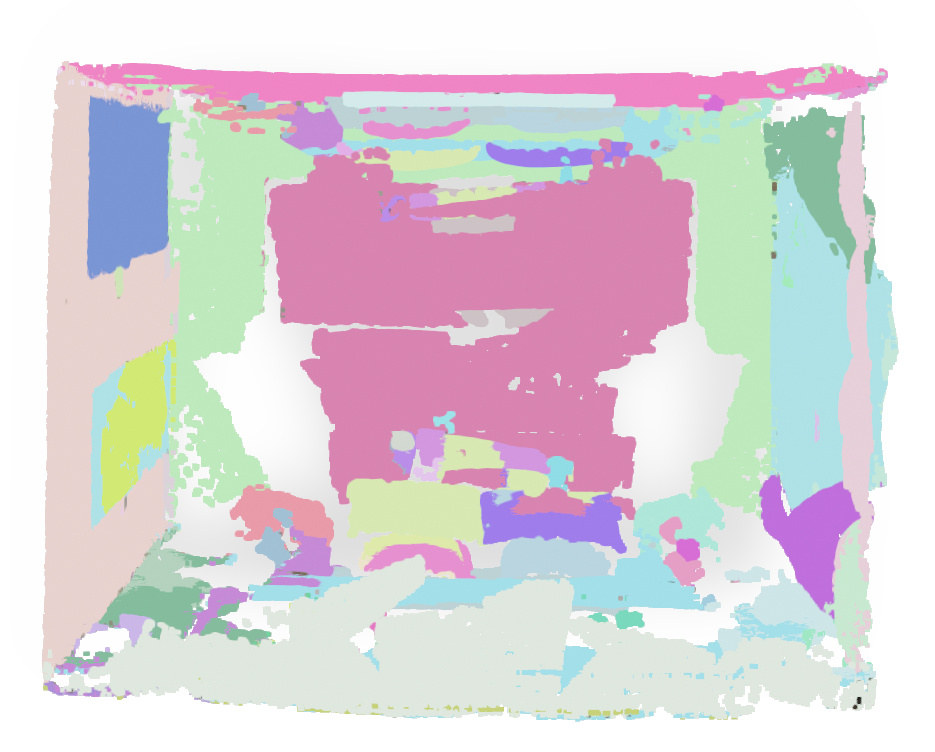} &
      \pcpanoremimage{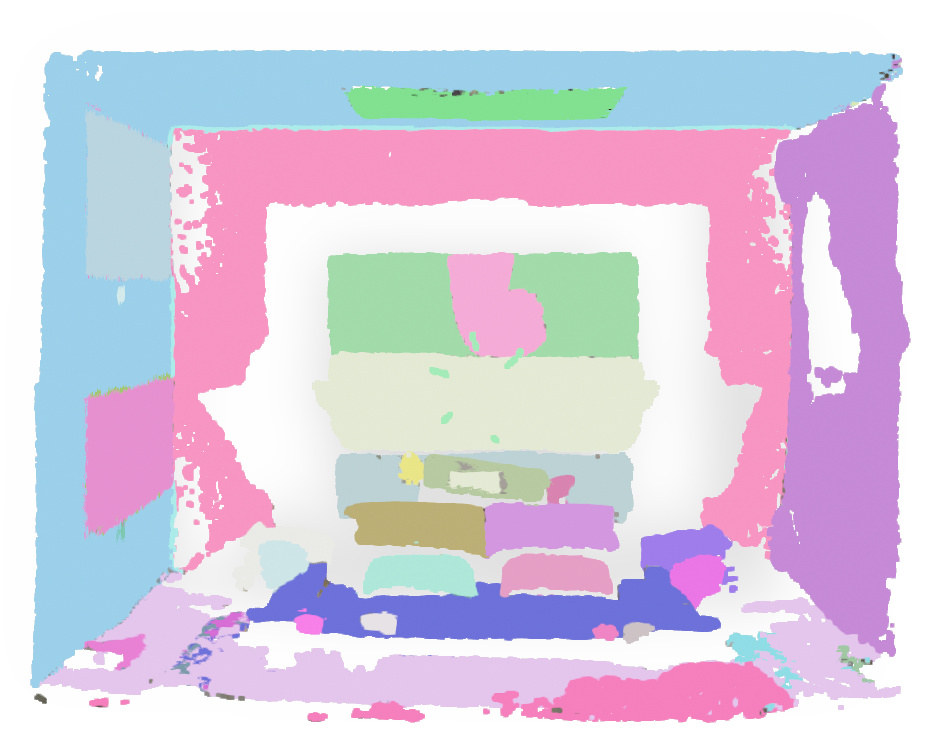}
  \end{tabular}
  \caption{\textbf{Additional zero-shot panoramic segmentation.}
  A Structured3D panorama~\citep{zheng2020structured3d} is decomposed into overlapping perspective
  patches; representative patches are shown above aligned RGB and instance-colored point clouds.}
  \label{fig:supp-panorama-pointcloud}
\end{figure}

%% file: arxiv/figures/pointcloud_driving_remaining.tex
\begin{figure}[!t]
  \centering
  \setlength{\tabcolsep}{1.2pt}
  \renewcommand{\arraystretch}{0.96}
  \newcommand{\pcdriveremrowlabel}[1]{\raisebox{-.5\height}{\rotatebox[origin=c]{90}{\scriptsize #1}}}
  \newcommand{\pcdriveremimage}[1]{\raisebox{-.5\height}{\includegraphics[width=0.40\textwidth]{#1}}}
  \begin{tabular}{@{}>{\centering\arraybackslash}m{4mm}cc@{}}
    \multicolumn{3}{c}{\includegraphics[width=0.92\textwidth]{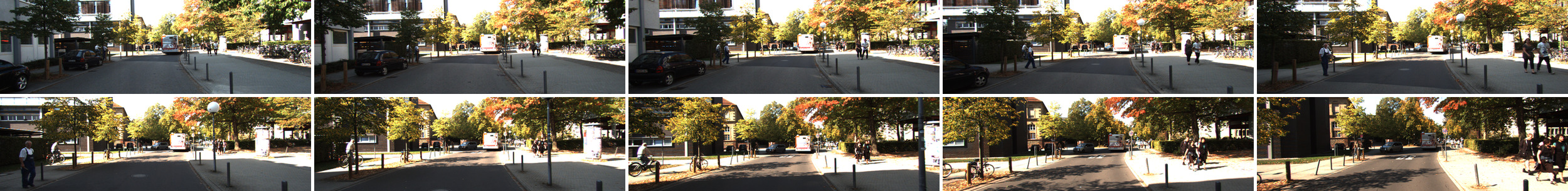}} \\[-0.2mm]
    & \scriptsize IGGT4D & \scriptsize\textbf{Ours} \\
    \pcdriveremrowlabel{RGB} &
      \pcdriveremimage{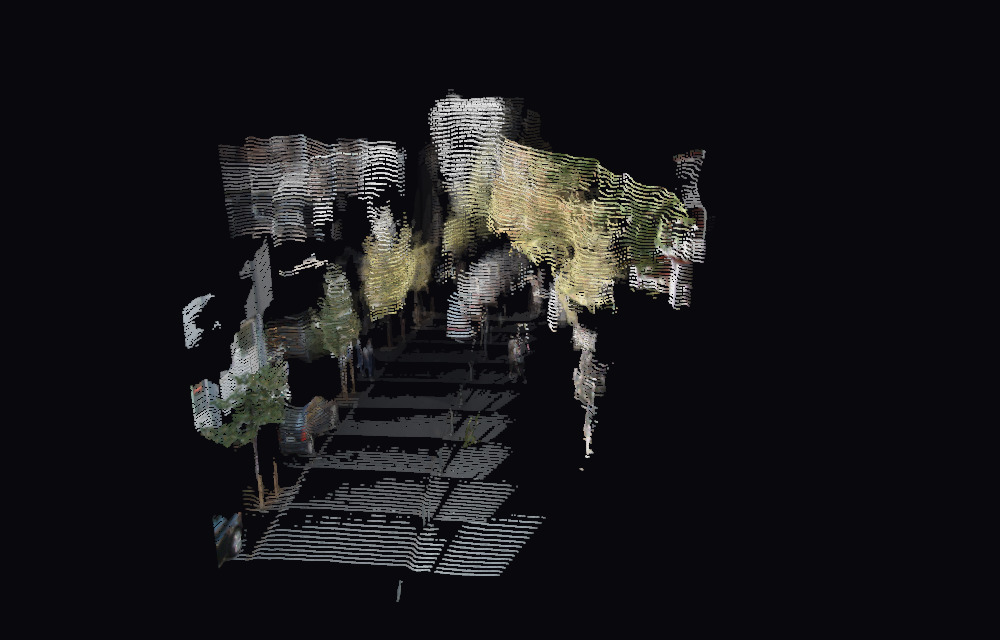} &
      \pcdriveremimage{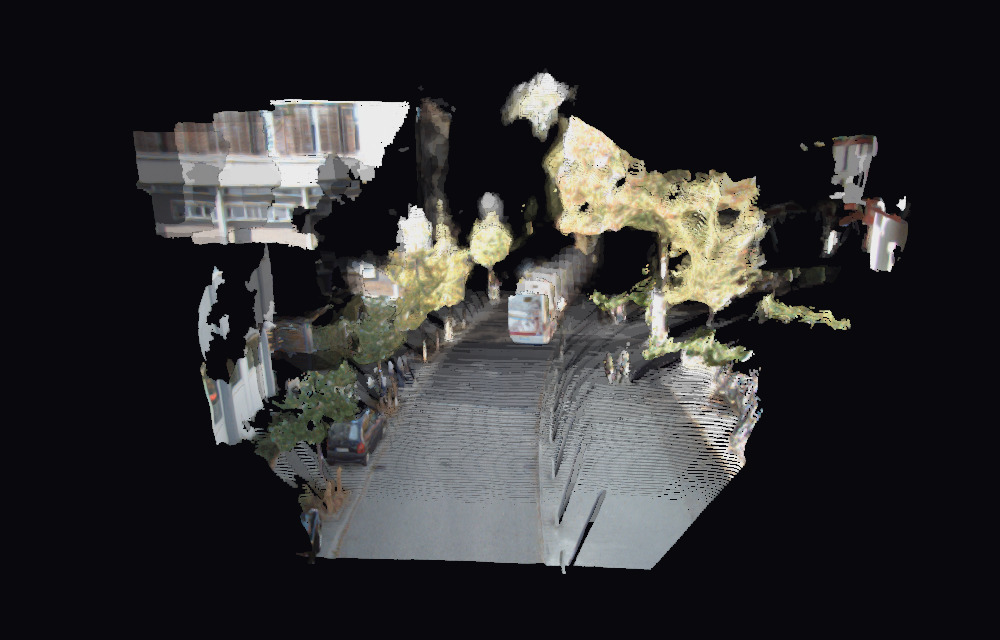} \\[-0.5mm]
    \pcdriveremrowlabel{Mask} &
      \pcdriveremimage{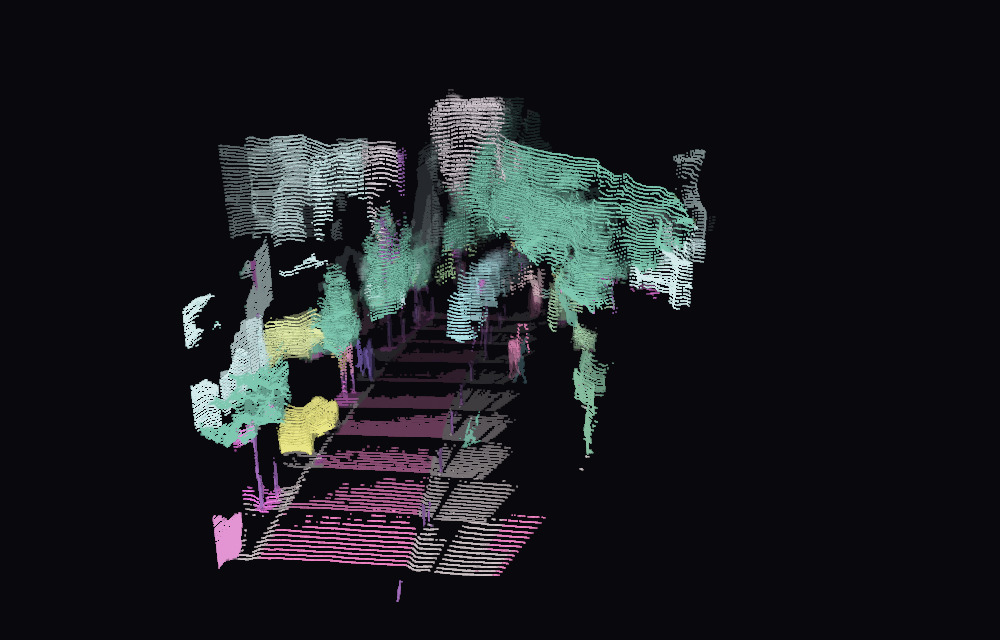} &
      \pcdriveremimage{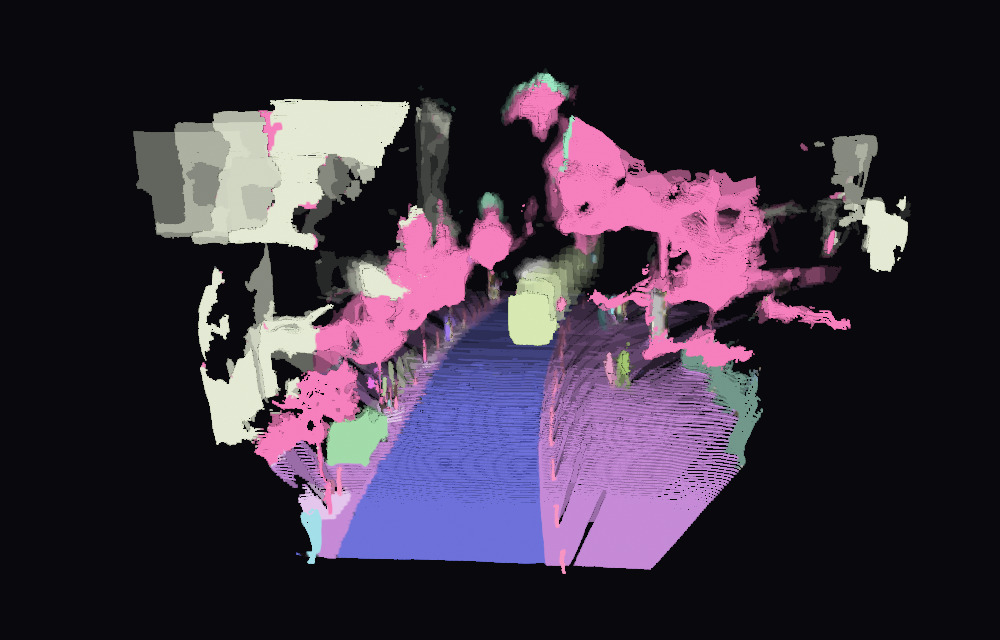}
  \end{tabular}
  \caption{\textbf{Additional 4D reconstruction and instance segmentation in a driving scene.}
  Selected observations are followed by aligned RGB and instance-colored point-cloud views.}
  \label{fig:supp-driving-pointcloud}
\end{figure}

%% file: arxiv/figures/additional_segmentation_combined.tex
\begin{figure}[p]
  \centering
  \setlength{\tabcolsep}{0pt}
  \renewcommand{\arraystretch}{1.02}
  \newcommand{\suppqualrowlabel}[1]{%
    \raisebox{\dimexpr4.7mm-.5\height+.5\depth\relax}{%
      \rotatebox[origin=c]{90}{\fontsize{5.3}{5.7}\selectfont #1}}}
  \newcommand{\suppqualimage}[1]{\includegraphics[width=0.435\textwidth]{#1}}
  \begin{tabular}{r@{\hspace{0.9mm}}c@{\hspace{2.0mm}}c}
    & {\fontsize{6.2}{6.6}\selectfont\textbf{(a)}} &
      {\fontsize{6.2}{6.6}\selectfont\textbf{(b)}} \\
    \suppqualrowlabel{RGB} & \suppqualimage{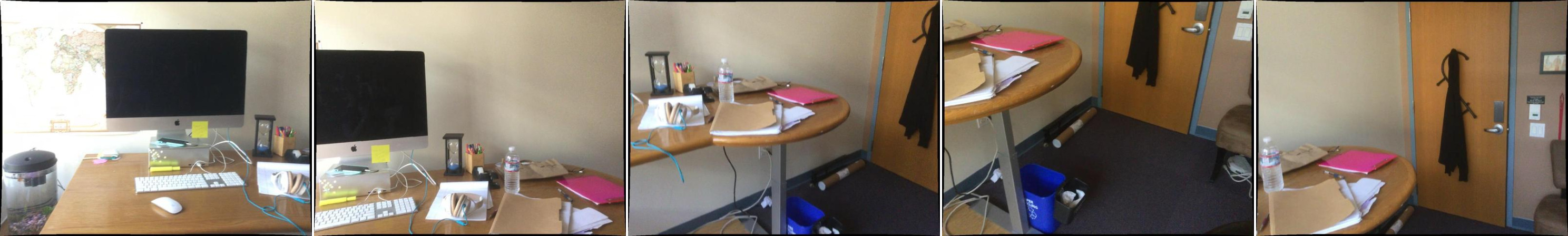} & \suppqualimage{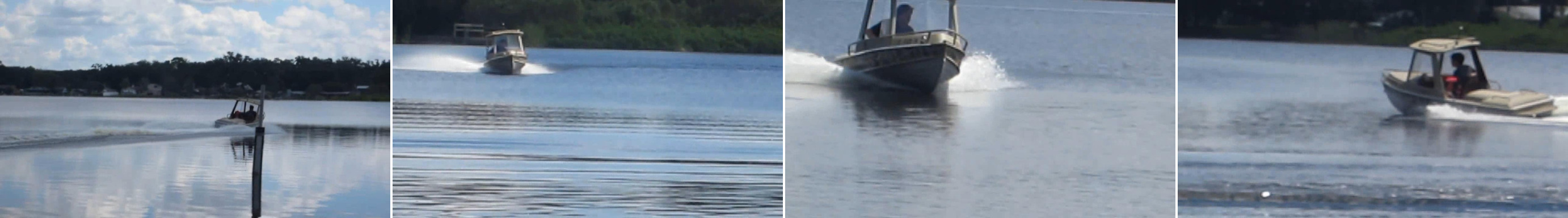} \\
    \suppqualrowlabel{SAM 2} & \suppqualimage{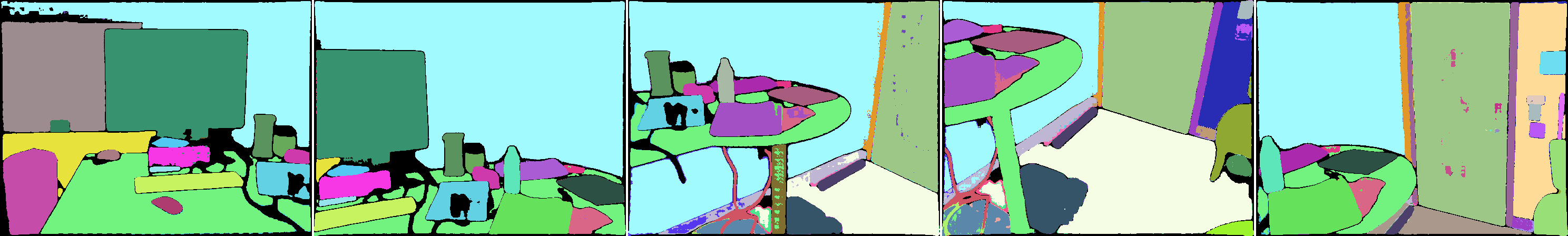} & \suppqualimage{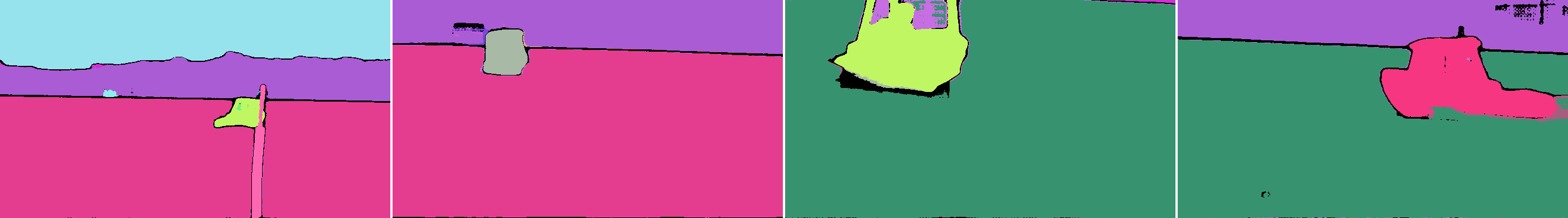} \\
    \suppqualrowlabel{IGGT} & \suppqualimage{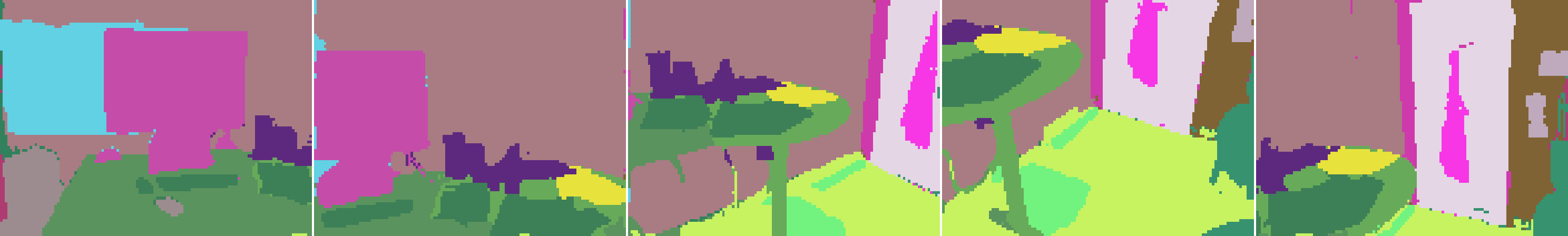} & \suppqualimage{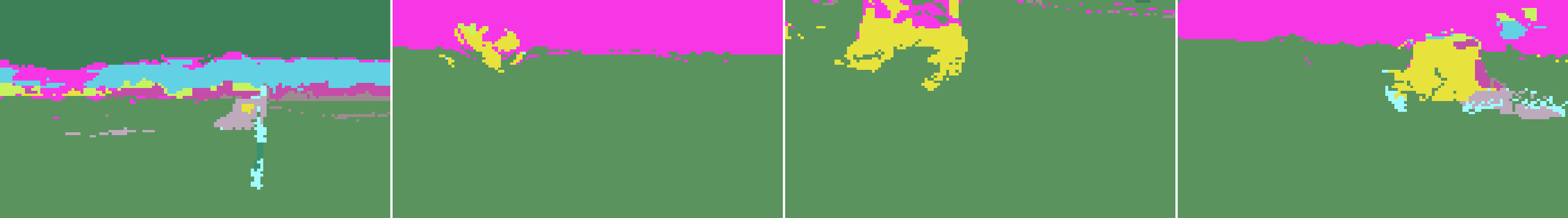} \\
    \suppqualrowlabel{IGGT4D} & \suppqualimage{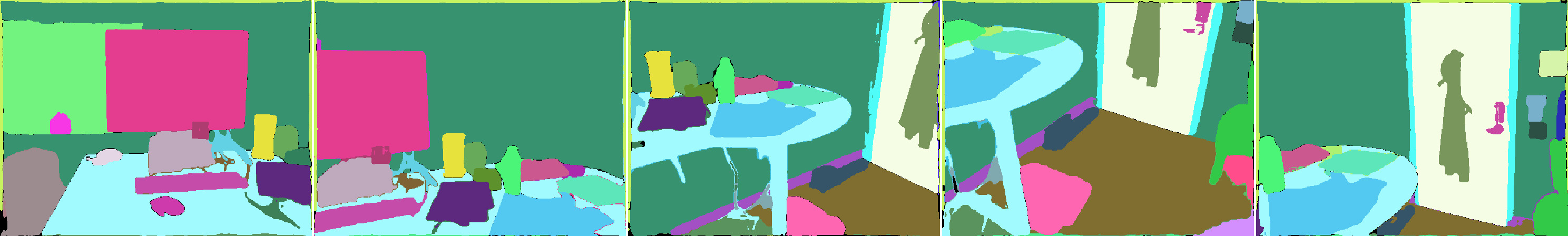} & \suppqualimage{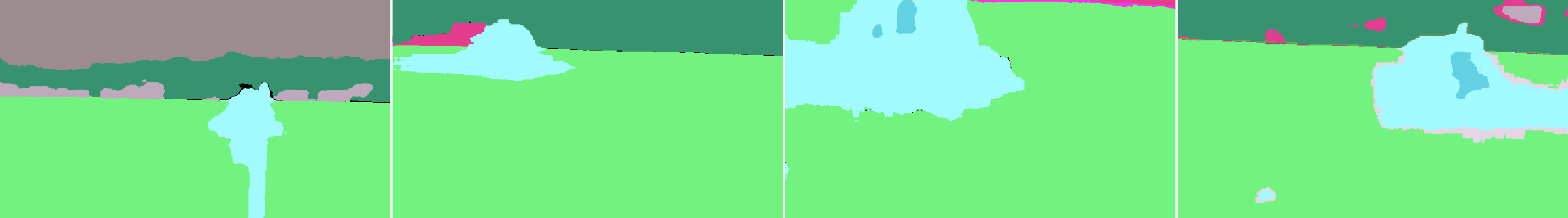} \\
    \suppqualrowlabel{\textbf{Ours}} & \suppqualimage{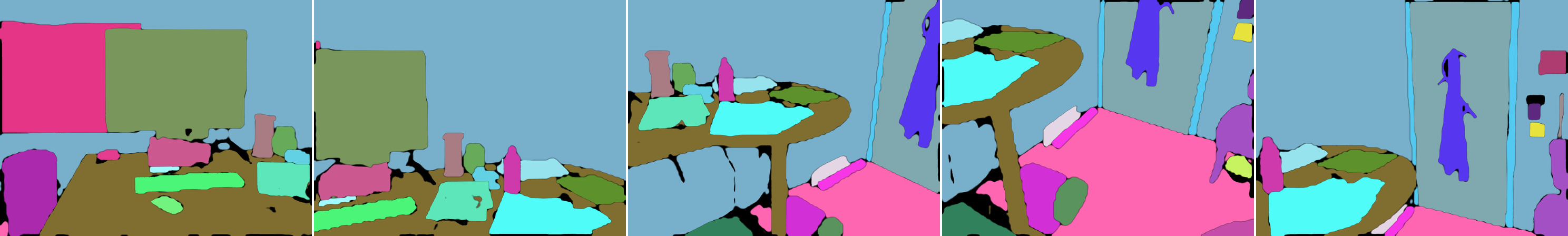} & \suppqualimage{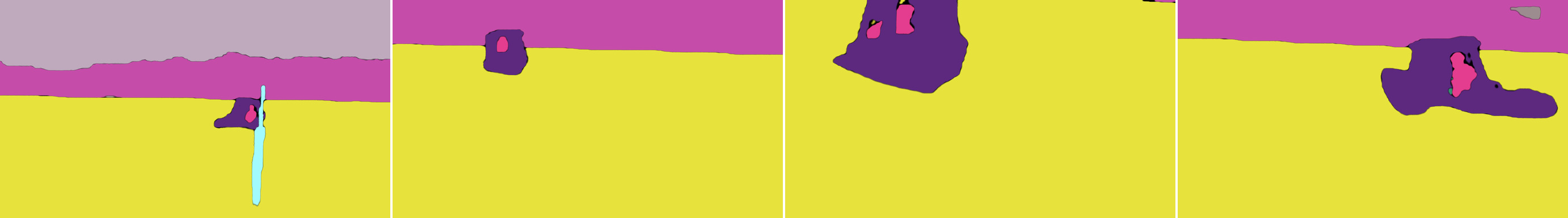} \\
    \suppqualrowlabel{GT} & \suppqualimage{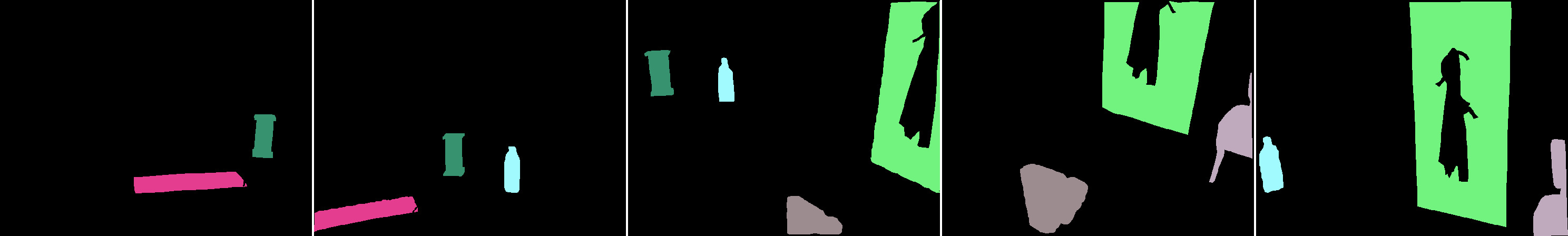} & \suppqualimage{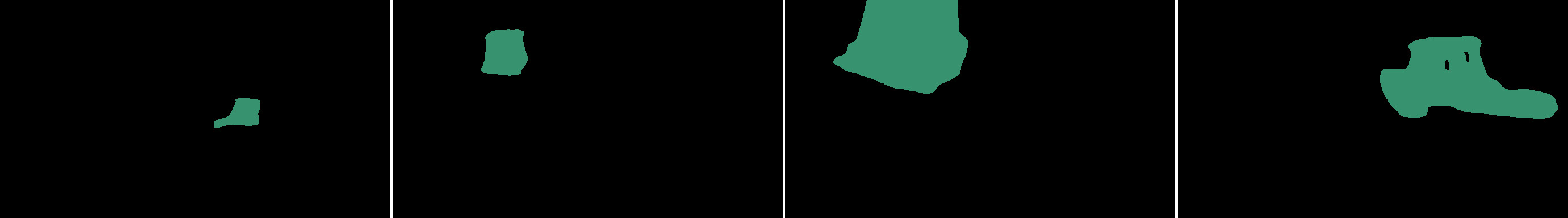} \\
    \noalign{\vskip -2.2mm}
    & {\fontsize{6.2}{6.6}\selectfont\textbf{(c)}} & {\fontsize{6.2}{6.6}\selectfont\textbf{(d)}} \\
    \suppqualrowlabel{RGB} & \suppqualimage{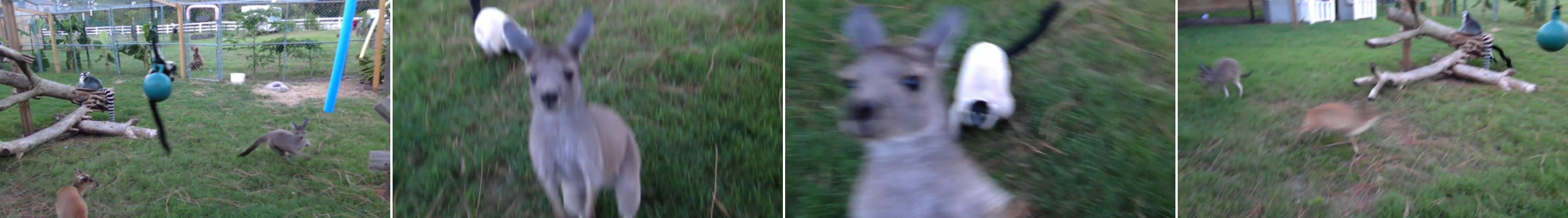} & \suppqualimage{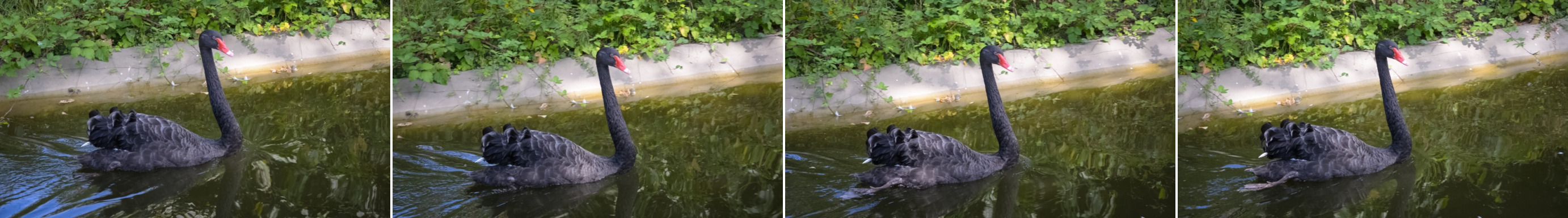} \\
    \suppqualrowlabel{SAM 2} & \suppqualimage{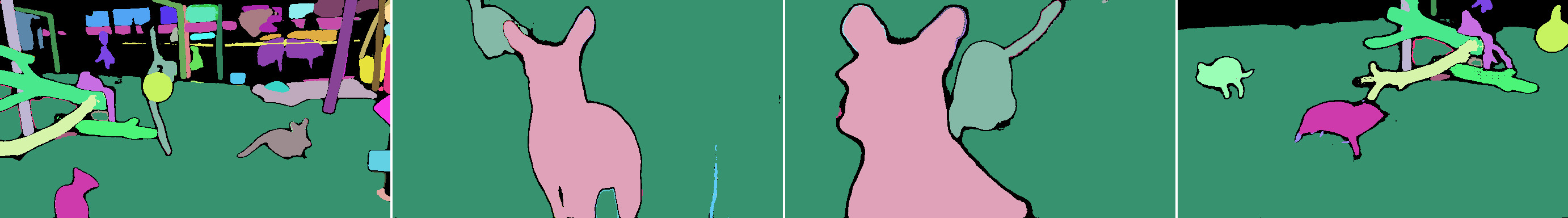} & \suppqualimage{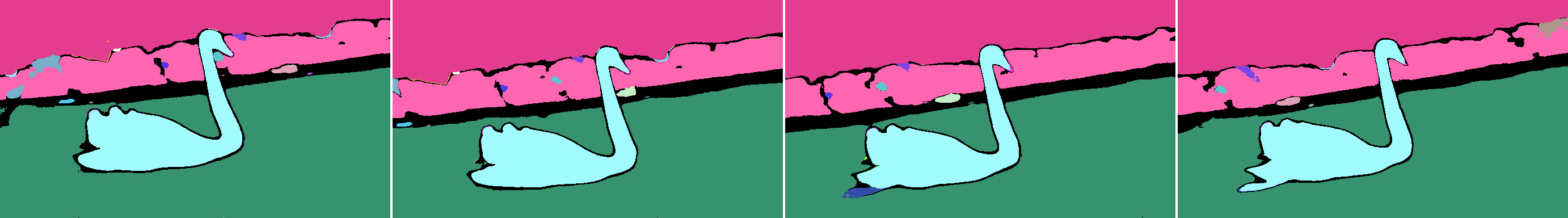} \\
    \suppqualrowlabel{IGGT} & \suppqualimage{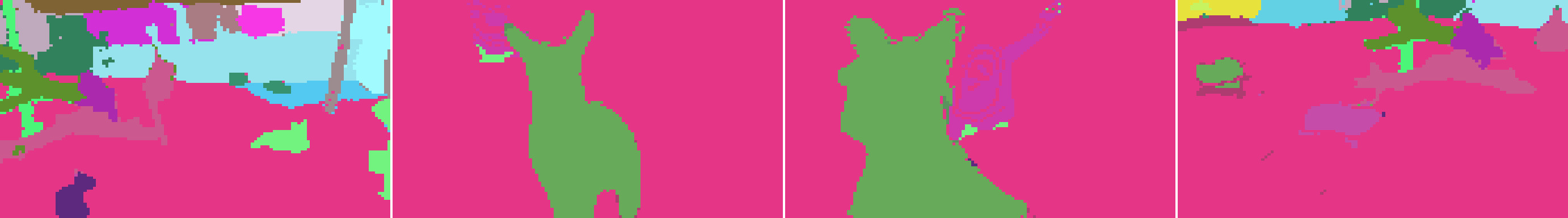} & \suppqualimage{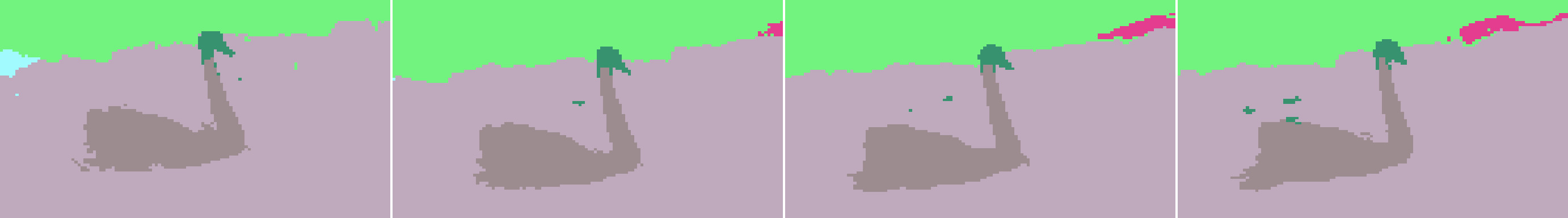} \\
    \suppqualrowlabel{IGGT4D} & \suppqualimage{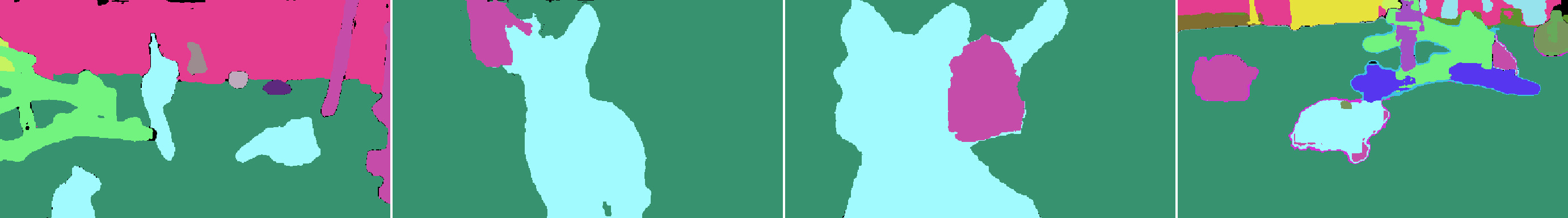} & \suppqualimage{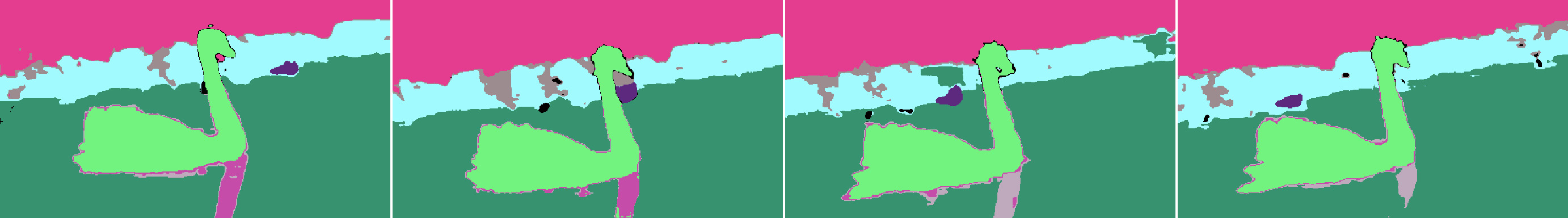} \\
    \suppqualrowlabel{\textbf{Ours}} & \suppqualimage{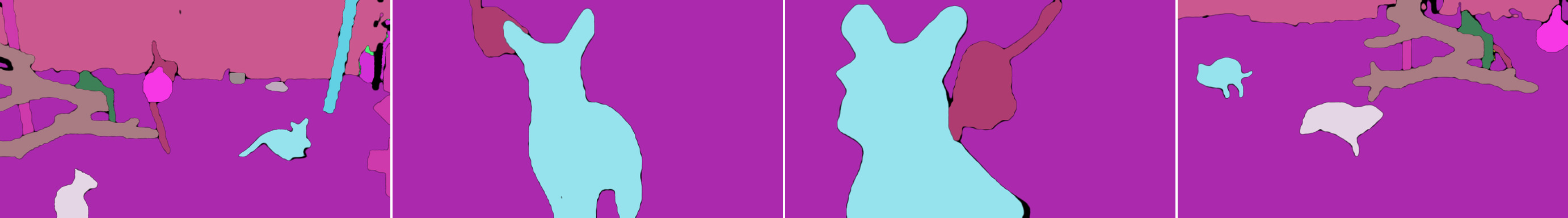} & \suppqualimage{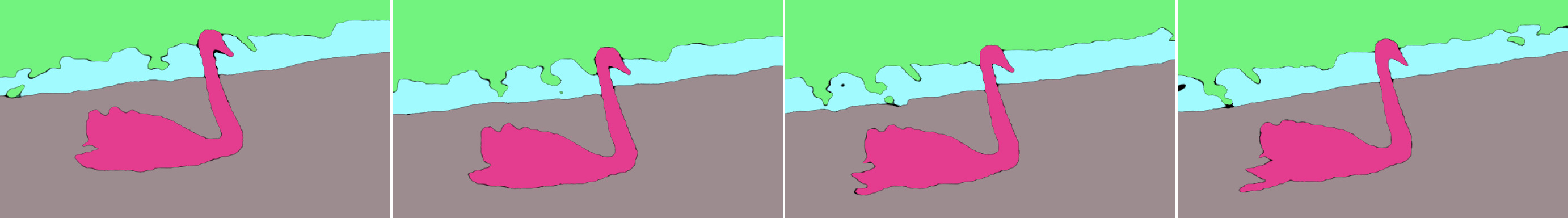} \\
    \suppqualrowlabel{GT} & \suppqualimage{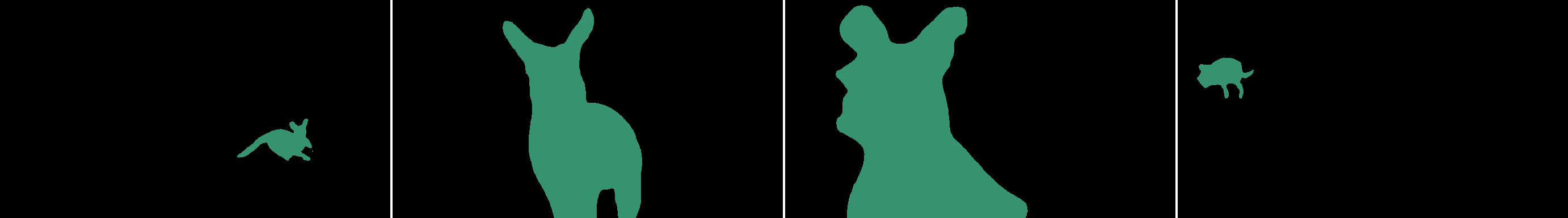} & \suppqualimage{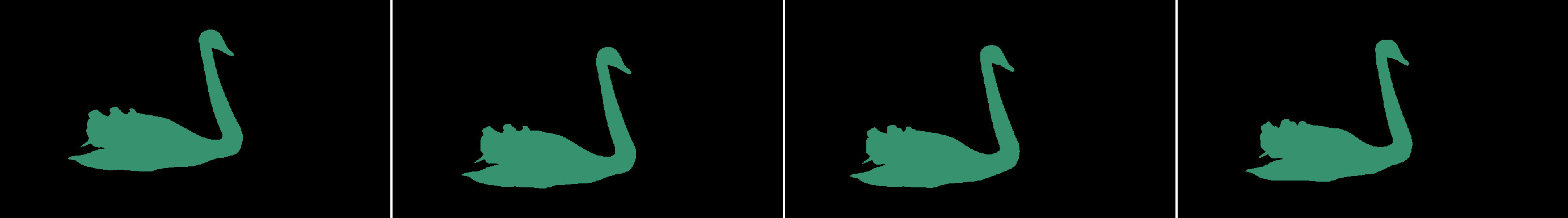} \\
    \noalign{\vskip -2.2mm}
    & {\fontsize{6.2}{6.6}\selectfont\textbf{(e)}} & {\fontsize{6.2}{6.6}\selectfont\textbf{(f)}} \\
    \suppqualrowlabel{RGB} & \suppqualimage{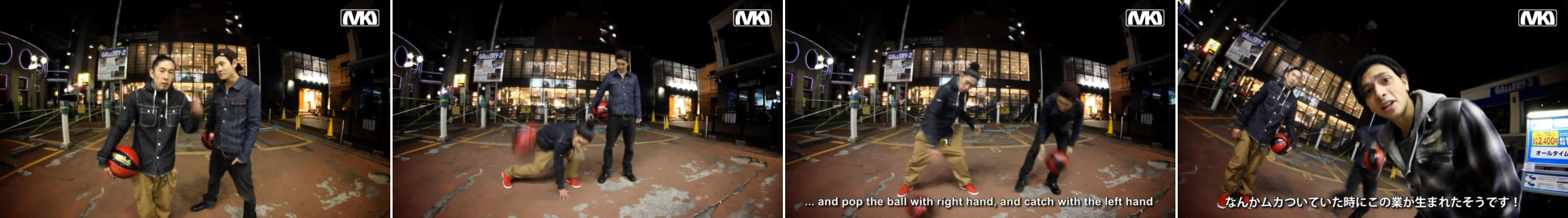} & \suppqualimage{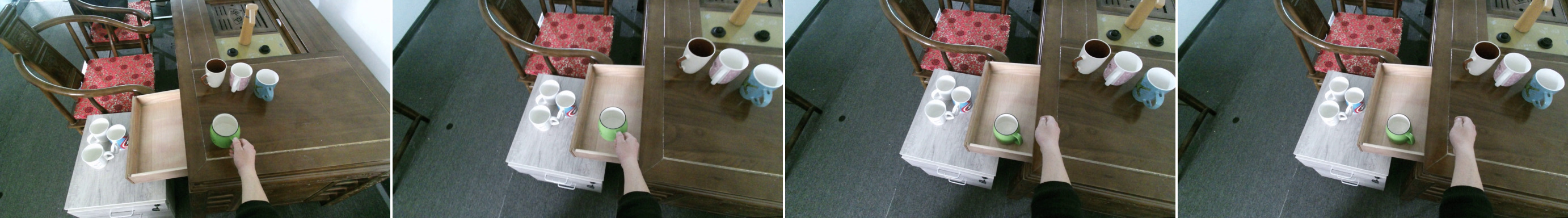} \\
    \suppqualrowlabel{SAM 2} & \suppqualimage{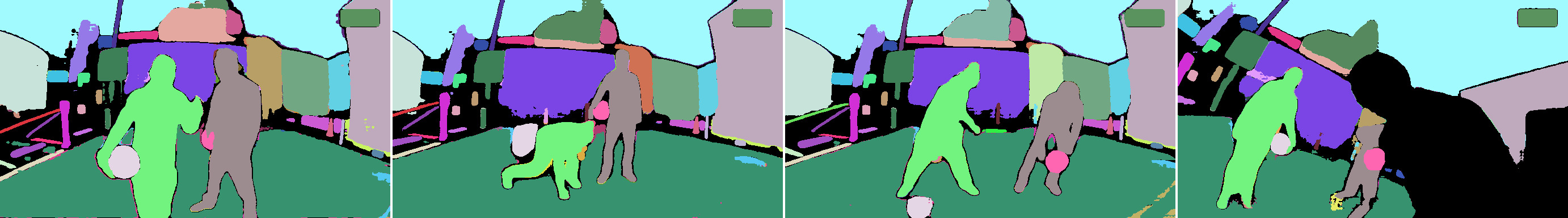} & \suppqualimage{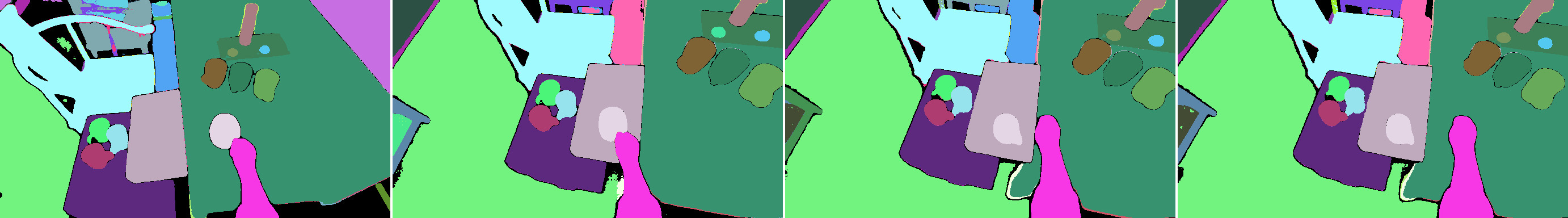} \\
    \suppqualrowlabel{IGGT} & \suppqualimage{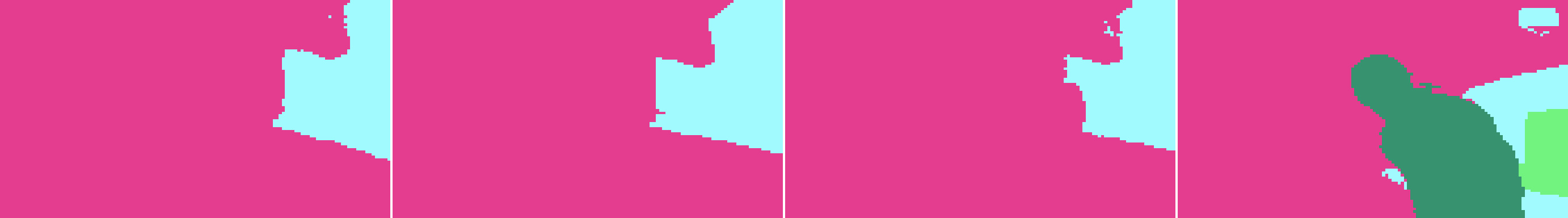} & \suppqualimage{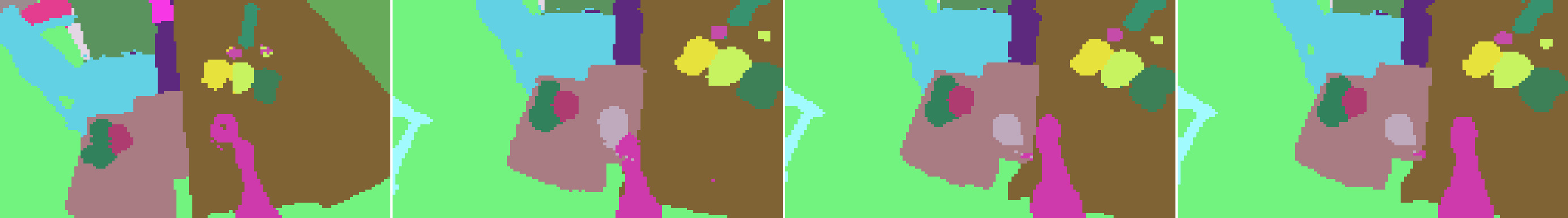} \\
    \suppqualrowlabel{IGGT4D} & \suppqualimage{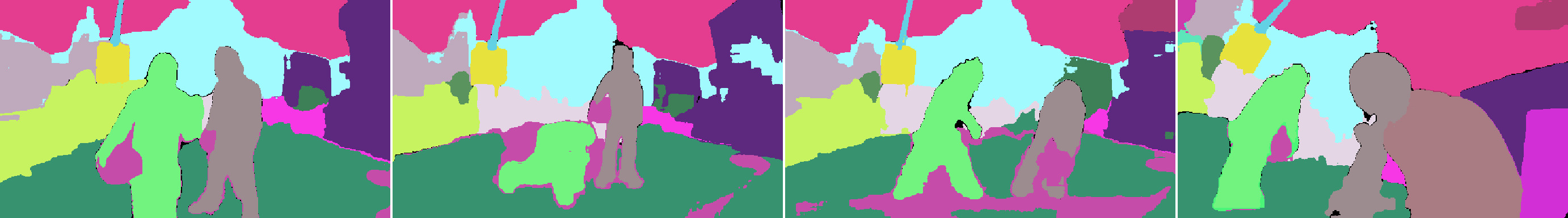} & \suppqualimage{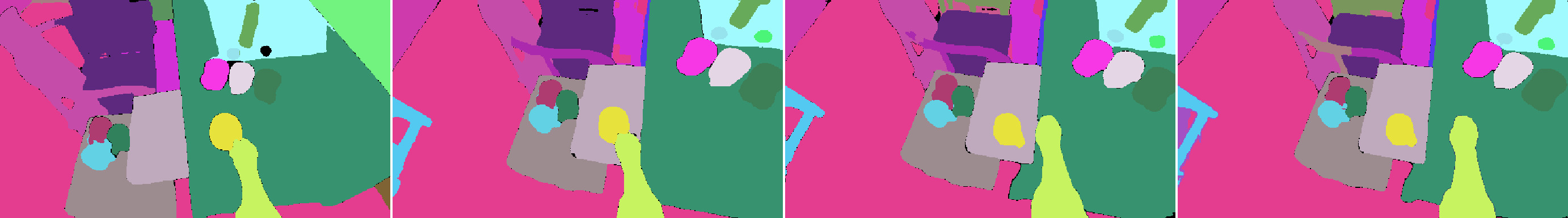} \\
    \suppqualrowlabel{\textbf{Ours}} & \suppqualimage{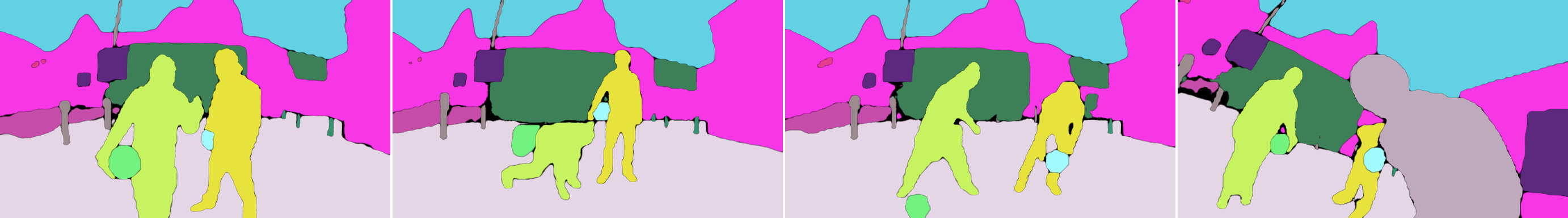} & \suppqualimage{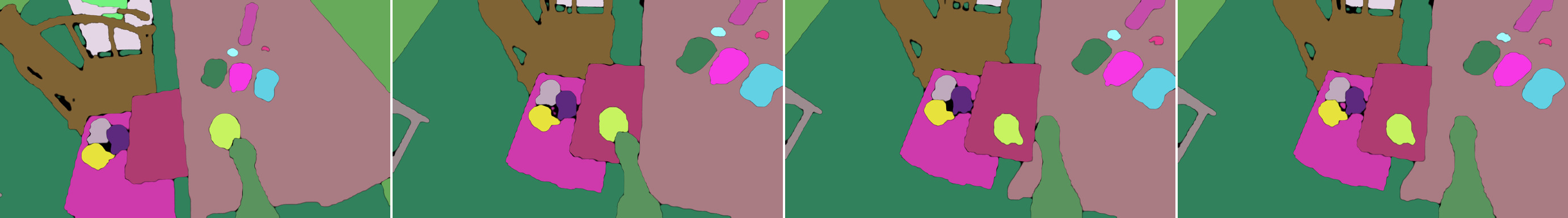} \\
    \suppqualrowlabel{GT} & \suppqualimage{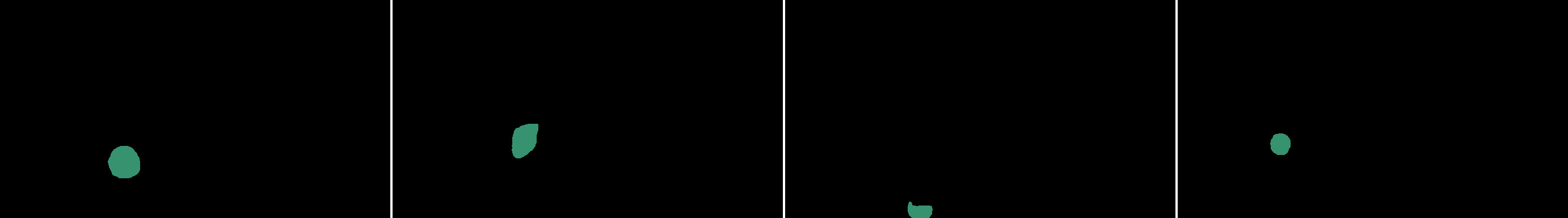} & \suppqualimage{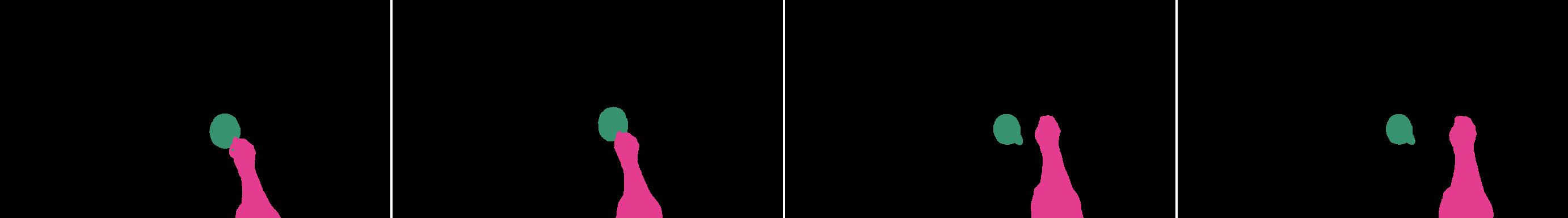}
  \end{tabular}
  \caption{\textbf{Additional qualitative comparisons of class-agnostic segmentation.}
  A consistent mask color denotes the same instance within each sequence.}
  \label{fig:supp-segmentation-qualitative}
\end{figure}

%% file: figures_final/pointcloud_static.tex
\begin{figure}[!tbp]
  \centering
  \setlength{\tabcolsep}{1.2pt}
  \renewcommand{\arraystretch}{0.98}
  \newcommand{\pcstaticrowlabel}[1]{\raisebox{-.5\height}{\rotatebox[origin=c]{90}{\scriptsize #1}}}
  \newcommand{\pcstaticimage}[1]{\raisebox{-.5\height}{\includegraphics[width=0.285\textwidth]{#1}}}
  \begin{tabular}{@{}>{\centering\arraybackslash}m{4mm}ccc@{}}
    \multicolumn{4}{c}{\scriptsize\textbf{(a)}} \\
    \multicolumn{4}{c}{
      \includegraphics[width=0.92\textwidth]{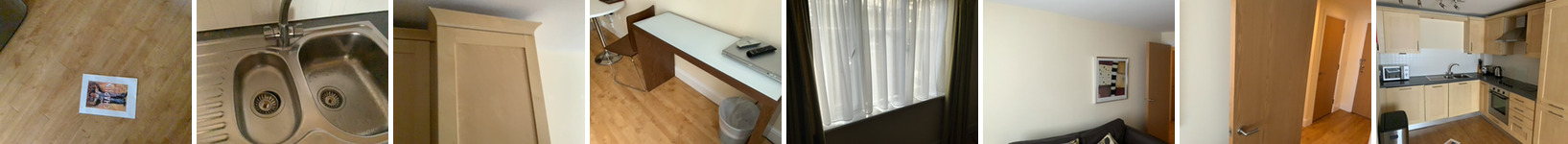}} \\[-0.2mm]
    & \scriptsize IGGT & \scriptsize IGGT4D & \scriptsize\textbf{Ours} \\
    \pcstaticrowlabel{RGB} &
      \pcstaticimage{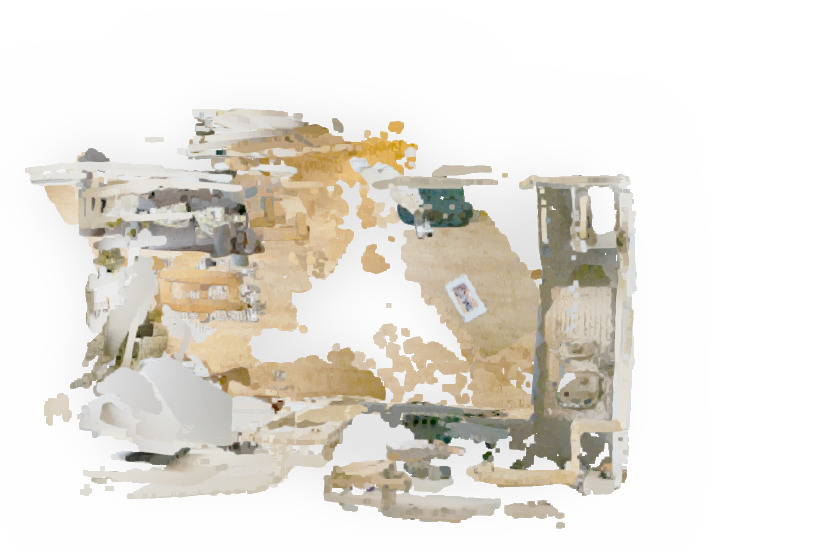} &
      \pcstaticimage{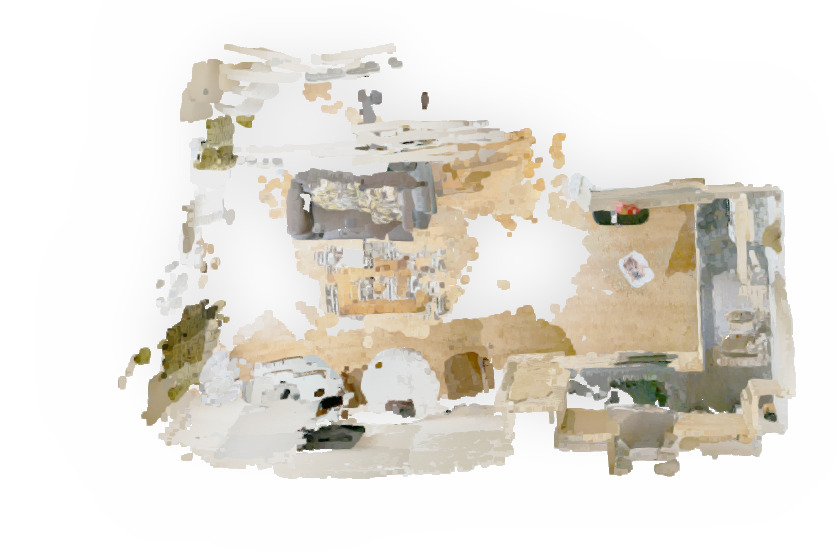} &
      \pcstaticimage{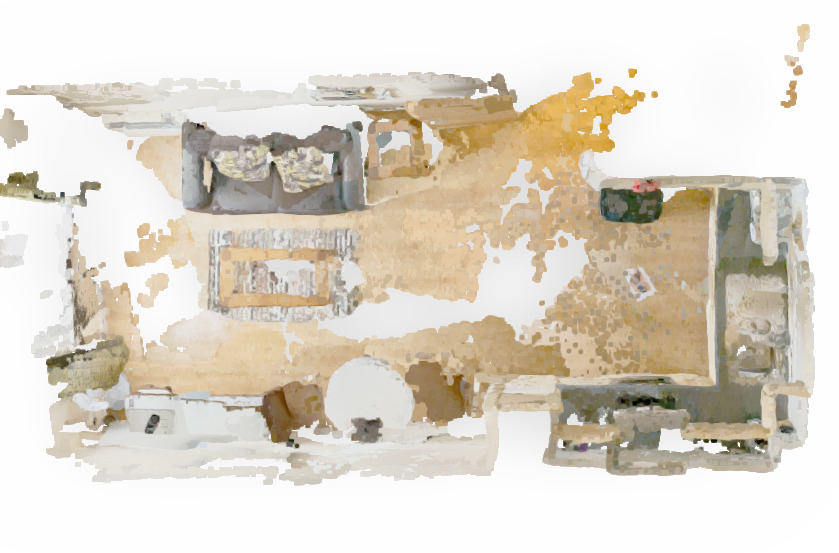} \\[-0.5mm]
    \pcstaticrowlabel{Mask} &
      \pcstaticimage{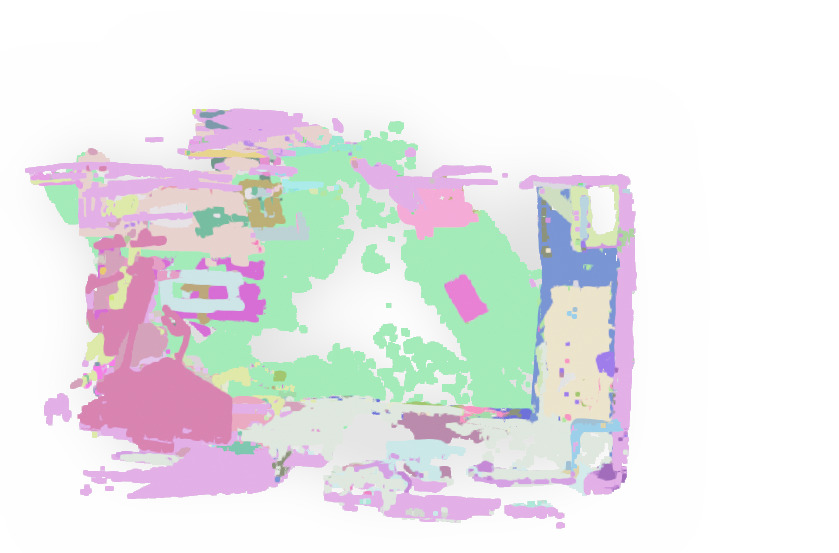} &
      \pcstaticimage{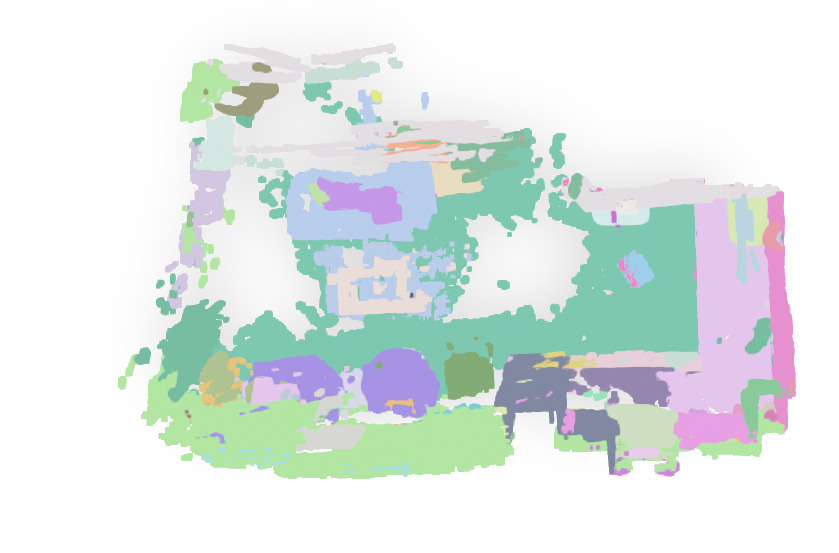} &
      \pcstaticimage{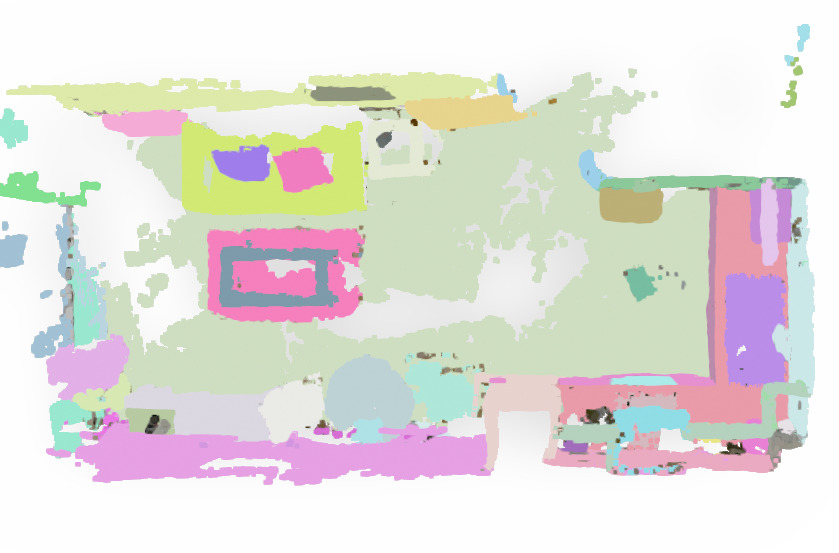} \\[1.0mm]
    \multicolumn{4}{c}{\scriptsize\textbf{(b)}} \\
    \multicolumn{4}{c}{
      \includegraphics[width=0.92\textwidth]{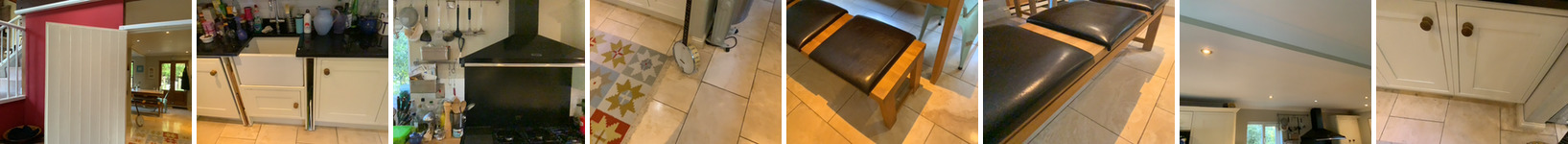}} \\[-0.2mm]
    & \scriptsize IGGT & \scriptsize IGGT4D & \scriptsize\textbf{Ours} \\
    \pcstaticrowlabel{RGB} &
      \pcstaticimage{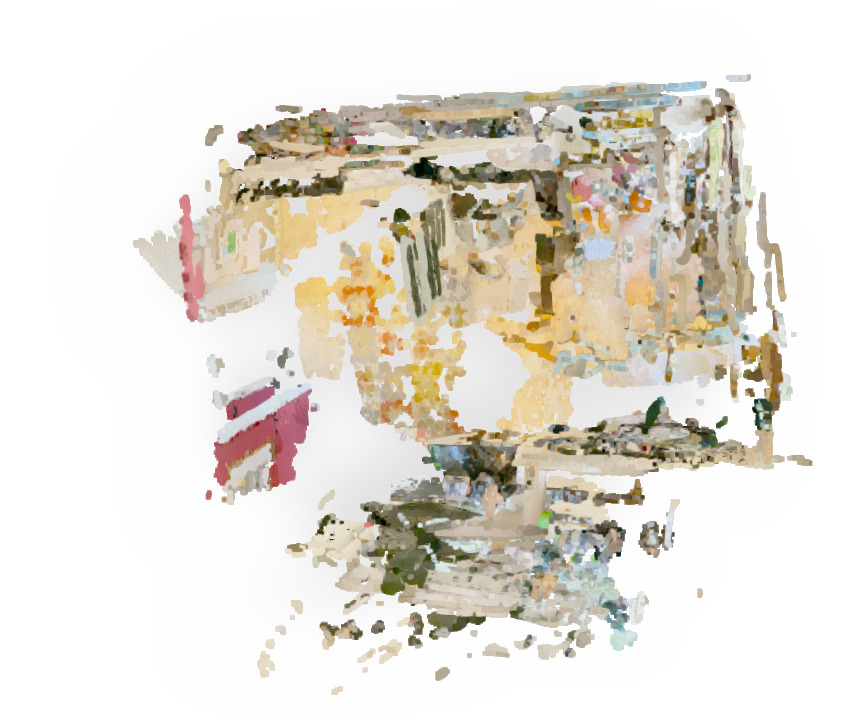} &
      \pcstaticimage{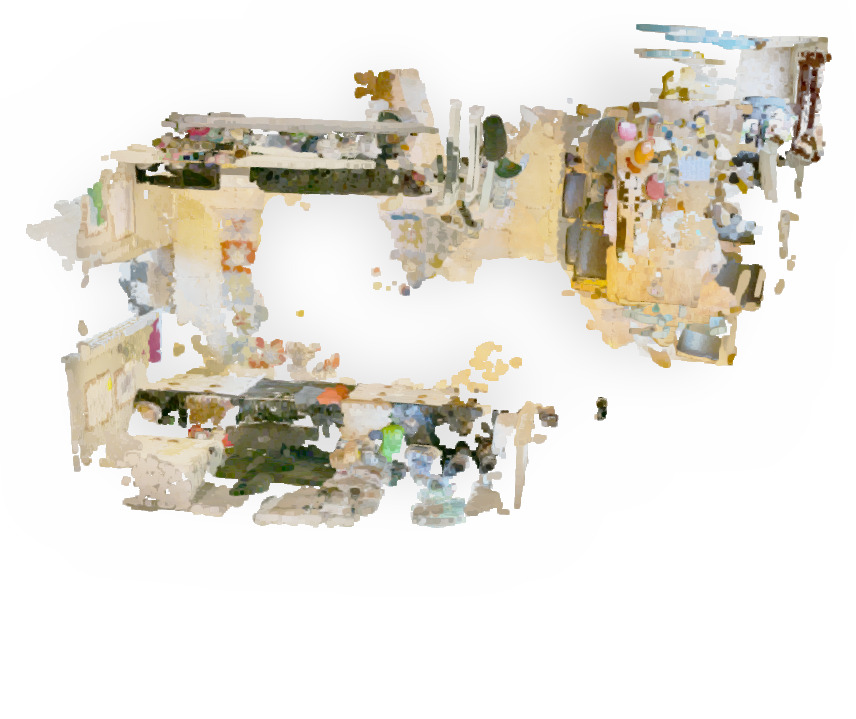} &
      \pcstaticimage{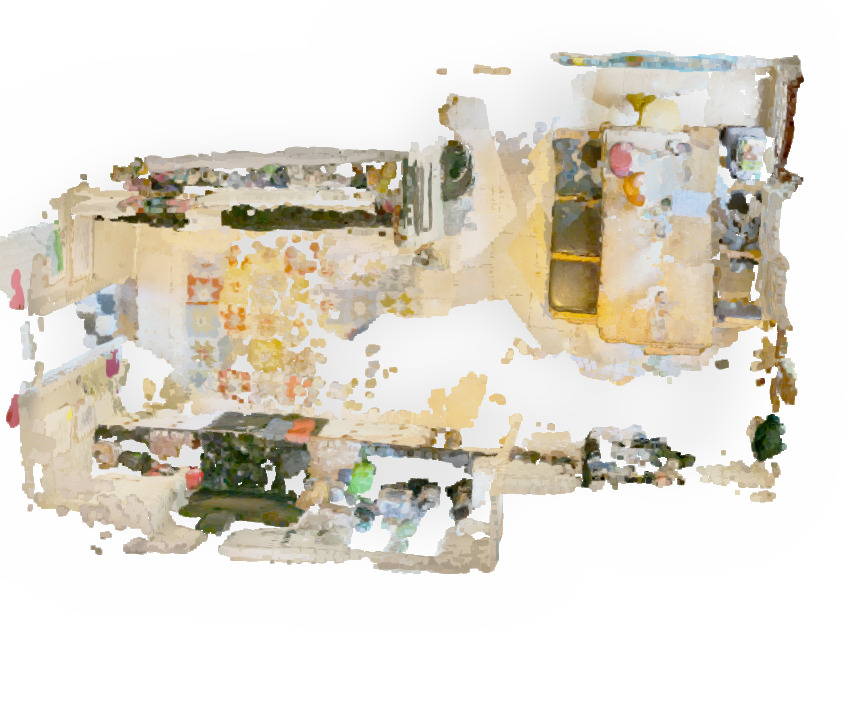} \\[-0.5mm]
    \pcstaticrowlabel{Mask} &
      \pcstaticimage{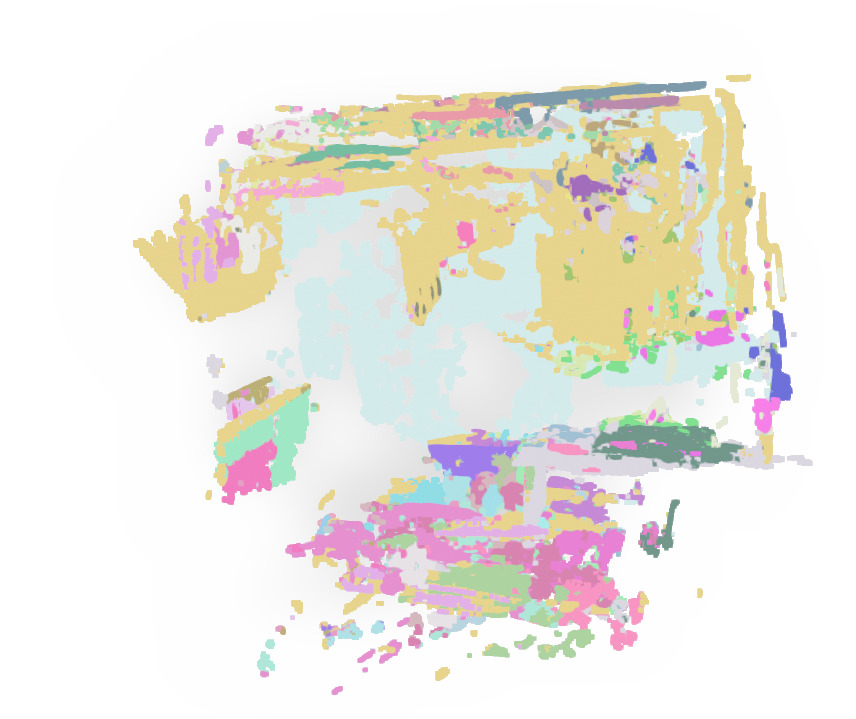} &
      \pcstaticimage{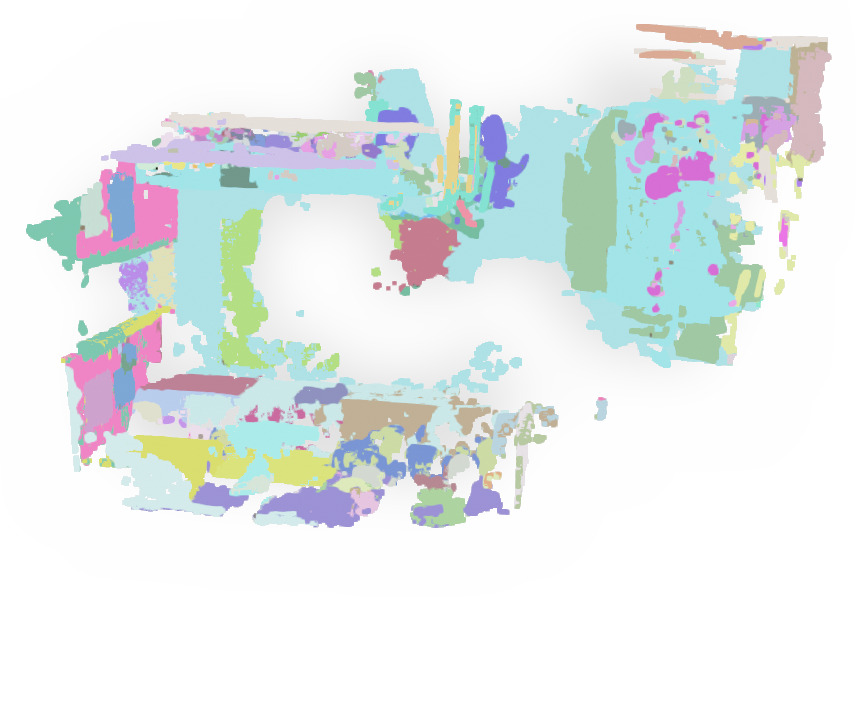} &
      \pcstaticimage{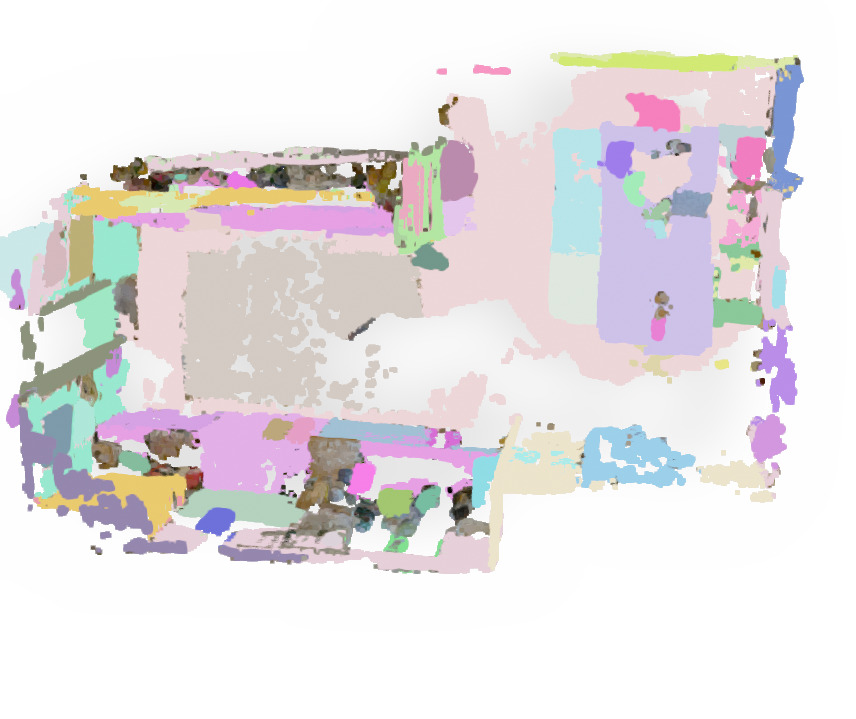}
  \end{tabular}
  \caption{\textbf{Feed-forward 3D reconstruction and instance segmentation.}
  Eight RGB observations from held-out ARKitScenes are followed by aligned top-view RGB and
  instance-colored point clouds; all methods are evaluated zero-shot.}
  \label{fig:supp-static-pointcloud}
\end{figure}

%% file: figures_final/pointcloud_dynamic.tex
\begin{figure}[!tbp]
  \centering
  \setlength{\tabcolsep}{1.2pt}
  \renewcommand{\arraystretch}{0.96}
  \newcommand{\pcdynamicrowlabel}[1]{\raisebox{-.5\height}{\rotatebox[origin=c]{90}{\scriptsize #1}}}
  \newcommand{\pcdynamicimage}[1]{\raisebox{-.5\height}{\includegraphics[width=0.40\textwidth]{#1}}}
  \begin{tabular}{@{}>{\centering\arraybackslash}m{4mm}cc@{}}
    \multicolumn{3}{c}{\scriptsize\textbf{(a)}} \\
    \multicolumn{3}{c}{
      \includegraphics[width=0.92\textwidth]{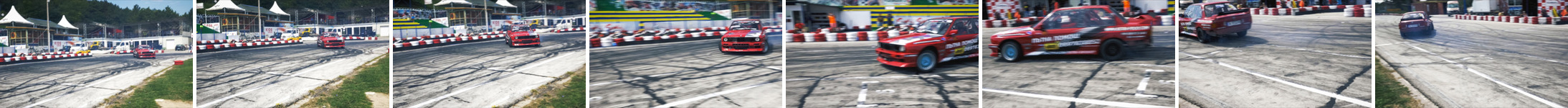}} \\[-0.2mm]
    & \scriptsize IGGT4D & \scriptsize\textbf{Ours} \\
    \pcdynamicrowlabel{RGB} &
      \pcdynamicimage{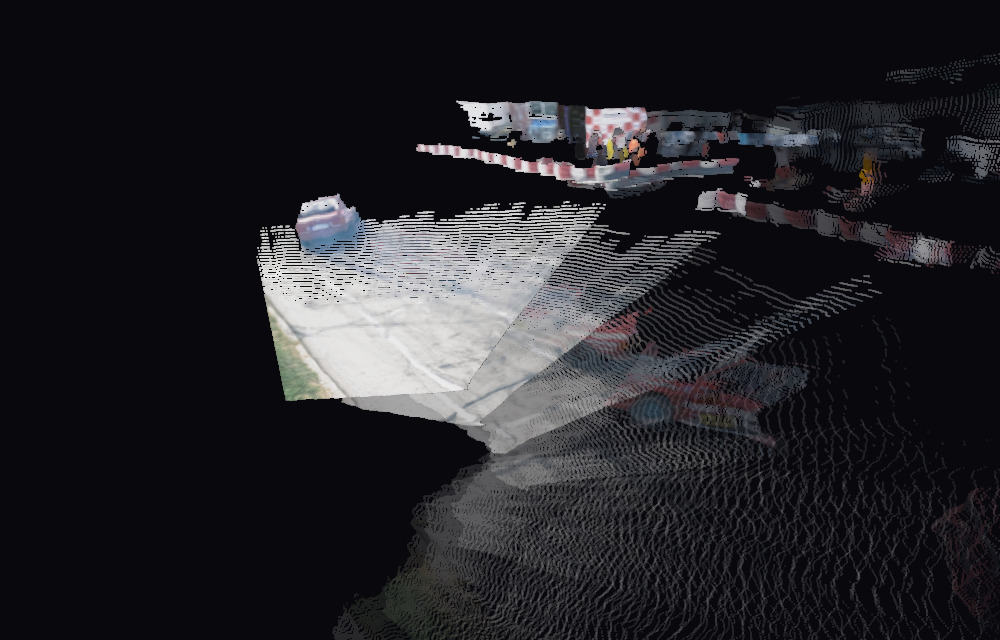} &
      \pcdynamicimage{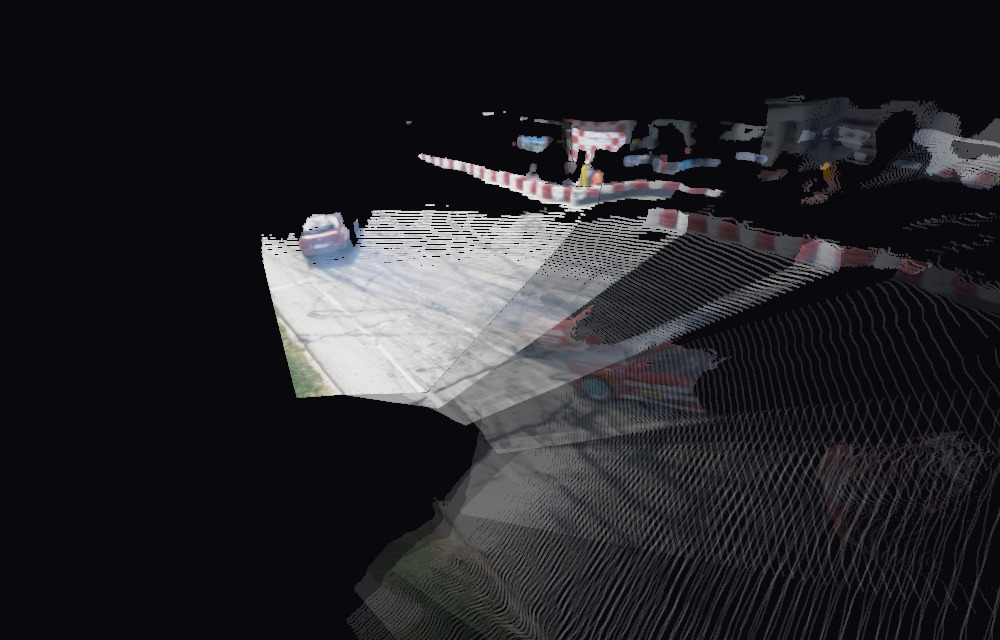} \\[-0.5mm]
    \pcdynamicrowlabel{Mask} &
      \pcdynamicimage{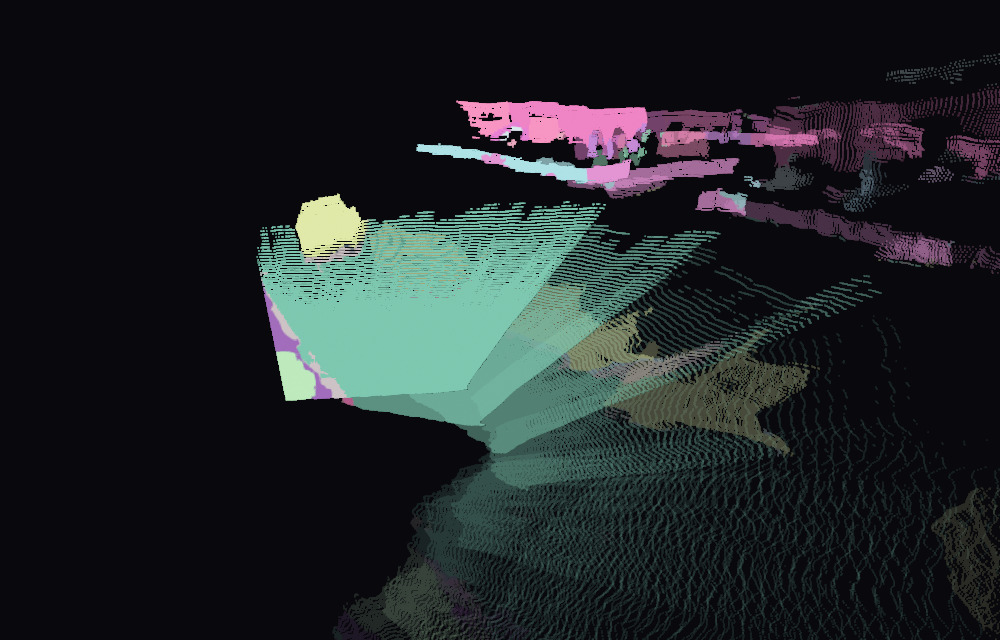} &
      \pcdynamicimage{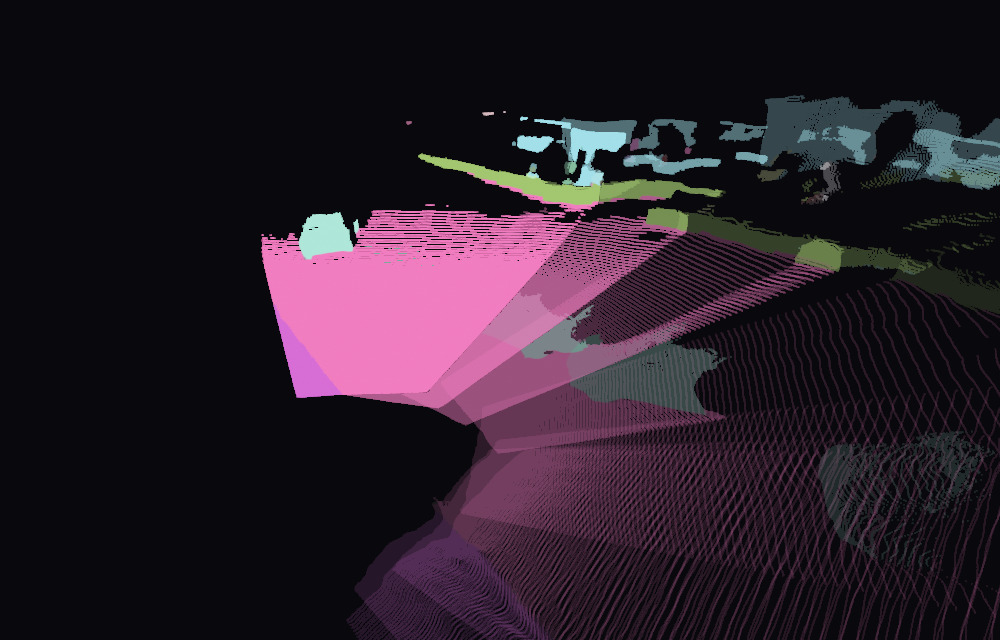} \\[2.2mm]
    \multicolumn{3}{c}{\scriptsize\textbf{(b)}} \\
    \multicolumn{3}{c}{
      \includegraphics[width=0.92\textwidth]{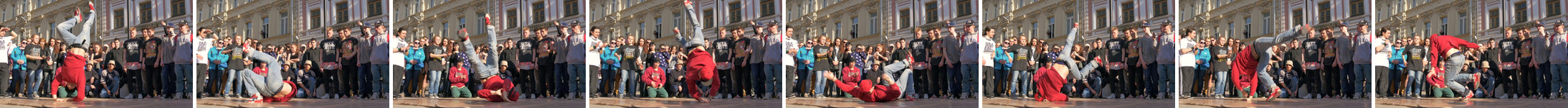}} \\[-0.2mm]
    & \scriptsize IGGT4D & \scriptsize\textbf{Ours} \\
    \pcdynamicrowlabel{RGB} &
      \pcdynamicimage{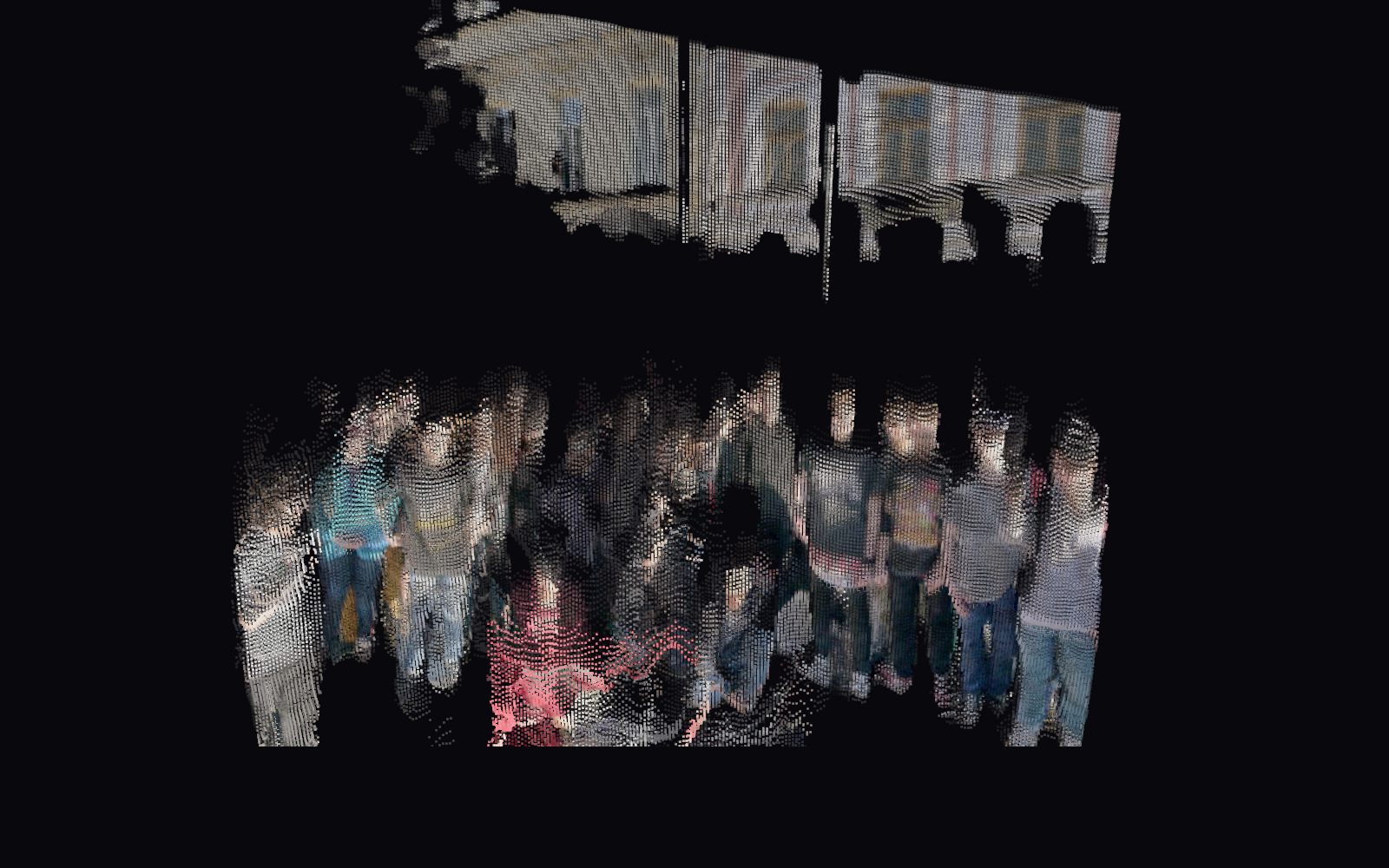} &
      \pcdynamicimage{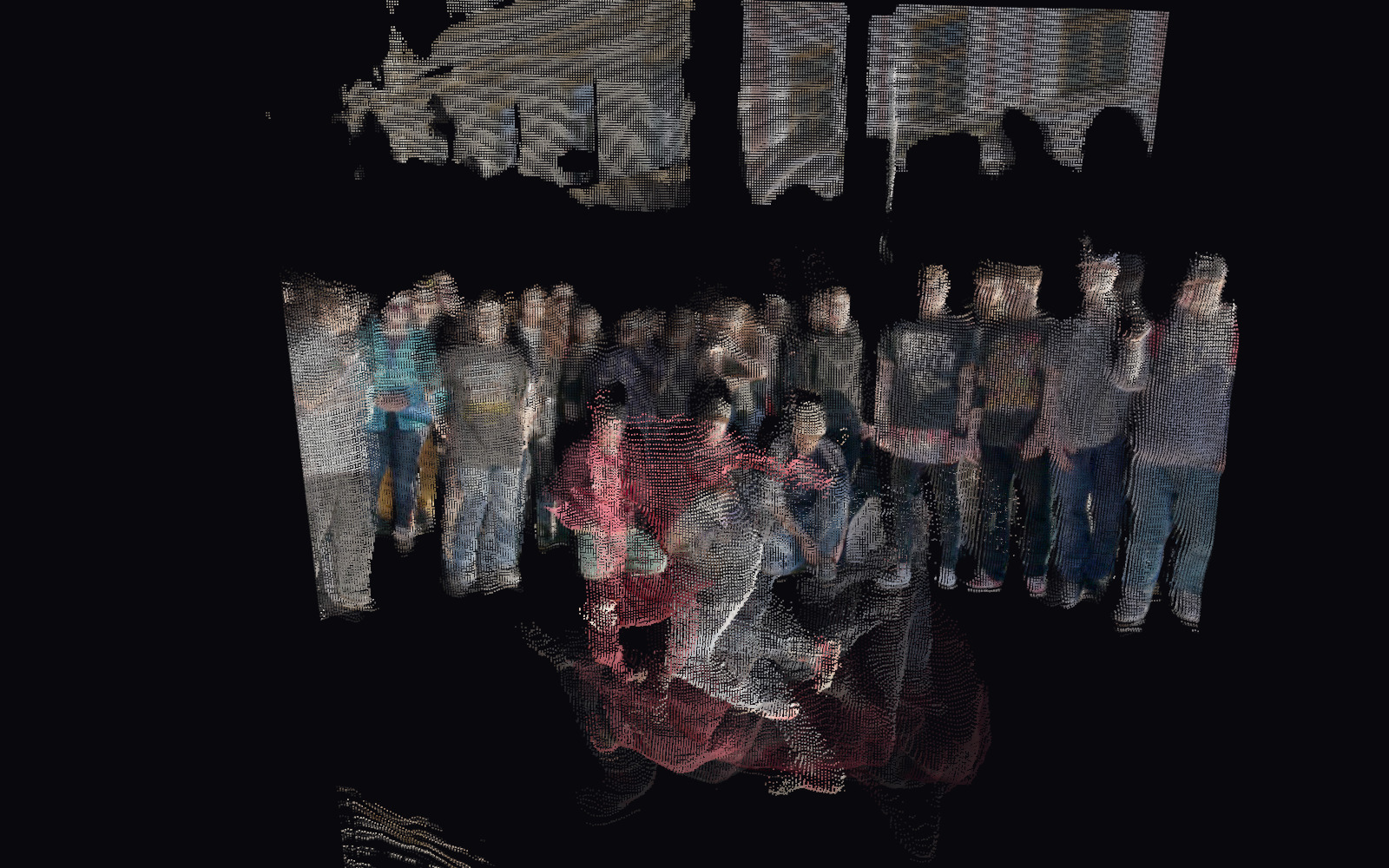} \\[-0.5mm]
    \pcdynamicrowlabel{Mask} &
      \pcdynamicimage{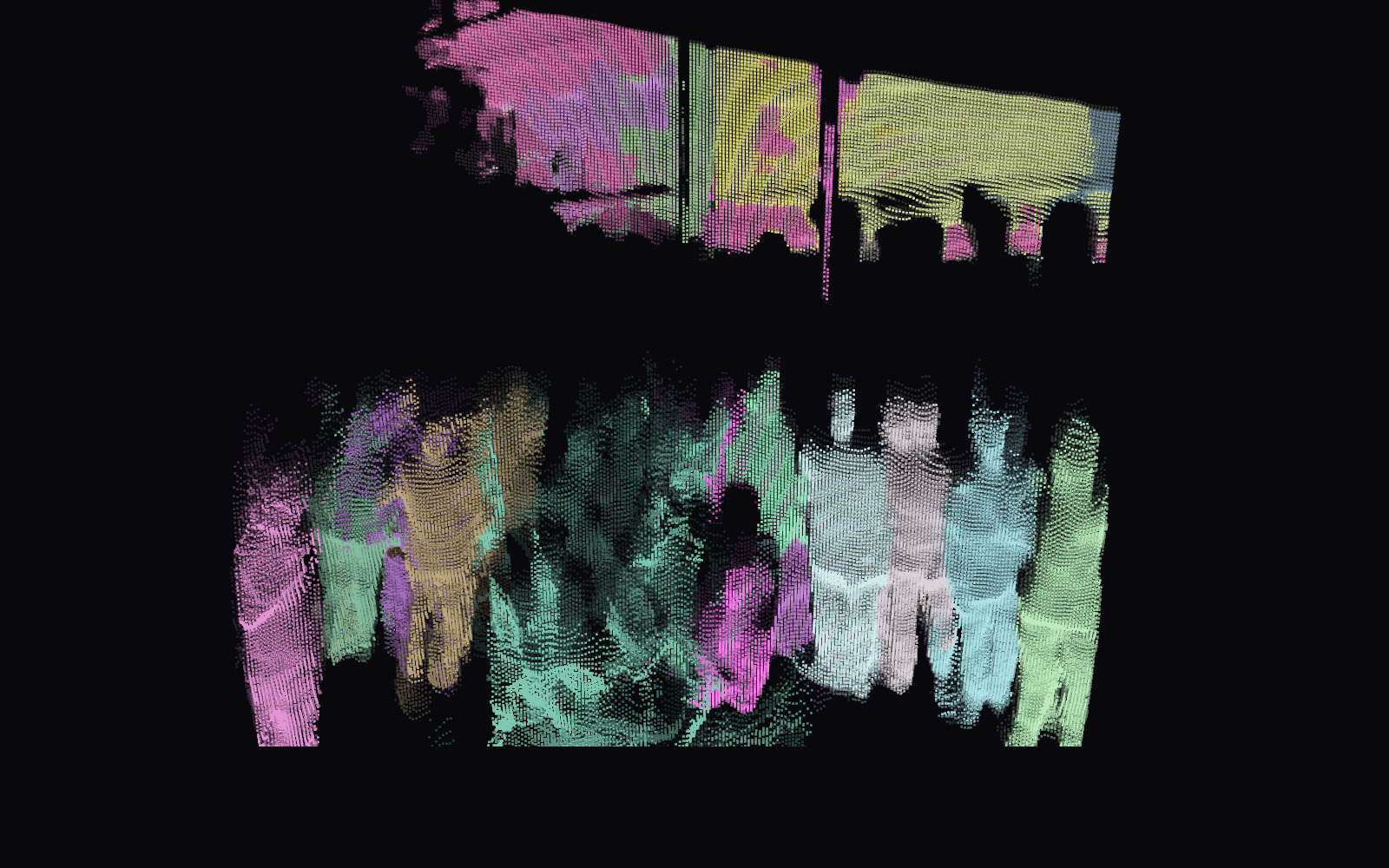} &
      \pcdynamicimage{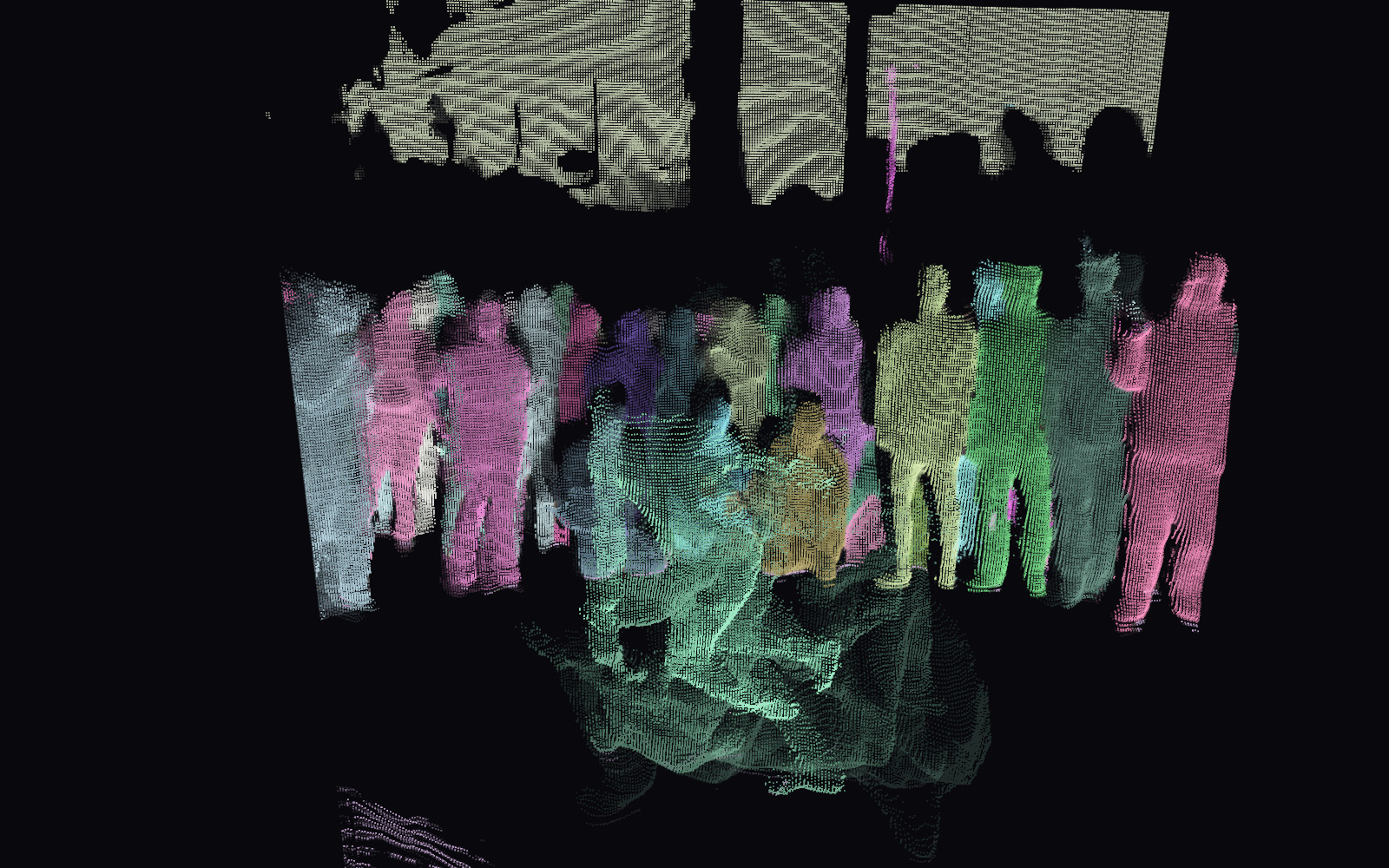}
  \end{tabular}
  \caption{\textbf{Feed-forward 4D reconstruction and instance segmentation in dynamic scenes.}
  Selected RGB observations are followed by aligned RGB and instance-colored point-cloud views;
  earlier observations are rendered with lower opacity.}
  \label{fig:supp-dynamic-pointcloud}
\end{figure}

%% file: arxiv/figures/additional_grounding_full_1.tex
\begin{figure}[p]
  \centering
  \setlength{\tabcolsep}{0pt}
  \renewcommand{\arraystretch}{1.04}
  \newcommand{\groundfullonerowlabel}[1]{%
    \raisebox{\dimexpr5.0mm-.5\height+.5\depth\relax}{%
      \rotatebox[origin=c]{90}{\fontsize{5.8}{6.2}\selectfont #1}}}
  \newcommand{\groundfulloneimage}[1]{\includegraphics[width=0.445\textwidth]{#1}}
  \begin{tabular}{r@{\hspace{1.0mm}}c@{\hspace{2.0mm}}c}
    & {\fontsize{7}{7.5}\selectfont\textbf{(a)}} & {\fontsize{7}{7.5}\selectfont\textbf{(b)}} \\
    & \parbox[c][9mm][c]{0.445\textwidth}{\centering\fontsize{6}{6.4}\selectfont\emph{``This is a white printer. It is on the counter.''}} &
      \parbox[c][9mm][c]{0.445\textwidth}{\centering\fontsize{6}{6.4}\selectfont\emph{``This is a small gray pillow that is crumpled up. It is in the corner of a black couch, near a wooden door.''}} \\
    \groundfullonerowlabel{RGB} & \groundfulloneimage{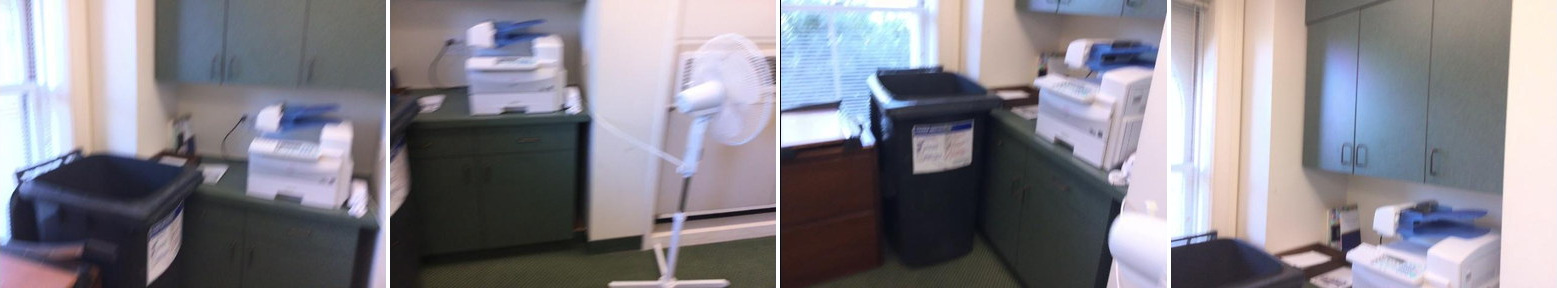} & \groundfulloneimage{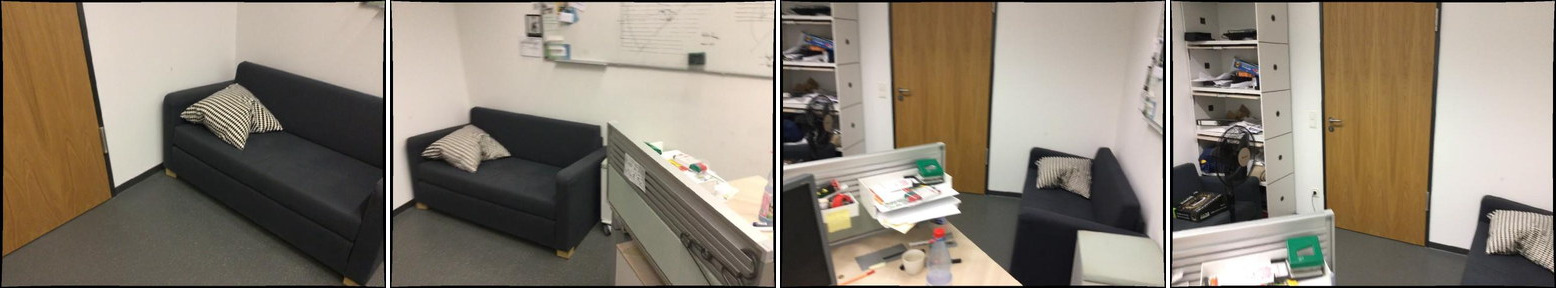} \\
    \groundfullonerowlabel{VLM+SAM2} & \groundfulloneimage{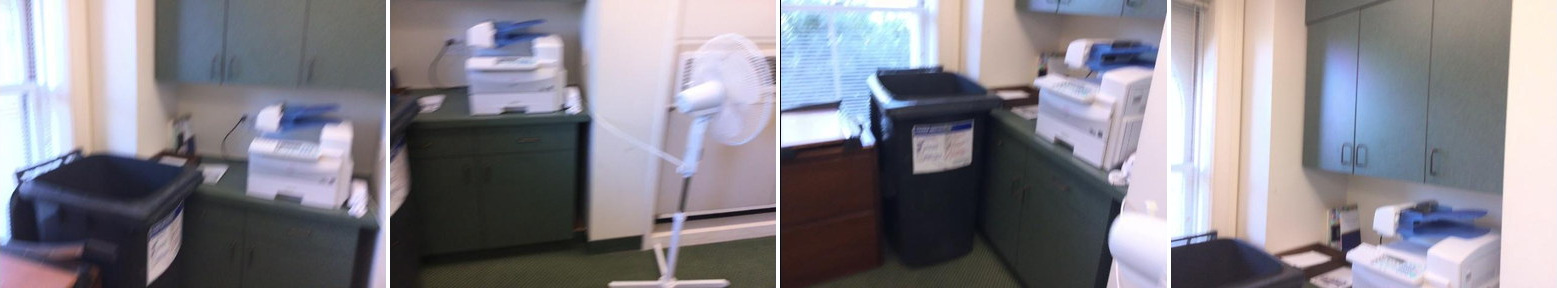} & \groundfulloneimage{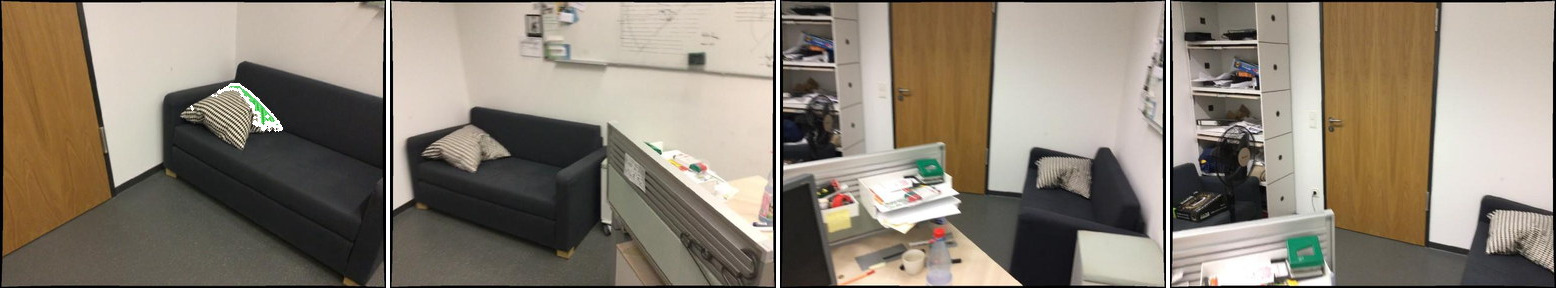} \\
    \groundfullonerowlabel{SAM3} & \groundfulloneimage{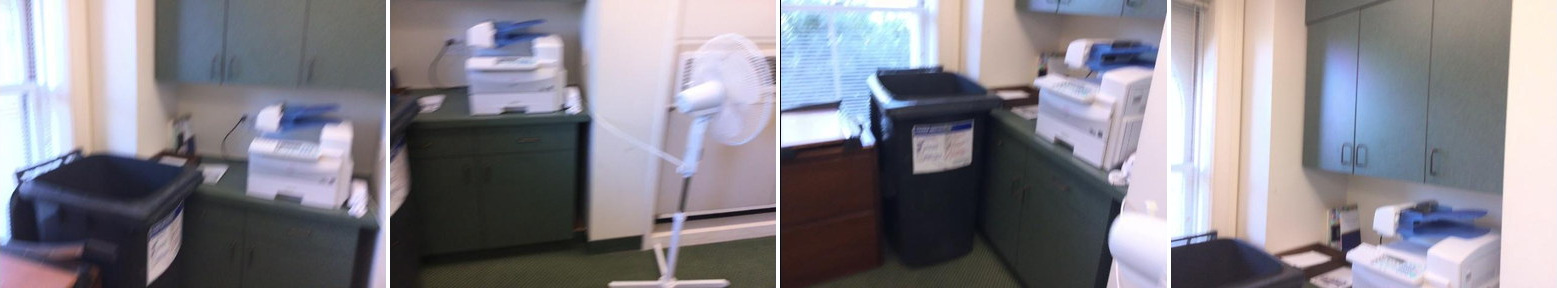} & \groundfulloneimage{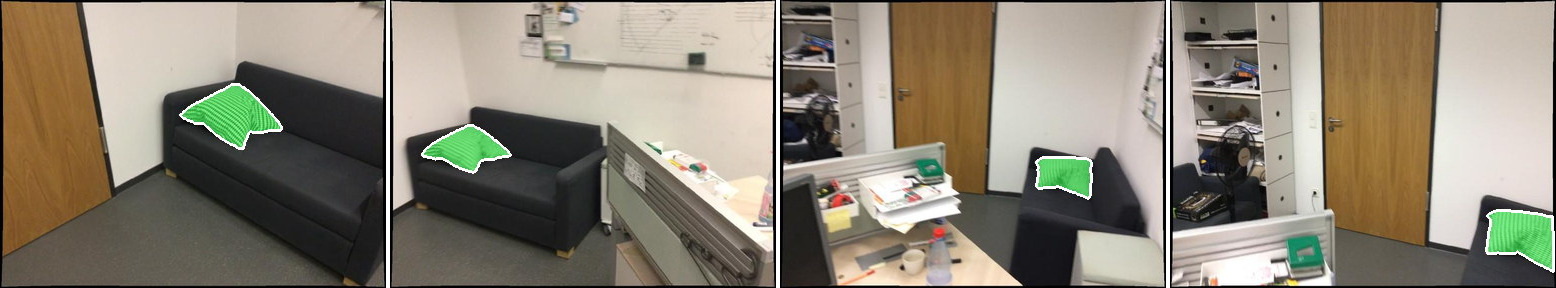} \\
    \groundfullonerowlabel{\textbf{Ours}} & \groundfulloneimage{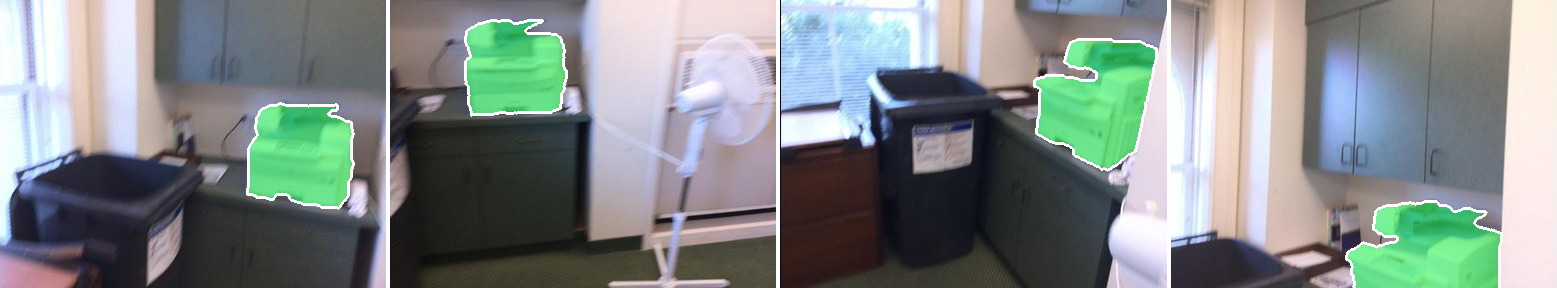} & \groundfulloneimage{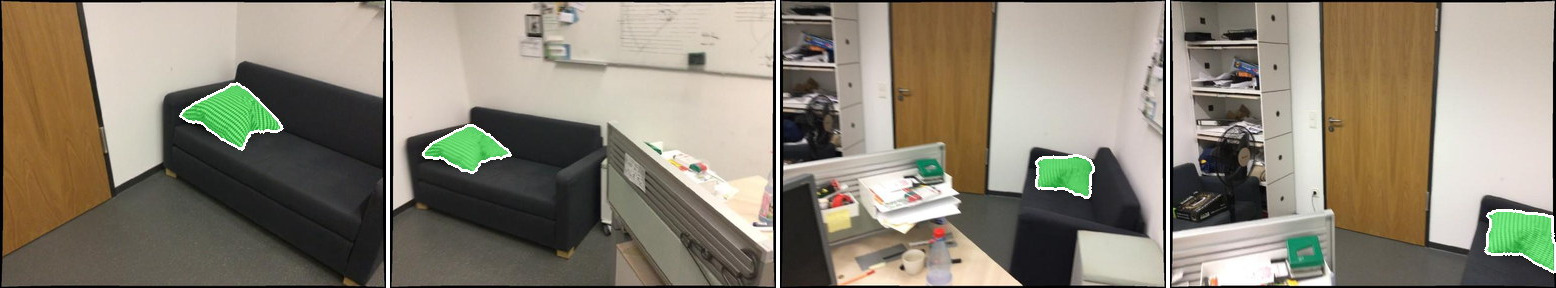} \\
    \groundfullonerowlabel{GT} & \groundfulloneimage{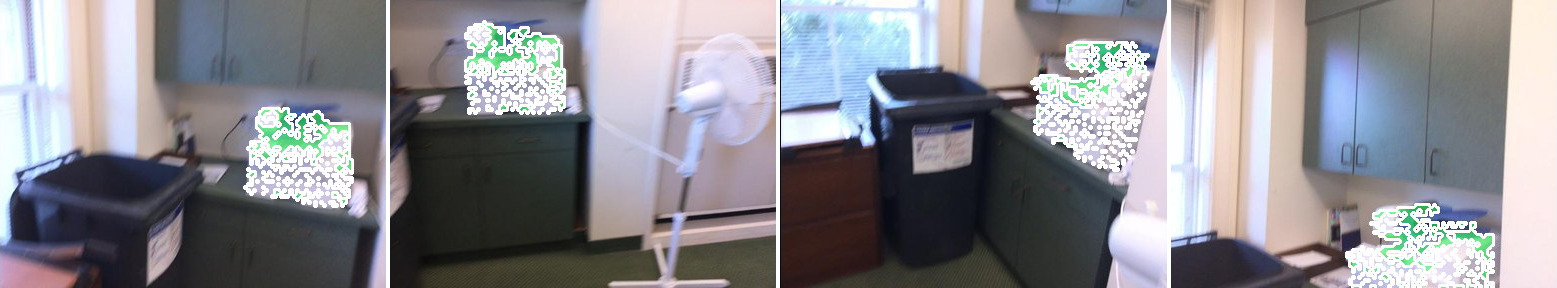} & \groundfulloneimage{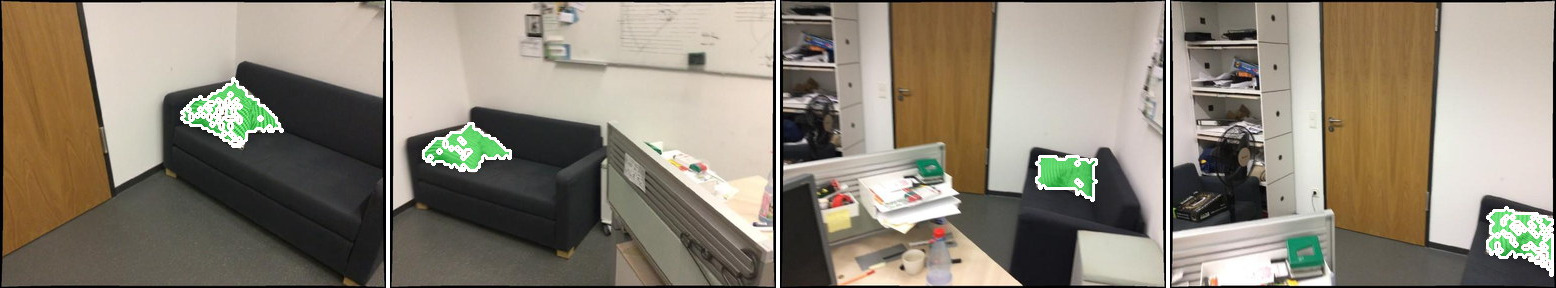} \\[1.8mm]
    & {\fontsize{7}{7.5}\selectfont\textbf{(c)}} & {\fontsize{7}{7.5}\selectfont\textbf{(d)}} \\
    & \parbox[c][9mm][c]{0.445\textwidth}{\centering\fontsize{6}{6.4}\selectfont\emph{``This is a silver laptop. It is on a desk, across from an armchair.''}} &
      \parbox[c][9mm][c]{0.445\textwidth}{\centering\fontsize{6}{6.4}\selectfont\emph{``TV with glossy black screen and slim bezel, designed for displaying digital video and images in indoor settings.''}} \\
    \groundfullonerowlabel{RGB} & \groundfulloneimage{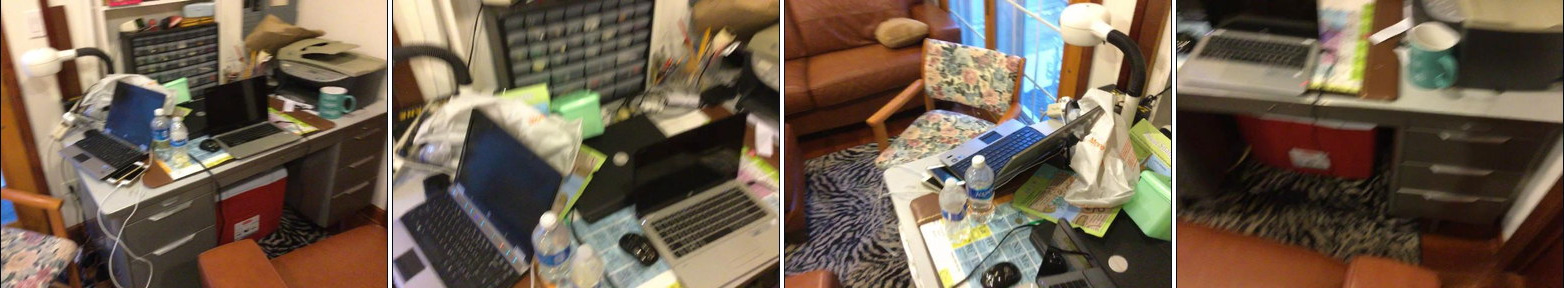} & \groundfulloneimage{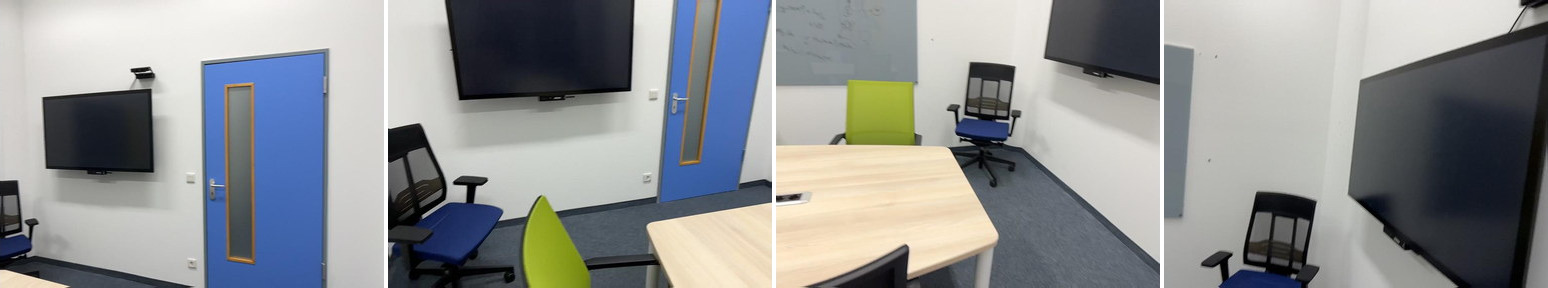} \\
    \groundfullonerowlabel{VLM+SAM2} & \groundfulloneimage{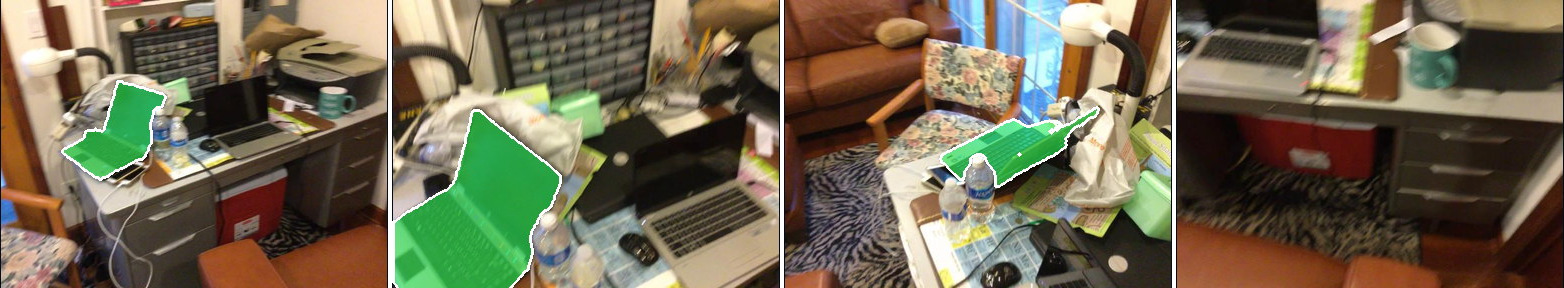} & \groundfulloneimage{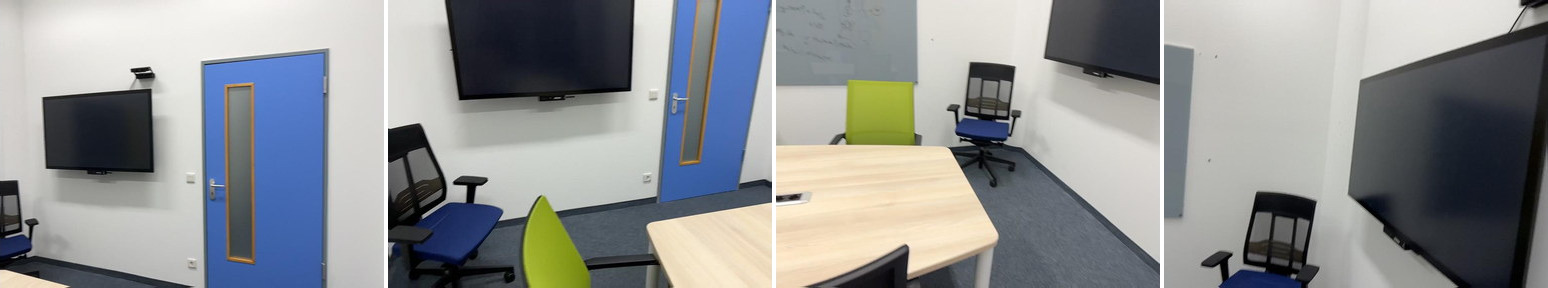} \\
    \groundfullonerowlabel{SAM3} & \groundfulloneimage{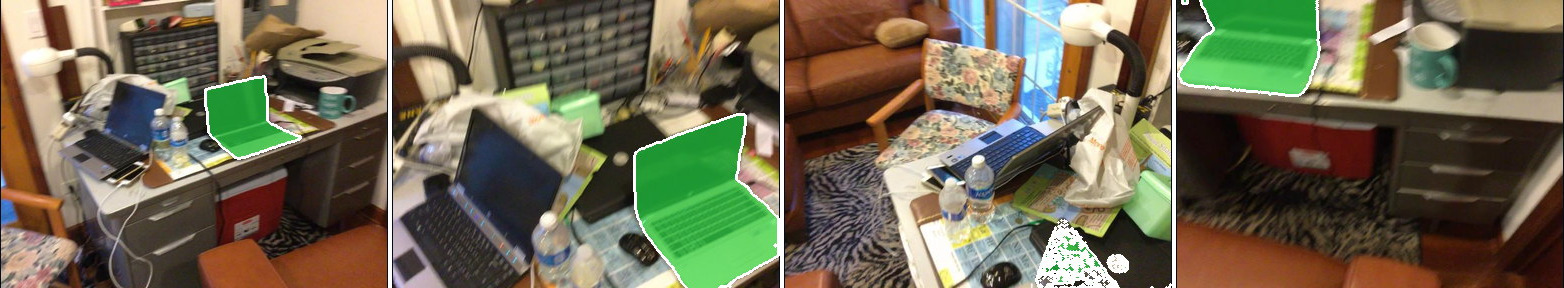} & \groundfulloneimage{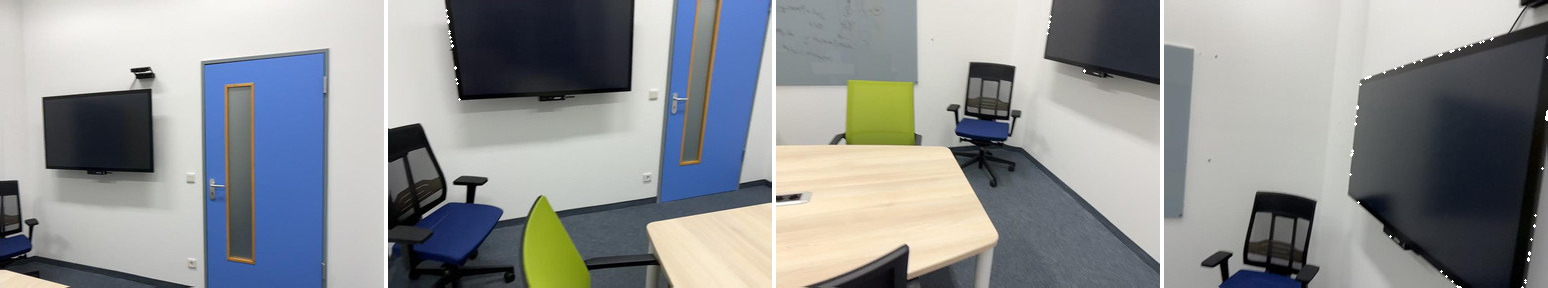} \\
    \groundfullonerowlabel{\textbf{Ours}} & \groundfulloneimage{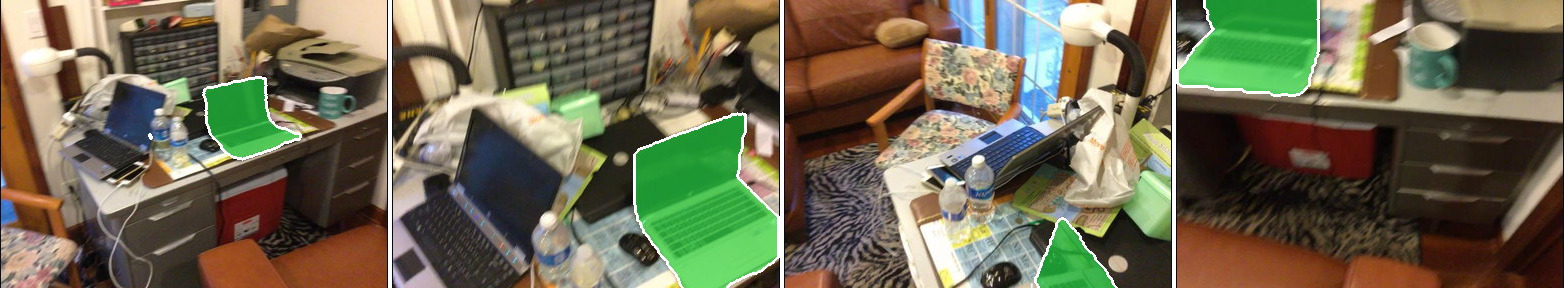} & \groundfulloneimage{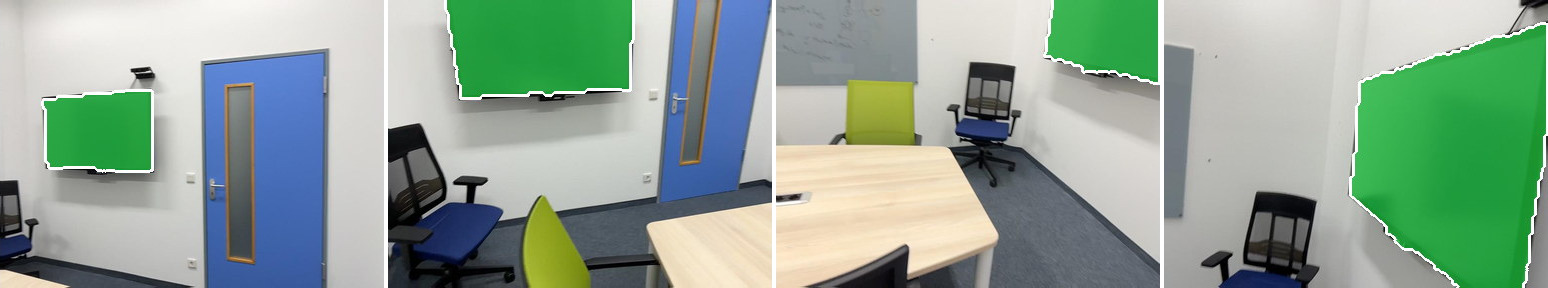} \\
    \groundfullonerowlabel{GT} & \groundfulloneimage{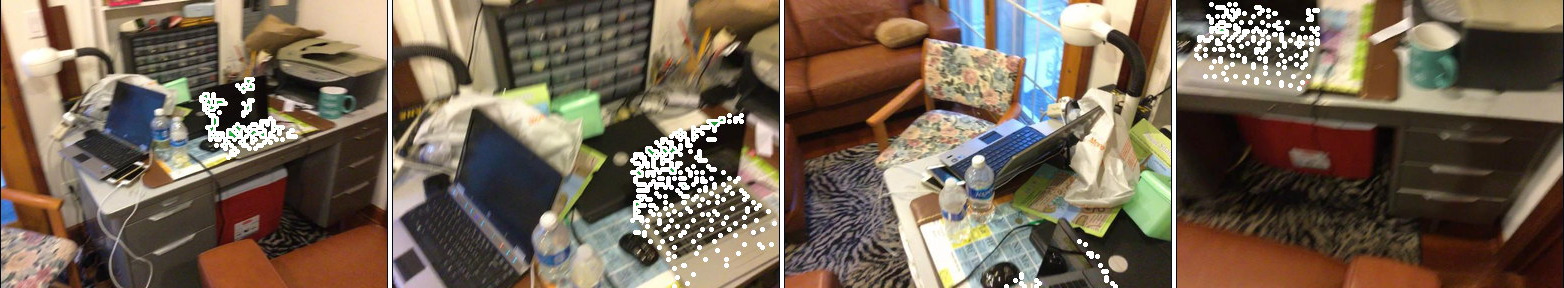} & \groundfulloneimage{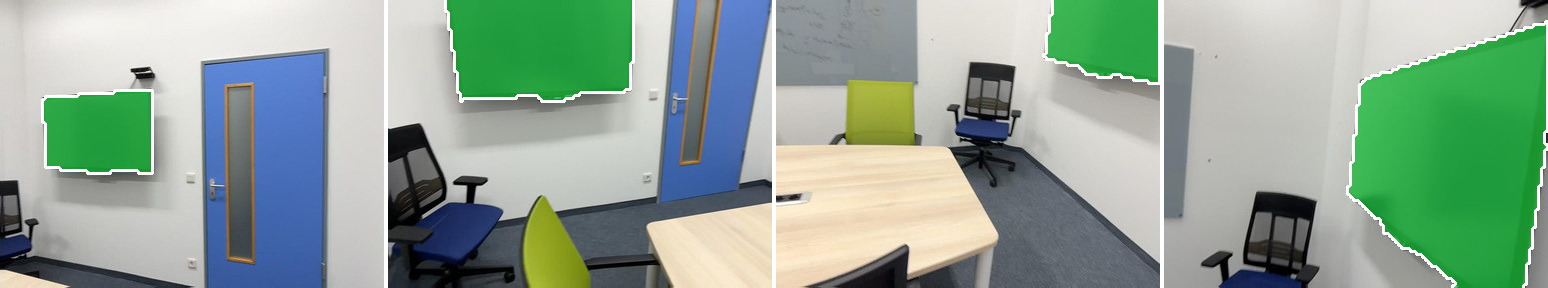}
  \end{tabular}
  \caption{\textbf{Additional qualitative comparisons of language-guided grounding (I).}
  Ground-truth masks are projected from the annotated 3D point clouds.}
  \label{fig:supp-grounding-qualitative-1}
\end{figure}

%% file: arxiv/figures/additional_grounding_full_2.tex
\begin{figure}[p]
  \centering
  \setlength{\tabcolsep}{0pt}
  \renewcommand{\arraystretch}{1.04}
  \newcommand{\groundfulltworowlabel}[1]{%
    \raisebox{\dimexpr5.0mm-.5\height+.5\depth\relax}{%
      \rotatebox[origin=c]{90}{\fontsize{5.8}{6.2}\selectfont #1}}}
  \newcommand{\groundfulltwoimage}[1]{\includegraphics[width=0.445\textwidth]{#1}}
  \begin{tabular}{r@{\hspace{1.0mm}}c@{\hspace{2.0mm}}c}
    & {\fontsize{7}{7.5}\selectfont\textbf{(a)}} & {\fontsize{7}{7.5}\selectfont\textbf{(b)}} \\
    & \parbox[c][9mm][c]{0.445\textwidth}{\centering\fontsize{6}{6.4}\selectfont\emph{``TV with glossy black screen and slim frame, designed for displaying digital video content in a home entertainment setup.''}} &
      \parbox[c][9mm][c]{0.445\textwidth}{\centering\fontsize{6}{6.4}\selectfont\emph{``The large fish swimming near the right side of the tank.''}} \\
    \groundfulltworowlabel{RGB} & \groundfulltwoimage{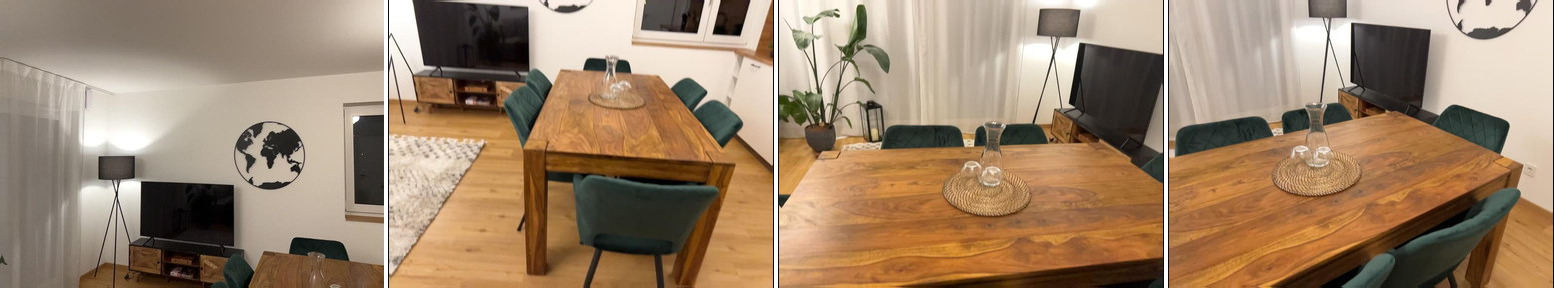} & \groundfulltwoimage{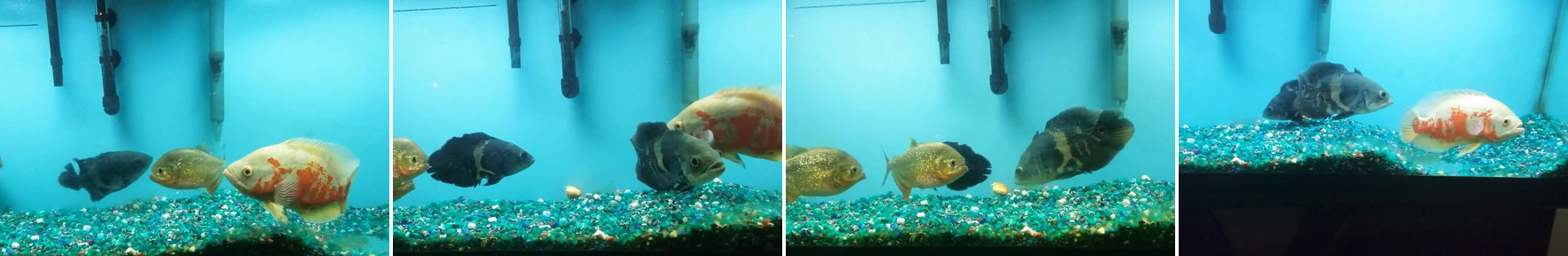} \\
    \groundfulltworowlabel{VLM+SAM2} & \groundfulltwoimage{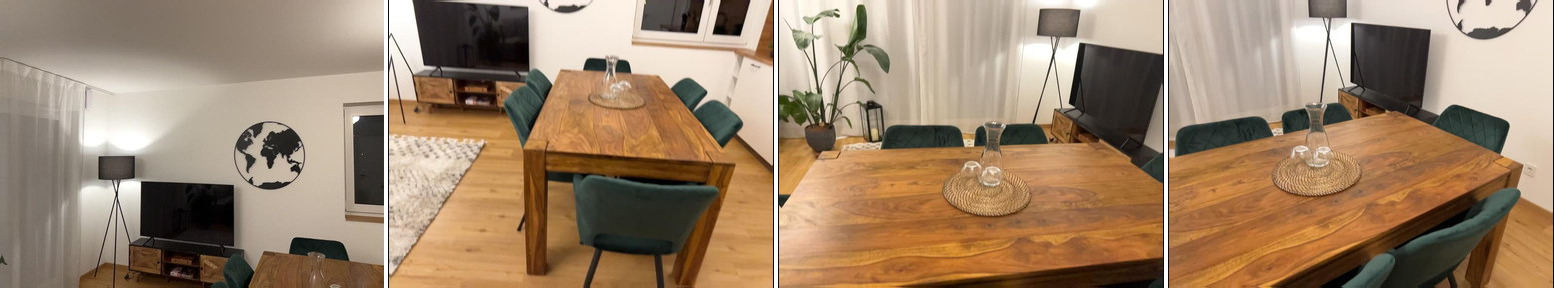} & \groundfulltwoimage{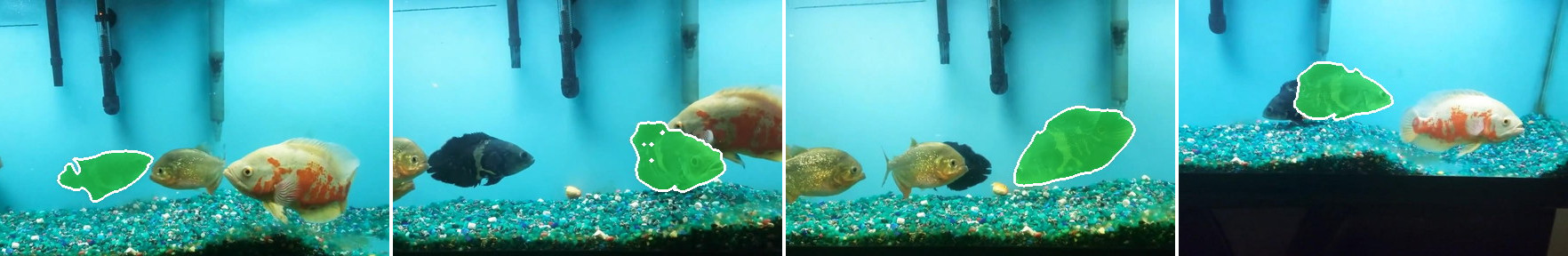} \\
    \groundfulltworowlabel{SAM3} & \groundfulltwoimage{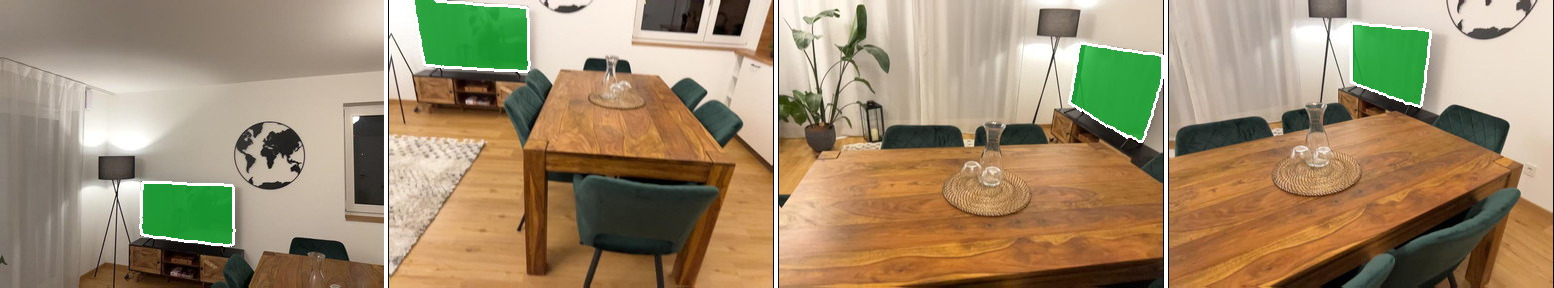} & \groundfulltwoimage{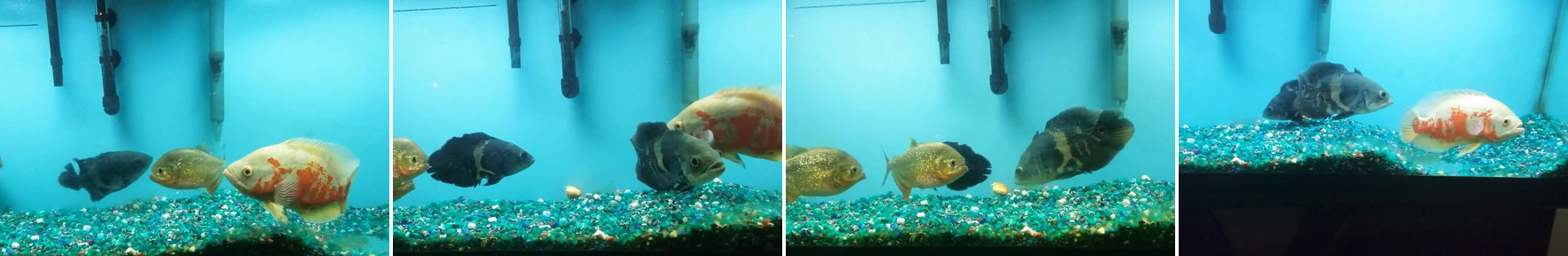} \\
    \groundfulltworowlabel{\textbf{Ours}} & \groundfulltwoimage{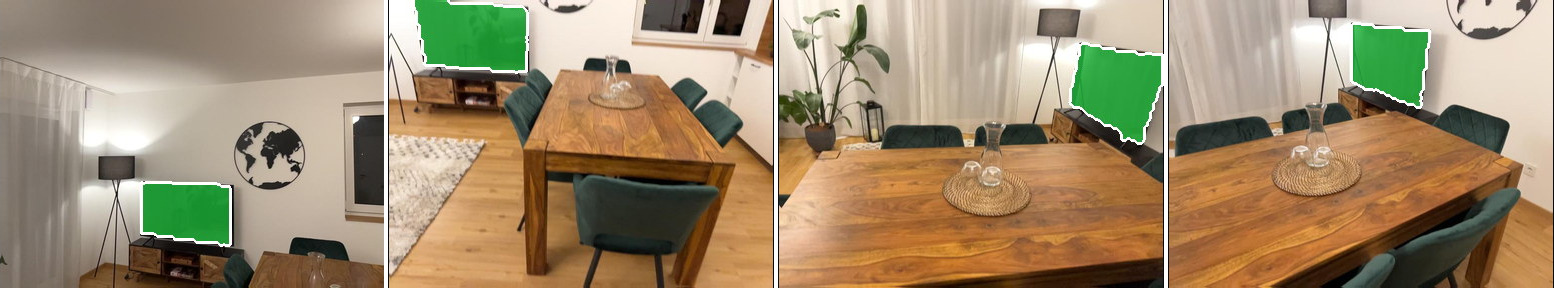} & \groundfulltwoimage{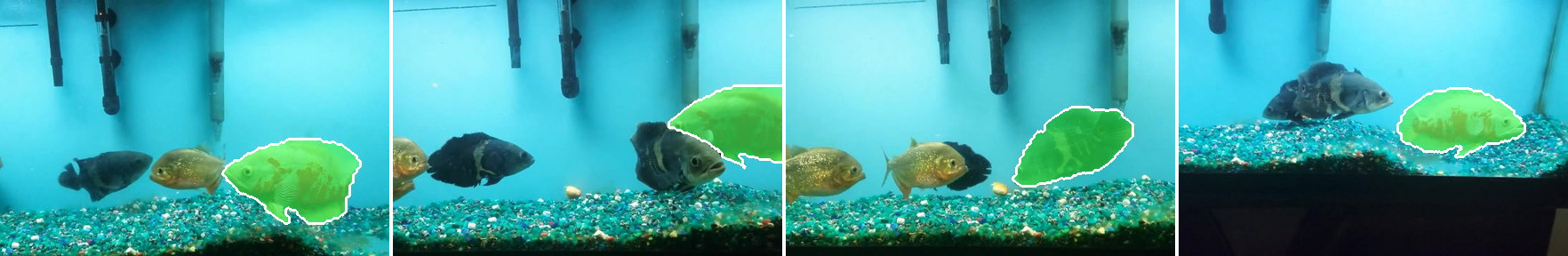} \\
    \groundfulltworowlabel{GT} & \groundfulltwoimage{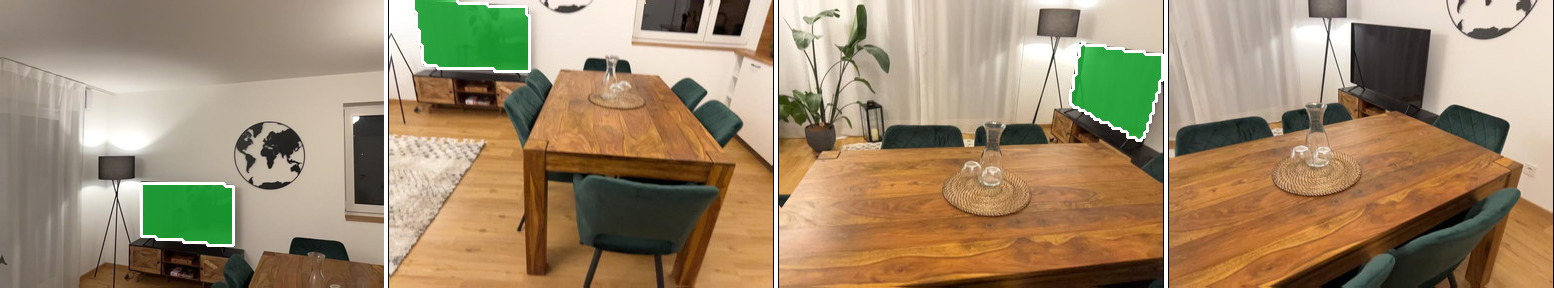} & \groundfulltwoimage{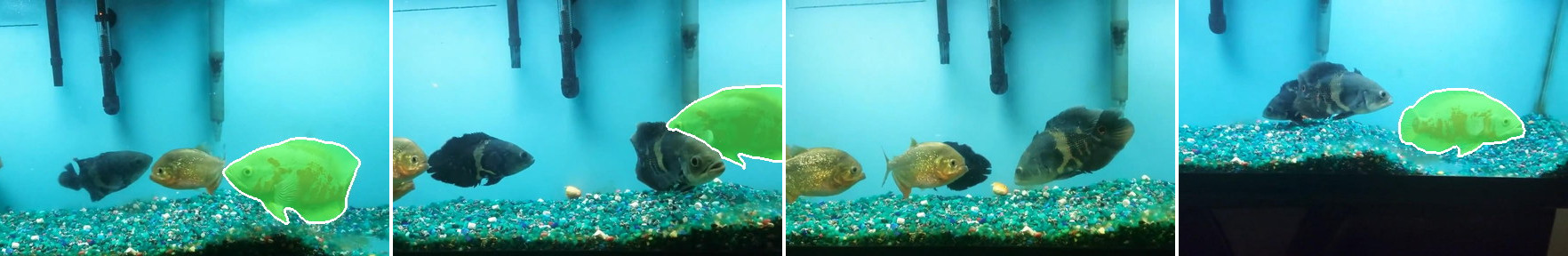} \\[1.8mm]
    & {\fontsize{7}{7.5}\selectfont\textbf{(c)}} & {\fontsize{7}{7.5}\selectfont\textbf{(d)}} \\
    & \parbox[c][9mm][c]{0.445\textwidth}{\centering\fontsize{6}{6.4}\selectfont\emph{``The zebra whose head is cut off by the left side of the image.''}} &
      \parbox[c][9mm][c]{0.445\textwidth}{\centering\fontsize{6}{6.4}\selectfont\emph{``A smallest goldfish.''}} \\
    \groundfulltworowlabel{RGB} & \groundfulltwoimage{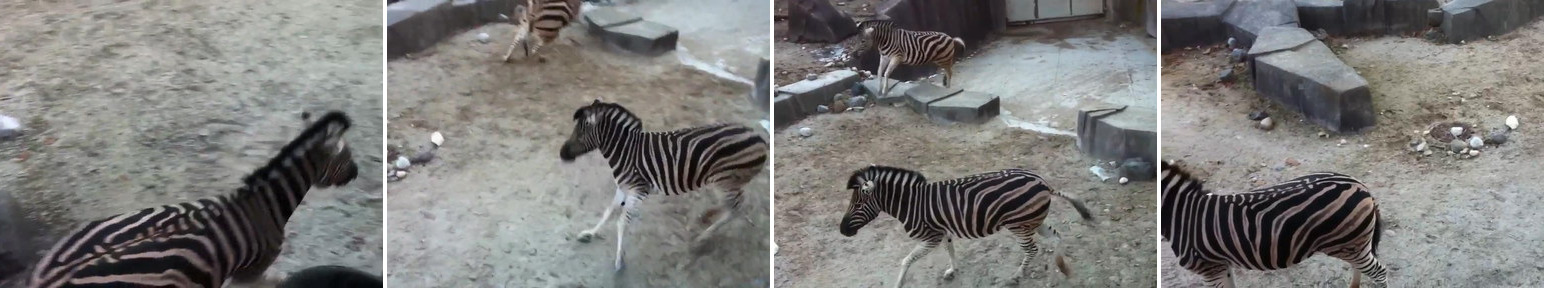} & \groundfulltwoimage{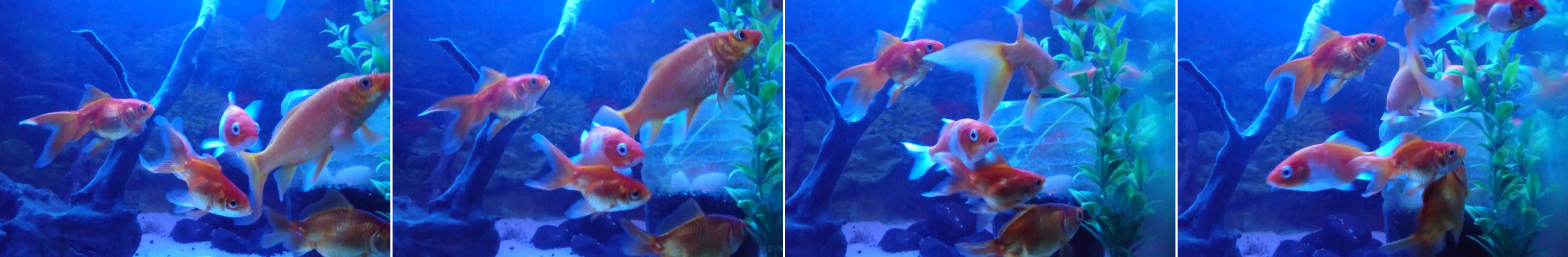} \\
    \groundfulltworowlabel{VLM+SAM2} & \groundfulltwoimage{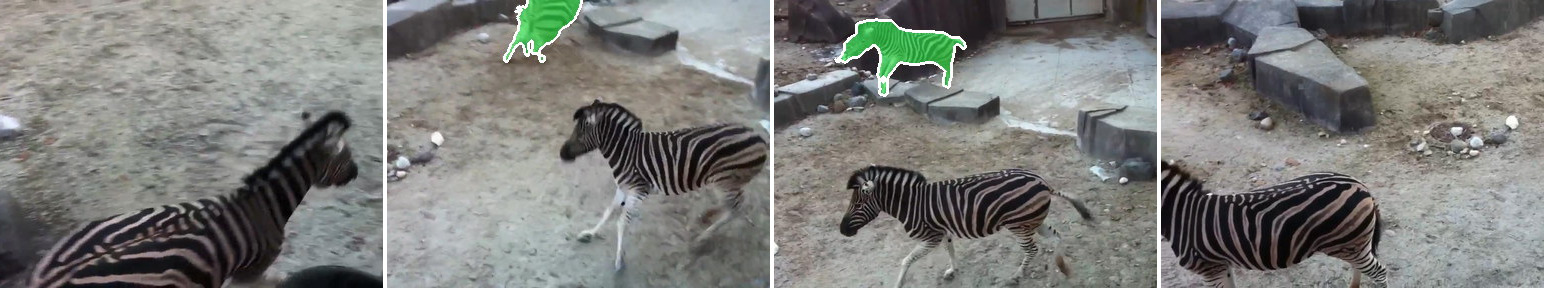} & \groundfulltwoimage{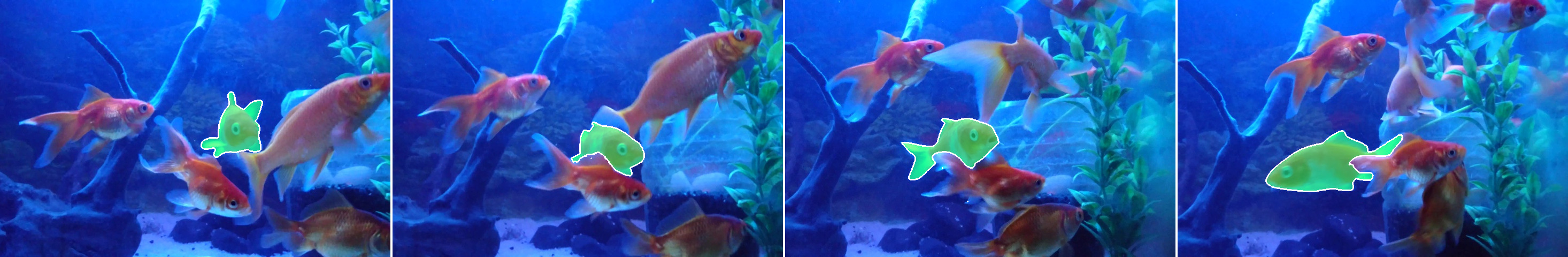} \\
    \groundfulltworowlabel{SAM3} & \groundfulltwoimage{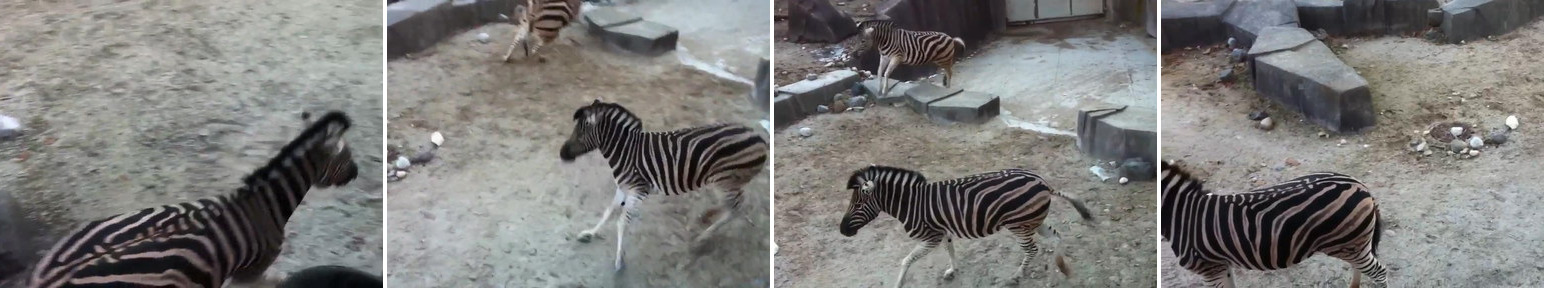} & \groundfulltwoimage{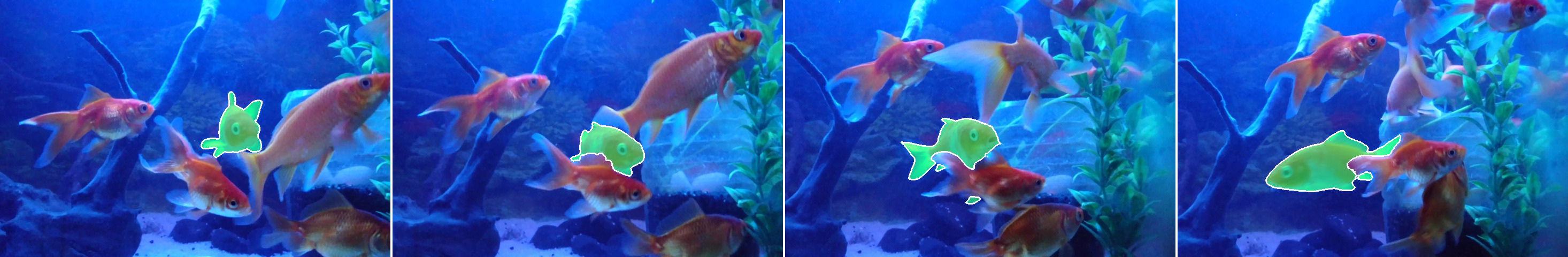} \\
    \groundfulltworowlabel{\textbf{Ours}} & \groundfulltwoimage{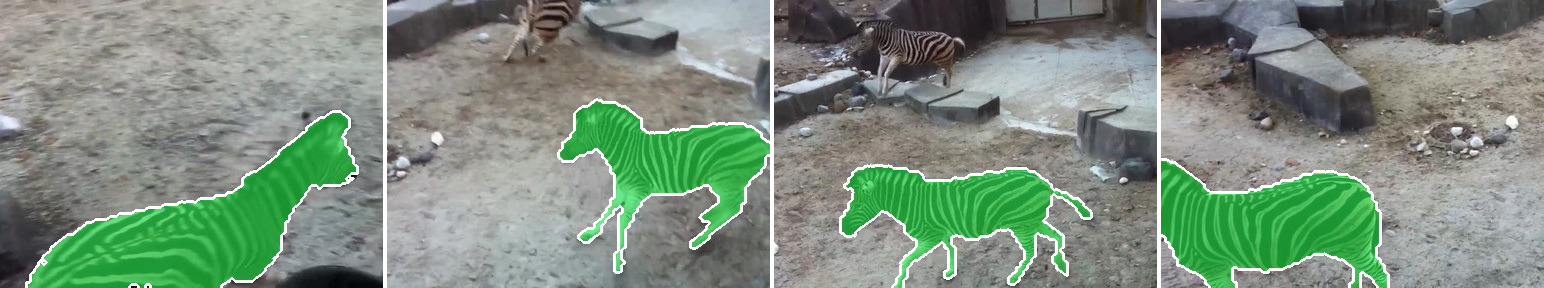} & \groundfulltwoimage{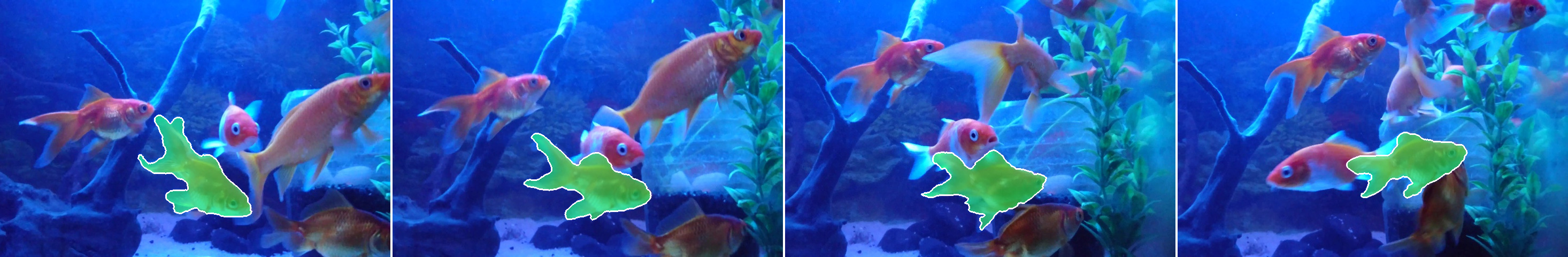} \\
    \groundfulltworowlabel{GT} & \groundfulltwoimage{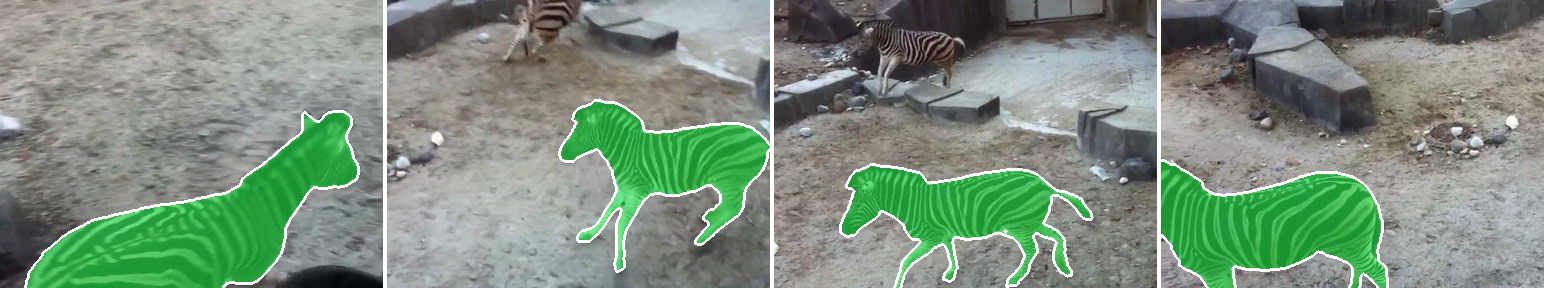} & \groundfulltwoimage{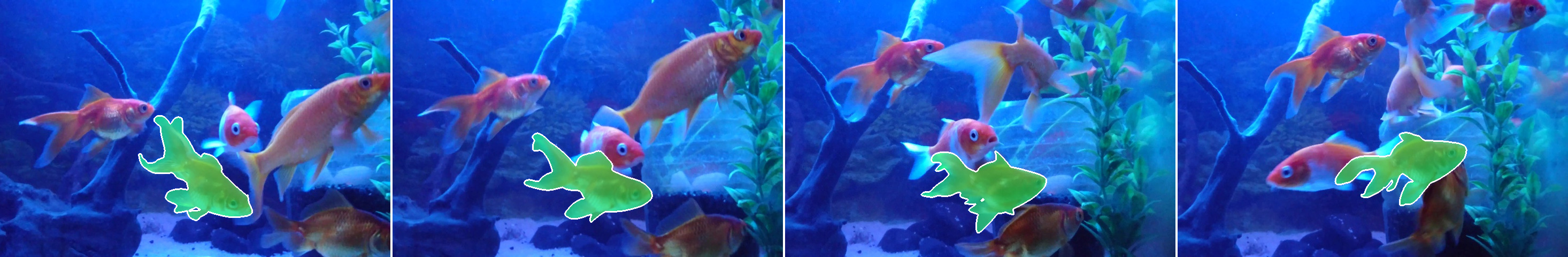}
  \end{tabular}
  \caption{\textbf{Additional qualitative comparisons of language-guided grounding (II).}}
  \label{fig:supp-grounding-qualitative-2}
\end{figure}

%% file: arxiv/figures/inthewild_appendix.tex
\begin{figure}[p]
  \centering
  \setlength{\tabcolsep}{0pt}
  \renewcommand{\arraystretch}{0.94}
  \newcommand{\wildsupprowlabel}[1]{%
    \raisebox{-.5\height}{\rotatebox[origin=c]{90}{\fontsize{6}{6.4}\selectfont #1}}}
  \newcommand{\wildsuppstrip}[1]{%
    \raisebox{-.5\height}{\includegraphics[width=0.92\textwidth]{figures_final/inthewild_qualitative/#1}}}
  \begin{tabular}{@{}>{\centering\arraybackslash}m{4mm}c@{}}
    \multicolumn{2}{c}{\fontsize{7}{7.5}\selectfont\textbf{(a) Road Intersection}} \\
    \wildsupprowlabel{RGB} & \wildsuppstrip{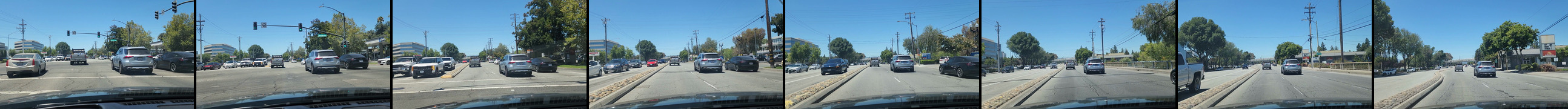} \\
    \wildsupprowlabel{Mask} & \wildsuppstrip{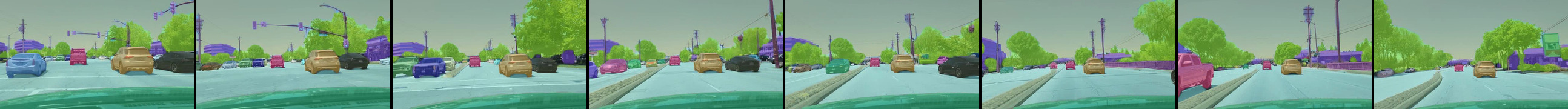} \\
    \multirow{2}{*}{\wildsupprowlabel{4D}} & \wildsuppstrip{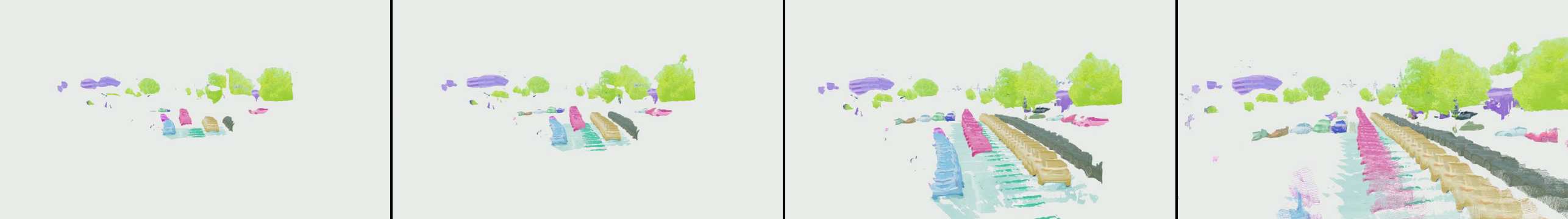} \\
    & \wildsuppstrip{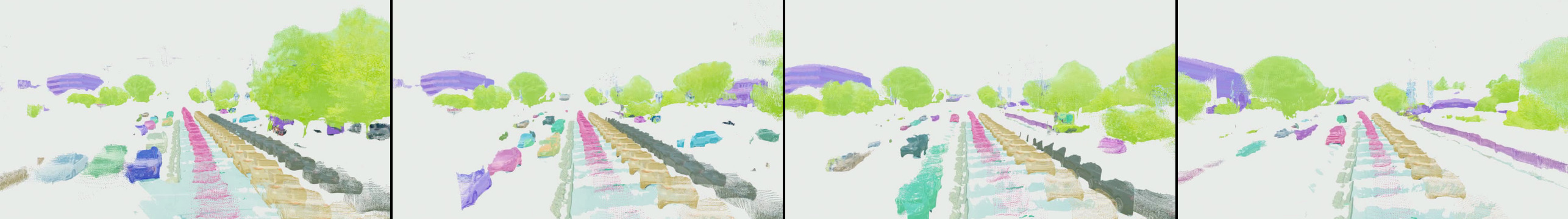} \\[0.7mm]
    \multicolumn{2}{c}{\fontsize{7}{7.5}\selectfont\textbf{(b) Downtown Street}} \\
    \wildsupprowlabel{RGB} & \wildsuppstrip{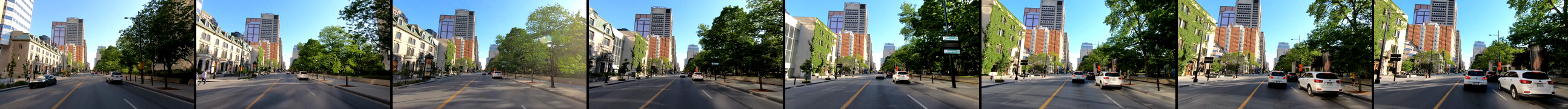} \\
    \wildsupprowlabel{Mask} & \wildsuppstrip{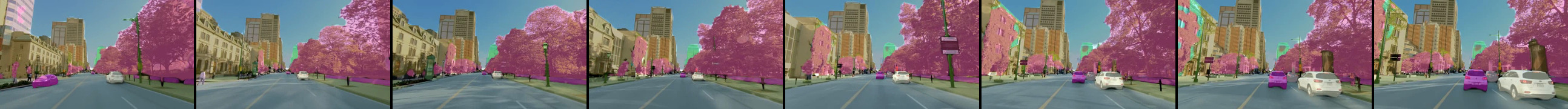} \\
    \multirow{2}{*}{\wildsupprowlabel{4D}} & \wildsuppstrip{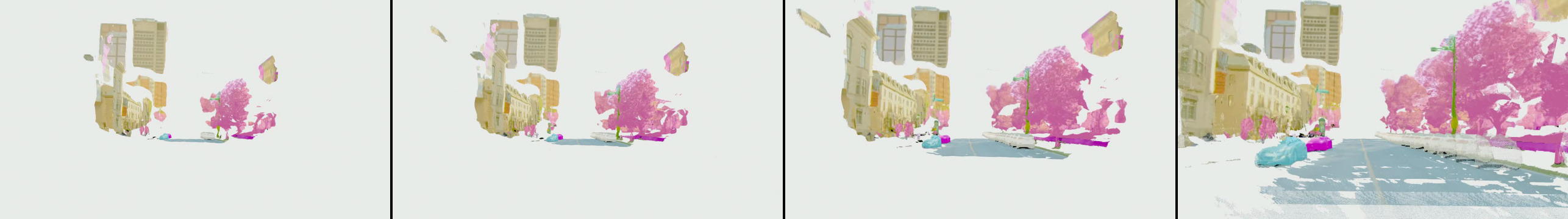} \\
    & \wildsuppstrip{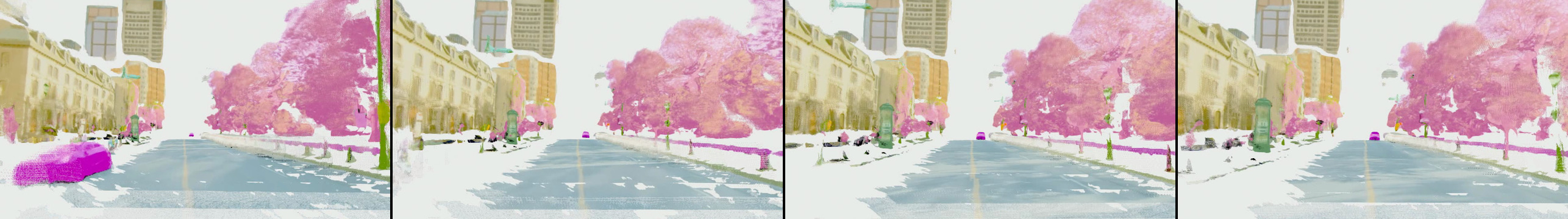} \\[0.7mm]
    \multicolumn{2}{c}{\fontsize{7}{7.5}\selectfont\textbf{(c) Cyclist at an Intersection}} \\
    \wildsupprowlabel{RGB} & \wildsuppstrip{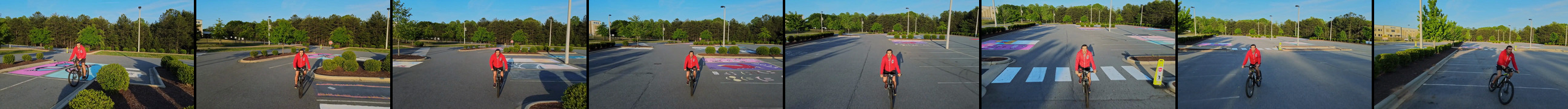} \\
    \wildsupprowlabel{Mask} & \wildsuppstrip{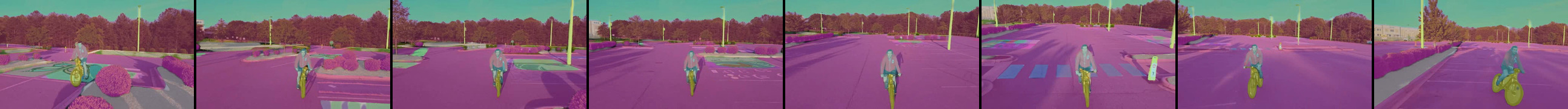} \\
    \multirow{2}{*}{\wildsupprowlabel{4D}} & \wildsuppstrip{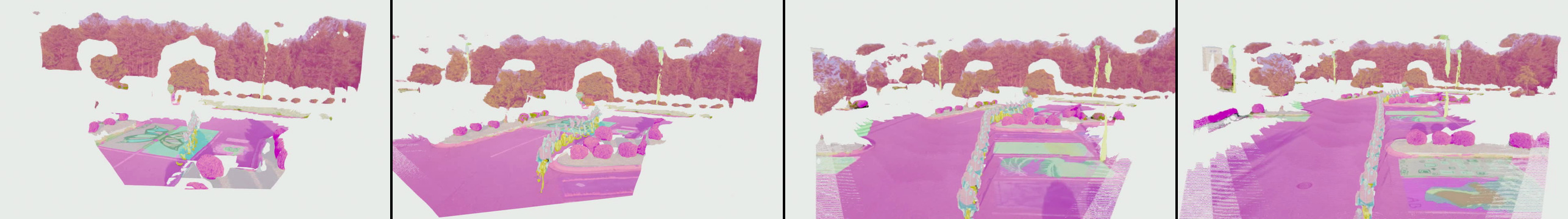} \\
    & \wildsuppstrip{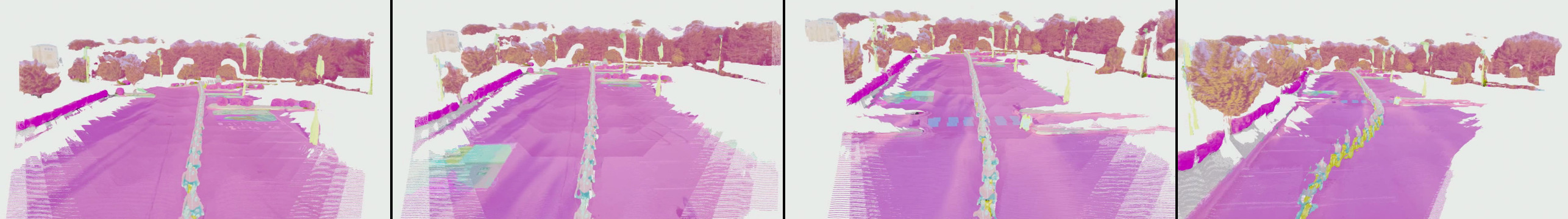}
  \end{tabular}
  \caption{\textbf{Additional in-the-wild 4D segmentation.}
  Uniformly sampled observations are shown with their masks and reconstructed 4D instances in the
  same temporal order.}
  \label{fig:supp-inthewild-qualitative}
\end{figure}

%% file: arxiv/main_arxiv.bbl
\begin{thebibliography}{128}
\providecommand{\natexlab}[1]{#1}
\providecommand{\url}[1]{\texttt{#1}}
\expandafter\ifx\csname urlstyle\endcsname\relax
  \providecommand{\doi}[1]{doi: #1}\else
  \providecommand{\doi}{doi: \begingroup \urlstyle{rm}\Url}\fi

\bibitem[Achlioptas et~al.(2020)Achlioptas, Abdelreheem, Xia, Elhoseiny, and
  Guibas]{achlioptas2020referit3d}
Panos Achlioptas, Ahmed Abdelreheem, Fei Xia, Mohamed Elhoseiny, and Leonidas
  Guibas.
\newblock Referit3d: Neural listeners for fine-grained 3d object identification
  in real-world scenes.
\newblock In \emph{European conference on computer vision}, pages 422--440.
  Springer, 2020.

\bibitem[Aygun et~al.(2021)Aygun, Osep, Weber, Maximov, Stachniss, Behley, and
  Leal-Taix{\'e}]{aygun20214d}
Mehmet Aygun, Aljosa Osep, Mark Weber, Maxim Maximov, Cyrill Stachniss, Jens
  Behley, and Laura Leal-Taix{\'e}.
\newblock 4d panoptic lidar segmentation.
\newblock In \emph{Proceedings of the IEEE/CVF Conference on Computer Vision
  and Pattern Recognition}, pages 5527--5537, 2021.

\bibitem[Bai et~al.(2025)Bai, Cai, Chen, Chen, Chen, Cheng, Deng, Ding, Gao,
  Ge, et~al.]{bai2025qwen3}
Shuai Bai, Yuxuan Cai, Ruizhe Chen, Keqin Chen, Xionghui Chen, Zesen Cheng,
  Lianghao Deng, Wei Ding, Chang Gao, Chunjiang Ge, et~al.
\newblock Qwen3-vl technical report.
\newblock \emph{arXiv preprint arXiv:2511.21631}, 2025.

\bibitem[Baruch et~al.(2021)Baruch, Chen, Dehghan, Dimry, Feigin, Fu, Gebauer,
  Joffe, Kurz, Schwartz, et~al.]{baruch2021arkitscenes}
Gilad Baruch, Zhuoyuan Chen, Afshin Dehghan, Tal Dimry, Yuri Feigin, Peter Fu,
  Thomas Gebauer, Brandon Joffe, Daniel Kurz, Arik Schwartz, et~al.
\newblock Arkitscenes: A diverse real-world dataset for 3d indoor scene
  understanding using mobile rgb-d data.
\newblock \emph{arXiv preprint arXiv:2111.08897}, 2021.

\bibitem[Bhalgat et~al.(2023)Bhalgat, Laina, Henriques, Zisserman, and
  Vedaldi]{bhalgat2023contrastive}
Yash Bhalgat, Iro Laina, Joao~F Henriques, Andrew Zisserman, and Andrea
  Vedaldi.
\newblock Contrastive lift: 3d object instance segmentation by slow-fast
  contrastive fusion.
\newblock \emph{arXiv preprint arXiv:2306.04633}, 2023.

\bibitem[Cabon et~al.(2025)Cabon, Stoffl, Antsfeld, Csurka, Chidlovskii,
  Revaud, and Leroy]{cabon2025must3r}
Yohann Cabon, Lucas Stoffl, Leonid Antsfeld, Gabriela Csurka, Boris
  Chidlovskii, Jerome Revaud, and Vincent Leroy.
\newblock Must3r: Multi-view network for stereo 3d reconstruction.
\newblock In \emph{2025 IEEE/CVF Conference on Computer Vision and Pattern
  Recognition (CVPR)}, pages 1050--1060. IEEE, 2025.

\bibitem[Carion et~al.(2020)Carion, Massa, Synnaeve, Usunier, Kirillov, and
  Zagoruyko]{carion2020end}
Nicolas Carion, Francisco Massa, Gabriel Synnaeve, Nicolas Usunier, Alexander
  Kirillov, and Sergey Zagoruyko.
\newblock End-to-end object detection with transformers.
\newblock In \emph{European conference on computer vision}, pages 213--229.
  Springer, 2020.

\bibitem[Carion et~al.(2026)Carion, Gustafson, Hu, Debnath, Hu, Suris
  Coll-Vinent, Ryali, Alwala, Khedr, Huang, et~al.]{carion2026sam}
Nicolas Carion, Laura Gustafson, Yuan-Ting Hu, Shoubhik Debnath, Ronghang Hu,
  Didac Suris Coll-Vinent, Chaitanya Ryali, Kalyan~Vasudev Alwala, Haitham
  Khedr, Andrew Huang, et~al.
\newblock Sam 3: Segment anything with concepts.
\newblock In \emph{International conference on learning representations},
  volume 2026, pages 138846--138923, 2026.

\bibitem[Cen et~al.(2023)Cen, Zhou, Fang, yang, Shen, Xie, Jiang, ZHANG, and
  Tian]{cen2023segment}
Jiazhong Cen, Zanwei Zhou, Jiemin Fang, chen yang, Wei Shen, Lingxi Xie,
  Dongsheng Jiang, XIAOPENG ZHANG, and Qi~Tian.
\newblock Segment anything in 3d with nerfs.
\newblock In A.~Oh, T.~Naumann, A.~Globerson, K.~Saenko, M.~Hardt, and
  S.~Levine, editors, \emph{Advances in Neural Information Processing Systems},
  volume~36, pages 25971--25990. Curran Associates, Inc., 2023.
\newblock \doi{10.52202/075280-1130}.
\newblock URL
  \url{https://proceedings.neurips.cc/paper_files/paper/2023/file/525d24400247f884c3419b0b7b1c4829-Paper-Conference.pdf}.

\bibitem[Chang et~al.(2017)Chang, Dai, Funkhouser, Halber, Niessner, Savva,
  Song, Zeng, and Zhang]{chang2017matterport3d}
Angel Chang, Angela Dai, Thomas Funkhouser, Maciej Halber, Matthias Niessner,
  Manolis Savva, Shuran Song, Andy Zeng, and Yinda Zhang.
\newblock Matterport3d: Learning from rgb-d data in indoor environments.
\newblock \emph{arXiv preprint arXiv:1709.06158}, 2017.

\bibitem[Chen et~al.(2020)Chen, Chang, and Nie{\ss}ner]{chen2020scanrefer}
Dave~Zhenyu Chen, Angel~X Chang, and Matthias Nie{\ss}ner.
\newblock Scanrefer: 3d object localization in rgb-d scans using natural
  language.
\newblock In \emph{European conference on computer vision}, pages 202--221.
  Springer, 2020.

\bibitem[Chen et~al.(2023)Chen, Zhu, Ding, Cao, Wang, Zhang, Li, Sun, Zang, and
  Mao]{chen2023sam}
Tianrun Chen, Lanyun Zhu, Chaotao Ding, Runlong Cao, Yan Wang, Shangzhan Zhang,
  Zejian Li, Lingyun Sun, Ying Zang, and Papa Mao.
\newblock Sam-adapter: Adapting segment anything in underperformed scenes.
\newblock In \emph{Proceedings of the IEEE/CVF International Conference on
  Computer Vision (ICCV) Workshops}, pages 3367--3375, October 2023.

\bibitem[Chen et~al.(2024)Chen, Yang, Huang, Wang, Xu, Lyu, Lin, and
  Pang]{chen2024grounded}
Yilun Chen, Shuai Yang, Haifeng Huang, Tai Wang, Runsen Xu, Ruiyuan Lyu, Dahua
  Lin, and Jiangmiao Pang.
\newblock Grounded 3d-llm with referent tokens.
\newblock \emph{arXiv preprint arXiv:2405.10370}, 2024.

\bibitem[Cheng et~al.(2021)Cheng, Choudhuri, Misra, Kirillov, Girdhar, and
  Schwing]{cheng2021mask2former}
Bowen Cheng, Anwesa Choudhuri, Ishan Misra, Alexander Kirillov, Rohit Girdhar,
  and Alexander~G Schwing.
\newblock Mask2former for video instance segmentation.
\newblock \emph{arXiv preprint arXiv:2112.10764}, 2021.

\bibitem[Cheng et~al.(2022)Cheng, Misra, Schwing, Kirillov, and
  Girdhar]{cheng2022masked}
Bowen Cheng, Ishan Misra, Alexander~G Schwing, Alexander Kirillov, and Rohit
  Girdhar.
\newblock Masked-attention mask transformer for universal image segmentation.
\newblock In \emph{2022 IEEE/CVF Conference on Computer Vision and Pattern
  Recognition (CVPR)}, pages 1280--1289. IEEE, 2022.

\bibitem[Cheng et~al.(2023)Cheng, Oh, Price, Schwing, and
  Lee]{cheng2023tracking}
Ho~Kei Cheng, Seoung~Wug Oh, Brian Price, Alexander Schwing, and Joon-Young
  Lee.
\newblock Tracking anything with decoupled video segmentation.
\newblock In \emph{2023 IEEE/CVF International Conference on Computer Vision
  (ICCV)}, pages 1316--1326. IEEE, 2023.

\bibitem[Choy et~al.(2026)Choy, Lee, Park, Cho, and Kautz]{choy2026spaceformer}
Chris Choy, Junha Lee, Chunghyun Park, Minsu Cho, and Jan Kautz.
\newblock Spaceformer: Fast proposal-free open-vocabulary 3d instance
  segmentation.
\newblock \emph{arXiv preprint arXiv:2604.20395}, 2026.

\bibitem[Choy et~al.(2019)Choy, Gwak, and Savarese]{choy20194d}
Christopher Choy, JunYoung Gwak, and Silvio Savarese.
\newblock 4d spatio-temporal convnets: Minkowski convolutional neural networks.
\newblock In \emph{2019 IEEE/CVF conference on computer vision and pattern
  recognition (CVPR)}, pages 3070--3079. IEEE, 2019.

\bibitem[Clark et~al.(2026)Clark, Zhang, Ma, Park, Tripathi, Lee, Salehi, Ren,
  Kim, Yang, Shao, Yang, Huang, Gao, Anderson, Zhang, Jain, Stoica, Farhadi,
  and Krishna]{clark2026molmo2}
Christopher Clark, Jieyu Zhang, Zixian Ma, Jae~Sung Park, Rohun Tripathi,
  Sangho Lee, Mohammadreza Salehi, Jason Ren, Chris~Dongjoo Kim, Yinuo Yang,
  Vincent Shao, Yue Yang, Weikai Huang, Ziqi Gao, Taira Anderson, Jianrui
  Zhang, Jitesh Jain, George Stoica, Ali Farhadi, and Ranjay Krishna.
\newblock Molmo2: Open weights and data for vision-language models with video
  understanding and grounding.
\newblock In \emph{Proceedings of the IEEE/CVF Conference on Computer Vision
  and Pattern Recognition (CVPR)}, pages 28652--28668, June 2026.

\bibitem[Darkhalil et~al.(2022)Darkhalil, Shan, Zhu, Ma, Kar, Higgins, Fidler,
  Fouhey, and Damen]{darkhalil2022epic}
Ahmad Darkhalil, Dandan Shan, Bin Zhu, Jian Ma, Amlan Kar, Richard Higgins,
  Sanja Fidler, David Fouhey, and Dima Damen.
\newblock Epic-kitchens visor benchmark: Video segmentations and object
  relations.
\newblock \emph{Advances in Neural Information Processing Systems},
  35:\penalty0 13745--13758, 2022.

\bibitem[Ding et~al.(2023{\natexlab{a}})Ding, Liu, He, Jiang, and
  Loy]{ding2023mevis}
Henghui Ding, Chang Liu, Shuting He, Xudong Jiang, and Chen~Change Loy.
\newblock Mevis: A large-scale benchmark for video segmentation with motion
  expressions.
\newblock In \emph{2023 IEEE/CVF International Conference on Computer Vision
  (ICCV)}, pages 2694--2703. IEEE, 2023{\natexlab{a}}.

\bibitem[Ding et~al.(2023{\natexlab{b}})Ding, Liu, He, Jiang, Torr, and
  Bai]{ding2023mose}
Henghui Ding, Chang Liu, Shuting He, Xudong Jiang, Philip~HS Torr, and Song
  Bai.
\newblock Mose: A new dataset for video object segmentation in complex scenes.
\newblock In \emph{2023 IEEE/CVF International Conference on Computer Vision
  (ICCV)}, pages 20167--20177. IEEE, 2023{\natexlab{b}}.

\bibitem[Drozdov et~al.(2026)Drozdov, Lemeshko, Gavrilov, Konushin, Rukhovich,
  and Kolodiazhnyi]{drozdov2026z3d}
Nikita Drozdov, Andrey Lemeshko, Nikita Gavrilov, Anton Konushin, Danila
  Rukhovich, and Maksim Kolodiazhnyi.
\newblock Z3d: Zero-shot 3d visual grounding from images.
\newblock In \emph{Proceedings of the 64th Annual Meeting of the Association
  for Computational Linguistics (Volume 2: Short Papers)}, pages 147--154,
  2026.

\bibitem[Fan et~al.(2024)Fan, Zhang, Cong, Wang, Li, Wen, Zhou, Kadambi, Wang,
  Xu, et~al.]{fan2024large}
Zhiwen Fan, Jian Zhang, Wenyan Cong, Peihao Wang, Renjie Li, Kairun Wen, Shijie
  Zhou, Achuta Kadambi, Zhangyang Wang, Danfei Xu, et~al.
\newblock Large spatial model: End-to-end unposed images to semantic 3d.
\newblock \emph{Advances in neural information processing systems},
  37:\penalty0 40212--40229, 2024.

\bibitem[Fiebelman et~al.(2025)Fiebelman, Cohen, Morgenstern, Hedman, and
  Averbuch-Elor]{fiebelman20254}
Gal Fiebelman, Tamir Cohen, Ayellet Morgenstern, Peter Hedman, and Hadar
  Averbuch-Elor.
\newblock 4-legs: 4d language embedded gaussian splatting.
\newblock \emph{Computer Graphics Forum}, 44\penalty0 (2):\penalty0 e70085,
  2025.
\newblock \doi{10.1111/cgf.70085}.

\bibitem[Gong et~al.(2026)Gong, Wu, Fu, Xiao, Zou, Wang, and
  Hoiem]{gong2026samv}
Jiangshan Gong, Yuqun Wu, Qiqian Fu, Yao Xiao, Chuhang Zou, Shenlong Wang, and
  Derek Hoiem.
\newblock {SAM-V}: Geometry-aware segment anything for multi-view instance
  segmentation.
\newblock \emph{arXiv preprint arXiv:2609.25490}, 2026.

\bibitem[He et~al.(2013)He, Sun, and Tang]{he2012guided}
Kaiming He, Jian Sun, and Xiaoou Tang.
\newblock Guided image filtering.
\newblock \emph{IEEE Transactions on Pattern Analysis and Machine
  Intelligence}, 35\penalty0 (6):\penalty0 1397--1409, 2013.
\newblock \doi{10.1109/TPAMI.2012.213}.

\bibitem[Hong et~al.(2023)Hong, Chen, Liu, Zhang, Guo, Chen, and
  Zhang]{hong2023lvos}
Lingyi Hong, Wenchao Chen, Zhongying Liu, Wei Zhang, Pinxue Guo, Zhaoyu Chen,
  and Wenqiang Zhang.
\newblock Lvos: A benchmark for long-term video object segmentation.
\newblock In \emph{2023 IEEE/CVF International Conference on Computer Vision
  (ICCV)}, pages 13434--13446. IEEE, 2023.

\bibitem[Hong et~al.(2025)Hong, Yu, Gu, Wang, Gan, Tang, Cheng, Qi, Ji, Pan,
  et~al.]{hong2025glm}
Wenyi Hong, Wenmeng Yu, Xiaotao Gu, Guo Wang, Guobing Gan, Haomiao Tang, Jiale
  Cheng, Ji~Qi, Junhui Ji, Lihang Pan, et~al.
\newblock Glm-4.5 v and glm-4.1 v-thinking: Towards versatile multimodal
  reasoning with scalable reinforcement learning.
\newblock \emph{arXiv preprint arXiv:2507.01006}, 2025.

\bibitem[Hu et~al.(2021)Hu, Shen, Wallis, Allen-Zhu, Li, Wang, Wang, and
  Chen]{hu2021lora}
Edward~J Hu, Yelong Shen, Phillip Wallis, Zeyuan Allen-Zhu, Yuanzhi Li, Shean
  Wang, Lu~Wang, and Weizhu Chen.
\newblock Lora: Low-rank adaptation of large language models.
\newblock \emph{arXiv preprint arXiv:2106.09685}, 2021.

\bibitem[Hu et~al.(2026)Hu, Lin, Long, Ran, Jiang, Wang, Zhu, Xu, Wang, and
  Pang]{hu2026g}
Wenbo Hu, Jingli Lin, Yilin Long, Yunlong Ran, Lihan Jiang, Yifan Wang,
  Chenming Zhu, Runsen Xu, Tai Wang, and Jiangmiao Pang.
\newblock {G$^2$VLM}: Geometry grounded vision language model with unified 3d
  reconstruction and spatial reasoning.
\newblock In \emph{Proceedings of the IEEE/CVF Conference on Computer Vision
  and Pattern Recognition}, pages 9535--9546, 2026.

\bibitem[Hu et~al.(2019)Hu, Chen, Hui, Huang, and Schwing]{hu2019sail}
Yuan-Ting Hu, Hong-Shuo Chen, Kexin Hui, Jia-Bin Huang, and Alexander~G
  Schwing.
\newblock Sail-vos: Semantic amodal instance level video object segmentation--a
  synthetic dataset and baselines.
\newblock In \emph{2019 IEEE/CVF Conference on Computer Vision and Pattern
  Recognition (CVPR)}, pages 3100--3110. IEEE, 2019.

\bibitem[Ji et~al.(2024)Ji, Wu, Fang, Cen, Yi, Liu, Tian, and
  Wang]{ji2024segment}
Shengxiang Ji, Guanjun Wu, Jiemin Fang, Jiazhong Cen, Taoran Yi, Wenyu Liu,
  Qi~Tian, and Xinggang Wang.
\newblock Segment any 4d gaussians.
\newblock \emph{arXiv preprint arXiv:2407.04504}, 2024.

\bibitem[Jiang et~al.(2020)Jiang, Zhao, Shi, Liu, Fu, and
  Jia]{jiang2020pointgroup}
Li~Jiang, Hengshuang Zhao, Shaoshuai Shi, Shu Liu, Chi-Wing Fu, and Jiaya Jia.
\newblock Pointgroup: Dual-set point grouping for 3d instance segmentation.
\newblock In \emph{2020 IEEE/CVF Conference on Computer Vision and Pattern
  Recognition (CVPR)}, pages 4866--4875. IEEE, 2020.

\bibitem[Karaev et~al.(2023)Karaev, Rocco, Graham, Neverova, Vedaldi, and
  Rupprecht]{karaev2023dynamicstereo}
Nikita Karaev, Ignacio Rocco, Benjamin Graham, Natalia Neverova, Andrea
  Vedaldi, and Christian Rupprecht.
\newblock Dynamicstereo: Consistent dynamic depth from stereo videos.
\newblock In \emph{2023 IEEE/CVF Conference on Computer Vision and Pattern
  Recognition (CVPR)}, pages 13229--13239. IEEE, 2023.

\bibitem[Ke et~al.(2023)Ke, Ye, Danelljan, liu, Tai, Tang, and
  Yu]{ke2023segment}
Lei Ke, Mingqiao Ye, Martin Danelljan, Yifan liu, Yu-Wing Tai, Chi-Keung Tang,
  and Fisher Yu.
\newblock Segment anything in high quality.
\newblock In A.~Oh, T.~Naumann, A.~Globerson, K.~Saenko, M.~Hardt, and
  S.~Levine, editors, \emph{Advances in Neural Information Processing Systems},
  volume~36, pages 29914--29934. Curran Associates, Inc., 2023.
\newblock \doi{10.52202/075280-1303}.
\newblock URL
  \url{https://proceedings.neurips.cc/paper_files/paper/2023/file/5f828e38160f31935cfe9f67503ad17c-Paper-Conference.pdf}.

\bibitem[Khoreva et~al.(2018)Khoreva, Rohrbach, and Schiele]{khoreva2018video}
Anna Khoreva, Anna Rohrbach, and Bernt Schiele.
\newblock Video object segmentation with language referring expressions.
\newblock In \emph{Asian conference on computer vision}, pages 123--141.
  Springer, 2018.

\bibitem[Kim et~al.(2020)Kim, Woo, Lee, and Kweon]{kim2020video}
Dahun Kim, Sanghyun Woo, Joon-Young Lee, and In~So Kweon.
\newblock Video panoptic segmentation.
\newblock In \emph{2020 IEEE/CVF Conference on Computer Vision and Pattern
  Recognition (CVPR)}, pages 9856--9865. IEEE, 2020.

\bibitem[Kirillov et~al.(2023)Kirillov, Mintun, Ravi, Mao, Rolland, Gustafson,
  Xiao, Whitehead, Berg, Lo, et~al.]{kirillov2023segment}
Alexander Kirillov, Eric Mintun, Nikhila Ravi, Hanzi Mao, Chloe Rolland, Laura
  Gustafson, Tete Xiao, Spencer Whitehead, Alexander~C Berg, Wan-Yen Lo, et~al.
\newblock Segment anything.
\newblock In \emph{2023 IEEE/CVF international conference on computer vision
  (ICCV)}, pages 3992--4003. IEEE, 2023.

\bibitem[Kolodiazhnyi et~al.(2024)Kolodiazhnyi, Vorontsova, Konushin, and
  Rukhovich]{kolodiazhnyi2024oneformer3d}
Maxim Kolodiazhnyi, Anna Vorontsova, Anton Konushin, and Danila Rukhovich.
\newblock Oneformer3d: One transformer for unified point cloud segmentation.
\newblock In \emph{2024 IEEE/CVF Conference on Computer Vision and Pattern
  Recognition (CVPR)}, pages 20943--20953. IEEE, 2024.

\bibitem[Kreuzberg et~al.(2023)Kreuzberg, Zulfikar, Mahadevan, Engelmann, and
  Leibe]{kreuzberg20224d}
Lars Kreuzberg, Idil~Esen Zulfikar, Sabarinath Mahadevan, Francis Engelmann,
  and Bastian Leibe.
\newblock 4d-stop: Panoptic segmentation of 4d lidar using spatio-temporal
  object proposal generation and aggregation.
\newblock In Leonid Karlinsky, Tomer Michaeli, and Ko~Nishino, editors,
  \emph{Computer Vision -- ECCV 2022 Workshops}, pages 537--553, Cham, 2023.
  Springer Nature Switzerland.
\newblock ISBN 978-3-031-25056-9.

\bibitem[Lai et~al.(2024)Lai, Tian, Chen, Li, Yuan, Liu, and Jia]{lai2024lisa}
Xin Lai, Zhuotao Tian, Yukang Chen, Yanwei Li, Yuhui Yuan, Shu Liu, and Jiaya
  Jia.
\newblock Lisa: Reasoning segmentation via large language model.
\newblock In \emph{2024 IEEE/CVF Conference on Computer Vision and Pattern
  Recognition (CVPR)}, pages 9579--9589. IEEE, 2024.

\bibitem[Le et~al.(2024)Le, Gou, Datta, Shi, Reid, Cai, and
  Rezatofighi]{le2024jrdb}
Duy~Tho Le, Chenhui Gou, Stavya Datta, Hengcan Shi, Ian Reid, Jianfei Cai, and
  Hamid Rezatofighi.
\newblock Jrdb-panotrack: An open-world panoptic segmentation and tracking
  robotic dataset in crowded human environments.
\newblock In \emph{2024 IEEE/CVF Conference on Computer Vision and Pattern
  Recognition (CVPR)}, pages 22325--22334. IEEE, 2024.

\bibitem[Leroy et~al.(2024)Leroy, Cabon, and Revaud]{leroy2024grounding}
Vincent Leroy, Yohann Cabon, and J{\'e}r{\^o}me Revaud.
\newblock Grounding image matching in 3d with mast3r.
\newblock In \emph{European conference on computer vision}, pages 71--91.
  Springer, 2024.

\bibitem[Li et~al.(2026{\natexlab{a}})Li, Huang, Chng, Zhan, Yan, and
  Xu]{li2026fast3dis}
Changyang Li, Xueqing Huang, Shin-Fang Chng, Huangying Zhan, Qingan Yan, and
  Yi~Xu.
\newblock Fast3dis: Feed-forward anchored scene transformer for 3d instance
  segmentation.
\newblock \emph{arXiv preprint arXiv:2603.25993}, 2026{\natexlab{a}}.

\bibitem[Li et~al.(2026{\natexlab{b}})Li, Yang, Zhou, Wu, Qian, and
  Zhang]{li20264dvlt}
Chaoyue Li, Boxue Yang, Shengyao Zhou, Haoyang Wu, Rui Qian, and Linfeng Zhang.
\newblock 4dvlt: Dynamic scene understanding with worldline-centered
  vision-language tracking.
\newblock \emph{arXiv preprint arXiv:2606.22631}, 2026{\natexlab{b}}.

\bibitem[Li et~al.(2024{\natexlab{a}})Li, Zhang, Sun, Zou, Liu, Li, Yang,
  Zhang, and Gao]{li2024segment}
Feng Li, Hao Zhang, Peize Sun, Xueyan Zou, Shilong Liu, Chunyuan Li, Jianwei
  Yang, Lei Zhang, and Jianfeng Gao.
\newblock Segment and recognize anything at any granularity.
\newblock In \emph{European Conference on Computer Vision}, pages 467--484.
  Springer, 2024{\natexlab{a}}.

\bibitem[Li et~al.(2025{\natexlab{a}})Li, Zou, Liu, Zhang, Hong, Cao, Lan,
  Zhang, Yu, Zhang, et~al.]{li2025iggt}
Hao Li, Zhengyu Zou, Fangfu Liu, Xuanyang Zhang, Fangzhou Hong, Yukang Cao,
  Yushi Lan, Manyuan Zhang, Gang Yu, Dingwen Zhang, et~al.
\newblock Iggt: Instance-grounded geometry transformer for semantic 3d
  reconstruction.
\newblock \emph{arXiv preprint arXiv:2510.22706}, 2025{\natexlab{a}}.

\bibitem[Li et~al.(2025{\natexlab{b}})Li, Feng, Guo, Huang, Ji, Bi, Piao,
  Zhang, Zhao, Chen, et~al.]{li2025sam3}
Jingjing Li, Yue Feng, Yuchen Guo, Jincai Huang, Wei Ji, Qi~Bi, Yongri Piao,
  Miao Zhang, Xiaoqi Zhao, Qiang Chen, et~al.
\newblock Sam3-i: Segment anything with instructions.
\newblock \emph{arXiv preprint arXiv:2512.04585}, 2025{\natexlab{b}}.

\bibitem[Li et~al.(2026{\natexlab{c}})Li, Li, Shen, Liu, Chang, and
  Shan]{li2026lenswalk}
Keliang Li, Yansong Li, Hongze Shen, Mengdi Liu, Hong Chang, and Shiguang Shan.
\newblock Lenswalk: Agentic video understanding by planning how you see in
  videos.
\newblock \emph{arXiv preprint arXiv:2603.24558}, 2026{\natexlab{c}}.

\bibitem[Li et~al.(2025{\natexlab{c}})Li, Li, Kong, Yang, and
  Liang]{li2025seeground}
Rong Li, Shijie Li, Lingdong Kong, Xulei Yang, and Junwei Liang.
\newblock Seeground: See and ground for zero-shot open-vocabulary 3d visual
  grounding.
\newblock In \emph{2025 IEEE/CVF Conference on Computer Vision and Pattern
  Recognition (CVPR)}, pages 3707--3717. IEEE, 2025{\natexlab{c}}.

\bibitem[Li et~al.(2025{\natexlab{d}})Li, Zhou, Zhou, Song, Herter, Qin, Huang,
  and Pfister]{li20254d}
Wanhua Li, Renping Zhou, Jiawei Zhou, Yingwei Song, Johannes Herter, Minghan
  Qin, Gao Huang, and Hanspeter Pfister.
\newblock 4d langsplat: 4d language gaussian splatting via multimodal large
  language models.
\newblock In \emph{2025 IEEE/CVF Conference on Computer Vision and Pattern
  Recognition (CVPR)}, pages 22001--22011. IEEE, 2025{\natexlab{d}}.

\bibitem[Li et~al.(2024{\natexlab{b}})Li, Yuan, Li, Ding, Wu, Zhang, Li, Chen,
  and Loy]{li2024omg}
Xiangtai Li, Haobo Yuan, Wei Li, Henghui Ding, Size Wu, Wenwei Zhang, Yining
  Li, Kai Chen, and Chen~Change Loy.
\newblock Omg-seg: Is one model good enough for all segmentation?
\newblock In \emph{2024 IEEE/CVF Conference on Computer Vision and Pattern
  Recognition (CVPR)}, pages 27948--27959. IEEE, 2024{\natexlab{b}}.

\bibitem[Li et~al.(2026{\natexlab{d}})Li, Gladkova, Xia, and
  Cremers]{li2026trase}
Yun-Jin Li, Mariia Gladkova, Yan Xia, and Daniel Cremers.
\newblock Trase: Tracking-free 4d segmentation and editing.
\newblock In \emph{2026 International Conference on 3D Vision (3DV)}, pages
  1670--1679. IEEE, 2026{\natexlab{d}}.

\bibitem[Ling et~al.(2024)Ling, Sheng, Tu, Zhao, Xin, Wan, Yu, Guo, Yu, Lu,
  et~al.]{ling2024dl3dv}
Lu~Ling, Yichen Sheng, Zhi Tu, Wentian Zhao, Cheng Xin, Kun Wan, Lantao Yu,
  Qianyu Guo, Zixun Yu, Yawen Lu, et~al.
\newblock Dl3dv-10k: A large-scale scene dataset for deep learning-based 3d
  vision.
\newblock In \emph{2024 IEEE/CVF Conference on Computer Vision and Pattern
  Recognition (CVPR)}, pages 22160--22169. IEEE, 2024.

\bibitem[Liu et~al.(2024)Liu, Zeng, Ren, Li, Zhang, Yang, Jiang, Li, Yang, Su,
  et~al.]{liu2024grounding}
Shilong Liu, Zhaoyang Zeng, Tianhe Ren, Feng Li, Hao Zhang, Jie Yang, Qing
  Jiang, Chunyuan Li, Jianwei Yang, Hang Su, et~al.
\newblock Grounding dino: Marrying dino with grounded pre-training for open-set
  object detection.
\newblock In \emph{European conference on computer vision}, pages 38--55.
  Springer, 2024.

\bibitem[Liu et~al.(2023)Liu, Kong, Cen, Chen, Zhang, Pan, Chen, and
  Liu]{liu2023segment}
Youquan Liu, Lingdong Kong, Jun Cen, Runnan Chen, Wenwei Zhang, Liang Pan, Kai
  Chen, and Ziwei Liu.
\newblock Segment any point cloud sequences by distilling vision foundation
  models.
\newblock \emph{Advances in Neural Information Processing Systems},
  36:\penalty0 37193--37229, 2023.

\bibitem[Liu et~al.(2022)Liu, Liu, Jiang, Lyu, Wan, Shen, Liang, Fu, Wang, and
  Yi]{liu2022hoi4d}
Yunze Liu, Yun Liu, Che Jiang, Kangbo Lyu, Weikang Wan, Hao Shen, Boqiang
  Liang, Zhoujie Fu, He~Wang, and Li~Yi.
\newblock Hoi4d: A 4d egocentric dataset for category-level human-object
  interaction.
\newblock In \emph{2022 IEEE/CVF Conference on Computer Vision and Pattern
  Recognition (CVPR)}, pages 20981--20990. IEEE, 2022.

\bibitem[Liu et~al.(2025)Liu, Wang, Zheng, Pan, Liang, Fu, and
  Xue]{liu2025reasongrounder}
Zhenyang Liu, Yikai Wang, Sixiao Zheng, Tongying Pan, Longfei Liang, Yanwei Fu,
  and Xiangyang Xue.
\newblock Reasongrounder: Lvlm-guided hierarchical feature splatting for
  open-vocabulary 3d visual grounding and reasoning.
\newblock In \emph{Proceedings of the IEEE/CVF Conference on Computer Vision
  and Pattern Recognition (CVPR)}, pages 3718--3727, June 2025.

\bibitem[Miao et~al.(2022)Miao, Wang, Wu, Li, Zhang, Wei, and
  Yang]{miao2022large}
Jiaxu Miao, Xiaohan Wang, Yu~Wu, Wei Li, Xu~Zhang, Yunchao Wei, and Yi~Yang.
\newblock Large-scale video panoptic segmentation in the wild: A benchmark.
\newblock In \emph{2022 IEEE/CVF Conference on Computer Vision and Pattern
  Recognition (CVPR)}, pages 21001--21011. IEEE, 2022.

\bibitem[Milletari et~al.(2016)Milletari, Navab, and Ahmadi]{milletari2016v}
Fausto Milletari, Nassir Navab, and Seyed-Ahmad Ahmadi.
\newblock V-net: Fully convolutional neural networks for volumetric medical
  image segmentation.
\newblock In \emph{2016 fourth international conference on 3D vision (3DV)},
  pages 565--571. Ieee, 2016.

\bibitem[{OpenAI}(2026)]{openai2026gpt56}
{OpenAI}.
\newblock {GPT-5.6 System Card}, July 2026.
\newblock URL \url{https://deploymentsafety.openai.com/gpt-5-6}.
\newblock Published July 9, 2026.

\bibitem[Peng et~al.(2023)Peng, Genova, Jiang, Tagliasacchi, Pollefeys, and
  Funkhouser]{peng2023openscene}
Songyou Peng, Kyle Genova, Chiyu Jiang, Andrea Tagliasacchi, Marc Pollefeys,
  and Thomas Funkhouser.
\newblock Openscene: 3d scene understanding with open vocabularies.
\newblock In \emph{2023 IEEE/CVF Conference on Computer Vision and Pattern
  Recognition (CVPR)}, pages 815--824. IEEE, 2023.

\bibitem[Perazzi et~al.(2016)Perazzi, Pont-Tuset, McWilliams, Van~Gool, Gross,
  and Sorkine-Hornung]{perazzi2016benchmark}
Federico Perazzi, Jordi Pont-Tuset, Brian McWilliams, Luc Van~Gool, Markus
  Gross, and Alexander Sorkine-Hornung.
\newblock A benchmark dataset and evaluation methodology for video object
  segmentation.
\newblock In \emph{Proceedings of the IEEE conference on computer vision and
  pattern recognition}, pages 724--732, 2016.

\bibitem[Pont-Tuset et~al.(2017)Pont-Tuset, Perazzi, Caelles, Arbel{\'a}ez,
  Sorkine-Hornung, and Van~Gool]{pont20172017}
Jordi Pont-Tuset, Federico Perazzi, Sergi Caelles, Pablo Arbel{\'a}ez, Alex
  Sorkine-Hornung, and Luc Van~Gool.
\newblock The 2017 davis challenge on video object segmentation.
\newblock \emph{arXiv preprint arXiv:1704.00675}, 2017.

\bibitem[Prabhudesai et~al.(2020)Prabhudesai, Tung, Javed, Sieb, Harley, and
  Fragkiadaki]{prabhudesai2020embodied}
Mihir Prabhudesai, Hsiao-Yu~Fish Tung, Syed~Ashar Javed, Maximilian Sieb,
  Adam~W Harley, and Katerina Fragkiadaki.
\newblock Embodied language grounding with 3d visual feature representations.
\newblock In \emph{2020 IEEE/CVF Conference on Computer Vision and Pattern
  Recognition (CVPR)}, pages 2217--2226. IEEE, 2020.

\bibitem[Qu et~al.(2026)Qu, Li, and Zhang]{qu2026segvggt}
Jinyuan Qu, Hongyang Li, and Lei Zhang.
\newblock Segvggt: Joint 3d reconstruction and instance segmentation from
  multi-view images.
\newblock \emph{arXiv preprint arXiv:2603.19926}, 2026.

\bibitem[Raistrick et~al.(2023)Raistrick, Lipson, Ma, Mei, Wang, Zuo, Kayan,
  Wen, Han, Wang, et~al.]{raistrick2023infinite}
Alexander Raistrick, Lahav Lipson, Zeyu Ma, Lingjie Mei, Mingzhe Wang, Yiming
  Zuo, Karhan Kayan, Hongyu Wen, Beining Han, Yihan Wang, et~al.
\newblock Infinite photorealistic worlds using procedural generation.
\newblock In \emph{2023 IEEE/CVF Conference on Computer Vision and Pattern
  Recognition (CVPR)}, pages 12630--12641. IEEE, 2023.

\bibitem[Raji{\v{c}} et~al.(2025)Raji{\v{c}}, Ke, Tai, Tang, Danelljan, and
  Yu]{rajivc2025segment}
Frano Raji{\v{c}}, Lei Ke, Yu-Wing Tai, Chi-Keung Tang, Martin Danelljan, and
  Fisher Yu.
\newblock Segment anything meets point tracking.
\newblock In \emph{Proceedings of the Winter Conference on Applications of
  Computer Vision}, pages 9284--9293, 2025.

\bibitem[Rasheed et~al.(2024)Rasheed, Maaz, Shaji, Shaker, Khan, Cholakkal,
  Anwer, Xing, Yang, and Khan]{rasheed2024glamm}
Hanoona Rasheed, Muhammad Maaz, Sahal Shaji, Abdelrahman Shaker, Salman Khan,
  Hisham Cholakkal, Rao~M Anwer, Eric Xing, Ming-Hsuan Yang, and Fahad~S Khan.
\newblock Glamm: Pixel grounding large multimodal model.
\newblock In \emph{2024 IEEE/CVF Conference on Computer Vision and Pattern
  Recognition (CVPR)}, pages 13009--13018. IEEE, 2024.

\bibitem[Ravi et~al.(2025)Ravi, Gabeur, Hu, Hu, Ryali, Ma, Khedr, R{\"a}dle,
  Rolland, Gustafson, et~al.]{ravi2025sam}
Nikhila Ravi, Valentin Gabeur, Yuan-Ting Hu, Ronghang Hu, Chaitanya Ryali,
  Tengyu Ma, Haitham Khedr, Roman R{\"a}dle, Chloe Rolland, Laura Gustafson,
  et~al.
\newblock Sam 2: Segment anything in images and videos.
\newblock In \emph{International Conference on Learning Representations},
  volume 2025, pages 28085--28128, 2025.

\bibitem[Ren et~al.(2024)Ren, Huang, Wei, Zhao, Fu, Feng, and
  Jin]{ren2024pixellm}
Zhongwei Ren, Zhicheng Huang, Yunchao Wei, Yao Zhao, Dongmei Fu, Jiashi Feng,
  and Xiaojie Jin.
\newblock Pixellm: Pixel reasoning with large multimodal model.
\newblock In \emph{2024 IEEE/CVF Conference on Computer Vision and Pattern
  Recognition (CVPR)}, pages 26364--26373. IEEE, 2024.

\bibitem[Schult et~al.(2023)Schult, Engelmann, Hermans, Litany, Tang, and
  Leibe]{schult2023mask3d}
Jonas Schult, Francis Engelmann, Alexander Hermans, Or~Litany, Siyu Tang, and
  Bastian Leibe.
\newblock Mask3d: Mask transformer for 3d semantic instance segmentation.
\newblock In \emph{2023 IEEE International Conference on Robotics and
  Automation (ICRA)}, pages 8216--8223. IEEE, 2023.

\bibitem[Siddiqui et~al.(2023)Siddiqui, Porzi, Bul{\'o}, M{\"u}ller,
  Nie{\ss}ner, Dai, and Kontschieder]{siddiqui2023panoptic}
Yawar Siddiqui, Lorenzo Porzi, Samuel~Rota Bul{\'o}, Norman M{\"u}ller,
  Matthias Nie{\ss}ner, Angela Dai, and Peter Kontschieder.
\newblock Panoptic lifting for 3d scene understanding with neural fields.
\newblock In \emph{2023 IEEE/CVF Conference on Computer Vision and Pattern
  Recognition (CVPR)}, pages 9043--9052. IEEE, 2023.

\bibitem[Steiner et~al.(2026)Steiner, Zheng, Howard-Jenkins, Xie, and
  Armeni]{steiner2026rescene4d}
Emily Steiner, Jianhao Zheng, Henry Howard-Jenkins, Chris Xie, and Iro Armeni.
\newblock Rescene4d: Temporally consistent semantic instance segmentation of
  evolving indoor 3d scenes.
\newblock \emph{arXiv preprint arXiv:2601.11508}, 2026.

\bibitem[Sun et~al.(2026)Sun, Jiang, Liu, Nam, Kang, Wang, Sui, Su, Liu, Wang,
  et~al.]{sun2026uni3r}
Xiangyu Sun, Haoyi Jiang, Liu Liu, Seungtae Nam, Gyeongjin Kang, Xinjie Wang,
  Wei Sui, Zhizhong Su, Wenyu Liu, Xinggang Wang, et~al.
\newblock Uni3r: Unified 3d reconstruction and semantic understanding via
  generalizable gaussian splatting from unposed multi-view images.
\newblock In \emph{Proceedings of the IEEE/CVF Conference on Computer Vision
  and Pattern Recognition}, pages 33280--33290, 2026.

\bibitem[Takmaz et~al.(2023)Takmaz, Fedele, Sumner, Pollefeys, Tombari, and
  Engelmann]{takmaz2023openmask3d}
Ay{\c{c}}a Takmaz, Elisabetta Fedele, Robert~W Sumner, Marc Pollefeys, Federico
  Tombari, and Francis Engelmann.
\newblock Openmask3d: Open-vocabulary 3d instance segmentation.
\newblock \emph{arXiv preprint arXiv:2306.13631}, 2023.

\bibitem[Tang et~al.(2026)Tang, Zhou, Ye, Liu, Huang, and
  He]{tang2026panopticquery}
Ruilin Tang, Yang Zhou, Zhong Ye, Wenxi Liu, Yan Huang, and Shengfeng He.
\newblock Panopticquery: Unified query-time reasoning for 4d scenes.
\newblock \emph{arXiv preprint arXiv:2604.05638}, 2026.

\bibitem[Team et~al.(2025)Team, Liu, Kuo, Du, Chen, Chen, Yuan, Zhang, Guo, Li,
  et~al.]{team2025vidi}
Vidi Team, Celong Liu, Chia-Wen Kuo, Dawei Du, Fan Chen, Guang Chen, Jiamin
  Yuan, Lingxi Zhang, Lu~Guo, Lusha Li, et~al.
\newblock Vidi: Large multimodal models for video understanding and editing.
\newblock \emph{arXiv preprint arXiv:2504.15681}, 2025.

\bibitem[Tokmakov et~al.(2023)Tokmakov, Li, and Gaidon]{tokmakov2023breaking}
Pavel Tokmakov, Jie Li, and Adrien Gaidon.
\newblock Breaking the “object” in video object segmentation.
\newblock In \emph{2023 IEEE/CVF Conference on Computer Vision and Pattern
  Recognition (CVPR)}, pages 22836--22845. IEEE, 2023.

\bibitem[Vu et~al.(2022)Vu, Kim, Luu, Nguyen, and Yoo]{vu2022softgroup}
Thang Vu, Kookhoi Kim, Tung~M Luu, Thanh Nguyen, and Chang~D Yoo.
\newblock Softgroup for 3d instance segmentation on point clouds.
\newblock In \emph{2022 IEEE/CVF Conference on Computer Vision and Pattern
  Recognition (CVPR)}, pages 2698--2707. IEEE, 2022.

\bibitem[Wang and Agapito(2025)]{wang20253d}
Hengyi Wang and Lourdes Agapito.
\newblock 3d reconstruction with spatial memory.
\newblock In \emph{2025 International Conference on 3D Vision (3DV)}, pages
  78--89. IEEE, 2025.

\bibitem[Wang et~al.(2025{\natexlab{a}})Wang, Chen, Karaev, Vedaldi, Rupprecht,
  and Novotny]{wang2025vggt}
Jianyuan Wang, Minghao Chen, Nikita Karaev, Andrea Vedaldi, Christian
  Rupprecht, and David Novotny.
\newblock Vggt: Visual geometry grounded transformer.
\newblock In \emph{2025 IEEE/CVF Conference on Computer Vision and Pattern
  Recognition (CVPR)}, pages 5294--5306. IEEE, 2025{\natexlab{a}}.

\bibitem[Wang et~al.(2026{\natexlab{a}})Wang, Chen, Zhang, Karaev,
  Sch{\"o}nberger, Labatut, Bojanowski, Novotny, Vedaldi, and
  Rupprecht]{wang2026vggt}
Jianyuan Wang, Minghao Chen, Shangzhan Zhang, Nikita Karaev, Johannes
  Sch{\"o}nberger, Patrick Labatut, Piotr Bojanowski, David Novotny, Andrea
  Vedaldi, and Christian Rupprecht.
\newblock {VGGT}-$\omega$.
\newblock \emph{arXiv preprint arXiv:2605.15195}, 2026{\natexlab{a}}.

\bibitem[Wang et~al.(2026{\natexlab{b}})Wang, Liu, Kuang, Wei, Liu, Li, Man,
  Chen, Tao, Liu, et~al.]{wang2026locateanything}
Shihao Wang, Shilong Liu, Yuanguo Kuang, Xinyu Wei, Yangzhou Liu, Zhiqi Li,
  Yunze Man, Guo Chen, Andrew Tao, Guilin Liu, et~al.
\newblock Locateanything: Fast and high-quality vision-language grounding with
  parallel box decoding.
\newblock In \emph{European Conference on Computer Vision}, pages 336--357.
  Springer, 2026{\natexlab{b}}.

\bibitem[Wang et~al.(2024{\natexlab{a}})Wang, Leroy, Cabon, Chidlovskii, and
  Revaud]{wang2024dust3r}
Shuzhe Wang, Vincent Leroy, Yohann Cabon, Boris Chidlovskii, and Jerome Revaud.
\newblock Dust3r: Geometric 3d vision made easy.
\newblock In \emph{2024 IEEE/CVF Conference on Computer Vision and Pattern
  Recognition (CVPR)}, pages 20697--20709. IEEE, 2024{\natexlab{a}}.

\bibitem[Wang et~al.(2021)Wang, Feiszli, Wang, and Tran]{wang2021unidentified}
Weiyao Wang, Matt Feiszli, Heng Wang, and Du~Tran.
\newblock Unidentified video objects: A benchmark for dense, open-world
  segmentation.
\newblock In \emph{2021 IEEE/CVF International Conference on Computer Vision
  (ICCV)}, pages 10756--10765. IEEE, 2021.

\bibitem[Wang et~al.(2024{\natexlab{b}})Wang, Zhang, Zohar, and
  Yeung-Levy]{wang2024videoagent}
Xiaohan Wang, Yuhui Zhang, Orr Zohar, and Serena Yeung-Levy.
\newblock Videoagent: Long-form video understanding with large language model
  as agent.
\newblock In \emph{European Conference on Computer Vision}, pages 58--76.
  Springer, 2024{\natexlab{b}}.

\bibitem[Wang et~al.(2026{\natexlab{c}})Wang, Zhou, Zhu, Chang, Zhou, Li, Chen,
  Pang, Shen, and He]{wang2026pi}
Yifan Wang, Jianjun Zhou, Haoyi Zhu, Wenzheng Chang, Yang Zhou, Zizun Li, Junyi
  Chen, Jiangmiao Pang, Chunhua Shen, and Tong He.
\newblock {$\pi^3$}: Permutation-equivariant visual geometry learning.
\newblock In \emph{International Conference on Learning Representations},
  volume 2026, pages 10481--10497, 2026{\natexlab{c}}.

\bibitem[Wang et~al.(2022)Wang, Lu, Li, Tao, Guo, Gong, and Liu]{wang2022cris}
Zhaoqing Wang, Yu~Lu, Qiang Li, Xunqiang Tao, Yandong Guo, Mingming Gong, and
  Tongliang Liu.
\newblock Cris: Clip-driven referring image segmentation.
\newblock In \emph{2022 IEEE/CVF Conference on Computer Vision and Pattern
  Recognition (CVPR)}, pages 11676--11685. IEEE, 2022.

\bibitem[Wang et~al.(2025{\natexlab{b}})Wang, Yu, Stengel-Eskin, Yoon, Cheng,
  Bertasius, and Bansal]{wang2025videotree}
Ziyang Wang, Shoubin Yu, Elias Stengel-Eskin, Jaehong Yoon, Feng Cheng, Gedas
  Bertasius, and Mohit Bansal.
\newblock Videotree: Adaptive tree-based video representation for llm reasoning
  on long videos.
\newblock In \emph{2025 IEEE/CVF Conference on Computer Vision and Pattern
  Recognition (CVPR)}, pages 3272--3282. IEEE, 2025{\natexlab{b}}.

\bibitem[Wang et~al.(2026{\natexlab{d}})Wang, Zhou, Wang, Li, Xiong, Savarese,
  Bansal, Ryoo, and Niebles]{wang2026active}
Ziyang Wang, Honglu Zhou, Shijie Wang, Junnan Li, Caiming Xiong, Silvio
  Savarese, Mohit Bansal, Michael~S. Ryoo, and Juan~Carlos Niebles.
\newblock Active video perception: Iterative evidence seeking for agentic long
  video understanding.
\newblock In \emph{Proceedings of the IEEE/CVF Conference on Computer Vision
  and Pattern Recognition (CVPR) Findings}, pages 9088--9099, June
  2026{\natexlab{d}}.

\bibitem[Weber et~al.(2021)Weber, Xie, Collins, Zhu, Voigtlaender, Adam, Green,
  Geiger, Leibe, Cremers, et~al.]{weber2021step}
Mark Weber, Jun Xie, Maxwell Collins, Yukun Zhu, Paul Voigtlaender, Hartwig
  Adam, Bradley Green, Andreas Geiger, Bastian Leibe, Daniel Cremers, et~al.
\newblock Step: Segmenting and tracking every pixel.
\newblock \emph{arXiv preprint arXiv:2102.11859}, 2021.

\bibitem[Wei et~al.(2022)Wei, Wang, Schuurmans, Bosma, ichter, Xia, Chi, Le,
  and Zhou]{wei2022chain}
Jason Wei, Xuezhi Wang, Dale Schuurmans, Maarten Bosma, brian ichter, Fei Xia,
  Ed~Chi, Quoc~V Le, and Denny Zhou.
\newblock Chain-of-thought prompting elicits reasoning in large language
  models.
\newblock In S.~Koyejo, S.~Mohamed, A.~Agarwal, D.~Belgrave, K.~Cho, and A.~Oh,
  editors, \emph{Advances in Neural Information Processing Systems}, volume~35,
  pages 24824--24837. Curran Associates, Inc., 2022.
\newblock \doi{10.52202/068431-1800}.
\newblock URL
  \url{https://proceedings.neurips.cc/paper_files/paper/2022/file/9d5609613524ecf4f15af0f7b31abca4-Paper-Conference.pdf}.

\bibitem[Wu et~al.(2026{\natexlab{a}})Wu, Wang, Ji, Yao, Du, Kang, Fu, and
  Cao]{wu2026mvggt}
Changli Wu, Haodong Wang, Jiayi Ji, Yutian Yao, Chunsai Du, Jihua Kang, Yanwei
  Fu, and Liujuan Cao.
\newblock Mvggt: Multimodal visual geometry grounded transformer for multiview
  3d referring expression segmentation.
\newblock \emph{arXiv preprint arXiv:2601.06874}, 2026{\natexlab{a}}.

\bibitem[Wu et~al.(2022{\natexlab{a}})Wu, Jiang, Sun, Yuan, and
  Luo]{wu2022language}
Jiannan Wu, Yi~Jiang, Peize Sun, Zehuan Yuan, and Ping Luo.
\newblock Language as queries for referring video object segmentation.
\newblock In \emph{2022 IEEE/CVF Conference on Computer Vision and Pattern
  Recognition (CVPR)}, pages 4964--4974. IEEE, 2022{\natexlab{a}}.

\bibitem[Wu et~al.(2022{\natexlab{b}})Wu, Jiang, Bai, Zhang, and
  Bai]{wu2022seqformer}
Junfeng Wu, Yi~Jiang, Song Bai, Wenqing Zhang, and Xiang Bai.
\newblock Seqformer: Sequential transformer for video instance segmentation.
\newblock In Shai Avidan, Gabriel Brostow, Moustapha Ciss{\'e}, Giovanni~Maria
  Farinella, and Tal Hassner, editors, \emph{Computer Vision -- ECCV 2022},
  pages 553--569, Cham, 2022{\natexlab{b}}. Springer Nature Switzerland.
\newblock ISBN 978-3-031-19815-1.

\bibitem[Wu et~al.(2024)Wu, Jiang, Liu, Yuan, Bai, and Bai]{wu2024general}
Junfeng Wu, Yi~Jiang, Qihao Liu, Zehuan Yuan, Xiang Bai, and Song Bai.
\newblock General object foundation model for images and videos at scale.
\newblock In \emph{2024 IEEE/CVF Conference on Computer Vision and Pattern
  Recognition (CVPR)}, pages 3783--3795. IEEE, 2024.

\bibitem[Wu et~al.(2026{\natexlab{b}})Wu, Nguyen, Planche, Tao, Sun, Gao, Zhao,
  Choudhuri, Zhang, Zheng, et~al.]{wu2026consistent}
Junyi Wu, Van~Nguyen Nguyen, Benjamin Planche, Jiachen Tao, Changchang Sun,
  Zhongpai Gao, Zhenghao Zhao, Anwesa Choudhuri, Gengyu Zhang, Meng Zheng,
  et~al.
\newblock Consistent instance field for dynamic scene understanding.
\newblock In \emph{Proceedings of the IEEE/CVF Conference on Computer Vision
  and Pattern Recognition}, pages 3783--3793, 2026{\natexlab{b}}.

\bibitem[Wu et~al.(2025)Wu, Bai, Li, Wu, Zhao, Lai, Liu, and
  Wang]{wu20254dlangvggt}
Xianfeng Wu, Yajing Bai, Minghan Li, Xianzu Wu, Xueqi Zhao, Zhongyuan Lai,
  Wenyu Liu, and Xinggang Wang.
\newblock 4dlangvggt: 4d language-visual geometry grounded transformer.
\newblock \emph{arXiv preprint arXiv:2512.05060}, 2025.

\bibitem[Wu et~al.(2023)Wu, Cheng, Zhang, Cheng, and Zhang]{wu2023eda}
Yanmin Wu, Xinhua Cheng, Renrui Zhang, Zesen Cheng, and Jian Zhang.
\newblock Eda: Explicit text-decoupling and dense alignment for 3d visual
  grounding.
\newblock In \emph{2023 IEEE/CVF Conference on Computer Vision and Pattern
  Recognition (CVPR)}, pages 19231--19242. IEEE, 2023.

\bibitem[Xia et~al.(2024)Xia, Han, Han, Pan, Song, and Huang]{xia2024gsva}
Zhuofan Xia, Dongchen Han, Yizeng Han, Xuran Pan, Shiji Song, and Gao Huang.
\newblock Gsva: Generalized segmentation via multimodal large language models.
\newblock In \emph{2024 IEEE/CVF Conference on Computer Vision and Pattern
  Recognition (CVPR)}, pages 3858--3869. IEEE, 2024.

\bibitem[Xiong et~al.(2024)Xiong, Varadarajan, Wu, Xiang, Xiao, Zhu, Dai, Wang,
  Sun, Iandola, et~al.]{xiong2024efficientsam}
Yunyang Xiong, Bala Varadarajan, Lemeng Wu, Xiaoyu Xiang, Fanyi Xiao, Chenchen
  Zhu, Xiaoliang Dai, Dilin Wang, Fei Sun, Forrest Iandola, et~al.
\newblock Efficientsam: Leveraged masked image pretraining for efficient
  segment anything.
\newblock In \emph{2024 IEEE/CVF Conference on Computer Vision and Pattern
  Recognition (CVPR)}, pages 16111--16121. IEEE, 2024.

\bibitem[Xu et~al.(2025)Xu, Wang, Ni, Hu, Yang, Zhu, and Li]{xu2025sam4d}
Jianyun Xu, Song Wang, Ziqian Ni, Chunyong Hu, Sheng Yang, Jianke Zhu, and
  Qiang Li.
\newblock Sam4d: Segment anything in camera and lidar streams.
\newblock In \emph{2025 IEEE/CVF International Conference on Computer Vision
  (ICCV)}, pages 28535--28545. IEEE, 2025.

\bibitem[Xu et~al.(2018)Xu, Yang, Fan, Yang, Yue, Liang, Price, Cohen, and
  Huang]{xu2018youtube}
Ning Xu, Linjie Yang, Yuchen Fan, Jianchao Yang, Dingcheng Yue, Yuchen Liang,
  Brian Price, Scott Cohen, and Thomas Huang.
\newblock Youtube-vos: Sequence-to-sequence video object segmentation.
\newblock In \emph{European conference on computer vision}, pages 603--619.
  Springer, 2018.

\bibitem[Xu et~al.(2024)Xu, Huang, Wang, Chen, Pang, and Lin]{xu2024vlm}
Runsen Xu, Zhiwei Huang, Tai Wang, Yilun Chen, Jiangmiao Pang, and Dahua Lin.
\newblock Vlm-grounder: A vlm agent for zero-shot 3d visual grounding.
\newblock \emph{arXiv preprint arXiv:2410.13860}, 2024.

\bibitem[Yan et~al.(2024)Yan, Wang, Yan, Jiang, Hu, Kang, Xie, and
  Gavves]{yan2024visa}
Cilin Yan, Haochen Wang, Shilin Yan, Xiaolong Jiang, Yao Hu, Guoliang Kang,
  Weidi Xie, and Efstratios Gavves.
\newblock Visa: Reasoning video object segmentation via large language models.
\newblock In \emph{European Conference on Computer Vision}, pages 98--115.
  Springer, 2024.

\bibitem[Yang et~al.(2024)Yang, Chen, Qian, Madaan, Iyengar, Fouhey, and
  Chai]{yang2024llm}
Jianing Yang, Xuweiyi Chen, Shengyi Qian, Nikhil Madaan, Madhavan Iyengar,
  David~F Fouhey, and Joyce Chai.
\newblock Llm-grounder: Open-vocabulary 3d visual grounding with large language
  model as an agent.
\newblock In \emph{2024 IEEE International Conference on Robotics and
  Automation (ICRA)}, pages 7694--7701. IEEE, 2024.

\bibitem[Yang et~al.(2025)Yang, Sax, Liang, Henaff, Tang, Cao, Chai, Meier, and
  Feiszli]{yang2025fast3r}
Jianing Yang, Alexander Sax, Kevin~J Liang, Mikael Henaff, Hao Tang, Ang Cao,
  Joyce Chai, Franziska Meier, and Matt Feiszli.
\newblock Fast3r: Towards 3d reconstruction of 1000+ images in one forward
  pass.
\newblock In \emph{2025 IEEE/CVF Conference on Computer Vision and Pattern
  Recognition (CVPR)}, pages 21924--21935. IEEE, 2025.

\bibitem[Yang et~al.(2023{\natexlab{a}})Yang, Gao, Li, Gao, Wang, and
  Zheng]{yang2023track}
Jinyu Yang, Mingqi Gao, Zhe Li, Shang Gao, Fangjing Wang, and Feng Zheng.
\newblock Track anything: Segment anything meets videos.
\newblock \emph{arXiv preprint arXiv:2304.11968}, 2023{\natexlab{a}}.

\bibitem[Yang et~al.(2023{\natexlab{b}})Yang, Wu, He, Zhao, and
  Liu]{yang2023sam3d}
Yunhan Yang, Xiaoyang Wu, Tong He, Hengshuang Zhao, and Xihui Liu.
\newblock Sam3d: Segment anything in 3d scenes.
\newblock \emph{arXiv preprint arXiv:2306.03908}, 2023{\natexlab{b}}.

\bibitem[Ye et~al.(2024)Ye, Danelljan, Yu, and Ke]{ye2024gaussian}
Mingqiao Ye, Martin Danelljan, Fisher Yu, and Lei Ke.
\newblock Gaussian grouping: Segment and edit anything in 3d scenes.
\newblock In \emph{European conference on computer vision}, pages 162--179.
  Springer, 2024.

\bibitem[Yeshwanth et~al.(2023)Yeshwanth, Liu, Nie{\ss}ner, and
  Dai]{yeshwanth2023scannet++}
Chandan Yeshwanth, Yueh-Cheng Liu, Matthias Nie{\ss}ner, and Angela Dai.
\newblock Scannet++: A high-fidelity dataset of 3d indoor scenes.
\newblock In \emph{2023 IEEE/CVF International Conference on Computer Vision
  (ICCV)}, pages 12--22. IEEE, 2023.

\bibitem[Yin et~al.(2024)Yin, Liu, Xiao, Cohen-Or, Huang, and
  Chen]{yin2024sai3d}
Yingda Yin, Yuzheng Liu, Yang Xiao, Daniel Cohen-Or, Jingwei Huang, and Baoquan
  Chen.
\newblock Sai3d: Segment any instance in 3d scenes.
\newblock In \emph{2024 IEEE/CVF Conference on Computer Vision and Pattern
  Recognition (CVPR)}, pages 3292--3302. IEEE, 2024.

\bibitem[Yoon et~al.(2026)Yoon, Jung, Nguyen, Lin, and
  Manocha]{yoon2026panoseg3r}
Heechan Yoon, Dongki Jung, Phuc Nguyen, Ming Lin, and Dinesh Manocha.
\newblock Panoseg3r: Feed-forward 3d semantic segmentation for panoramic images
  with an automatic data curation pipeline.
\newblock \emph{arXiv preprint arXiv:2609.22687}, 2026.

\bibitem[Yuan et~al.(2026{\natexlab{a}})Yuan, Li, Zhang, Sun, Huang, Xu, Ji,
  Tong, Qi, Feng, et~al.]{yuan2026sa2va}
Haobo Yuan, Xiangtai Li, Tao Zhang, Yueyi Sun, Zilong Huang, Shilin Xu,
  Shunping Ji, Yunhai Tong, Lu~Qi, Jiashi Feng, et~al.
\newblock Sa2va: Marrying sam2 with mllm for dense grounded understanding of
  images and videos.
\newblock \emph{IEEE Transactions on Pattern Analysis and Machine
  Intelligence}, 2026{\natexlab{a}}.

\bibitem[Yuan et~al.(2026{\natexlab{b}})Yuan, Li, Li, Lin, Li, Tang, Xiao,
  Zhuang, and Zhang]{yuan2026instructsam}
Yuqian Yuan, Wentong Li, Zhaocheng Li, Yutong Lin, Juncheng Li, Siliang Tang,
  Jun Xiao, Yueting Zhuang, and Wenqiao Zhang.
\newblock Instructsam: Segment any instance with any instructions.
\newblock \emph{arXiv preprint arXiv:2605.26102}, 2026{\natexlab{b}}.

\bibitem[Zhang et~al.(2026{\natexlab{a}})Zhang, Zhan, Zhang, Wang, Yu, Wang,
  Wang, and Li]{zhang2026mtpano}
Jingdong Zhang, Xiaohang Zhan, Lingzhi Zhang, Yizhou Wang, Zhengming Yu,
  Jionghao Wang, Wenping Wang, and Xin Li.
\newblock Mtpano: Multi-task panoramic scene understanding via label-free
  integration of dense prediction priors.
\newblock In \emph{Proceedings of the Special Interest Group on Computer
  Graphics and Interactive Techniques Conference Conference Papers}, pages
  1--11, 2026{\natexlab{a}}.

\bibitem[Zhang et~al.(2026{\natexlab{b}})Zhang, Konz, Kramer, and
  Mazurowski]{zhang2024quantifying}
Yixin Zhang, Nicholas Konz, Kevin Kramer, and Maciej~A. Mazurowski.
\newblock Quantifying the limits of segmentation foundation models: Modeling
  challenges in segmenting tree-like and low-contrast objects.
\newblock In \emph{2026 IEEE/CVF Winter Conference on Applications of Computer
  Vision (WACV)}, pages 5205--5215, 2026{\natexlab{b}}.
\newblock \doi{10.1109/WACV61042.2026.00505}.

\bibitem[Zhao et~al.(2021)Zhao, Cai, Sheng, and Xu]{zhao20213dvg}
Lichen Zhao, Daigang Cai, Lu~Sheng, and Dong Xu.
\newblock 3dvg-transformer: Relation modeling for visual grounding on point
  clouds.
\newblock In \emph{2021 IEEE/CVF International Conference on Computer Vision
  (ICCV)}, pages 2908--2917. IEEE, 2021.

\bibitem[Zheng et~al.(2020)Zheng, Zhang, Li, Tang, Gao, and
  Zhou]{zheng2020structured3d}
Jia Zheng, Junfei Zhang, Jing Li, Rui Tang, Shenghua Gao, and Zihan Zhou.
\newblock Structured3d: A large photo-realistic dataset for structured 3d
  modeling.
\newblock In \emph{European Conference on Computer Vision}, pages 519--535.
  Springer, 2020.

\bibitem[Zhou et~al.(2025)Zhou, Ren, Weng, Zhang, Wang, Xu, Fan, You, Wang,
  Guibas, et~al.]{zhou2025feature4x}
Shijie Zhou, Hui Ren, Yijia Weng, Shuwang Zhang, Zhen Wang, Dejia Xu, Zhiwen
  Fan, Suya You, Zhangyang Wang, Leonidas Guibas, et~al.
\newblock Feature4x: Bridging any monocular video to 4d agentic ai with
  versatile gaussian feature fields.
\newblock In \emph{2025 IEEE/CVF Conference on Computer Vision and Pattern
  Recognition (CVPR)}, pages 14179--14190. IEEE, 2025.

\bibitem[Zhou et~al.(2018)Zhou, Tucker, Flynn, Fyffe, and
  Snavely]{zhou2018stereo}
Tinghui Zhou, Richard Tucker, John Flynn, Graham Fyffe, and Noah Snavely.
\newblock Stereo magnification: Learning view synthesis using multiplane
  images.
\newblock \emph{arXiv preprint arXiv:1805.09817}, 2018.

\bibitem[Zhou et~al.(2024)Zhou, Gu, Chiang, Xiang, and Su]{zhou2024point}
Yuchen Zhou, Jiayuan Gu, Tung~Yen Chiang, Fanbo Xiang, and Hao Su.
\newblock Point-sam: Promptable 3d segmentation model for point clouds.
\newblock \emph{arXiv preprint arXiv:2406.17741}, 2024.

\bibitem[Zhu et~al.(2024)Zhu, Wang, Zhang, Chen, and Liu]{zhu2024scanreason}
Chenming Zhu, Tai Wang, Wenwei Zhang, Kai Chen, and Xihui Liu.
\newblock Scanreason: Empowering 3d visual grounding with reasoning
  capabilities.
\newblock In \emph{European Conference on Computer Vision}, pages 151--168.
  Springer, 2024.

\bibitem[Zou et~al.(2023)Zou, Yang, Zhang, Li, Li, Wang, Wang, Gao, and
  Lee]{zou2023segment}
Xueyan Zou, Jianwei Yang, Hao Zhang, Feng Li, Linjie Li, Jianfeng Wang, Lijuan
  Wang, Jianfeng Gao, and Yong~Jae Lee.
\newblock Segment everything everywhere all at once.
\newblock \emph{Advances in neural information processing systems},
  36:\penalty0 19769--19782, 2023.

\bibitem[Zou et~al.(2026)Zou, Li, Jiao, Liu, Xiao, Zhou, Hong, Su, Zhang, and
  Liu]{zou2026iggt4d}
Zhengyu Zou, Hao Li, Kuixuan Jiao, Liu Liu, Tingyang Xiao, Xiaolin Zhou,
  Fangzhou Hong, Zhizhong Su, Dingwen Zhang, and Ziwei Liu.
\newblock Iggt4d: Streaming 4d instance-grounded geometry transformer.
\newblock \emph{arXiv preprint arXiv:2607.19228}, 2026.

\bibitem[Zust et~al.(2025)Zust, Cabon, Marrie, Antsfeld, Chidlovskii, Revaud,
  and Csurka]{zust2025panst3r}
Lojze Zust, Yohann Cabon, Juliette Marrie, Leonid Antsfeld, Boris Chidlovskii,
  Jerome Revaud, and Gabriela Csurka.
\newblock Panst3r: Multi-view consistent panoptic segmentation.
\newblock In \emph{2025 IEEE/CVF International Conference on Computer Vision
  (ICCV)}, pages 5856--5886. IEEE, 2025.

\end{thebibliography}
